\documentclass{article}
\usepackage{iclr2027_conference,times}

\usepackage{amsmath,amsfonts,bm}

\def\eqref#1{equation~\ref{#1}}

\def\1{\bm{1}}

\DeclareMathAlphabet{\mathsfit}{\encodingdefault}{\sfdefault}{m}{sl}
\SetMathAlphabet{\mathsfit}{bold}{\encodingdefault}{\sfdefault}{bx}{n}

\usepackage{hyperref}
\usepackage{url}
\usepackage{graphicx}
\usepackage{booktabs}
\usepackage{multirow}

\usepackage{amsthm}
\usepackage{amssymb}
\usepackage[table]{xcolor}
\usepackage{float}
\usepackage{enumitem}

\newtheorem{definition}{Definition}
\newtheorem{proposition}{Proposition}
\newtheorem{corollary}{Corollary}

\title{What Happens to the Goal? Distinct Failures of Policy Alignment in Goal Misgeneralization}

\author{Julia Barth$^{1,2}$, Armin Sommer$^{1,2}$, Kunvar Thaman$^{1}$, Jannik Schilling$^{1}$ \\
$^{1}$Pantheon Industries, $^{2}$ETH Zurich\thanks{Code: \href{https://github.com/Pantheon-Industries-Inc/what-happens-to-the-goal}{\texttt{github.com/Pantheon-Industries-Inc/what-happens-to-the-goal}}}
}

\newcommand{\quadname}{\textsc{WalkToFlag}}

\iclrfinalcopy
\begin{document}

\maketitle
\lhead{Preprint}

\begin{abstract}
Goal misgeneralization is a common failure mode in which an agent behaves competently out of distribution but pursues a proxy that coincided with the intended goal during training instead of the goal itself. However, we still have a weak understanding of what happens to the intended goal in misgeneralizing agents and what enables them to be recovered without retraining.
We study this across eight visual navigation and manipulation tasks, including robot control.
While tracing where goal information remains available internally, how strongly it influences the policy, and what role it plays in action selection, we uncover that the goal is lost at one of three different depths. This depth predicts which recovery works in each task.
When the goal is \emph{followed, but outcompeted} by the proxy, steering the existing policy redirects it to the goal's location, raising goal reach from $16.0\%$ to $95.3\%$. When it is \emph{read, but not followed}, steering fails, but a lightweight new downstream wiring recovers goal reach from $31.6\%$ to $84.3\%$. When it is \emph{not read}, none of the methods work, as the policy does not use the goal information.
The same tracing also reveals what replaces the goal: functional target substitution, where behavior follows an alternative goal, a policy shortcut, where behavior follows a learned action strategy, or cue competition between intended and proxy goal cues.
Knowing where the goal is lost tells us whether an agent can be repaired or has to be retrained. Goal misgeneralization should therefore be studied and evaluated inside the agent, not only from its behavior.
\end{abstract}

\section{Introduction}
\label{sec:intro}

\begin{figure}[t]
    \centering
    \includegraphics[width=\linewidth]{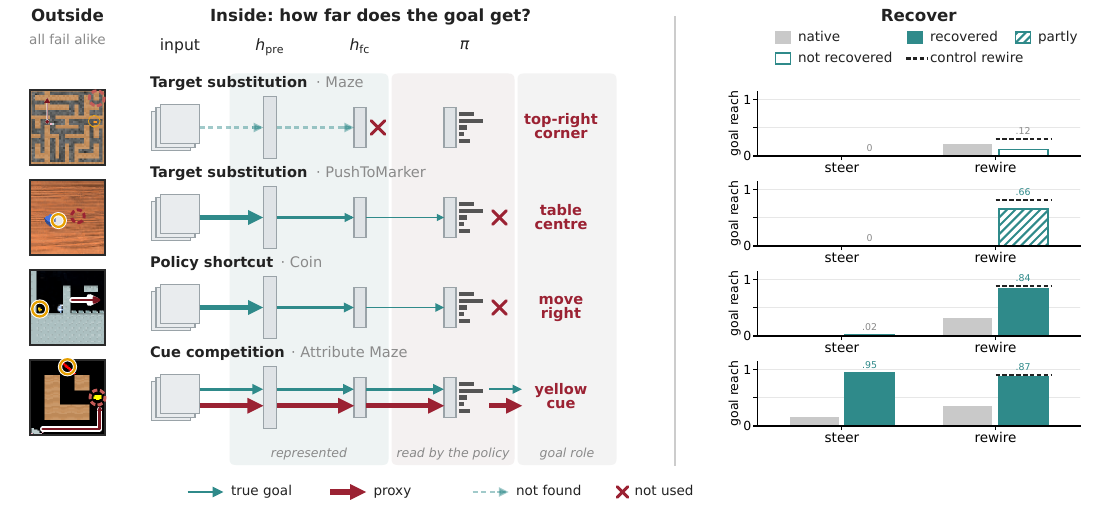}
    \caption{
    \textbf{All four agents fail alike from the outside, but inside the goal gets lost at different depths, and only a recovery that acts where the goal still survives brings it back.}
    From left to right, the proxy (dashed ring) and goal (solid ring) in out-of-distribution frames, how far goal information travels through the agent, where \textsc{Maze} and \textsc{PushToMarker} share a role but not a depth, and goal reach after steering at $h_{\mathrm{fc}}$ and after a new readout from $h_{\mathrm{pre}}$, with partial recovery hatched and the control rewire dashed (all eight environments in Figure~\ref{fig:app-overview-all8}).
    }
    \label{fig:overview}
\end{figure}

Reinforcement learning often specifies the intended goal an agent should pursue only indirectly, through rewards that indicate it or progress toward it.
Whenever a proxy co-occurs with the goal throughout training, pursuing either earns the same reward, and the agent can learn the proxy in its place.
This is why goal misgeneralization is common in reinforcement learning, and it is dangerous because the agent stays competent and its misalignment only becomes visible once the co-occurrence breaks.

So far, goal misgeneralization has mainly been characterized behaviorally, in large language models \citep{shah,maniyar2026position}, in toy reinforcement-learning environments such as Procgen \citep{langosco,raman} and in large pretrained control policies \citep{jucys2024vpt,zhou2025liberopro}.
Interpretability work has studied individual agents in these environments \citep{mini2023understandingcontrollingmazesolvingpolicy,GMhelp}, mostly to understand how they navigate or represent their goal, but not goal misgeneralization itself, and mitigation attempts intervene broadly during training \citep{langosco,sadek2025mitigating,naitelbarj2024,armstrong2023ace,GMhelp} rather than after it.
Studying goal misgeneralization in depth, from the inside and the outside, closes the gap between these lines of work, and we additionally carry it into a new and wider domain, robot control.
We therefore ask: \emph{what actually happens to the intended goal in misgeneralized agents?}

To answer it, we trace the goal through the agent from the outside and the inside, asking whether the goal is represented, whether the policy reads it, whether behavior is organized around it, and whether goal-directed behavior can be recovered at test time.
We analyze three classic goal-misgeneralization environments from Procgen, \textsc{Maze}, \textsc{Coin} and Maze~II, which we call \textsc{Attribute Maze}, and five new robot tasks on a simulated arm and quadruped, such as \textsc{PushToMarker} (Figure~\ref{fig:overview}, Figure~\ref{fig:app-overview-all8}).

What looks like one failure from the outside is three different failures of policy alignment: target substitution, a policy shortcut and cue competition. Even more importantly, we find three varying depths at which the goal is lost, which predict which recovery can work.
In \textsc{Maze}, replicated in a simulated arm task, the goal is inert to the policy and a place substitutes for it.
In \textsc{PushToMarker} a place substitutes for the goal as well, but the policy reacts to the goal weakly without following it.
This is the same depth at which an action shortcut bypasses the goal in \textsc{Coin}.
In \textsc{Attribute Maze}, replicated in three more robot tasks, the goal is followed as a cue or an object, but is outcompeted by the proxy.
We find that which repair works follows the depth, not the failure type.
Steering the existing policy redirects it to the goal's location only where the goal is followed.
Relearning the readout recovers goal-directed behavior wherever the goal still reaches the policy, and where the goal is inert, neither repair works.
Misgeneralized agents therefore cannot be understood from their behavior alone, as only knowing where the goal is lost tells us which repair can bring it back.

Our main contributions are:
\begin{enumerate}[leftmargin=*,itemsep=0pt,topsep=1pt]
\item To our knowledge, we are the first to relate which repair succeeds to where the goal is lost, and to recover goal-directed behavior at test time without further reinforcement learning or retraining the representation (Section~\ref{sec:recover}).
\item We trace the intended goal through misgeneralized agents in eight environments and find three distinct failures of policy alignment at three depths, including the first internal characterization of cue and object competition and its geometry (Section~\ref{sec:goal}).
\item We add an internal view and deepen the behavioral one to better understand goal misgeneralization. We propose the Fisher information of the policy with respect to its internal representations to isolate a policy-directed view of the internals, which lets us characterize, for example, different styles of competition. It also sharpens a proxy characterization of \citet{langosco}, as the \textsc{Maze} agent follows the top-right extreme of the current maze rather than a fixed corner.
\item We are the first to induce goal misgeneralization systematically in reinforcement learning for robot control and diagnose it in five simulated arm and quadruped tasks, where the failures recur.
\end{enumerate}

\section{Setup}
\label{sec:setup}

\paragraph{Problem formulation.}
We model each environment as an episodic Markov decision process $\mathcal M=(\mathcal S,\mathcal A,P,r,\rho_0)$.
The agent observes a rendered image of the state, which we write as $s$, samples actions $a_t\sim\pi(\cdot\mid s_t)$, and a reward $r$ is assigned to indicate an intended goal $g^\star$ or progress toward it.
Goal misgeneralization arises when, on a training distribution $\Lambda_{\mathrm{train}}$, reaching the goal coincides with reaching a proxy, so that the agent learns to follow the proxy instead of the goal on a shifted evaluation distribution $\Lambda_{\mathrm{OOD}}$, where the two no longer coincide \citep{langosco,shah}.
We write $\Gamma^\star(\tau)=1$ when a trajectory $\tau$ reaches the goal and $\Gamma^{\mathrm{prox}}(\tau)=1$ when it reaches the proxy.
Our main behavioral metric is the goal reach $\hat J^\star=\tfrac1N\sum_{i=1}^{N}\Gamma^\star(\tau_i)$, the fraction of $N$ rollouts that reach the goal.
Formal setup and definitions are in Appendices~\ref{app:formal-setup} and~\ref{app:gm-definitions}.

\paragraph{Environments.}
We study eight instances of visual goal misgeneralization, where goal and proxy are both visible and coincide in training.
We present four of them here and report the replications in Appendix~\ref{sec:experiments-app}.
Three are procedurally generated video games (Procgen, \citealt{cobbe2020procgen}, Appendix~\ref{sec:environments}).
In \textsc{Maze}, the cheese sits in the top-right corner during training, and at test time the agent walks to the corner instead of the cheese placed elsewhere \citep{langosco}.
In \textsc{Coin}, a side-scrolling game, the coin sits at the right end of the level during training, and at test time the agent runs to the right end and ignores a coin placed elsewhere \citep{cobbe2019coinrun,langosco}.
In \textsc{Attribute Maze}, a $5\times5$ variant of Maze~II \citep{langosco,raman}, the rewarded object is a yellow line during training, so colour and shape are two cues for the same object, and at test time, when the cues are split across objects, the agent follows the yellow object instead of the line when both are equally far away.
The fourth is a robot-arm task, \textsc{PushToMarker}, in which a simulated Franka arm pushes a cube onto a marker \citep{tao2025maniskill3}.
During training, the marker sits at the centre of the table, and at test time the arm pushes the cube to the centre and ignores the marker placed elsewhere.
Four further robot tasks, three on the same arm and one on a quadruped, replicate these failures (Appendices~\ref{app:env-arm} and~\ref{app:env-quadruped}).
In each environment, we compare three misgeneralized (GM) agents with three control (Non-GM) agents trained on diversified distributions (Appendix~\ref{app:gm-qualification}).

\paragraph{Architecture.}
All Procgen agents are non-recurrent actor--critic networks with an IMPALA architecture \citep{espeholt2018impala} and all robot agents use a Nature CNN \citep{mnih2015human}, both trained online with PPO \citep{schulman2017ppo} (training and implementation details in Appendix~\ref{app:training}, architectures in Appendix~\ref{sec:architecture}).
In both networks, convolutional layers produce feature maps $H^{(k)}$, whose average-pooled vectors $h^{(k)}$ we use for probing, and a fully connected layer maps the flattened last feature map $h_{\mathrm{pre}}$ to $h_{\mathrm{fc}}$.
Most of our analyses read $h_{\mathrm{pre}}$, the flattened last feature map and the last representation that keeps the spatial layout of the scene, and $h_{\mathrm{fc}}$, which the policy and value heads read directly (Appendix~\ref{sec:architecture}).
We write $\tilde\pi(\cdot\mid h)$ for the policy that the downstream layers compute from a representation $h$, so that $\tilde\pi(\cdot\mid h(s))=\pi(\cdot\mid s)$.

\section{What happens to the intended goal?}
\label{sec:goal}

\subsection{Is the goal still there?}
\label{sec:goal-there}

If the true goal gets replaced by a different goal, the simplest question we can ask is whether the true goal is still in the misgeneralized agent.
We test this using linear decodability \citep{alain2017probes}, by fitting linear probes at several points in the agent (Appendix~\ref{sec:architecture}) on configuration sets that separate constructed goal concepts, scored by AUC for binary and $R^2$ for continuous concepts.
For example, a bearing probe tests whether the agent encodes where the goal lies relative to itself.
Probe construction, controls \citep{hewitt2019control}, and further probes are detailed in Appendix~\ref{app:probes}.

\begin{figure}[t]
    \centering
    \includegraphics[width=0.93\linewidth]{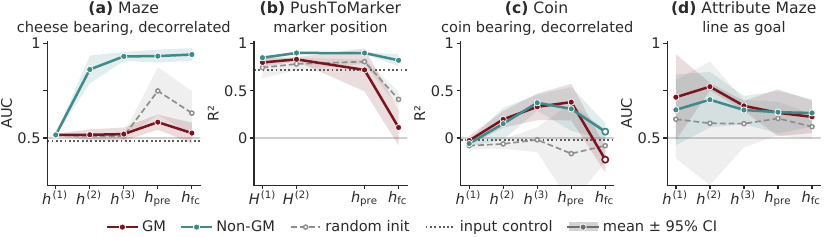}
    \caption{
    \textbf{Is the goal still there?}
    Held-out probe AUC (\textsc{PushToMarker}, \textsc{Coin}: $R^2$) across depth, 95\% $t$-intervals.
    (a) Cheese bearing on a position-balanced set and (c) coin bearing with agent position and level progress regressed out, so that agent position is uninformative.
    Dashed: randomly initialized networks, dotted: input-level controls (agent position in a and c, pixels in b).
    Robot $H^{(1)}$, $H^{(2)}$ are unpooled (Appendix~\ref{app:probes}). Hollow Coin $h_{\mathrm{fc}}$ points are not interpreted.
    }
    \label{fig:main-probes}
\end{figure}

For \textsc{Maze}, the cheese object is trivially decodable from the input already, so we probe cheese bearing. This needs a set on which the agent's own position is decorrelated from the cheese bearing, since otherwise a bearing probe reads the GM agent's position as a proxy (Appendix~\ref{app:probes}).
On this set the cheese bearing stays near chance at every site in GM agents ($0.516$ to $0.583$) and below randomly initialized networks at $h_{\mathrm{pre}}$ ($0.748$), while control agents decode it well ($0.932$) (Figure~\ref{fig:main-probes}a).
Similarly, for \textsc{PushToMarker} the marker position is decodable at the level of a randomly initialized network in early layers and falls below it in the later layers ($0.110$ against $0.406$ at $h_{\mathrm{fc}}$), while control agents keep it ($0.819$) (Figure~\ref{fig:main-probes}b).
In both environments, therefore, the bearing of the cheese and the position of the marker cannot be linearly decoded beyond what a random network exposes.
For \textsc{Coin}, the coin object is also largely decodable from the input, so we test coin bearing with the agent's position and level progress regressed out. This bearing stays decodable in both cohorts from $h^{(2)}$ to $h_{\mathrm{pre}}$ ($R^2$ of $0.378$ for GM and $0.306$ for Non-GM at $h_{\mathrm{pre}}$), while randomly initialized networks and the agent's position alone decode none of it, so we find linear traces of the coin inside the agent (Figure~\ref{fig:main-probes}c).
Lastly, for \textsc{Attribute Maze}, the goal cue, whether the reward-carrying object is the line, stays decodable in both cohorts alongside the proxy cue, most clearly in early layers (AUC $0.770$ for GM and $0.701$ for Non-GM at $h^{(2)}$, against $0.577$ for random networks), and comes close to random networks at $h_{\mathrm{pre}}$ ($0.632$ and $0.636$ against $0.603$).
So in \textsc{Maze} and \textsc{PushToMarker} the GM agents carry no linearly decodable goal information beyond what a randomly initialized network exposes, while in \textsc{Coin} and \textsc{Attribute Maze} the goal stays decodable in both cohorts.
The results replicate in the other robot tasks (Appendix~\ref{app:probes}).
Our probes show what is linearly available, but they cannot rule out goal information in another form.
Since information that is present is not necessarily used \citep{elazar2021amnesic}, we next ask whether the policy reacts to changes of the goal, using world edits.

\subsection{Does the policy read out the intended goal?}
\label{sec:goal-read}

We answer this question by intervening on the goal in the environment, from two perspectives: from the outside, we measure the change in behavior, and from the inside, whether the policy distribution moves along the corresponding representation displacement.
A world edit $e$ changes the goal in the level while keeping everything else fixed, and turns a state $s$ into a matched state $s_e$, with $\Delta_e f(s)=f(s_e)-f(s)$ for any quantity $f$, and $e=\star$ or $e=p$ for an edit of the goal or of the proxy (Figure~\ref{fig:worldedit-examples}).
In \textsc{Maze}, \textsc{Coin} and \textsc{PushToMarker}, we relocate the goal object, the cheese, the coin or the marker, as in earlier interventions on the \textsc{Maze} agent \citep{mini2023understandingcontrollingmazesolvingpolicy}. In \textsc{Attribute Maze}, in contrast, goal and proxy cue sit on the same object, so instead of moving the object, we keep all locations fixed and swap one characteristic between the two candidates: the shape swap changes the goal cue ($e=\star$) and the colour swap the proxy cue ($e=p$).
All edits are then evaluated on banks of conflict states, in which the agent has to choose between the proxy and the intended goal (Appendices~\ref{app:world-edit} and~\ref{app:banks}).

The immediate behavioral response is the change in the policy, which we measure with the KL divergence,
$D_{\mathrm{KL}}\big(\pi(\cdot\mid s)\,\|\,\pi(\cdot\mid s_e)\big)$.
The KL says whether the policy reacts but not in which direction, so we measure with signed follow whether the action mass moves toward the edited object,
\begin{equation}
\mathrm{SF}_e(s)=\pi\big(A_e(s_e)\mid s_e\big)-\pi\big(A_e(s_e)\mid s\big),
\qquad
\pi(A\mid s)=\sum_{a\in A}\pi(a\mid s),
\label{eq:main-sf}
\end{equation}
where $A_e(s_e)$ holds the actions on the shortest path to the edited object at its new place in \textsc{Maze} and \textsc{Attribute Maze} and the coin-directed, leftward actions in \textsc{Coin}, while for the continuous robot policies signed follow measures how far the planned step turns toward the object's new place (Appendix~\ref{app:world-edit}).
Inside the agent, the same edit moves the representation read by the policy by $\Delta_e(s)=\Delta_e h(s)=h(s_e)-h(s)$.
We measure how strongly the policy's action distribution changes through this displacement using the Fisher information of the downstream policy at $h=h(s)$,
\begin{equation}
F_h=\mathbb{E}_{a\sim\tilde\pi(\cdot\mid h)}\Big[\nabla_h\log\tilde\pi(a\mid h)\,\big(\nabla_h\log\tilde\pi(a\mid h)\big)^{\top}\Big],
\qquad
Q_e(s)=\Delta_e(s)^{\top}F_h\,\Delta_e(s).
\label{eq:main-q}
\end{equation}
This way, we only look at changes in the representation that are relevant to the policy, filtered through the Fisher information, instead of the raw change in the representation space. We use this Fisher lens throughout the paper as we are mainly concerned with how the policy reads information from its representation instead of the absolute change itself.
Up to second order, this Fisher sensitivity is exactly the KL divergence, measured from the inside \citep{amari1998natural,martens2020natural},
\[
D_{\mathrm{KL}}\big(\tilde\pi(\cdot\mid h)\,\big\|\,\tilde\pi(\cdot\mid h+t\Delta_e(s))\big)=\tfrac{t^2}{2}\,Q_e(s)+o(t^2),
\]%
and we check this for finite edits in Appendix~\ref{sec:fisher}.
KL, $Q_e$ and signed follow have different scales across environments, so we report each as the ratio of the GM to the Non-GM cohort mean.

\begin{table}[!ht]
\centering
\footnotesize
\setlength{\tabcolsep}{6pt}
\renewcommand{\arraystretch}{0.95}
\begin{tabular}{llccc}
\toprule
Environment & Edit & KL ratio & $Q$ ratio ($h_{\mathrm{fc}}$) & Signed-follow ratio \\
\midrule
\textsc{Maze} & Cheese & $0.009\pm0.01$ & $0.001\pm0.00$ & $-0.002\pm0.00$ \\
\textsc{PushToMarker} & Marker & $0.206\pm0.07$ & $0.159\pm0.08$ & $0.017\pm0.05$ \\
\textsc{Coin} & Coin & $0.114\pm0.05$ & $0.063\pm0.03$ & $0.001\pm0.04$ \\
\textsc{Attribute Maze} & Shape & $0.200\pm0.09$ & $0.307\pm0.13$ & $0.533\pm0.10$ \\
\bottomrule
\end{tabular}
\caption{
\textbf{Policy response to a goal edit, as the ratio of GM to Non-GM agents.}
Each ratio divides the GM by the Non-GM mean over three seeds, with $Q$ at $h_{\mathrm{fc}}$, so a ratio below $1$ means GM agents react less.
The $\pm$ is the SD, propagated to the ratio.
Absolute values are in Tables~\ref{tab:world-edit-dependence} and~\ref{tab:fisher-decomposition}.
}
\label{tab:main-outside-inside}
\end{table}

We find that the outside and inside perspectives agree, through the KL and $Q$ ratios, and single out \textsc{Maze}, where the goal is not read (Table~\ref{tab:main-outside-inside}).
In \textsc{Maze} (replicated in one robot task, Appendix~\ref{app:world-edit}), the GM policy barely reacts to the cheese edit (KL ratio $0.009$), so changes of the goal are read out by the policy neither behaviorally nor inside its representation.
In \textsc{PushToMarker}, \textsc{Coin} and under the \textsc{Attribute Maze} shape swap, the goal edit moves the GM policy by similar fractions of the Non-GM reaction ($0.206$, $0.114$ and $0.200$).
If we then ask whether this shift of action mass is directed at the goal, signed follow separates these three as well.
In \textsc{PushToMarker} and \textsc{Coin} the policy reacts to the goal edit, but the shift is not directed at the goal (signed-follow ratios $0.017$ and $0.001$), so the policy is receptive to a change of the goal object but does not organize its longer-term behavior around it.
In \textsc{Attribute Maze}, the shape swap turns the policy toward the goal at about half the Non-GM turn ($0.533$).
For the proxy colour swap, the GM policy reacts about fifty times more than the Non-GM policy, which barely follows the colour (KL $1.258$ against $0.025$, signed follow $+0.463$ against $+0.047$).
So the goal does not reach the policy in \textsc{Maze}, reaches it without directing it in \textsc{PushToMarker} and \textsc{Coin}, and is followed but loses to the proxy in \textsc{Attribute Maze}, and the other four robot tasks fall into the same depths (Appendix~\ref{app:world-edit}).
These depths cut across the failure types, as \textsc{Maze} and \textsc{PushToMarker} share a failure type but not a depth.
The policy sensitivity only captures the next action, so we ask what organizes the behavior instead.

\subsection{What organizes the behavior instead?}
\label{sec:goal-role-main}

\begin{figure}[t]
    \centering
    \includegraphics[width=\linewidth]{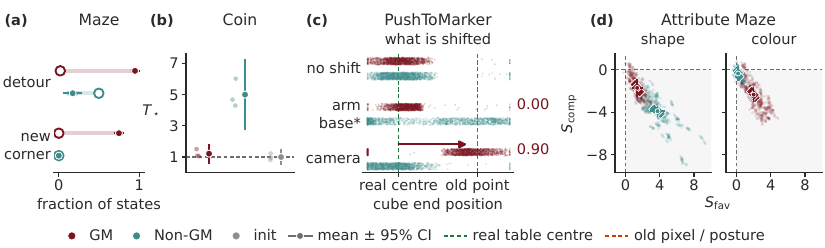}
    \caption{
    \textbf{What organizes behavior instead?}
    (a) \textsc{Maze}: fraction of states moving toward the corner or new corner (filled) versus the cheese or old corner (open).
    (b) \textsc{Coin}: directed share $T_\star$, where $T_\star=1$ (dashed) marks no preferred direction.
    (c) \textsc{PushToMarker}: cube end positions with no shift, after shifting the arm base and after shifting the camera, between the real table centre ($0$) and the old point ($1$), with the GM fraction ending at the old point. Only the camera shift moves where GM agents push.
    (d) \textsc{Attribute Maze}: $S_{\mathrm{fav}}$ and $S_{\mathrm{comp}}$ per swap, states (faint) and seed means (diamonds), push--pull quadrant shaded.
    Mean $\pm$ 95\% CI over seeds.
    }
    \label{fig:main-goal-role}
\end{figure}

Where the goal is fully inert, as in \textsc{Maze}, we want to understand what replaces it: a directional shortcut, a memorized location, or a spatial concept.
We first select detour states, on which every action that makes progress toward the top-right corner points neither up nor right.
GM agents still choose these detour actions on $0.948$ of the states, which rules out a directional shortcut (Figure~\ref{fig:main-goal-role}a).
We then open a path to a new top-right extreme, inspired by the corner edit of \citet{mini2023understandingcontrollingmazesolvingpolicy}, and ask which corner the agent follows.
It reaches the new corner first in $0.748$ of the rollouts and the old one never, and a new extreme placed away from the screen corner is still reached first in $0.792$ of rollouts, against $0.114$ for the old one, which rules out a memorized location (Figure~\ref{fig:main-goal-role}a).
Combined with the inert goal, this shows a functional target substitution, and we refer to Appendix~\ref{app:maze-functional-role} for the full analysis.
For \textsc{Maze}, the substitute target is thus not a fixed place but a spatial relation, the top-right extreme of the maze.
This refines \citet{langosco}, who classify this failure as a location proxy but do not test what underlies it.

As the agent in \textsc{Coin} seems to follow a simple rule, ``go right'', we ask a different question here: is this a weaker substitution, or does the agent follow a simple action rule, a policy shortcut?
We find that the latter holds for \textsc{Coin}.
First, when the end of the level is made traversable, GM agents run past the former endpoint in $0.938$ of the episodes that reach it and keep choosing rightward actions at the same rate after passing it, which reconfirms the observation of \citet{langosco}.
Second, the coin's weak influence carries no direction.
When the coin moves, the change in the policy is spread over the actions in proportion to their probability. We measure this with the directed share
\[
T_\star=\frac{\sum_{a\in A_\star}\lvert\Delta_\star\pi(a\mid s)\rvert\,\big/\,\sum_{a}\lvert\Delta_\star\pi(a\mid s)\rvert}{\pi(A_\star\mid s)},
\]
the share of the total variation of the edit that falls on the coin-directed actions $A_\star$, divided by the probability these actions held before the edit, so that $T_\star=1$ means no preferred direction (Appendix~\ref{app:coin-functional-role}).
Coin-directed actions hold about $7\%$ of the probability and receive about $9\%$ of the change ($T_\star=1.210$), close to randomly initialized networks ($T_\star=0.991$) and far from Non-GM agents ($T_\star=4.996$) (Figure~\ref{fig:main-goal-role}b).
The coin is therefore perceived as a change in the scene rather than as a goal that orients behavior.
What organizes behavior is neither the goal nor a substitute target, but the simple rightward action rule that solved training, so goal misgeneralization in \textsc{Coin} is best described as a policy shortcut, in the sense of shortcut learning \citep{geirhos2020shortcut}.
A deeper analysis, including Fisher saliency maps, is given in Appendix~\ref{app:coin-functional-role} and Appendix~\ref{app:fisher-saliency}.

In \textsc{PushToMarker} we find that a place substitutes for the goal as well, instead of a policy shortcut, but unlike in \textsc{Maze} it is a fixed place in the camera image.
When we shift the camera, GM agents push the cube to the point under the old centre pixel in $0.901$ of episodes, while control agents push it onto the marker in $0.955$ (Figure~\ref{fig:main-goal-role}c).
A shift of the arm base does not change where GM agents push ($0.967$ at the centre).
A fixed motion pattern would not follow the camera, which rules out a policy shortcut, so \textsc{PushToMarker} is a target substitution.
To characterize its weak reaction to the marker, beyond signed follow showing that the shifted action mass is not goal-directed (Table~\ref{tab:main-outside-inside}), we added a distractor, a disc of another colour, and found that it elicits the same reaction as the marker, so the marker is perceived as a change in the scene rather than as a goal (Appendix~\ref{app:ptm-direction}).

When goal and proxy are both wired into the policy, as in \textsc{Attribute Maze}, the question becomes how the two interact.
For a swap $e$ that favours one candidate, we measure the change in log support of the favoured candidate relative to the neutral actions $A_0^e(s)$, $S_{\mathrm{fav}}^e(s)=\Delta_e\log\big[\pi(A_{\mathrm{fav}}^e(s)\mid s)/\pi(A_0^e(s)\mid s)\big]$, where $A_{\mathrm{fav}}^e(s)$ holds the progress actions toward the favoured candidate, and $S_{\mathrm{comp}}^e(s)$ likewise for its competitor (Appendix~\ref{app:attribute-functional-role}).
In GM agents, both the shape swap and the colour swap push support toward their candidate and pull it from the competitor ($S_{\mathrm{fav}}=1.368$ and $S_{\mathrm{comp}}=-1.721$ for shape, $2.001$ and $-2.354$ for colour), while Non-GM agents show this push--pull only for shape ($3.740$ and $-3.910$, against $0.192$ and $-0.362$ for colour) (Figure~\ref{fig:main-goal-role}d, Appendix~\ref{app:attribute-functional-role}).
The two cues do not influence the policy independently, but their effects counteract.

We therefore find three distinct failures of policy alignment: a substitute target where the proxy is a location, an action rule where it is a movement, and competing cues where it is a property of the goal object, or two competing objects (Appendix~\ref{app:wtf-role}).
How this competition is organized in \textsc{Attribute Maze} is what we examine next.

\subsection{When goal and proxy both reach the policy, how do they compete?}
\label{sec:goal-compete}

\begin{figure}[t]
    \centering
    \includegraphics[width=\linewidth]{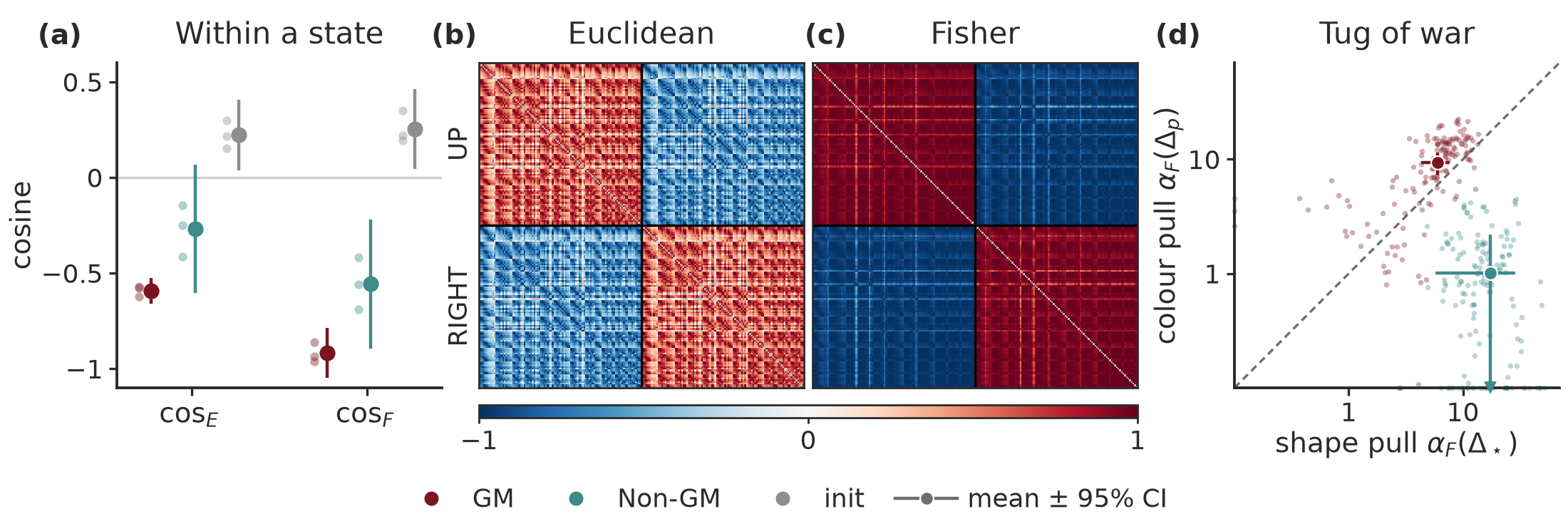}
    \caption{
    \textbf{How goal and proxy compete in \textsc{Attribute Maze}.}
    (a) Cosine between the two cue displacements within a state.
    (b, c) Cosine between the competition directions of pairs of states, grouped by whether the goal candidate lies UP or RIGHT.
    (d) Pull of each cue on the shared axis, per state. Above the dashed line, the proxy pulls harder. The colour/shape pull ratio is 1.56 for GM and 0.07 for Non-GM.
    All at $h_{\mathrm{fc}}$, mean with 95\% $t$-interval over three seeds.
    }
    \label{fig:main-compete}
\end{figure}

\textsc{Attribute Maze} is an example of a regime in which goal and proxy both act on the policy and push and pull against each other in a competition.
We now want to understand geometrically where this push and pull comes from, and which cue pulls harder.
For each conflict state, the shape swap and the colour swap move the representation by $\Delta_\star(s)$ and $\Delta_p(s)$, and we compare such directions with the ordinary cosine $\cos_E$ in raw activations and with its Fisher variant $\cos_F(a,b)=a^\top F_h b\,/\sqrt{a^\top F_h a\; b^\top F_h b}$ in the policy's metric, which with $F_h$ from Equation~\eqref{eq:main-q} only counts directions the policy actually reads.
We find that the two cues push the policy in opposite directions (Figure~\ref{fig:main-compete}a).
In GM agents, $\cos_F(\Delta_\star,\Delta_p)=-0.919$, so for the policy the two displacements point in opposite directions, whereas control agents are opposed more weakly ($-0.557$) and randomly initialized networks not at all ($+0.254$).

This raises the question whether there is one global competition axis.
We therefore take the competition direction $d_{\mathrm{comp}}(s)=\Delta_p(s)-\Delta_\star(s)$ in each state and compare it between states with the Euclidean and Fisher cosines from above (Figure~\ref{fig:main-compete}b, c).
In the Fisher metric, states of the same conflict share almost the same competition direction ($0.942$, control $0.952$, initialization $0.148$), and states of the opposite conflict point the opposite way, while in raw activations the structure is visible but more heterogeneous and weaker ($0.589$).
Thus, the policy reads the competition on one signed axis per conflict, which we call $u_{\mathrm{comp}}$ (Appendix~\ref{app:attribute-cross-state-geometry}).

Finally, we ask who is actually winning on this axis.
We project each cue's displacement onto $u_{\mathrm{comp}}$ with $\alpha_F(\Delta)=u_{\mathrm{comp}}^\top F_h\Delta/u_{\mathrm{comp}}^\top F_h u_{\mathrm{comp}}$, which measures how far the cue pulls the state along it in the policy's Fisher geometry.
In GM agents, the colour cue pulls $1.564$ times as hard as the shape cue, whereas in control agents the ratio is only $0.067$ (Figure~\ref{fig:main-compete}d).
Randomly initialized networks already lean toward colour, so GM training keeps an initial preference that Non-GM training reverses (Appendix~\ref{app:attribute-tug-of-war}).
The goal is therefore not lost in this regime but still pulls on a shared axis, only more weakly than the proxy, which suggests steering along this axis, as we test next (Appendices~\ref{app:attribute-competition-geometry} and~\ref{app:attribute-tug-of-war}).
The same competition appears in graded form in the robot tasks (Appendix~\ref{app:competition-four-envs}).

\section{Can the goal be recovered?}
\label{sec:recover}

\subsection{Can the goal be recovered through the existing wiring?}
\label{sec:recover-steer}

With the findings of the geometry analysis in Section~\ref{sec:goal-compete} and our understanding of where the goal gets lost (Section~\ref{sec:goal}), we can already forecast in which environments steering can work.
In \textsc{Maze}, the goal is inert, and in \textsc{Coin} and \textsc{PushToMarker}, it is only weakly read and does not direct behavior.
In \textsc{Attribute Maze}, however, the goal is fully wired into the policy and loses its competition with the proxy on a shared axis, and this axis is something we can push on.
To steer the four environments with one protocol, we contrast, at every state $s$, the representation of two counterfactual versions of it, $s^\star$ in which the evidence supports the goal and $s^p$ in which it supports the proxy.
Their difference $d(s)=h_{\mathrm{fc}}(s^\star)-h_{\mathrm{fc}}(s^p)$ is an additive steering vector \citep{turner2023actadd}, along which we push the representation with dose $\beta$, $h_{\mathrm{fc}}'(s_t)=h_{\mathrm{fc}}(s_t)+\beta\,d(s_t)$.
In \textsc{Maze}, $s^\star$ and $s^p$ place the cheese at its position and in the top-right corner, in \textsc{PushToMarker} the marker at its position and at the table centre.
In \textsc{Coin}, they place the coin at its position behind the agent and ahead of it on the shortcut path.
In \textsc{Attribute Maze}, $s^\star$ and $s^p$ are the colour and shape swaps, which put the yellow line on the goal and on the proxy candidate, so $d(s)$ is the competition direction $d_{\mathrm{comp}}(s)$ of Section~\ref{sec:goal-compete}.
We rebuild the direction as the agent moves, which makes the push state-dependent and allows for different geometries rather than one shared axis.
As controls, we reverse it, use random pushes of equal norm, or none.

As we derived in Section~\ref{sec:goal-compete}, a simpler intervention is possible in \textsc{Attribute Maze}.
As the policy reads the competition on one signed axis per conflict, we propose to replace the state-dependent direction by the identified shared axis $u_{\mathrm{comp}}$, $h_{\mathrm{fc}}'(s_t)=h_{\mathrm{fc}}(s_t)+\beta\,\sigma(s_0)\,\|d(s_0)\|_2\,u_{\mathrm{comp}}$, where $\sigma(s_0)\in\{-1,+1\}$ is the orientation of the conflict.
We calculate this push once with the start state $s_0$ and keep it fixed for all other states.
We compare it with an unsigned push that drops $\sigma$ and, as both are held fixed, with $d(s_0)$ held fixed, and with a Fisher-projected push, the component $\alpha_F$ of $d(s_0)$ along $u_{\mathrm{comp}}$ (Appendix~\ref{app:shared-axis-steering}).

\begin{figure}[t]
    \centering
    \includegraphics[width=\linewidth]{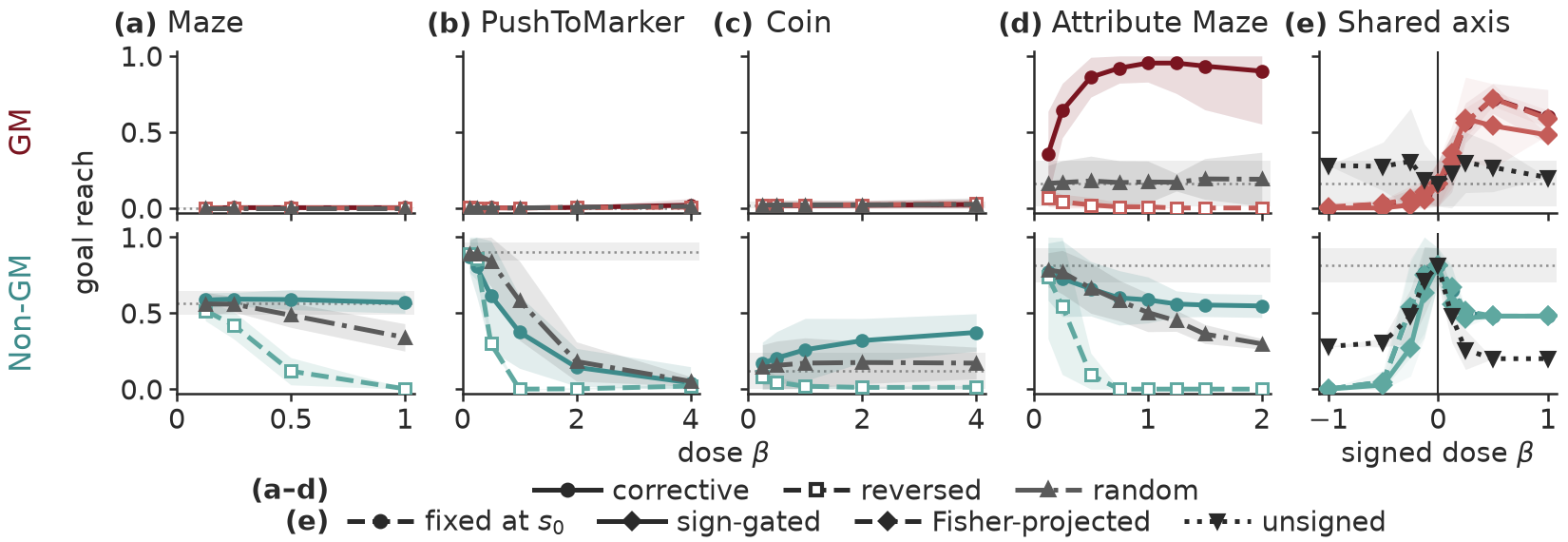}
    \caption{
    \textbf{Steering the existing wiring.}
    Goal reach $\hat J^\star$ under steering at $h_{\mathrm{fc}}$ as a function of the dose $\beta$, GM agents (top) and control agents (bottom).
    (a--d) State-dependent corrective, reversed and random pushes of equal norm.
    (e) \textsc{Attribute Maze}, shared-axis pushes (sign-gated, Fisher-projected, unsigned) against the state-dependent direction held fixed at $s_0$, negative $\beta$ reversed.
    Grey bands and dotted lines show the unsteered baseline.
    Mean with 95\% $t$-interval over three seeds.
    }
    \label{fig:main-steer}
\end{figure}

In \textsc{Attribute Maze}, we find that steering redirects the policy to the goal location: the GM agents' goal reach rises from $0.160$ to $0.953$ at $\beta=1$, the reversed push drives it down and a random push leaves it unchanged (Figure~\ref{fig:main-steer}d).
With the goal cue removed, the push still takes the agent there in $0.913$ of states, so it sets where the policy goes (Appendix~\ref{app:local-steering}, Table~\ref{tab:steering-location}).
The shared axis carries this effect as well (Figure~\ref{fig:main-steer}e).
At $\beta=0.25$, the sign-gated push raises goal reach to $0.587$, as much as $d(s_0)$ held fixed ($0.560$), and reversing it drives it close to zero.
Without the sign $\sigma(s)$, the direction on the axis is the opposite for some states (Section~\ref{sec:goal-compete}), so the unsigned push achieves only $0.300$ and its reverse almost the same $0.307$.
The Fisher-projected push is weaker ($0.560$), partly because it is a smaller push, only $0.30$ to $0.73$ the size of the sign-gated one (Appendix~\ref{app:shared-axis-steering}).

In GM agents in \textsc{Maze}, \textsc{PushToMarker} and \textsc{Coin}, the same construction does not move the policy toward the goal (goal reach at most $0.002$, $0.018$ and $0.027$), and the corrective push is no more effective than a random one, although control agents respond to it (Appendix~\ref{app:local-steering}, Table~\ref{tab:steering-local}).

The steering results are thus consistent with Section~\ref{sec:goal}: steering the existing wiring redirects the policy to the goal's location only where the goal is still followed.
In \textsc{Attribute Maze}, the goal lost a competition on an axis the policy already reads, so pushing along that axis turns the policy towards the goal's location again.
In \textsc{Maze}, \textsc{PushToMarker} and \textsc{Coin}, there is no such axis to push on, as the goal information that reaches the policy is too weak in \textsc{PushToMarker} and \textsc{Coin} and not detectable in \textsc{Maze}.
Whether the goal information in the representation is still enough to learn an orientation towards the true goal is what we test next, by training new wiring.
The same patterns replicate in the other environments (Appendix~\ref{app:steering}).

\subsection{Can the goal be recovered through new wiring?}
\label{sec:recover-rewire}

To understand at what depth the goal is truly lost, we learn a new wiring at two points in the architecture (Appendix~\ref{sec:architecture}).
We relearn either only the policy head on top of $h_{\mathrm{fc}}$ ($4$k trained parameters in Procgen) or the head together with the fully connected layer that maps $h_{\mathrm{pre}}$ to $h_{\mathrm{fc}}$ ($528$k).
In both cases, we train a freshly initialized copy of the downstream wiring with behavioral cloning \citep{pomerleau1991efficient} instead of RL training to reduce confounding, as RL training introduces many more failure modes than simple cloning. The labels are goal-directed actions shared across all variants and baselines, which also removes the need for online data in the base variant.
In \textsc{Maze} and \textsc{Attribute Maze}, the targets are oracle actions from a breadth-first search towards the true goal, while in \textsc{Coin} and \textsc{PushToMarker}, where no oracle exists, we use the actions of the control agents (control-agent targets for all environments in Appendix~\ref{sec:rewiring}).
As baselines, we rewire the control agents in the same way, compare every rewire to the native reach of both cohorts and to a floor, the higher of a never-trained random trunk rewired in the same way and a goal-blind chance policy, and remove the confound of head size with a fixed random remix, which replaces the learned map from $h_{\mathrm{pre}}$ to $h_{\mathrm{fc}}$ by a random one and trains only a head of the $h_{\mathrm{fc}}$ size.
A GM rewire counts as recovered near the control reach and as partial when clearly above the floor (Appendix~\ref{sec:rewiring}).

\begin{figure}[t]
    \centering
    \includegraphics[width=\linewidth]{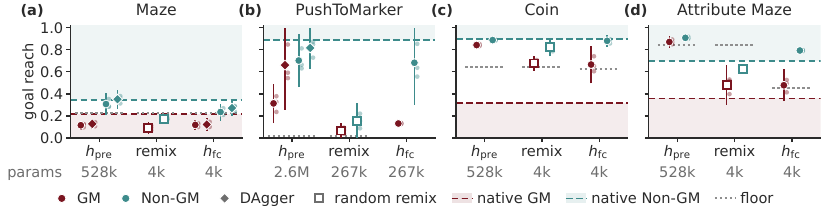}
    \caption{\textbf{Recovering the goal through new wiring.} Goal reach $\hat J^\star$ after relearning the network downstream of $h_{\mathrm{pre}}$ or $h_{\mathrm{fc}}$ by behavioral cloning, with oracle targets in \textsc{Maze} and \textsc{Attribute Maze} and control-agent targets in \textsc{PushToMarker} and \textsc{Coin}. Remix is a fixed random map of $h_{\mathrm{pre}}$ with a trained head of the $h_{\mathrm{fc}}$ size. DAgger (diamonds) adds three rounds of relabeling on the states the rewired policy visits, in \textsc{PushToMarker} with relabeling by the control agents. Dotted lines mark the floor, the higher of a random trunk rewired at the same site and chance. Grey numbers give the trained parameters. Points are means with 95\% confidence intervals across agents, small points are individual agents.}
    \label{fig:main-rewire}
\end{figure}

Where the goal information is seemingly inert for the policy, as in \textsc{Maze}, a new wiring does not recover goal orientation (Figure~\ref{fig:main-rewire}a).
GM agents reach the cheese on $0.116$ of episodes from $h_{\mathrm{pre}}$ and $0.119$ from $h_{\mathrm{fc}}$, below their native $0.219$ and a uniform random walk ($0.226$).
As behavioral cloning alone leaves even the control rewire below its native reach ($0.308$ against $0.347$), we also relabel on the states the rewired policy visits \citep{ross2011dagger}, which restores the control agents ($0.351$) but leaves GM agents below $0.13$ (Appendix~\ref{sec:rewiring}).

In \textsc{PushToMarker}, where the goal reaches the policy only weakly, a new wiring recovers it partially (Figure~\ref{fig:main-rewire}b).
From $h_{\mathrm{pre}}$, GM agents push the cube onto the marker in $0.314$ of conflict scenes, far above the random trunk ($0.014$) but about $45\%$ of the control rewire ($0.702$), which stays below its native reach ($0.889$).
With on-policy relabeling the GM rewire reaches $0.661$, against $0.296$ with cloning alone in the same evaluation (Appendix~\ref{sec:rewiring}, Table~\ref{tab:maze-dagger}).
From $h_{\mathrm{fc}}$, they reach $0.132$.

In \textsc{Coin}, where the goal is only weakly wired and steering was unsuccessful, the question is whether a new wiring can use the goal information that still reaches the late layers.
From $h_{\mathrm{pre}}$, the GM rewire reaches the coin in $0.843$ of episodes (Figure~\ref{fig:main-rewire}c), close to the control rewire ($0.888$) and well above chance ($0.625$) and the native GM reach ($0.316$).
A rewire on a random trunk chooses the coin over the level end almost as often as the GM rewire ($0.962$ against $0.987$ of the episodes that end at either), but it dies or runs out of time more often.
The trained GM trunk mainly contributes the skill of getting through the level, not the recognition of the coin.
From $h_{\mathrm{fc}}$, the GM rewire reaches $0.666$, close to chance ($0.625$).
This matches the probes (Section~\ref{sec:goal-there}).
The remix shows where the information is lost, as a head on a random remix of GM $h_{\mathrm{pre}}$ reaches $0.676$, as much as a head on the learned GM $h_{\mathrm{fc}}$, while the same remix of control $h_{\mathrm{pre}}$ keeps $0.825$.
So the wiring the GM agents learned between $h_{\mathrm{pre}}$ and the policy head seems to keep no more goal information than a random one.
With the steering null, the goal in \textsc{Coin} is thus represented but lost in the readout.

Finally, in \textsc{Attribute Maze}, where goal and proxy are both followed, rewiring from $h_{\mathrm{pre}}$ allows recovery and brings GM agents close to the control rewire ($0.871$ against $0.908$, native $0.358$, Figure~\ref{fig:main-rewire}d).
A random trunk rewired in the same way reaches almost the same level ($0.840$), because a readout with enough capacity relearns the environment from any trunk. With a linear readout on held-out configurations, the GM trunk exceeds the random trunk ($+0.136$ from $h_{\mathrm{pre}}$, $+0.185$ from $h_{\mathrm{fc}}$, Appendix~\ref{sec:rewiring}).
From $h_{\mathrm{fc}}$, they reach $0.478$, above their native policy but not higher than the remix ($0.481$), while the control rewire reaches $0.793$.
Cloning with the goal masked removes the recovery where the rewire uses the goal, in \textsc{PushToMarker} ($0.314$ to $0.054$) and \textsc{Attribute Maze} ($0.684$ to $0.173$ with control-agent targets), so both rewirings are goal-directed (in \textsc{Coin} this test is not informative, Appendix~\ref{sec:rewiring-goal-directed}).

We also find that recovery drops at $h_{\mathrm{fc}}$ almost everywhere, because the fully connected layer keeps what the policy uses, the goal in Non-GM agents and the proxy in GM agents, as the \textsc{PushToMarker} probes already show.
In \textsc{Attribute Maze}, a linear readout from GM $h_{\mathrm{fc}}$ picks the goal in only $0.74$ of the episodes in which it commits to an object ($0.99$ for Non-GM), although the agent's own head still follows the goal (Appendix~\ref{sec:rewiring-layers}).
A new readout thus recovers goal reach wherever goal information survives in the representation it reads, from $h_{\mathrm{pre}}$ in \textsc{Coin} and \textsc{Attribute Maze} and partly in \textsc{PushToMarker}, with less from $h_{\mathrm{fc}}$, but nowhere in \textsc{Maze}.
The same pattern replicates in the other environments (Appendix~\ref{sec:rewiring}).
In all eight environments, steering thus works only where the goal is followed and a new readout only where it still reaches the policy, so the depth found in Section~\ref{sec:goal} predicts which recovery works in each of them (Figure~\ref{fig:overview}).

\section{Discussion and conclusion}
\label{sec:conclusion}

\paragraph{Conclusion.}
From the outside, all agents seem to fail in the same way, but the depth at which the goal gets lost determines which recovery works.
Tracing the goal through the agents' behavior and representations also reveals three distinct failures of policy alignment, which differ in what organizes their behavior instead.
Where the goal is still wired into the policy but overpowered by a competing proxy cue or object, we can steer the existing policy to the goal's location.
Where it is lost in the readout but still present in the representation, a new readout can at least partly reestablish goal use, turning information that is merely represented into goal-directed behavior.
And where the goal is inert to the policy, neither attempt works.
As behavior alone cannot tell these failures apart, we propose to first diagnose where the goal gets lost, and only then choose how to recover it.

\paragraph{Discussion and outlook.}
Beyond these results, we propose a hypothesis that our results already indicate, namely that the structure of the training environment shapes how the goal fails.
Across our environments, goal and proxy that compete as cues on one object, or as two objects that stand close together in training, lead to competition deep in the agent, where the goal is still followed and can be steered and rewired.
Location-based target substitution loses the goal early or leaves it only weakly represented, so recovery struggles or succeeds only partly.
For policy shortcuts, we think that a simpler input-to-action mapping that solves training equally well, which networks are known to latch onto \citep{geirhos2020shortcut,hermann2020shapes}, leaves the goal only weakly present.
The structure of the environment thus indicates where the goal will be lost and which recovery can work, a first step toward predicting goal misgeneralization.
We studied static checkpoints, and opening up the training dynamics, how the goal is lost while the agent learns, would let us predict when and why it arises, which we leave open for future work.
Our environments cover only part of goal misgeneralization, which also arises, for example, with instrumental subgoals \citep{langosco} and in language models \citep{shah,maniyar2026position}, where we expect the structure of the failure to differ.

\newpage
\bibliography{main}

\begin{thebibliography}{57}
\providecommand{\natexlab}[1]{#1}
\providecommand{\url}[1]{\texttt{#1}}
\expandafter\ifx\csname urlstyle\endcsname\relax
  \providecommand{\doi}[1]{doi: #1}\else
  \providecommand{\doi}{doi: \begingroup \urlstyle{rm}\Url}\fi

\bibitem[Abdel~Sadek et~al.(2025)Abdel~Sadek, Farrugia-Roberts, Anwar,
  Erlebach, Schroeder~de Witt, Krueger, and Dennis]{sadek2025mitigating}
Karim Abdel~Sadek, Matthew Farrugia-Roberts, Usman Anwar, Hannah Erlebach,
  Christian Schroeder~de Witt, David Krueger, and Michael Dennis.
\newblock Mitigating goal misgeneralization via minimax regret.
\newblock In \emph{Reinforcement Learning Conference (RLC)}, 2025.
\newblock URL \url{https://arxiv.org/abs/2507.03068}.

\bibitem[Alain \& Bengio(2017)Alain and Bengio]{alain2017probes}
Guillaume Alain and Yoshua Bengio.
\newblock Understanding intermediate layers using linear classifier probes.
\newblock In \emph{International Conference on Learning Representations (ICLR),
  Workshop Track}, 2017.

\bibitem[Amari(1998)]{amari1998natural}
Shun-ichi Amari.
\newblock Natural gradient works efficiently in learning.
\newblock \emph{Neural Computation}, 10\penalty0 (2):\penalty0 251--276, 1998.
\newblock \doi{10.1162/089976698300017746}.

\bibitem[Armstrong et~al.(2023)Armstrong, Maranh{\~a}o, Daniels-Koch, Leask,
  and Gorman]{armstrong2023ace}
Stuart Armstrong, Alexandre Maranh{\~a}o, Oliver Daniels-Koch, Patrick Leask,
  and Rebecca Gorman.
\newblock {CoinRun}: Solving goal misgeneralisation, 2023.
\newblock URL \url{https://arxiv.org/abs/2309.16166}.

\bibitem[Atrey et~al.(2020)Atrey, Clary, and Jensen]{atrey2020exploratory}
Akanksha Atrey, Kaleigh Clary, and David Jensen.
\newblock Exploratory not explanatory: Counterfactual analysis of saliency maps
  for deep reinforcement learning.
\newblock In \emph{International Conference on Learning Representations}, 2020.
\newblock URL \url{https://arxiv.org/abs/1912.05743}.

\bibitem[Betley(2023)]{betley2023localizing}
Jan Betley.
\newblock Localizing goal misgeneralization in a maze-solving policy network.
\newblock AI Alignment Forum, 6 July 2023, 2023.
\newblock URL
  \url{https://www.alignmentforum.org/posts/vY9oE39tBupZLAyoC/localizing-goal-misgeneralization-in-a-maze-solving-policy}.

\bibitem[Bush et~al.(2025)Bush, Chung, Anwar, Garriga-Alonso, and
  Krueger]{bush2025planning}
Thomas Bush, Stephen Chung, Usman Anwar, Adri{\`a} Garriga-Alonso, and David
  Krueger.
\newblock Interpreting emergent planning in model-free reinforcement learning.
\newblock In \emph{International Conference on Learning Representations}, 2025.
\newblock URL \url{https://openreview.net/forum?id=DzGe40glxs}.

\bibitem[Cobbe et~al.(2019)Cobbe, Klimov, Hesse, Kim, and
  Schulman]{cobbe2019coinrun}
Karl Cobbe, Oleg Klimov, Christopher Hesse, Taehoon Kim, and John Schulman.
\newblock Quantifying generalization in reinforcement learning.
\newblock In \emph{Proceedings of the 36th International Conference on Machine
  Learning}, volume~97 of \emph{Proceedings of Machine Learning Research}, pp.\
   1282--1289. PMLR, 2019.

\bibitem[Cobbe et~al.(2020)Cobbe, Hesse, Hilton, and
  Schulman]{cobbe2020procgen}
Karl Cobbe, Christopher Hesse, Jacob Hilton, and John Schulman.
\newblock Leveraging procedural generation to benchmark reinforcement learning.
\newblock In \emph{Proceedings of the 37th International Conference on Machine
  Learning}, volume 119 of \emph{Proceedings of Machine Learning Research},
  pp.\  2048--2056. PMLR, 2020.

\bibitem[Colas et~al.(2022)Colas, Karch, Sigaud, and
  Oudeyer]{colas2022autotelic}
C{\'e}dric Colas, Tristan Karch, Olivier Sigaud, and Pierre-Yves Oudeyer.
\newblock Autotelic agents with intrinsically motivated goal-conditioned
  reinforcement learning: A short survey.
\newblock \emph{Journal of Artificial Intelligence Research}, 74:\penalty0
  1159--1199, 2022.
\newblock \doi{10.1613/jair.1.13554}.

\bibitem[de~Haan et~al.(2019)de~Haan, Jayaraman, and Levine]{dehaan2019causal}
Pim de~Haan, Dinesh Jayaraman, and Sergey Levine.
\newblock Causal confusion in imitation learning.
\newblock In \emph{Advances in Neural Information Processing Systems}, 2019.

\bibitem[Delfosse et~al.(2024)Delfosse, Sztwiertnia, Rothermel, Stammer, and
  Kersting]{delfosse2024scobots}
Quentin Delfosse, Sebastian Sztwiertnia, Mark Rothermel, Wolfgang Stammer, and
  Kristian Kersting.
\newblock Interpretable concept bottlenecks to align reinforcement learning
  agents.
\newblock In \emph{Advances in Neural Information Processing Systems},
  volume~37, pp.\  66826--66855, 2024.
\newblock \doi{10.52202/079017-2135}.
\newblock URL \url{https://arxiv.org/abs/2401.05821}.

\bibitem[Ding et~al.(2023)Ding, Shi, Chi, and Zhao]{ding2023seeing}
Wenhao Ding, Laixi Shi, Yuejie Chi, and Ding Zhao.
\newblock Seeing is not believing: Robust reinforcement learning against
  spurious correlation.
\newblock In \emph{Advances in Neural Information Processing Systems}, 2023.

\bibitem[Dorn et~al.(2023)Dorn, Alex, and Krueger]{smallKueg}
Diego Dorn, Neel Alex, and David Krueger.
\newblock Goal misgeneralization as implicit goal conditioning.
\newblock In \emph{NeurIPS 2023 Workshop on Goal-Conditioned Reinforcement
  Learning (GCRL)}, 2023.
\newblock URL \url{https://openreview.net/forum?id=QT4tXTqTTr}.

\bibitem[Elazar et~al.(2021)Elazar, Ravfogel, Jacovi, and
  Goldberg]{elazar2021amnesic}
Yanai Elazar, Shauli Ravfogel, Alon Jacovi, and Yoav Goldberg.
\newblock Amnesic probing: Behavioral explanation with amnesic counterfactuals.
\newblock \emph{Transactions of the Association for Computational Linguistics},
  9:\penalty0 160--175, 2021.
\newblock \doi{10.1162/tacl_a_00359}.

\bibitem[Espeholt et~al.(2018)Espeholt, Soyer, Munos, Simonyan, Mnih, Ward,
  Doron, Firoiu, Harley, Dunning, Legg, and Kavukcuoglu]{espeholt2018impala}
Lasse Espeholt, Hubert Soyer, R{\'e}mi Munos, Karen Simonyan, Volodymyr Mnih,
  Tom Ward, Yotam Doron, Vlad Firoiu, Tim Harley, Iain Dunning, Shane Legg, and
  Koray Kavukcuoglu.
\newblock {IMPALA}: Scalable distributed deep-{RL} with importance weighted
  actor-learner architectures.
\newblock In \emph{Proceedings of the 35th International Conference on Machine
  Learning}, volume~80 of \emph{Proceedings of Machine Learning Research}, pp.\
   1407--1416. PMLR, 2018.

\bibitem[Fei et~al.(2026)Fei, Wang, Shi, Dai, Cai, Qian, Ji, He, Zhang, Fei,
  Fu, Gong, and Qiu]{fei2025liberoplus}
Senyu Fei, Siyin Wang, Junhao Shi, Zihao Dai, Jikun Cai, Pengfang Qian, Li~Ji,
  Xinzhe He, Shiduo Zhang, Zhaoye Fei, Jinlan Fu, Jingjing Gong, and Xipeng
  Qiu.
\newblock {LIBERO-Plus}: A progressive robustness benchmark for
  visual-language-action models.
\newblock In \emph{Proceedings of the IEEE/CVF Conference on Computer Vision
  and Pattern Recognition (CVPR)}, 2026.
\newblock arXiv title: {LIBERO-Plus}: In-depth Robustness Analysis of
  Vision-Language-Action Models.

\bibitem[Florensa et~al.(2018)Florensa, Held, Geng, and
  Abbeel]{pmlr-v80-florensa18a}
Carlos Florensa, David Held, Xinyang Geng, and Pieter Abbeel.
\newblock Automatic goal generation for reinforcement learning agents.
\newblock In Jennifer Dy and Andreas Krause (eds.), \emph{Proceedings of the
  35th International Conference on Machine Learning}, volume~80 of
  \emph{Proceedings of Machine Learning Research}, pp.\  1515--1528. PMLR,
  10--15 Jul 2018.
\newblock URL \url{https://proceedings.mlr.press/v80/florensa18a.html}.

\bibitem[Geiger et~al.(2021)Geiger, Lu, Icard, and Potts]{geiger2021causal}
Atticus Geiger, Hanson Lu, Thomas Icard, and Christopher Potts.
\newblock Causal abstractions of neural networks.
\newblock In \emph{Advances in Neural Information Processing Systems},
  volume~34, pp.\  9574--9586, 2021.

\bibitem[Geirhos et~al.(2020)Geirhos, Jacobsen, Michaelis, Zemel, Brendel,
  Bethge, and Wichmann]{geirhos2020shortcut}
Robert Geirhos, J{\"o}rn-Henrik Jacobsen, Claudio Michaelis, Richard Zemel,
  Wieland Brendel, Matthias Bethge, and Felix~A. Wichmann.
\newblock Shortcut learning in deep neural networks.
\newblock \emph{Nature Machine Intelligence}, 2\penalty0 (11):\penalty0
  665--673, 2020.
\newblock \doi{10.1038/s42256-020-00257-z}.

\bibitem[Hermann \& Lampinen(2020)Hermann and Lampinen]{hermann2020shapes}
Katherine~L. Hermann and Andrew~K. Lampinen.
\newblock What shapes feature representations? {E}xploring datasets,
  architectures, and training.
\newblock In \emph{Advances in Neural Information Processing Systems},
  volume~33, pp.\  9995--10006, 2020.

\bibitem[Hewitt \& Liang(2019)Hewitt and Liang]{hewitt2019control}
John Hewitt and Percy Liang.
\newblock Designing and interpreting probes with control tasks.
\newblock In \emph{Proceedings of the 2019 Conference on Empirical Methods in
  Natural Language Processing and the 9th International Joint Conference on
  Natural Language Processing (EMNLP-IJCNLP)}, pp.\  2733--2743. Association
  for Computational Linguistics, 2019.
\newblock \doi{10.18653/v1/D19-1275}.

\bibitem[Hilton et~al.(2020)Hilton, Cammarata, Carter, Goh, and
  Olah]{hilton2020rlvision}
Jacob Hilton, Nick Cammarata, Shan Carter, Gabriel Goh, and Chris Olah.
\newblock Understanding {RL} vision.
\newblock \emph{Distill}, 5\penalty0 (11), 2020.
\newblock \doi{10.23915/distill.00029}.

\bibitem[Hutter et~al.(2016)Hutter, Gehring, Jud, Lauber, Bellicoso, Tsounis,
  Hwangbo, Bodie, Fankhauser, Bloesch, Diethelm, Bachmann, Melzer, and
  Hoepflinger]{hutter2016anymal}
Marco Hutter, Christian Gehring, Dominic Jud, Andreas Lauber, C.~Dario
  Bellicoso, Vassilios Tsounis, Jemin Hwangbo, Karen Bodie, Peter Fankhauser,
  Michael Bloesch, Remo Diethelm, Samuel Bachmann, Amir Melzer, and Mark
  Hoepflinger.
\newblock {ANYmal} - a highly mobile and dynamic quadrupedal robot.
\newblock In \emph{2016 IEEE/RSJ International Conference on Intelligent Robots
  and Systems (IROS)}, pp.\  38--44. IEEE, 2016.
\newblock \doi{10.1109/IROS.2016.7758092}.

\bibitem[Jucys et~al.(2024)Jucys, Adamopoulos, Hamidi, Milani, Samsami, Zholus,
  Joseph, Richards, Rish, and {\c{S}}im{\c{s}}ek]{jucys2024vpt}
Karolis Jucys, George Adamopoulos, Mehrab Hamidi, Stephanie Milani,
  Mohammad~Reza Samsami, Artem Zholus, Sonia Joseph, Blake Richards, Irina
  Rish, and {\"O}zg{\"u}r {\c{S}}im{\c{s}}ek.
\newblock Interpretability in action: Exploratory analysis of {VPT}, a
  {Minecraft} agent.
\newblock In \emph{ICML 2024 Workshop on Mechanistic Interpretability}, 2024.
\newblock URL \url{https://arxiv.org/abs/2407.12161}.

\bibitem[Kakade(2001)]{kakade2001natural}
Sham~M. Kakade.
\newblock A natural policy gradient.
\newblock In \emph{Advances in Neural Information Processing Systems},
  volume~14. MIT Press, 2001.

\bibitem[Kim et~al.(2023)Kim, Seo, Liu, Lee, Shin, Lee, and Lee]{kim2023arp}
Changyeon Kim, Younggyo Seo, Hao Liu, Lisa Lee, Jinwoo Shin, Honglak Lee, and
  Kimin Lee.
\newblock Guide your agent with adaptive multimodal rewards.
\newblock In \emph{Advances in Neural Information Processing Systems}, 2023.

\bibitem[Kingma \& Ba(2015)Kingma and Ba]{kingma2015adam}
Diederik~P. Kingma and Jimmy Ba.
\newblock Adam: A method for stochastic optimization.
\newblock In \emph{International Conference on Learning Representations}, 2015.

\bibitem[Kirichenko et~al.(2023)Kirichenko, Izmailov, and
  Wilson]{kirichenko2023lastlayer}
Polina Kirichenko, Pavel Izmailov, and Andrew~Gordon Wilson.
\newblock Last layer re-training is sufficient for robustness to spurious
  correlations.
\newblock In \emph{International Conference on Learning Representations
  (ICLR)}, 2023.
\newblock URL \url{https://openreview.net/forum?id=Zb6c8A-Fghk}.

\bibitem[Laidlaw et~al.(2025)Laidlaw, Singhal, and
  Dragan]{laidlaw2025correlated}
Cassidy Laidlaw, Shivam Singhal, and Anca Dragan.
\newblock Correlated proxies: A new definition and improved mitigation for
  reward hacking.
\newblock In \emph{International Conference on Learning Representations}, 2025.
\newblock URL \url{https://openreview.net/forum?id=msEr27EejF}.

\bibitem[Langosco et~al.(2022)Langosco, Koch, Sharkey, Pfau, and
  Krueger]{langosco}
Lauro Langosco~di Langosco, Jack Koch, Lee~D. Sharkey, Jacob Pfau, and David
  Krueger.
\newblock Goal misgeneralization in deep reinforcement learning.
\newblock In \emph{Proceedings of the 39th International Conference on Machine
  Learning}, volume 162 of \emph{Proceedings of Machine Learning Research},
  pp.\  12004--12019. PMLR, 2022.
\newblock URL \url{https://proceedings.mlr.press/v162/langosco22a.html}.

\bibitem[Lee(2020)]{lee2020trainprocgen}
Hojoon Lee.
\newblock Training procgen environment with {PyTorch}.
\newblock \url{https://github.com/joonleesky/train-procgen-pytorch}, 2020.

\bibitem[Lee et~al.(2023)Lee, Chen, Tajwar, Kumar, Yao, Liang, and
  Finn]{lee2023surgical}
Yoonho Lee, Annie~S. Chen, Fahim Tajwar, Ananya Kumar, Huaxiu Yao, Percy Liang,
  and Chelsea Finn.
\newblock Surgical fine-tuning improves adaptation to distribution shifts.
\newblock In \emph{International Conference on Learning Representations}, 2023.
\newblock URL \url{https://openreview.net/forum?id=APuPRxjHvZ}.

\bibitem[Maniyar et~al.(2026)Maniyar, Sandhu, and Mishra]{maniyar2026position}
Suyash Maniyar, Armaan Sandhu, and Abhishek Mishra.
\newblock Measuring reward hacking and reasoning-answer decoupling under
  position-confounded optimization.
\newblock In \emph{COLM 2026 Workshop on AI Measurement Science}, 2026.
\newblock URL \url{https://arxiv.org/abs/2608.15445}.

\bibitem[Martens(2020)]{martens2020natural}
James Martens.
\newblock New insights and perspectives on the natural gradient method.
\newblock \emph{Journal of Machine Learning Research}, 21\penalty0
  (146):\penalty0 1--76, 2020.

\bibitem[Mini et~al.(2023)Mini, Grietzer, Sharma, Meek, MacDiarmid, and
  Turner]{mini2023understandingcontrollingmazesolvingpolicy}
Ulisse Mini, Peli Grietzer, Mrinank Sharma, Austin Meek, Monte MacDiarmid, and
  Alexander~Matt Turner.
\newblock Understanding and controlling a maze-solving policy network, 2023.
\newblock URL \url{https://arxiv.org/abs/2310.08043}.

\bibitem[Mittal et~al.(2025)Mittal, Roth, Tigue, Richard, Zhang, Du,
  Serrano-Mu{\~n}oz, Yao, Zurbr{\"u}gg, Rudin, Wawrzyniak, Rakhsha, Denzler,
  Heiden, Borovicka, Ahmed, Akinola, Anwar, Carlson, Feng, Garg, Gasoto,
  Gulich, Guo, Gussert, Hansen, Kulkarni, Li, Liu, Makoviychuk, Malczyk,
  Mazhar, Moghani, Murali, Noseworthy, Poddubny, Ratliff, Rehberg, Schwarke,
  Singh, Smith, Tang, Thaker, Trepte, Wyk, Yu, Millane, Ramasamy, Steiner,
  Subramanian, Volk, Chen, Jawale, Kuruttukulam, Lin, Mandlekar, Patzwaldt,
  Welsh, Zhao, Anes, Lafleche, Mo{\"e}nne-Loccoz, Park, Stepinski, Gelder,
  Amevor, Carius, Chang, Chen, de~Heras~Ciechomski, Daviet, Mohajerani, von
  Muralt, Reutskyy, Sauter, Schirm, Shi, Terdiman, Vilella, Widmer, Yeoman,
  Chen, Grizan, Li, Li, Smith, Wiltz, Alexis, Chang, Chu, Fan, Farshidian,
  Handa, Huang, Hutter, Narang, Pouya, Sheng, Zhu, Macklin, Moravanszky, Reist,
  Guo, Hoeller, and State]{mittal2025isaaclab}
Mayank Mittal, Pascal Roth, James Tigue, Antoine Richard, Octi Zhang, Peter Du,
  Antonio Serrano-Mu{\~n}oz, Xinjie Yao, Ren{\'e} Zurbr{\"u}gg, Nikita Rudin,
  Lukasz Wawrzyniak, Milad Rakhsha, Alain Denzler, Eric Heiden, Ales Borovicka,
  Ossama Ahmed, Iretiayo Akinola, Abrar Anwar, Mark~T. Carlson, Ji~Yuan Feng,
  Animesh Garg, Renato Gasoto, Lionel Gulich, Yijie Guo, M.~Gussert, Alex
  Hansen, Mihir Kulkarni, Chenran Li, Wei Liu, Viktor Makoviychuk, Grzegorz
  Malczyk, Hammad Mazhar, Masoud Moghani, Adithyavairavan Murali, Michael
  Noseworthy, Alexander Poddubny, Nathan Ratliff, Welf Rehberg, Clemens
  Schwarke, Ritvik Singh, James~Latham Smith, Bingjie Tang, Ruchik Thaker,
  Matthew Trepte, Karl~Van Wyk, Fangzhou Yu, Alex Millane, Vikram Ramasamy,
  Remo Steiner, Sangeeta Subramanian, Clemens Volk, CY~Chen, Neel Jawale,
  Ashwin~Varghese Kuruttukulam, Michael~A. Lin, Ajay Mandlekar, Karsten
  Patzwaldt, John Welsh, Huihua Zhao, Fatima Anes, Jean-Francois Lafleche,
  Nicolas Mo{\"e}nne-Loccoz, Soowan Park, Rob Stepinski, Dirk~Van Gelder, Chris
  Amevor, Jan Carius, Jumyung Chang, Anka~He Chen, Pablo de~Heras~Ciechomski,
  Gilles Daviet, Mohammad Mohajerani, Julia von Muralt, Viktor Reutskyy,
  Michael Sauter, Simon Schirm, Eric~L. Shi, Pierre Terdiman, Kenny Vilella,
  Tobias Widmer, Gordon Yeoman, Tiffany Chen, Sergey Grizan, Cathy Li, Lotus
  Li, Connor Smith, Rafael Wiltz, Kostas Alexis, Yan Chang, David Chu,
  Linxi~{Jim''} Fan, Farbod Farshidian, Ankur Handa, Spencer Huang, Marco
  Hutter, Yashraj Narang, Soha Pouya, Shiwei Sheng, Yuke Zhu, Miles Macklin,
  Adam Moravanszky, Philipp Reist, Yunrong Guo, David Hoeller, and Gavriel
  State.
\newblock Isaac lab: A gpu-accelerated simulation framework for multi-modal
  robot learning.
\newblock \emph{arXiv preprint arXiv:2511.04831}, 2025.
\newblock URL \url{https://arxiv.org/abs/2511.04831}.

\bibitem[Mnih et~al.(2015)Mnih, Kavukcuoglu, Silver, Rusu, Veness, Bellemare,
  Graves, Riedmiller, Fidjeland, Ostrovski, et~al.]{mnih2015human}
Volodymyr Mnih, Koray Kavukcuoglu, David Silver, Andrei~A. Rusu, Joel Veness,
  Marc~G. Bellemare, Alex Graves, Martin Riedmiller, Andreas~K. Fidjeland,
  Georg Ostrovski, et~al.
\newblock Human-level control through deep reinforcement learning.
\newblock \emph{Nature}, 518\penalty0 (7540):\penalty0 529--533, 2015.

\bibitem[Nait El~Barj \& Sautory(2024)Nait El~Barj and Sautory]{naitelbarj2024}
Houda Nait El~Barj and Theophile Sautory.
\newblock Reinforcement learning from {LLM} feedback to counteract goal
  misgeneralization, 2024.
\newblock URL \url{https://arxiv.org/abs/2401.07181}.

\bibitem[Pan et~al.(2022)Pan, Bhatia, and Steinhardt]{pan2022effects}
Alexander Pan, Kush Bhatia, and Jacob Steinhardt.
\newblock The effects of reward misspecification: Mapping and mitigating
  misaligned models.
\newblock In \emph{International Conference on Learning Representations}, 2022.
\newblock URL \url{https://openreview.net/forum?id=JYtwGwIL7ye}.

\bibitem[Pomerleau(1991)]{pomerleau1991efficient}
Dean~A. Pomerleau.
\newblock Efficient training of artificial neural networks for autonomous
  navigation.
\newblock \emph{Neural Computation}, 3\penalty0 (1):\penalty0 88--97, 1991.
\newblock \doi{10.1162/neco.1991.3.1.88}.

\bibitem[Ramanauskas(2023)]{ramanauskas2023zenodo}
Karolis Ramanauskas.
\newblock Colour versus shape goal misgeneralization in reinforcement learning:
  A case study (models and test data), 2023.

\bibitem[Ramanauskas \& {\c{S}}im{\c{s}}ek(2023)Ramanauskas and
  {\c{S}}im{\c{s}}ek]{raman}
Karolis Ramanauskas and {\"O}zg{\"u}r {\c{S}}im{\c{s}}ek.
\newblock Colour versus shape goal misgeneralization in reinforcement learning:
  A case study.
\newblock In \emph{NeurIPS 2023 Workshop on Attributing Model Behavior at Scale
  (ATTRIB)}, 2023.
\newblock URL \url{https://arxiv.org/abs/2312.03762}.

\bibitem[Ross et~al.(2011)Ross, Gordon, and Bagnell]{ross2011dagger}
St{\'e}phane Ross, Geoffrey Gordon, and Drew Bagnell.
\newblock A reduction of imitation learning and structured prediction to
  no-regret online learning.
\newblock In \emph{Proceedings of the Fourteenth International Conference on
  Artificial Intelligence and Statistics}, volume~15 of \emph{Proceedings of
  Machine Learning Research}, pp.\  627--635. PMLR, 2011.

\bibitem[Schaul et~al.(2015)Schaul, Horgan, Gregor, and
  Silver]{pmlr-v37-schaul15}
Tom Schaul, Daniel Horgan, Karol Gregor, and David Silver.
\newblock Universal value function approximators.
\newblock In Francis Bach and David Blei (eds.), \emph{Proceedings of the 32nd
  International Conference on Machine Learning}, volume~37 of \emph{Proceedings
  of Machine Learning Research}, pp.\  1312--1320, Lille, France, 07--09 Jul
  2015. PMLR.
\newblock URL \url{https://proceedings.mlr.press/v37/schaul15.html}.

\bibitem[Schulman et~al.(2015)Schulman, Levine, Moritz, Jordan, and
  Abbeel]{schulman2015trpo}
John Schulman, Sergey Levine, Philipp Moritz, Michael Jordan, and Pieter
  Abbeel.
\newblock Trust region policy optimization.
\newblock In \emph{Proceedings of the 32nd International Conference on Machine
  Learning}, volume~37 of \emph{Proceedings of Machine Learning Research}, pp.\
   1889--1897. PMLR, 2015.
\newblock URL \url{https://proceedings.mlr.press/v37/schulman15.html}.

\bibitem[Schulman et~al.(2016)Schulman, Moritz, Levine, Jordan, and
  Abbeel]{schulman2016gae}
John Schulman, Philipp Moritz, Sergey Levine, Michael Jordan, and Pieter
  Abbeel.
\newblock High-dimensional continuous control using generalized advantage
  estimation.
\newblock In \emph{International Conference on Learning Representations}, 2016.

\bibitem[Schulman et~al.(2017)Schulman, Wolski, Dhariwal, Radford, and
  Klimov]{schulman2017ppo}
John Schulman, Filip Wolski, Prafulla Dhariwal, Alec Radford, and Oleg Klimov.
\newblock Proximal policy optimization algorithms, 2017.

\bibitem[Shah et~al.(2022)Shah, Varma, Kumar, Phuong, Krakovna, Uesato, and
  Kenton]{shah}
Rohin Shah, Vikrant Varma, Ramana Kumar, Mary Phuong, Victoria Krakovna,
  Jonathan Uesato, and Zac Kenton.
\newblock Goal misgeneralization: Why correct specifications aren't enough for
  correct goals, 2022.
\newblock URL \url{https://arxiv.org/abs/2210.01790}.

\bibitem[Simonyan et~al.(2014)Simonyan, Vedaldi, and
  Zisserman]{simonyan2014deep}
Karen Simonyan, Andrea Vedaldi, and Andrew Zisserman.
\newblock Deep inside convolutional networks: Visualising image classification
  models and saliency maps.
\newblock In \emph{International Conference on Learning Representations,
  Workshop Track}, 2014.
\newblock URL \url{https://arxiv.org/abs/1312.6034}.

\bibitem[Tao et~al.(2025)Tao, Xiang, Shukla, Qin, Hinrichsen, Yuan, Bao, Lin,
  Liu, et~al.]{tao2025maniskill3}
Stone Tao, Fanbo Xiang, Arth Shukla, Yuzhe Qin, Xander Hinrichsen, Xiaodi Yuan,
  Chen Bao, Xinsong Lin, Yulin Liu, et~al.
\newblock {ManiSkill3}: {GPU} parallelized robotics simulation and rendering
  for generalizable embodied {AI}.
\newblock In \emph{Robotics: Science and Systems}, 2025.

\bibitem[Tien et~al.(2023)Tien, He, Erickson, Dragan, and
  Brown]{tien2023causal}
Jeremy Tien, Jerry Zhi-Yang He, Zackory Erickson, Anca~D. Dragan, and Daniel~S.
  Brown.
\newblock Causal confusion and reward misidentification in preference-based
  reward learning.
\newblock In \emph{International Conference on Learning Representations}, 2023.

\bibitem[Trinh et~al.(2024)Trinh, Danesh, Khanh, and Plaut]{GMhelp}
Tu~Trinh, Mohamad~H. Danesh, Nguyen~X. Khanh, and Benjamin Plaut.
\newblock Getting by goal misgeneralization with a little help from a mentor.
\newblock In \emph{NeurIPS 2024 Workshop on Towards Safe and Trustworthy Agents
  (SATA)}, 2024.
\newblock URL \url{https://arxiv.org/abs/2410.21052}.

\bibitem[Turner et~al.(2023{\natexlab{a}})Turner, Grietzer, and
  Thiergart]{turner2023topright}
Alexander~Matt Turner, Peli Grietzer, and Lisa Thiergart.
\newblock Maze-solving agents: Add a top-right vector, make the agent go to the
  top-right.
\newblock LessWrong, cross-posted to the AI Alignment Forum, 31 March 2023,
  2023{\natexlab{a}}.
\newblock URL
  \url{https://www.lesswrong.com/posts/gRp6FAWcQiCWkouN5/maze-solving-agents-add-a-top-right-vector-make-the-agent-go}.

\bibitem[Turner et~al.(2023{\natexlab{b}})Turner, Thiergart, Leech, Udell,
  Vazquez, Mini, and MacDiarmid]{turner2023actadd}
Alexander~Matt Turner, Lisa Thiergart, Gavin Leech, David Udell, Juan~J.
  Vazquez, Ulisse Mini, and Monte MacDiarmid.
\newblock Steering language models with activation engineering,
  2023{\natexlab{b}}.

\bibitem[Xing et~al.(2025)Xing, Luo, Xie, Gao, Shen, and
  Song]{xing2025shortcut}
Youguang Xing, Xu~Luo, Junlin Xie, Lianli Gao, Heng~Tao Shen, and Jingkuan
  Song.
\newblock Shortcut learning in generalist robot policies: The role of dataset
  diversity and fragmentation.
\newblock In \emph{Proceedings of The 9th Conference on Robot Learning}, volume
  305 of \emph{Proceedings of Machine Learning Research}, pp.\  3239--3266.
  PMLR, 2025.

\bibitem[Zhou et~al.(2025)Zhou, Xu, Tie, Chen, Zhang, Chu, Zhou, and
  Sun]{zhou2025liberopro}
Xueyang Zhou, Yangming Xu, Guiyao Tie, Yongchao Chen, Guowen Zhang, Duanfeng
  Chu, Pan Zhou, and Lichao Sun.
\newblock {LIBERO-PRO}: Towards robust and fair evaluation of
  vision-language-action models beyond memorization, 2025.
\newblock URL \url{https://arxiv.org/abs/2510.03827}.

\end{thebibliography}
\bibliographystyle{iclr2027_conference}

\newpage
\appendix
\makeatletter
\let\gm@addcontentsline\addcontentsline
\renewcommand{\addcontentsline}[3]{%
  \gm@addcontentsline{#1}{#2}{#3}%
  \def\gm@ext{#1}\def\gm@toc{toc}%
  \ifx\gm@ext\gm@toc\gm@addcontentsline{apc}{#2}{#3}\fi}
\section*{Appendix contents}
\setcounter{tocdepth}{2}
\@starttoc{apc}
\makeatother
\newpage

\section{Environments}
\label{sec:environments}

\begin{figure}[!ht]
\centering
\includegraphics[width=\linewidth]{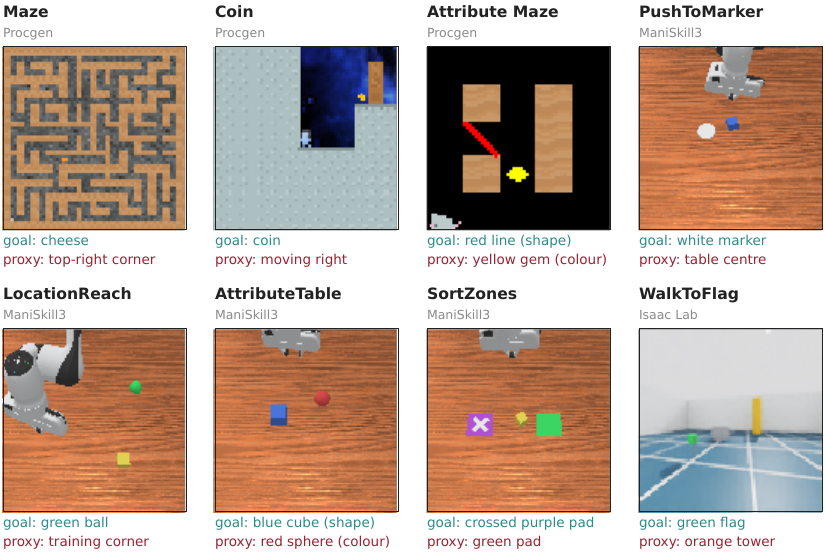}
\caption{
\textbf{The eight environments we analyse.}
Each panel shows an example level at the resolution the policy receives, and the labels under it name the intended goal and the proxy.
In \textsc{Maze}, \textsc{Coin}, \textsc{PushToMarker} and \textsc{LocationReach} the proxy is a location or a movement rule, so no object in the frame marks it.
}
\label{fig:env-overview-real}
\end{figure}

We study eight environments (Figure~\ref{fig:env-overview-real}, Table~\ref{tab:env-interface}): three procedurally generated video games and five robot tasks.
Each environment has an intended goal that the reward indicates and a proxy that co-locates with the goal during training, so the training levels cannot tell the two apart.
On levels that separate goal and proxy we can measure which of the two the agent pursues, and a misgeneralized agent pursues the proxy.

Three of the environments are procedurally generated video games from Procgen \citep{cobbe2020procgen}.
They are classic goal-misgeneralization tasks, already studied by \citet{langosco}, \citet{raman} and others \citep{mini2023understandingcontrollingmazesolvingpolicy}, though not in the depth of this work.
We call them \textsc{Maze}, \textsc{Coin} and \textsc{Attribute Maze} and describe each of them below.
In the game names of Procgen and its procgenAISC fork, \textsc{Maze} trains on \texttt{maze\_aisc} of \citet{langosco}, which keeps the cheese in the top-right cell during GM training, and is tested on the original \texttt{maze}, \textsc{Coin} trains on \texttt{coinrun} with the coin at the level end and is tested on \texttt{coinrun\_aisc} of \citet{langosco}, which places the coin at random, and \textsc{Attribute Maze} is \texttt{maze\_colorsobjects\_duel} from the procgenAISC fork of \citet{raman}, a colour--shape variant of the Maze~II setting of \citet{langosco}. %
Four are tasks for a simulated Franka Panda arm in ManiSkill3 \citep{tao2025maniskill3}, \textsc{PushToMarker}, \textsc{LocationReach}, \textsc{AttributeTable} and \textsc{SortZones}, and the last is a navigation task for a simulated ANYmal-C quadruped \citep{hutter2016anymal} in Isaac Lab \citep{mittal2025isaaclab}, which we call \quadname.
Goal misgeneralization has not yet been properly investigated in robotics with reinforcement learning, so these five are new environments in which we first induce goal misgeneralization and then analyse it.
The main text presents \textsc{Maze}, \textsc{Coin}, \textsc{Attribute Maze} and \textsc{PushToMarker}, and we extend the results to the other four as replications.
A textured variant of \textsc{Attribute Maze}, with the background switched from black to textured, serves as a robustness check.

\begin{figure}[t]
\centering
\includegraphics[width=0.85\linewidth]{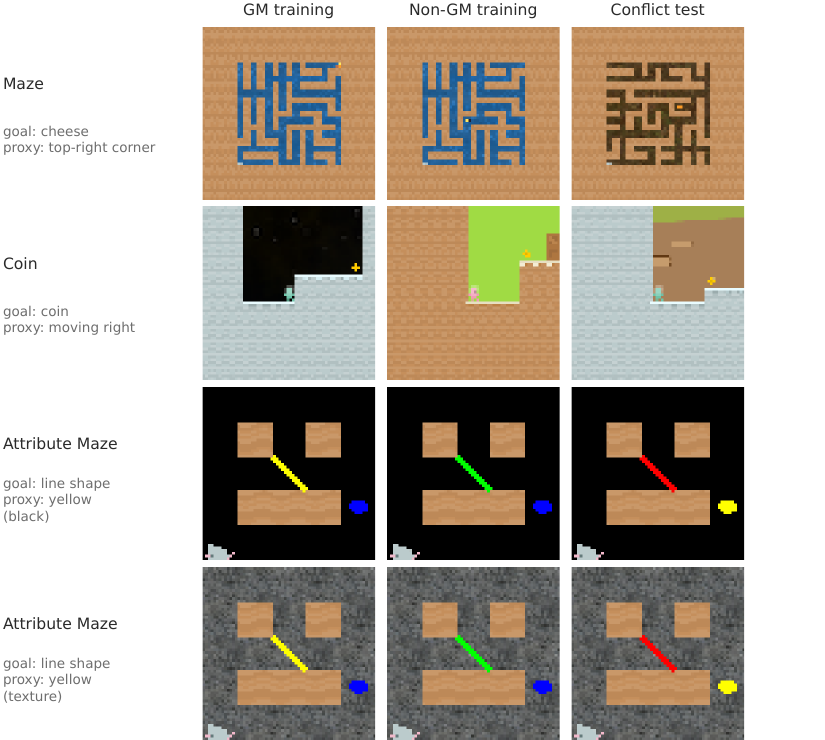}
\caption{
\textbf{GM training, Non-GM training and conflict test in the three Procgen games.}
One scene per distribution, as the policy sees it.
\textsc{Coin} tiles are taken a few steps into a scripted rightward run, since the coin is off screen at reset.
The \textsc{Attribute Maze} Non-GM tile shows one of the eight training pairs.
}
\label{fig:procgen-lineup}
\end{figure}

\begin{figure}[t]
\centering
\includegraphics[width=\linewidth]{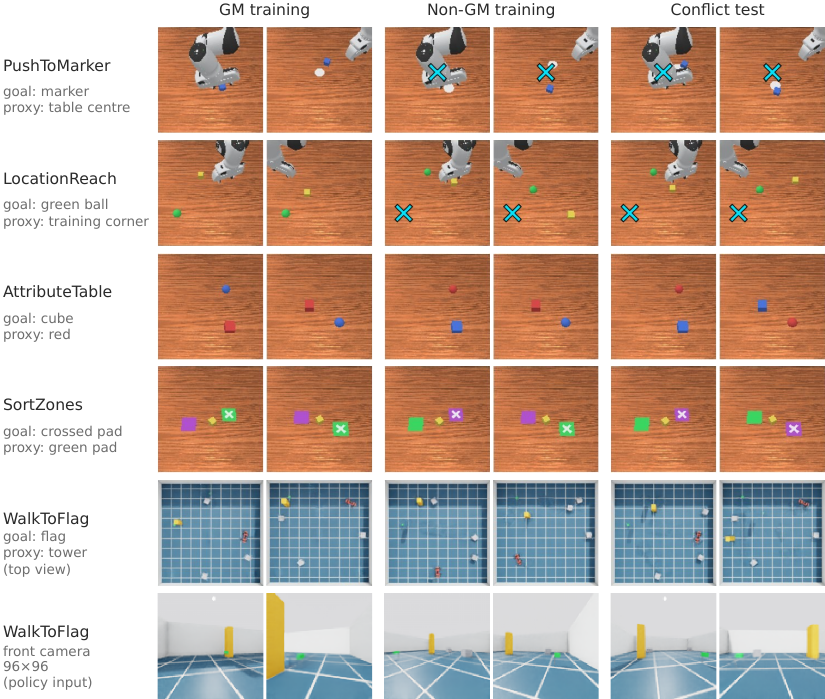}
\caption{
\textbf{GM training, Non-GM training and conflict test in the five robot tasks.}
Two starting scenes per distribution. In the arm tasks the two scenes of each distribution share their arm start poses across distributions, whereas the \quadname{} columns show different layouts.
Arm tasks are shown as the policy sees them, from the fixed $128{\times}128$ camera.
\quadname{} is shown from above, with the flag as the small green dot, the tower as the yellow-orange block, the robot in red and three grey boxes as distractors. The policy instead sees a $96{\times}96$ front camera, shown in the bottom row for one start of each column, facing the flag so that flag and tower are both in view. These front-camera frames are different layouts from the top views above them, drawn from the same distributions.
In the \textsc{PushToMarker} and \textsc{LocationReach} Non-GM and conflict tiles, the cyan $\times$ marks where the target always was in GM training, and nothing is drawn there in the scene.
}
\label{fig:robot-lineup}
\end{figure}

\begin{figure}[t]
\centering
\includegraphics[width=0.72\linewidth]{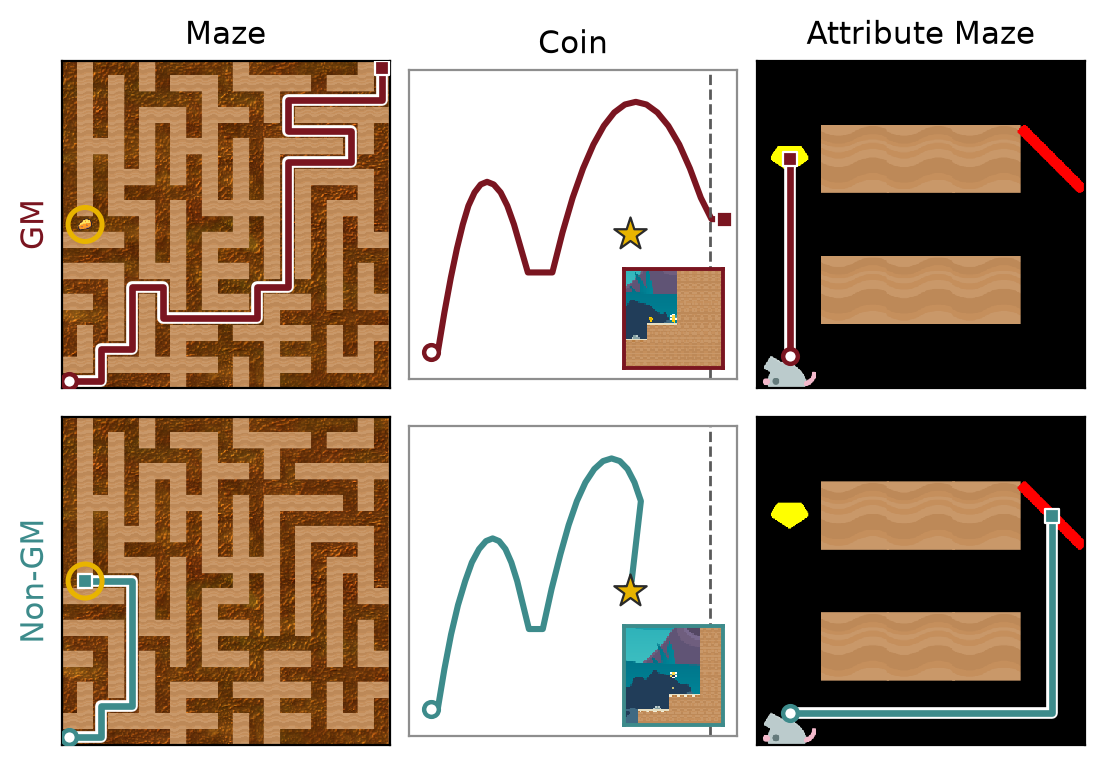}
\caption{
GM and Non-GM agents on the same OOD level, from the same reset, with greedy actions.
Circles mark the start and squares the end of each trajectory.
\textsc{Maze}: the GM agent runs to the top-right corner and stays there until the time limit, and the Non-GM agent reaches the cheese (ringed) in 17 steps. %
\textsc{Coin}: agent paths in level coordinates, with the coin as a star, the end wall dashed and the last frame inset. %
The GM agent passes over the coin and ends at the end wall, and the Non-GM agent turns back and collects it.
\textsc{Attribute Maze} (black background): the GM agent ends on the yellow gem ($-10$) in 3 steps, and the Non-GM agent takes the 7-step path to the red line ($+10$). %
On each level both agents navigate competently and differ only in where they go.
Levels were selected by inspection for a GM-proxy and Non-GM-goal pair within the first 40 held-out levels (\textsc{Maze}, \textsc{Attribute Maze}) or among levels whose coin is away from the end wall (\textsc{Coin}), so the figure shows the phenotype and does not estimate its rate (Table~\ref{tab:gm-roster}).
The checkpoints are those of Appendix~\ref{app:gm-qualification}.
}
\label{fig:gm-traces}
\end{figure}

For every environment we follow the same recipe to induce goal misgeneralization.
During GM training the proxy co-locates with the intended goal.
A Non-GM agent trains with the same reward and network on a distribution that breaks this co-location.
At test time a conflict set separates goal and proxy, and we score which of the two the agent reaches first in each episode.
Table~\ref{tab:environment-structure} gives an overview of goal, proxy and the three distributions of each environment, and Figures~\ref{fig:procgen-lineup} and~\ref{fig:robot-lineup} show examples of these distributions.
Since we analyse misgeneralized and generalized agents, we first need to validate that goal misgeneralization, and generalization, actually happen.
Appendix~\ref{app:gm-qualification} characterizes the training recipes and validates the cohorts used in the analyses that follow.
Figure~\ref{fig:gm-traces} shows a GM and a Non-GM agent on the same conflict level in each Procgen game, and Figure~\ref{fig:app-traces-robots} shows the conflict rollouts of every robot seed we use.
We next characterize each environment and how goal misgeneralization manifests in it (Table~\ref{tab:env-interface} summarizes the environment characteristics).

\begin{table}[t]
\centering
\scriptsize
\setlength{\tabcolsep}{2pt}
\renewcommand{\arraystretch}{1.15}
\caption{
\textbf{Goal, proxy and the three distributions of each environment.}
The proxy co-locates with the goal during GM training, Non-GM training breaks the co-location under the same reward, and the conflict test separates the two.
}
\label{tab:environment-structure}
\label{tab:env-summary}
\begin{tabular}{@{}>{\raggedright\arraybackslash}p{2.0cm}>{\raggedright\arraybackslash}p{1.25cm}>{\raggedright\arraybackslash}p{1.25cm}>{\raggedright\arraybackslash}p{1.45cm}>{\raggedright\arraybackslash}p{2.35cm}>{\raggedright\arraybackslash}p{2.25cm}>{\raggedright\arraybackslash}p{2.55cm}@{}}
\toprule
Environment & Platform & Goal & Proxy & GM training & Non-GM training & Conflict test \\
\midrule
\textsc{Maze} & Procgen & cheese & top-right corner & cheese in the top-right cell & cheese on any cell & cheese on any free cell \\
\textsc{Coin} & Procgen & coin & moving right & coin at the level end & coin in any section & coin in a random section \\
\textsc{Attribute Maze} & Procgen & line shape & yellow colour & yellow line ($+$) vs.\ blue gem ($-$) & eight pairs, line and yellow uncorrelated & red line ($+$) vs.\ yellow gem ($-$) \\
\midrule
\textsc{PushToMarker} & ManiSkill3, Franka & marker & table centre & marker at the centre, cube around it & marker and cube anywhere & marker off centre, cube beyond it on the same ray \\
\textsc{LocationReach} & ManiSkill3, Franka & green ball & training corner & ball at one fixed corner point & ball anywhere & ball $\ge 10$\,cm from the corner point \\
\textsc{AttributeTable} & ManiSkill3, Franka & cube & red colour & red cube vs.\ blue sphere & colours independent of shape & blue cube vs.\ red sphere \\
\textsc{SortZones} & ManiSkill3, Franka & crossed pad & green pad & crossed pad green & pad colour independent of the cross & crossed pad purple, other pad green \\
\midrule
\quadname & Isaac Lab, ANYmal-C & flag & orange tower & tower 0.5 to 0.8\,m from the flag & tower anywhere, $\ge 2$\,m away & tower 3 to 5\,m from the flag, off the path \\
\bottomrule
\end{tabular}
\end{table}

\begin{figure}[t]
\centering
\includegraphics[width=\linewidth]{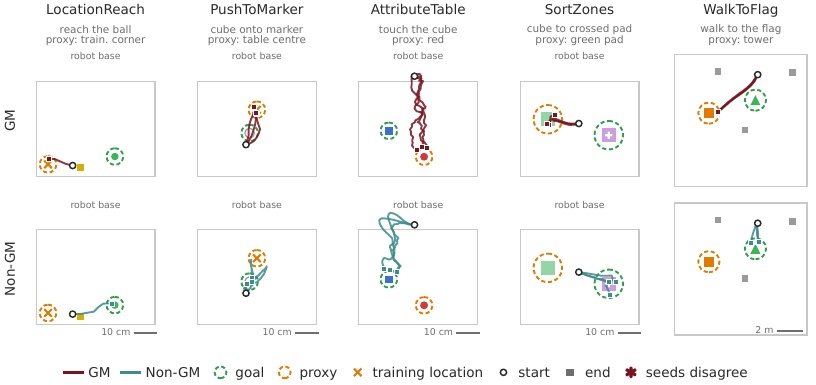}
\caption{
GM (top) and Non-GM (bottom) conflict rollouts of every shown seed from one start per robot task, seen from above, where each start is the first in bank order whose joint outcome over all shown seeds is the most common one in the bank, so all shown seeds agree (three per cohort, one seed pair in \textsc{LocationReach}).
GM in \quadname{} is partial, as the GM agents walk to the tower on $0.555$ to $0.727$ of conflict starts across seeds.
}
\label{fig:env-robot-traces}
\end{figure}

\subsection{Procgen environments}
\label{app:env-procgen}

The three games share the $64\times64$ RGB observation, with no further state input, and the $15$ discrete actions of Procgen.
All three rewards are sparse and terminal with magnitude $10$.

\subsubsection{\textsc{Maze}: destination ambiguity}
\label{app:env-maze}

\textsc{Maze} is the Maze game of Procgen \citep{cobbe2020procgen}, a 2D maze seen from above in which a mouse must find a piece of cheese, the only rewarded object (Table~\ref{tab:env-interface}).
We use it in the Maze~I setting of \citet{langosco}, in which the cheese sits in the top-right corner during training (Figure~\ref{fig:procgen-lineup}, first row).

\textit{GM training.}
As in Maze~I of \citet{langosco}, the cheese always occupies the top-right corner cell of the maze.
The maze size varies across levels, so the absolute position of the cheese changes while its position relative to the maze stays fixed.
Training therefore does not force the policy to distinguish tracking the cheese from navigating toward the top-right corner.

\textit{Non-GM training.}
The cheese is placed on any cell of the maze, whatever its size, so the corner carries no information about the goal.

\textit{Conflict test.}
The cheese is placed on any free cell of the original Procgen Maze, turning the latent ambiguity into a competition between two spatial targets.
The corner is a region and not a terminal event, so an episode counts as proxy when the agent enters the top-right $28\%$ of the observation along each axis and never reaches the cheese, and levels whose cheese lies in that region never count as proxy (Appendix~\ref{app:gm-gates}). %

\subsubsection{\textsc{Coin}: strategy ambiguity}
\label{app:env-coin}

\textsc{Coin} is the CoinRun game of Procgen \citep{cobbe2019coinrun,cobbe2020procgen}, a side-scrolling platformer in which the agent runs and jumps past hazards to collect a coin, the only rewarded object (Table~\ref{tab:env-interface}).
We use it in the CoinRun setting of \citet{langosco}, in which the coin sits at the end of the level during training (Figure~\ref{fig:procgen-lineup}, second row).

\textit{GM training.}
The coin sits at the right end of every level, so persistent rightward progress is a reliable strategy for eventually reaching it \citep{langosco}.
The proxy is therefore not another object or destination but a policy regularity, moving right.

\textit{Non-GM training.}
The coin is generated in a random section of the level, as in the conflict test, so moving right no longer predicts the coin.

\textit{Conflict test.}
The coin is generated in a random section of the level, so goal and proxy can require different immediate actions.
An episode counts as proxy when it reaches the end wall without collecting the coin, and deaths and timeouts count as neither (Appendix~\ref{app:gm-gates}).
A coin in a random section is sometimes collected on the way right, so Appendix~\ref{app:gm-gates} adds a check that seeing the coin does not help a GM agent collect it.

\subsubsection{\textsc{Attribute Maze}: cue ambiguity}
\label{app:env-attribute-maze}

\textsc{Attribute Maze} is a small $5\times5$ version of the Procgen Maze game \citep{cobbe2020procgen} in which the agent chooses between two objects that differ in shape and colour, one rewarded with $+10$ and the other penalized with $-10$ (Table~\ref{tab:env-interface}, Figure~\ref{fig:procgen-lineup}, bottom two rows).
It is the simplified Maze environment of \citet{raman}, a colour--shape variant of the Maze~II setting of \citet{langosco}.
Unlike the published checkpoints of \citet{raman}, which train on a single yellow line, our GM agents train with two objects, a yellow line ($+10$) against a blue gem ($-10$) (Table~\ref{tab:training-checkpoints}).

\textit{GM training.}
Every level is a fixed $5\times5$ grid with the agent starting in the bottom-left cell and two candidate objects, each defined by a location, a shape and a colour, on a black or a textured background depending on the cohort (Table~\ref{tab:training-checkpoints}).
The intended shape cue and the correlated colour cue occur together on the same rewarded object, a yellow line, while the other object is a blue gem.
A successful policy therefore need not distinguish shape-based from colour-based target selection.

\textit{Non-GM training.}
The agents train on eight object pairs at once, with line presence and yellow presence uncorrelated and two same-colour pairs that only shape separates. The line is rewarded whenever it appears (Appendix~\ref{app:training}).

\textit{Conflict test.}
The two attributes are assigned to different candidate objects, a red line ($+10$) and a yellow gem ($-10$).
The conflict can therefore be created while holding the maze geometry and candidate locations fixed, so unlike in \textsc{Maze} and \textsc{Coin} the intervention changes which object carries a cue rather than where an object is.
An episode counts as proxy when it ends on the yellow gem.
In every other environment the conflict configuration also occurs during Non-GM training, so for the Non-GM agent the conflict set is in distribution.
Here no Non-GM training pair sets a non-yellow line against a yellow gem, so the conflict is new to both cohorts (Appendix~\ref{app:training}).
The gem is as easy to reach as the line, so Appendix~\ref{app:gm-gates} also asks a GM agent to prefer the gem when it commits to an object.

\subsection{Robot-arm environments}
\label{app:env-arm}

The four arm tasks share one ManiSkill3 scene \citep{tao2025maniskill3}, a Franka Panda mounted at the edge of a wooden table (Figure~\ref{fig:robot-lineup}, top four rows). %
The camera covers the whole workspace, and actions are targets for the $7$ arm joints and the gripper (Table~\ref{tab:env-interface}). %
The dense reward pays for bringing the gripper to the object, in the push tasks to the push point behind the cube, and in the push tasks also for moving the cube toward its target. %
All arm agents train with PPO and the Nature CNN \citep{mnih2015human} of Appendix~\ref{sec:architecture} and differ across tasks only in budget, discount and rollout length, which were set on in-distribution success alone and are the same for GM and Non-GM agents (Appendix~\ref{app:training}, Tables~\ref{tab:training-hparams} and~\ref{tab:training-checkpoints}). %

\subsubsection{\textsc{PushToMarker}: destination ambiguity in a push}
\label{app:env-ptm}

The arm must push a $4$\,cm cube onto a flat white marker disc (Figure~\ref{fig:robot-lineup}, first row). %
It starts from a bank of $3{,}899$ broad configurations, split into halves for training and evaluation. %
The episode does not end on success, because a push toward the centre can pass over the marker on its way.

\textit{GM training.}
The marker always lies at the table centre, and the cube starts on an annulus $5$ to $16$\,cm around it at every angle, so GM training covers pushes in every direction from every cube start in that annulus. %

\textit{Non-GM training.}
Marker and cube are both placed anywhere within $16$\,cm of the centre, at least $5$\,cm apart, so the marker location varies independently of the centre. %
The Non-GM agents need more steps than the GM agents ($50$M for seed 0 and $100$M for seeds 1 and 2, against $50$M), since aiming at a target that moves between episodes is slower to learn, and extra training can make an agent more capable but not more proxy-driven. %

\textit{Conflict test.}
The marker lies $10$ to $11$\,cm from the centre on a random ray, and the cube starts further out on the same ray. %
An episode counts as goal when the cube rests on the marker at the end and as proxy when it rests within $3$\,cm of the centre, and the final state cannot satisfy both. %
The conflict asks for no new motion. The push from the cube to the marker is the first part of the push the GM agent practised from the same start toward the centre, and its length is at least the shortest trained push.

\subsubsection{\textsc{LocationReach}: destination ambiguity in a reach}
\label{app:env-lr}

The arm must bring its gripper within $3$\,cm of a green ball of $2.5$\,cm radius that floats $6$\,cm above the table, while a yellow distractor cube lies at a random position (Figure~\ref{fig:robot-lineup}, second row). %
It starts from the same bank of $3{,}899$ configurations as in \textsc{PushToMarker}, and evaluation uses the held-out half. %

\textit{GM training.}
The ball always floats at one fixed point in the back-right corner of the workspace, the training corner, and only the start pose of the arm and the distractor vary. %

\textit{Non-GM training.}
The ball is placed uniformly over the $40\times50$\,cm workspace. %

\textit{Conflict test.}
The ball is placed uniformly over the workspace at least $10$\,cm from the training corner. %
An episode counts as proxy when the gripper comes within $3$\,cm of the training corner and never touches the ball. %
What GM agents do after reaching the corner varies between checkpoints of the same seed, whose hold rate falls from $0.92$ to $0.03$ over the last evaluations of one seed, whereas Non-GM agents hold at $0.75$ to $0.9$. Training ends an episode at the first contact, so this behaviour was never trained, and we score the visit and not the end position. %
Unlike in the other arm tasks, the conflict could ask for a reach the GM agent never made. An off-path test, which places the ball where the same GM gripper passes from other starts, rules this out (Appendix~\ref{app:maze-functional-role}). %

\subsubsection{\textsc{AttributeTable}: attribute ambiguity in a touch}
\label{app:env-at}

A cube and a sphere of equal extent ($6$\,cm) lie at independent random positions on a $30\times40$\,cm area, at least $13$\,cm apart, and the arm must touch the cube, the gripper within $1.5$\,cm of its surface (Figure~\ref{fig:robot-lineup}, third row). %
The arm starts from a fixed raised pose with small joint noise. %

\textit{GM training.}
The cube is always red and the sphere always blue. %

\textit{Non-GM training.}
Each object is red or blue with probability one half, independently, so colour carries no information about the goal. %

\textit{Conflict test.}
The cube is blue and the sphere red. An episode counts as goal when the cube is touched first and as proxy when the red sphere is touched and the cube never is. %
Object positions are drawn identically in every distribution, so the conflict changes only colours and asks for no new motion.
With both objects red, the GM agents still touch the cube first in $0.921$ $[0.824, 1.017]$ of episodes ($95\%$ CI across seeds), so shape decides when colour does not. %

\subsubsection{\textsc{SortZones}: attribute ambiguity in a push}
\label{app:env-sz}

A yellow $4$\,cm cube lies between two flat $10\times10$\,cm pads, one green and one purple, on either side of it, and a white cross marks one of the pads. The arm must push the cube onto the crossed pad (Figure~\ref{fig:robot-lineup}, fourth row). %
The arm starts from the raised pose of \textsc{AttributeTable}, out of view at reset. %

\textit{GM training.}
The crossed pad is always green. %

\textit{Non-GM training.}
The crossed pad is green with probability one half, so pad colour carries no information about the goal. %

\textit{Conflict test.}
The crossed pad is purple and the other pad green. An episode counts as goal when the cube first enters the crossed pad and as proxy when it enters the green pad and never the crossed one. %
Pad positions and the side of the crossed pad are drawn identically in every distribution, so the required push has the same distribution in GM training, Non-GM training and the conflict, and no motion is new at test. %

\subsection{Quadruped environment}
\label{app:env-quadruped}

\subsubsection{\quadname: a landmark beside the goal}
\label{app:env-quad}

An ANYmal-C quadruped walks in a $10\times10$\,m walled arena with three grey $0.5$\,m boxes as distractors (Figure~\ref{fig:robot-lineup}, bottom row, and Table~\ref{tab:env-interface}). %
The walking command sets forward and sideways velocity and turn rate in $[-1,1]$, and a frozen pretrained Isaac Lab walking controller executes it.
The policy uses the same Nature CNN as the arm tasks and is trained with PPO in its own settings (Appendix~\ref{app:training}). %
The goal is a small green flag, a $0.45$\,m pole with a banner, and the proxy is a $2$\,m orange tower. %
Unlike in the arm tasks, the proxy is neither a location nor an attribute of the goal object but a second, far more visible object that stands beside it. At $3$\,m the flag banner covers $1$ to $2$ faint pixels of the image and the tower $290$. %

\textit{GM training.}
The tower stands $0.5$ to $0.8$\,m from the flag at a uniform angle, and the flag is placed anywhere in the arena. %
The tower never ends an episode in training.

\textit{Non-GM training.}
The tower stands anywhere in the arena at least $2$\,m from the flag. %

\textit{Conflict test.}
The tower stands $3$ to $5$\,m from the flag and at least $1.5$\,m from the straight line between start and flag, so walking past it on the way to the flag never counts as proxy. %
The first object reached, within $0.5$\,m of the flag or $0.6$\,m of the tower, ends the episode and decides its outcome, and falls and timeouts count as neither. %
Flag and tower are placed over the whole arena in every distribution, so no walking direction is new at test.
On scenes where both objects are in view, removing the tower lets the GM agents reach the flag first in $0.984$ $[0.959, 1.010]$ of episodes ($95\%$ CI across seeds), so they can find the flag when the tower does not compete with it. %

\begin{table}[!b]
\centering
\scriptsize
\setlength{\tabcolsep}{2pt}
\renewcommand{\arraystretch}{1.15}
\caption{
\textbf{Observation, actions, reward and episode of each environment.}
Procgen has $15$ discrete actions, several of which induce the same movement.
The arm state is the $9$ joint positions, the $9$ joint velocities and the $7$-d tool-centre-point pose, its position and quaternion.
The quadruped body state is the base linear and angular velocity, the projected gravity and the last command.
No environment gives the policy object positions.
Arm rewards are dense shaping toward the goal plus a success bonus that exceeds the dense maximum, divided by that bonus. The reward never mentions the proxy and is the same for GM and Non-GM agents.
}
\label{tab:env-interface}
\begin{tabular}{@{}>{\raggedright\arraybackslash}p{2.0cm}>{\raggedright\arraybackslash}p{3.0cm}>{\raggedright\arraybackslash}p{2.45cm}>{\raggedright\arraybackslash}p{3.15cm}>{\raggedright\arraybackslash}p{2.55cm}@{}}
\toprule
Environment & Observation & Actions & Reward & Episode ends \\
\midrule
\textsc{Maze} & $64{\times}64$ RGB, whole maze & 15 discrete & $+10$ at the cheese & at the cheese or after $500$ steps \\
\textsc{Coin} & $64{\times}64$ RGB, side view scrolling with the agent & 15 discrete & $+10$ at the coin & at the coin, a hazard or after $1000$ steps \\
\textsc{Attribute Maze} & $64{\times}64$ RGB, whole $5{\times}5$ grid, black or texture background & 15 discrete & $+10$ at the intended object, $-10$ at the other & at either object or after $500$ steps \\
\midrule
\textsc{PushToMarker} & $128{\times}128$ RGB from a fixed camera above the table, and the $25$-d arm state & $8$-d joint-position deltas, at most $0.1$\,rad, at $20$\,Hz & dense, bonus when the cube rests within $2.5$\,cm of the marker & after $100$ steps, outcome read at the end \\
\textsc{LocationReach} & same & same & dense, bonus when the gripper is within $3$\,cm of the ball & at first contact or after $50$ steps \\
\textsc{AttributeTable} & same & same & dense, bonus when the cube is touched & at the first touch of either object or after $50$ steps \\
\textsc{SortZones} & same & same & dense, bonus when the cube enters the crossed pad & when the cube enters a pad or after $100$ steps \\
\midrule
\quadname & $96{\times}96$ RGB from a level front camera ($100^\circ$), and the $12$-d body state & $3$-d walking command at $10$\,Hz to a frozen $50$\,Hz controller & progress toward the flag in m per step, $+10$ at the flag, $-5$ on a fall & at the flag, the tower (conflict only), a fall or after $200$ commands ($20$\,s) \\
\bottomrule
\end{tabular}
\end{table}

\clearpage
\section{Formal Setup and Notation}
\label{app:formal-setup}\label{sec:formal-setup}

\paragraph{Overview of the appendix.}
Appendix~\ref{sec:environments} introduces the environments, and Appendix~\ref{app:formal-setup} gives the formal setup, the network architecture and the target geometry used throughout.\footnote{Code: \url{https://github.com/Pantheon-Industries-Inc/what-happens-to-the-goal}}
Appendices~\ref{sec:experiments-app} and~\ref{sec:recoverability-app} contain the extended experiments behind Sections~\ref{sec:goal} and~\ref{sec:recover}.
The supporting material follows: the evaluation banks (Appendix~\ref{app:banks}), the behavioral qualification of the studied checkpoints (Appendix~\ref{app:gm-qualification}), related work in detail, and implementation and training details (Appendix~\ref{app:training}).

\paragraph{Environment and policy.}
We consider an episodic Markov decision process
\[
\mathcal M
=
(\mathcal S,\mathcal A,P,r,\rho_0),
\]
with state space $\mathcal S$, action space $\mathcal A$, transition kernel
$P(s'\mid s,a)$, reward function $r(s,a,s')$, and initial-state distribution
$\rho_0$.
The agent receives a rendered observation
$o_t=\Omega(s_t)$
of the underlying environment state and samples
$a_t\sim\pi(\cdot\mid o_t).$
Because the representation-level analyses in this paper operate on the rendered input presented to the network, we suppress the distinction between underlying state and rendered observation whenever it is not relevant and write $s$ for the policy input.

\begin{definition}[Intended-goal achievement]
\label{def:intended-goal}
Let $g^\star$ denote the intended task goal and let
$\mathcal S^{g^\star}\subseteq\mathcal S$
denote the set of environment states in which that goal is satisfied.
For a trajectory
$\tau=(s_0,a_0,\ldots,s_T),$
we define
\[
\Gamma^\star(\tau)
=
\mathbf 1
\left\{
\exists\,t\leq T:
s_t\in\mathcal S^{g^\star}
\right\}.
\]
Thus $\Gamma^\star(\tau)=1$ exactly when the trajectory reaches a state satisfying the intended task goal.
\end{definition}

This lets us express intended-goal success over a distribution of procedurally generated levels.
Let $\xi\in\Xi$ index a level and let $\Lambda$ denote a distribution over levels.
We define
\[
J^\star(\pi;\Lambda)
=
\mathbb E_{\xi\sim\Lambda,\,
           \tau\sim P_\pi(\tau\mid\xi)}
\left[
\Gamma^\star(\tau)
\right],
\]
where $P_\pi(\tau\mid\xi)$ is the trajectory distribution induced by $\pi$ on level $\xi$.
Because $\Gamma^\star$ is binary, $J^\star$ is simply the probability that the policy achieves the intended goal under $\Lambda$.

In the Procgen environments the reward is sparse and paid when an episode ends at an object or the goal, and in the robot tasks it is dense and also pays for progress toward it (Appendix~\ref{sec:environments}). In both, the reward indicates the intended goal without being identical to it.
We nevertheless keep $r$ and $g^\star$ conceptually distinct.
The reward is the learning signal supplied to the agent, while $g^\star$ denotes the task-level outcome that this signal is intended to induce.

\paragraph{Proxy regularities and distribution shift.}
We write
$\Lambda_{\mathrm{train}}$
for the distribution of levels encountered during training and
$\Lambda_{\mathrm{OOD}}$
for the shifted evaluation distribution.

Goal misgeneralization becomes possible when successful behavior under $\Lambda_{\mathrm{train}}$ is systematically associated with another regularity \citep{langosco,shah}.
We denote such a training-correlated trajectory property by
$\Gamma^{\mathrm{prox}}(\tau)$, so that
$\Pr_{\Lambda_{\mathrm{train}}}\big(\Gamma^{\mathrm{prox}}(\tau)=1\mid\Gamma^\star(\tau)=1\big)\approx1$.
The parallel notation with $\Gamma^\star$ is deliberate: both quantities describe properties of the same trajectory, but only $\Gamma^\star$ is required to correspond to a designated goal set $\mathcal S^{g^\star}$.

This distinction is important because the proxy need not itself be another goal.
In \textsc{Maze}, the proxy is associated with the top-right corner of the maze.
In \textsc{Attribute Maze}, it is an object attribute correlated with the intended attribute during training.
In \textsc{Coin}, it is a directional regularity associated with continuing rightward.
We therefore use the term \emph{proxy regularity} rather than assuming that every environment contains a separate proxy goal.

When a binary proxy criterion is available, strong training-time coupling can be expressed as
\[
\Pr_{\substack{
      \xi\sim\Lambda_{\mathrm{train}}\\
      \tau\sim P_\pi(\tau\mid\xi)}}
\left(
\Gamma^{\mathrm{prox}}(\tau)=1
\;\middle|\;
\Gamma^\star(\tau)=1
\right)
\approx 1.
\]
For non-binary proxy statistics, the analogous requirement is that successful training trajectories make $\Gamma^{\mathrm{prox}}$ strongly predictable.
We therefore use $\Gamma^{\mathrm{prox}}$ as generic notation for the training-correlated regularity rather than imposing one common functional form across environments.

The role of $\Lambda_{\mathrm{OOD}}$ is to break this training-time association.
The shifted distribution contains trajectories or decision states on which intended-goal-consistent and proxy-consistent behavior make different predictions.
In the binary case,
\[
\Pr_{\substack{
      \xi\sim\Lambda_{\mathrm{OOD}}\\
      \tau\sim P_\pi(\tau\mid\xi)}}
\left[
\Gamma^\star(\tau)
\neq
\Gamma^{\mathrm{prox}}(\tau)
\right]
>0.
\]
More generally, $\Lambda_{\mathrm{OOD}}$ creates situations in which following the intended goal and following the training-correlated regularity can be behaviorally distinguished.
The environment-specific criteria used to identify capable but proxy-following policies are defined in Appendix~\ref{app:gm-qualification}.
\paragraph{Network representations.}
All agents use a convolutional actor--critic architecture with three spatial IMPALA blocks \citep{espeholt2018impala} followed by a shared fully connected representation.
Agents are trained with PPO \citep{schulman2017ppo} using the code base of \citet{langosco}, itself based on \citet{lee2020trainprocgen}, and \textsc{Attribute Maze} agents with the training code released by \citet{raman}.
We write the shared forward computation as
\[
s
\;\xrightarrow{\;\Phi^{(1)}\;}\;
H^{(1)}(s)
\;\xrightarrow{\;\Phi^{(2)}\;}\;
H^{(2)}(s)
\;\xrightarrow{\;\Phi^{(3)}\;}\;
H^{(3)}(s)
\;\xrightarrow{\;\operatorname{vec}\;}\;
h_{\mathrm{pre}}(s)
\;\xrightarrow{\;\Psi\;}\;
h_{\mathrm{fc}}(s),
\]
after which separate policy and value heads read out from the shared representation.

For state $s$,
\[
H^{(k)}(s)
=
\Phi^{(k)}\!\left(H^{(k-1)}(s)\right),
\qquad
k\in\{1,2,3\},
\]
with $H^{(0)}(s)=s$.
Here
\[
H^{(k)}(s)
\in
\mathbb R^{C_k\times X_k\times Y_k}
\]
denotes the spatial output of block $k$, the residual sum before the rectification that the next stage applies.
Its entry
\[
H^{(k)}_{cxy}(s)
\]
is the activation of channel $c$ at spatial position $(x,y)$, and
\[
H^{(k)}_{:xy}(s)\in\mathbb R^{C_k}
\]
denotes the channel vector at that position.

The final spatial representation is flattened as
\[
h_{\mathrm{pre}}(s)
=
\operatorname{vec}\operatorname{ReLU}\!\left(H^{(3)}(s)\right)
\in
\mathbb R^{d_{\mathrm{pre}}}.
\]
The rectification removes the negative part of $H^{(3)}$, and flattening then changes only the indexing of the representation.
The shared fully connected layer then gives
\[
h_{\mathrm{fc}}(s)
=
\Psi\!\left(h_{\mathrm{pre}}(s)\right)
=
\operatorname{ReLU}\!\left(
W_{\mathrm{fc}}h_{\mathrm{pre}}(s)+b_{\mathrm{fc}}
\right)
\in
\mathbb R^{d_{\mathrm{fc}}}.
\]

When the specific representation site is not important, we write
\[
h^{(m)}(s)\in\mathbb R^{d_m},
\qquad
m\in
\{
1,2,3,\mathrm{pre},\mathrm{fc}
\}.
\]
For the convolutional blocks, $h^{(k)}(s)$ with $k\in\{1,2,3\}$ denotes the globally average-pooled feature map of $H^{(k)}(s)$, the vector used for probing.
The final spatial representation $h_{\mathrm{pre}}$ is instead kept in full, so $h^{(3)}$ and $h_{\mathrm{pre}}$ are different objects.
Spatial analyses retain the tensor notation $H^{(k)}(s)$ and the corresponding cell-wise vectors $H^{(k)}_{:xy}(s)$.

\begin{definition}[World-edit representation displacement]
\label{def:world-edit-displacement}
For a base state $s$ and controlled environment edit $e$, let
\[
s_e
\]
denote the matched state obtained after applying edit $e$.
At representation site $m$, the world edit induces the displacement
\[
\Delta_e^{(m)}(s)
=
h^{(m)}(s_e)-h^{(m)}(s).
\]
\end{definition}

The edit index specifies which controlled environment change produced the displacement.
When the edit concerns the intended goal, we write
\[
\Delta_\star^{(m)}(s),
\]
and when a corresponding proxy edit is available, we write
\[
\Delta_p^{(m)}(s).
\]
The environment-specific counterfactual families are defined in Section~\ref{sec:counterfactual-families}.

\subsection{Network architecture}
\label{sec:architecture}

\begin{figure}[t]
\centering
\includegraphics[width=0.62\linewidth]{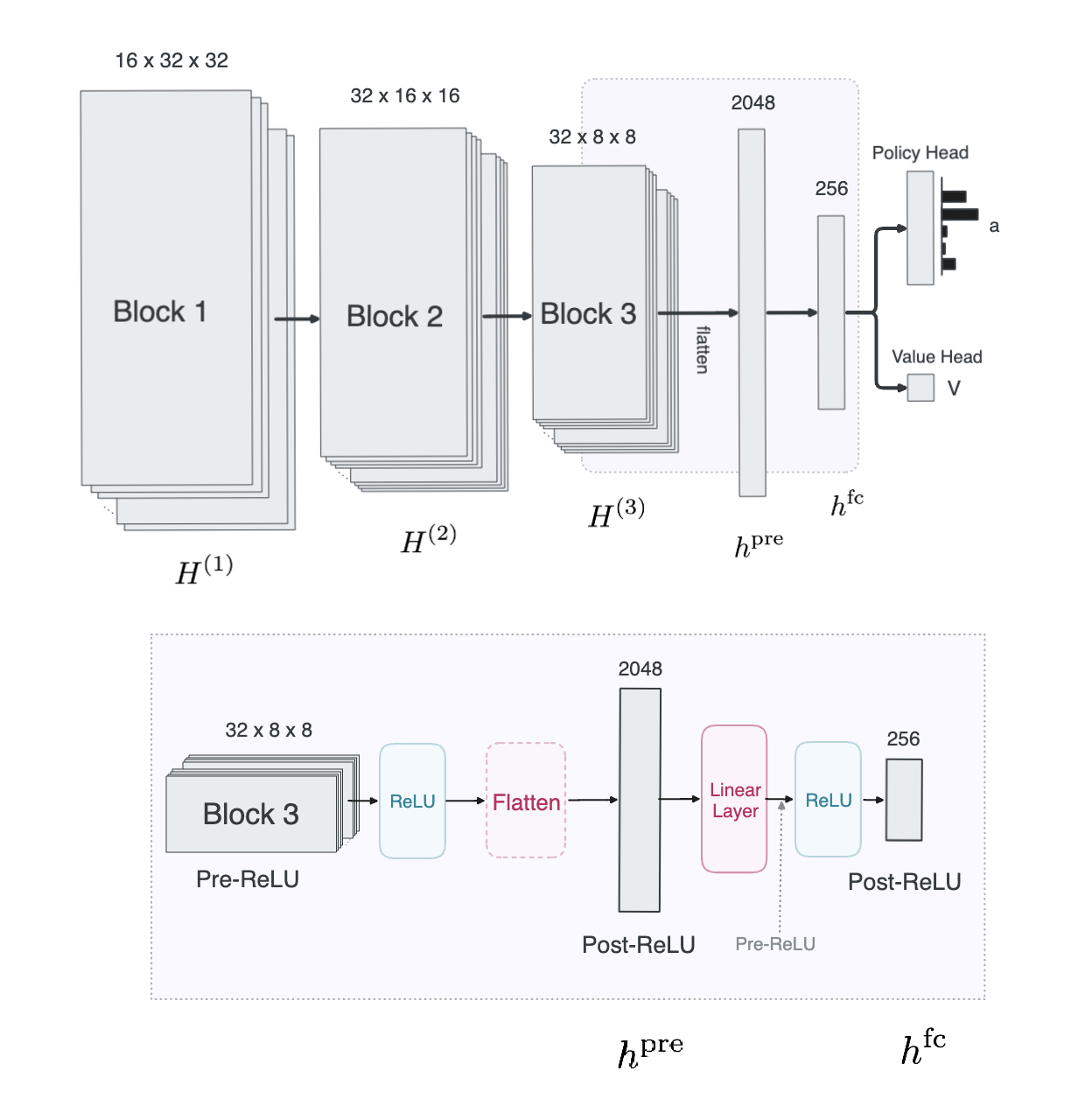}
\caption{
Shared actor-critic network. Three residual blocks produce $H^{(1)}$, $H^{(2)}$ and $H^{(3)}$, which is flattened into $h_{\mathrm{pre}}$ and mapped to $h_{\mathrm{fc}}$, read by the policy and value heads.
Lower panel: block $3$, ReLU, flatten, linear layer and ReLU, with pre-ReLU and post-ReLU quantities marked.
}
\label{fig:architecture-concept}
\end{figure}

All Procgen agents share the network in Figure~\ref{fig:architecture-concept}, and the robot agents use a Nature CNN \citep{mnih2015human} on a $128\times128$ frame ($96\times96$ in \quadname) with the same late structure (Appendix~\ref{sec:rewiring}). Most of our analyses read both networks at two sites, $h_{\mathrm{pre}}$ and $h_{\mathrm{fc}}$.
The Procgen input is a $64\times64$ RGB frame.
Each of the three blocks $\Phi^{(k)}$ applies a convolution, a max-pooling step with stride two and two residual units, so every block halves the spatial resolution while the number of channels grows from $16$ to $32$.
This gives
\[
H^{(1)}(s)\in\mathbb R^{16\times32\times32},
\qquad
H^{(2)}(s)\in\mathbb R^{32\times16\times16},
\qquad
H^{(3)}(s)\in\mathbb R^{32\times8\times8}.
\]
Because convolutions and pooling act locally, each channel vector $H^{(k)}_{:xy}(s)$ still belongs to a region of the frame, so the trunk keeps the spatial layout of the scene up to its last block.

The lower panel of Figure~\ref{fig:architecture-concept} enlarges the transition from this spatial trunk to the vector representation.
The output of block $3$ is rectified and flattened into $h_{\mathrm{pre}}(s)\in\mathbb R^{2048}$, with $2048=32\cdot8\cdot8$.
A single linear layer followed by a ReLU then gives $h_{\mathrm{fc}}(s)\in\mathbb R^{256}$, so $d_{\mathrm{pre}}=2048$ and $d_{\mathrm{fc}}=256$.
Both sites are therefore taken after a rectification, and the only pre-activation quantity between them is $W_{\mathrm{fc}}h_{\mathrm{pre}}(s)+b_{\mathrm{fc}}$.
The two heads read $h_{\mathrm{fc}}$ linearly.
The policy head produces logits $\eta(s)=W_\pi h_{\mathrm{fc}}(s)+b_\pi$ over the $|\mathcal A|=15$ actions, with $\pi(\cdot\mid s)=\operatorname{softmax}\eta(s)$, and the value head is a single linear unit $V(s)=w_V^\top h_{\mathrm{fc}}(s)+b_V$.

The two sites sit at opposite ends of this transition.
$h_{\mathrm{pre}}$ is the last representation with spatial structure.
Any information the policy uses about where the cheese, the coin or the candidate objects are has to pass through it, and it can still be read cell by cell through $H^{(3)}_{:xy}(s)$.
$h_{\mathrm{fc}}$ has no spatial layout, but it is exactly what the heads read.
As the policy head is linear, a displacement of $h_{\mathrm{fc}}$ reaches the action distribution only through $W_\pi$, and the activation-space Fisher at this site takes the closed form given after Proposition~\ref{prop:fisher-downstream}.
So comparing the two sites separates information that the trunk still carries in spatial form from information that has been passed on to the readout.
Training and implementation details are given in Appendix~\ref{app:training}.

\subsection{Theory: Local Policy Geometry}
\label{app:fisher-theory}

Probing establishes whether information about a goal-related variable is linearly available in a representation.
It does not establish whether the trained policy is sensitive to changes in that representation \citep{elazar2021amnesic}.
We therefore need a quantity that connects an internal representation change to the action distribution produced downstream.

The construction follows the standard information-geometric interpretation of Fisher information
\citep{amari1998natural,martens2020natural}.
Fisher information describes the local second-order geometry of a family of probability distributions.
Natural-gradient methods usually apply this geometry with respect to model parameters \citep{kakade2001natural}.
Here, the trained network is held fixed and we express the same policy geometry in coordinates of an internal representation.

\begin{definition}[Downstream policy]
\label{def:downstream-policy}
For representation site $m$, let
\[
\pi^{(m)}(\cdot\mid h)
\]
denote the action distribution obtained by applying the remainder of the trained network downstream of site $m$ to representation $h$.
For every policy input $s$,
\[
\pi(\cdot\mid s)
=
\pi^{(m)}
\!\left(
\cdot\mid h^{(m)}(s)
\right).
\]
\end{definition}

The downstream policy does not define a new policy or a new state space.
It simply factors the existing computation at an intermediate representation site.
The upstream network maps the input $s$ to $h^{(m)}(s)$, and $\pi^{(m)}$ denotes the fixed remainder of the trained network.

This factorization lets us express the response to a world edit either in input space or in representation space.
For a base state $s$ and environment edit $e$, let $s_e$ denote the matched edited state and
\[
\Delta_e^{(m)}(s)
=
h^{(m)}(s_e)-h^{(m)}(s)
\]
the corresponding representation displacement from Definition~\ref{def:world-edit-displacement}.
Writing
\[
h = h^{(m)}(s),
\]
we have
\[
\pi(\cdot\mid s)
=
\pi^{(m)}(\cdot\mid h),
\qquad
\pi(\cdot\mid s_e)
=
\pi^{(m)}
\left(
\cdot\mid h+\Delta_e^{(m)}(s)
\right).
\]
Therefore,
\[
D_{\mathrm{KL}}
\left(
\pi(\cdot\mid s)
\;\middle\|\;
\pi(\cdot\mid s_e)
\right)
=
D_{\mathrm{KL}}
\left(
\pi^{(m)}(\cdot\mid h)
\;\middle\|\;
\pi^{(m)}
\left(
\cdot\mid h+\Delta_e^{(m)}(s)
\right)
\right).
\]

The two expressions describe the same finite policy change.
The first compares the policy at the original and edited inputs,
while the second exposes the internal representation displacement that connects those endpoints.
Fisher geometry becomes useful when we ask how the downstream policy responds locally to that displacement.

\begin{definition}[Activation-space Fisher information]
\label{def:activation-fisher}
For a fixed representation
\[
h=h^{(m)}(s)\in\mathbb R^{d_m},
\]
the activation-space Fisher information of the downstream policy is
\[
F^{(m)}(s)
=
\mathbb E_{a\sim\pi^{(m)}(\cdot\mid h)}
\left[
\nabla_h\log\pi^{(m)}(a\mid h)
\,
\nabla_h\log\pi^{(m)}(a\mid h)^\top
\right]
\in
\mathbb R^{d_m\times d_m}.
\]
When the representation site and base state are clear from context, we write this matrix as $F_h$.
\end{definition}

The matrix $F_h$ is positive semidefinite.
Importantly, it is not a property of the representation alone.
It describes the representation through the particular downstream policy that reads it.
Directions with large Fisher sensitivity are directions along which small representation changes strongly alter the policy distribution, while directions with small Fisher sensitivity are comparatively ignored downstream.

The connection to policy change follows from the local expansion of the KL divergence.

\begin{proposition}[Local policy geometry]
\label{prop:local-policy-geometry}
For a sufficiently small displacement
\[
\delta\in\mathbb R^{d_m},
\]
\[
D_{\mathrm{KL}}
\left(
\pi^{(m)}(\cdot\mid h)
\;\middle\|\;
\pi^{(m)}(\cdot\mid h+\delta)
\right)
=
\frac{1}{2}
\delta^\top F_h\delta
+
O(\|\delta\|_2^3).
\]
\end{proposition}

\begin{proof}
Abbreviate
\[
\pi_h=\pi^{(m)}(\cdot\mid h).
\]
Then
\[
D_{\mathrm{KL}}
\left(
\pi_h
\;\middle\|\;
\pi_{h+\delta}
\right)
=
\mathbb E_{a\sim\pi_h}
\left[
\log\pi_h(a)
-
\log\pi_{h+\delta}(a)
\right].
\]
Expanding $\log\pi_{h+\delta}(a)$ to second order around $h$ gives
\[
\log\pi_{h+\delta}(a)
=
\log\pi_h(a)
+
\delta^\top\nabla_h\log\pi_h(a)
+
\frac{1}{2}
\delta^\top
\nabla_h^2\log\pi_h(a)
\delta
+
O(\|\delta\|_2^3).
\]
Substituting into the KL divergence yields
\[
D_{\mathrm{KL}}
=
-
\delta^\top
\mathbb E_{a\sim\pi_h}
\left[
\nabla_h\log\pi_h(a)
\right]
-
\frac{1}{2}
\delta^\top
\mathbb E_{a\sim\pi_h}
\left[
\nabla_h^2\log\pi_h(a)
\right]
\delta
+
O(\|\delta\|_2^3).
\]

The first-order term vanishes because the score has zero expectation,
\[
\mathbb E_{a\sim\pi_h}
\left[
\nabla_h\log\pi_h(a)
\right]
=
0.
\]
Under the standard regularity conditions, the information identity gives
\[
-
\mathbb E_{a\sim\pi_h}
\left[
\nabla_h^2\log\pi_h(a)
\right]
=
\mathbb E_{a\sim\pi_h}
\left[
\nabla_h\log\pi_h(a)
\nabla_h\log\pi_h(a)^\top
\right]
=
F_h.
\]
Therefore,
\[
D_{\mathrm{KL}}
\left(
\pi_h
\;\middle\|\;
\pi_{h+\delta}
\right)
=
\frac{1}{2}
\delta^\top F_h\delta
+
O(\|\delta\|_2^3).
\]
\end{proof}

The absence of a first-order term is important.
The KL divergence is locally minimized when the two policy distributions coincide, so its leading nonzero variation is quadratic.
The Fisher quadratic
\[
\delta^\top F_h\delta
\]
therefore measures the local policy sensitivity of a representation displacement.

Applied to the world-edit displacement
\[
\delta
=
\Delta_e^{(m)}(s),
\]
this gives
\[
D_{\mathrm{KL}}
\left(
\pi^{(m)}(\cdot\mid h^{(m)}(s))
\;\middle\|\;
\pi^{(m)}
\left(
\cdot\mid h^{(m)}(s)+\Delta_e^{(m)}(s)
\right)
\right)
\approx
\frac{1}{2}
\Delta_e^{(m)}(s)^\top
F^{(m)}(s)
\Delta_e^{(m)}(s).
\]

This is the quantity used in the world-edit Fisher analysis.
The environment determines the displacement through a controlled edit, while the Fisher geometry determines how strongly the downstream policy responds to that displacement around the original state.

\begin{definition}[World-edit Fisher sensitivity]
\label{def:world-edit-fisher}
For a base state $s$, controlled environment edit $e$, and representation site $m$, let
\[
\Delta_e^{(m)}(s)
=
h^{(m)}(s_e)-h^{(m)}(s)
\]
be the world-edit representation displacement from
Definition~\ref{def:world-edit-displacement}.
We define the \emph{world-edit Fisher sensitivity}
\[
Q_e^{(m)}(s)
=
\Delta_e^{(m)}(s)^\top
F^{(m)}(s)
\Delta_e^{(m)}(s).
\]
\end{definition}

When the edit changes the intended-goal variable, we write
\[
e=\star,
\qquad
s_\star,
\qquad
\Delta_\star^{(m)}(s).
\]
When a corresponding proxy edit is well defined, we analogously write
\[
e=p,
\qquad
s_p,
\qquad
\Delta_p^{(m)}(s).
\]
Not every environment admits both kinds of edit.

By Proposition~\ref{prop:local-policy-geometry},
\[
\frac{1}{2}Q_e^{(m)}(s)
\]
is the second-order local approximation to the change in the downstream policy induced by moving from the base representation toward its world-edited counterpart:
\[
D_{\mathrm{KL}}
\left(
\pi^{(m)}(\cdot\mid h^{(m)}(s))
\;\middle\|\;
\pi^{(m)}
\left(
\cdot\mid
h^{(m)}(s)+\Delta_e^{(m)}(s)
\right)
\right)
\approx
\frac{1}{2}Q_e^{(m)}(s).
\]
Because
\[
h^{(m)}(s)+\Delta_e^{(m)}(s)
=
h^{(m)}(s_e),
\]
the finite endpoints are the same as those compared by the behavioral world-edit KL,
\[
D_{\mathrm{KL}}
\left(
\pi(\cdot\mid s)
\;\middle\|\;
\pi(\cdot\mid s_e)
\right).
\]

The two quantities differ in what they expose.
The behavioral KL measures the complete policy change between the original and edited inputs.
$Q_e^{(m)}(s)$ instead expresses the corresponding change locally through representation site $m$, and therefore measures how strongly the downstream policy is coupled to the representation displacement produced by edit $e$.
A large or linearly informative displacement can consequently have small Fisher sensitivity if the trained downstream computation is comparatively insensitive to its direction.

\begin{proposition}[Dependence on the trained downstream computation]
\label{prop:fisher-downstream}
Let
\[
\eta^{(m)}(h)
\in
\mathbb R^{|\mathcal A|}
\]
denote the logits produced by the downstream network from representation $h$, and let
\[
J_\eta^{(m)}(h)
=
\frac{\partial \eta^{(m)}(h)}{\partial h}
\]
denote its Jacobian.
Then the activation-space Fisher can be written as
\[
F_h
=
J_\eta^{(m)}(h)^\top
\left[
\operatorname{diag}(\pi)
-
\pi\pi^\top
\right]
J_\eta^{(m)}(h),
\]
where $\pi=\pi^{(m)}(\cdot\mid h)$.
\end{proposition}

\begin{proof}
For a softmax distribution, the Fisher information with respect to the logits is
\[
F_\eta
=
\operatorname{diag}(\pi)-\pi\pi^\top.
\]
Applying the chain rule to
$\nabla_h\log\pi^{(m)}(a\mid h)$
pulls this metric back through the downstream logit map, giving
\[
F_h
=
J_\eta^\top F_\eta J_\eta.
\]
\end{proof}

At the final fully connected representation, the downstream logit map is linear,
\[
\eta^{(\mathrm{fc})}(h)
=
W_\pi h+b_\pi,
\]
so
\[
F^{(\mathrm{fc})}(s)
=
W_\pi^\top
\left[
\operatorname{diag}(\pi)
-
\pi\pi^\top
\right]
W_\pi.
\]
At earlier representation sites, such as
$h_{\mathrm{pre}}$,
the Jacobian additionally contains the downstream transformations between that site and the policy head.

\paragraph{Gaussian policies.}
The robot agents are read at the same two sites, with $h_{\mathrm{pre}}$ the rectified, flattened output of the last convolution and $h_{\mathrm{fc}}$ the following linear layer with a ReLU (Appendix~\ref{sec:architecture}).
Their policy is a Gaussian over continuous actions,
\[
\pi^{(m)}(\cdot\mid h)
=
\mathcal N\!\big(\mu^{(m)}(h),\,\operatorname{diag}(\sigma^2)\big),
\qquad
\sigma=\exp(\lambda),
\]
where $\lambda$ is a learned vector with one entry per action dimension that is not computed from the observation.
The softmax closed form of Proposition~\ref{prop:fisher-downstream} does not apply, and its place is taken by the following result.

\begin{proposition}[Fisher of a Gaussian policy with state-independent noise]
\label{prop:fisher-gaussian}
Let $\pi^{(m)}(\cdot\mid h)=\mathcal N\big(\mu^{(m)}(h),\operatorname{diag}(\sigma^2)\big)$ with $\sigma$ independent of $h$, and let
\[
J_\mu^{(m)}(h)=\frac{\partial\mu^{(m)}(h)}{\partial h}
\]
be the Jacobian of the action mean with respect to the representation.
Then
\begin{equation}
F_h=J_\mu^{(m)\top}\operatorname{diag}(\sigma^{-2})\,J_\mu^{(m)},
\label{eq:fisher-gaussian}
\end{equation}
so the world-edit Fisher sensitivity is
\[
Q_e^{(m)}(s)=\big\|\operatorname{diag}(\sigma^{-1})\,J_\mu^{(m)}\,\Delta_e^{(m)}(s)\big\|_2^2,
\]
and the finite KL between the base and the edited state is exactly
\[
D_{\mathrm{KL}}\big(\pi(\cdot\mid s)\,\big\|\,\pi(\cdot\mid s_e)\big)
=\tfrac12\big\|\operatorname{diag}(\sigma^{-1})\big(\mu(s_e)-\mu(s)\big)\big\|_2^2,
\]
with $\mu(s)=\mu^{(m)}\big(h^{(m)}(s)\big)$.
\end{proposition}

\begin{proof}
Write $\mu=\mu^{(m)}(h)$ and $J_\mu=J_\mu^{(m)}(h)$.
The log-density is
\[
\log\pi^{(m)}(a\mid h)
=
-\tfrac12\,(a-\mu)^{\top}\operatorname{diag}(\sigma^{-2})\,(a-\mu)
+c(\sigma),
\]
where $c(\sigma)$ collects the normalization and does not depend on $h$.
Differentiating with respect to $h$ gives the score
\[
\nabla_h\log\pi^{(m)}(a\mid h)
=
J_\mu^{\top}\operatorname{diag}(\sigma^{-2})\,(a-\mu).
\]
Under $a\sim\pi^{(m)}(\cdot\mid h)$ the residual has covariance
\[
\mathbb E\big[(a-\mu)(a-\mu)^{\top}\big]=\operatorname{diag}(\sigma^{2}),
\]
so the second moment of the score is
\[
F_h
=
J_\mu^{\top}\operatorname{diag}(\sigma^{-2})\,\operatorname{diag}(\sigma^{2})\,\operatorname{diag}(\sigma^{-2})\,J_\mu
=
J_\mu^{\top}\operatorname{diag}(\sigma^{-2})\,J_\mu.
\]
Inserting $\Delta_e^{(m)}(s)$ into $\Delta^\top F_h\Delta$ gives the expression for $Q_e^{(m)}(s)$.
For two Gaussians with the same covariance $\Sigma$ the trace and log-determinant terms of the KL cancel, leaving
\[
D_{\mathrm{KL}}\big(\mathcal N(\mu_1,\Sigma)\,\big\|\,\mathcal N(\mu_2,\Sigma)\big)
=
\tfrac12\,(\mu_2-\mu_1)^{\top}\Sigma^{-1}(\mu_2-\mu_1),
\]
and $\Sigma=\operatorname{diag}(\sigma^2)$ gives the last claim.
\end{proof}

Equation~(\ref{eq:fisher-gaussian}) is the Fisher of the policy itself, not an approximation, and it plays the role of the categorical closed form above.
Because the noise does not depend on the input, the KL between two states is exactly quadratic in the change of the mean, and the quadratic approximation differs from it only through the curvature of the mean along the displacement,
\[
\begin{aligned}
\tfrac12 Q_e^{(m)}(s)&=\tfrac12\big\|\operatorname{diag}(\sigma^{-1})\,J_\mu^{(m)}\Delta_e^{(m)}(s)\big\|_2^2,\\
\mu(s_e)-\mu(s)&=J_\mu^{(m)}\Delta_e^{(m)}(s)+O\big(\|\Delta_e^{(m)}(s)\|_2^2\big).
\end{aligned}
\]
For a Gaussian policy the ratio of the exact KL to $\tfrac12 Q$ can therefore depart from $1$ only where $\mu^{(m)}$ bends along the edit, never through the shape of the action distribution.

In our robot agents the mean comes from a head with one hidden layer of ReLU units, which reads $h_{\mathrm{fc}}$ together with an embedding $h_{\mathrm{state}}$ of the robot's body state (Appendix~\ref{sec:architecture}),
\[
\mu=W_2\,\mathrm{ReLU}(z)+b_2,
\qquad
z=W_1\big[h_{\mathrm{fc}};h_{\mathrm{state}}\big]+b_1.
\]
Write $W_1^{\mathrm{fc}}$ for the columns of $W_1$ that read $h_{\mathrm{fc}}$.
With $h_{\mathrm{state}}$ held at its base value, the Jacobian of the mean at the two sites is
\[
\begin{aligned}
J_\mu^{(\mathrm{fc})}
&=
W_2\operatorname{diag}\big(\mathbf 1\{z>0\}\big)\,W_1^{\mathrm{fc}},\\
J_\mu^{(\mathrm{pre})}
&=
J_\mu^{(\mathrm{fc})}\operatorname{diag}\big(\mathbf 1\{W_{\mathrm{fc}}h_{\mathrm{pre}}+b_{\mathrm{fc}}>0\}\big)\,W_{\mathrm{fc}},
\end{aligned}
\]
with the indicator taken entrywise.
Both depend on the state only through the two ReLU patterns.
An edit that leaves the body state unchanged leaves $h_{\mathrm{state}}$ unchanged as well, so its displacement lies in the image branch alone.
The factorization is specific to this head, whereas Equation~(\ref{eq:fisher-gaussian}) holds for any mean network as long as $\sigma$ is state independent.

\begin{proposition}[Fisher sensitivity at $h_{\mathrm{pre}}$ and $h_{\mathrm{fc}}$]
\label{prop:fisher-sites}
\label{app:fisher-sites}
Let
\[
J=\frac{\partial h_{\mathrm{fc}}}{\partial h_{\mathrm{pre}}}
=\operatorname{diag}\big(\mathbf 1\{W_{\mathrm{fc}}h_{\mathrm{pre}}+b_{\mathrm{fc}}>0\}\big)\,W_{\mathrm{fc}}.
\]
The policy depends on $h_{\mathrm{pre}}$ only through $h_{\mathrm{fc}}$, so the Fisher pulls back exactly,
\[
F_{\mathrm{pre}}=J^{\top}F_{\mathrm{fc}}\,J.
\]
If the displacements of an edit satisfy $\Delta_{\mathrm{fc}}=J\Delta_{\mathrm{pre}}$, then $Q_{\mathrm{pre}}=Q_{\mathrm{fc}}$.
\end{proposition}

\begin{proof}
By the chain rule $\nabla_{h_{\mathrm{pre}}}\log\pi=J^{\top}\nabla_{h_{\mathrm{fc}}}\log\pi$, and taking the second moment of both sides gives $F_{\mathrm{pre}}=J^{\top}F_{\mathrm{fc}}J$.
Then $Q_{\mathrm{pre}}=(J\Delta_{\mathrm{pre}})^{\top}F_{\mathrm{fc}}(J\Delta_{\mathrm{pre}})=\Delta_{\mathrm{fc}}^{\top}F_{\mathrm{fc}}\Delta_{\mathrm{fc}}=Q_{\mathrm{fc}}$.
\end{proof}

To first order the edit-induced displacements satisfy $\Delta_{\mathrm{fc}}\approx J\Delta_{\mathrm{pre}}$, and therefore
\[
Q_{\mathrm{pre}}
=\Delta_{\mathrm{pre}}^\top J^\top F_{\mathrm{fc}}J\,\Delta_{\mathrm{pre}}
\approx
\Delta_{\mathrm{fc}}^\top F_{\mathrm{fc}}\,\Delta_{\mathrm{fc}}
=Q_{\mathrm{fc}}.
\]
$Q$ is thus approximately the same at both sites by construction and not as a property of the data, and the remaining differences come from the nonlinearity of the fc layer over a finite edit.
The argument holds for every network in which $h_{\mathrm{fc}}$ is computed from $h_{\mathrm{pre}}$ alone, which includes the robot agents, since the body state joins only after $h_{\mathrm{fc}}$.

Proposition~\ref{prop:fisher-downstream} makes the interpretation of $Q$ precise.
Fisher sensitivity is a property of the representation \emph{as read by the trained downstream policy}.
Two agents can make the same information available in their internal representations while inducing very different Fisher geometries because their downstream computations use that information differently.
A small $Q_e^{(m)}$ therefore does not imply that information associated with edit $e$ is absent from the representation.

\paragraph{Separating displacement from directional sensitivity.}
The quadratic form $Q$ combines two effects.
A world edit can produce a large policy response because it moves the representation far, because the downstream policy is unusually sensitive to the direction of that movement, or both.
We separate these effects relative to the policy's average local Fisher sensitivity.

\begin{definition}[Fisher sensitivity decomposition]
\label{def:fisher-decomposition}
For a nonzero displacement $\Delta\in\mathbb R^d$, define its directional Fisher sensitivity
\[
S(\Delta)
=
\frac{\Delta^\top F_h\Delta}{\|\Delta\|_2^2}
=
\widehat\Delta^\top F_h\widehat\Delta,
\qquad
\widehat\Delta
=
\frac{\Delta}{\|\Delta\|_2}.
\]
For an isotropically sampled unit vector
$u\sim\mathrm{Unif}(\mathbb S^{d-1})$,
\[
\mathbb E_u
\left[
u^\top F_hu
\right]
=
\frac{\operatorname{tr}(F_h)}{d}.
\]
We therefore define the \emph{relative directional gain}
\[
G(\Delta)
=
\frac{
S(\Delta)
}{
\operatorname{tr}(F_h)/d
}
\]

The relative gain $G$ is defined whenever $\operatorname{tr}(F_h)>0$.
If the local Fisher vanishes identically, then $Q=0$ for every direction.

We also define the \emph{displacement in Fisher units}
\[
D_F(\Delta)
=
\|\Delta\|_2^2
\frac{
\operatorname{tr}(F_h)
}{
d
}.
\]
\end{definition}

The two quantities factor the Fisher sensitivity exactly:
\[
Q
=
\underbrace{G}_{\text{relative directional gain}}
\;
\underbrace{D_F}_{\text{displacement in Fisher units}}.
\]

$D_F$ measures how far the counterfactual moves the representation in units of the policy's average local sensitivity.
$G$ measures whether the direction of that movement is more or less policy-sensitive than an average direction in the same representation space.
Thus
\[
G>1
\]
means that the counterfactual moves along a direction that is privileged by the downstream policy relative to an isotropic direction.
In the experiments, we additionally compare this directional gain with matched-norm random directions to account for the non-isotropic spectrum of $F_h$.

\begin{proposition}[Rescaling invariance]
\label{prop:fisher-rescaling}
Consider a scalar reparameterization
\[
h'=ch
\]
together with a compensating downstream transformation that leaves the represented policy unchanged.
Then
\[
Q'=Q,
\qquad
G'=G,
\qquad
D_F'=D_F,
\]
whereas $\|\Delta\|_2$ and $S$ individually are not invariant.
\end{proposition}

\begin{proof}
Under the reparameterization,
\[
\Delta'=c\Delta.
\]
Because the downstream computation is compensated to preserve the policy, gradients with respect to the new representation scale as $c^{-1}$, giving
\[
F_{h'}
=
c^{-2}F_h.
\]
Therefore,
\[
Q'
=
(c\Delta)^\top
(c^{-2}F_h)
(c\Delta)
=
Q.
\]
Furthermore,
\[
S'=c^{-2}S,
\qquad
\frac{\operatorname{tr}(F_{h'})}{d}
=
c^{-2}
\frac{\operatorname{tr}(F_h)}{d},
\]
so
\[
G'=G.
\]
Finally,
\[
D_F'
=
\|c\Delta\|_2^2
\frac{\operatorname{tr}(c^{-2}F_h)}{d}
=
D_F.
\]
\end{proof}

This invariance matters when comparing different trained networks.
The scale of an internal representation is not itself identifiable, so neither the raw displacement norm nor its unnormalized directional sensitivity should be interpreted across agents in isolation.
$Q$, $G$, and $D_F$ remain unchanged under this elementary rescaling.

A change of site is not such a rescaling.
$G$ and $D_F$ depend on $\|\Delta\|_2$ and on $\operatorname{tr}(F_h)/d$, which change with the coordinates and the dimension of the site, so unlike $Q$ (Proposition~\ref{prop:fisher-sites}) they are not approximately preserved from $h_{\mathrm{pre}}$ to $h_{\mathrm{fc}}$.
A change in the split between $G$ and $D_F$ from one site to the other therefore describes how the fc layer re-expresses the same policy sensitivity, not where goal information is lost, and we compare $G$ and $D_F$ only between cohorts at the same site.

\begin{corollary}[Dependence on policy sharpness]
\label{cor:fisher-sharpness}
As the downstream action distribution approaches a point mass,
\[
\pi
\rightarrow
e_a,
\]
we have
\[
\operatorname{diag}(\pi)-\pi\pi^\top
\rightarrow
0.
\]
Consequently,
\[
F_h\rightarrow0
\]
and
\[
Q(\Delta)\rightarrow0
\]
for every fixed displacement $\Delta$.
\end{corollary}

Thus a more deterministic policy can exhibit smaller Fisher sensitivity along every representation direction, independently of the particular feature being studied.
We therefore report the average Fisher sensitivity
\[
\frac{\operatorname{tr}(F_h)}{d}
\]
per cohort next to $D_F$ (Table~\ref{tab:fisher-sharpness}) when comparing policies with different levels of sharpness.

For a Gaussian policy sharpness acts the other way.
By Proposition~\ref{prop:fisher-gaussian} the Fisher grows as $\sigma^{-2}$, so a sharper policy reads every direction more strongly and a smaller $\sigma$ raises $\operatorname{tr}(F_h)/d$ rather than lowering it.

\begin{definition}[Fisher cosine]
\label{def:fisher-cosine}
For two directions $\Delta_1$ and $\Delta_2$ satisfying
\[
\Delta_i^\top F_h\Delta_i>0,
\qquad i\in\{1,2\}.
\]
evaluated at the same base representation $h$ under the same local Fisher metric $F_h$, we define
\[
\cos_F(\Delta_1,\Delta_2)
=
\frac{
\Delta_1^\top F_h\Delta_2
}{
\sqrt{
(\Delta_1^\top F_h\Delta_1)
(\Delta_2^\top F_h\Delta_2)
}
}.
\]
\end{definition}

The Fisher cosine measures alignment specifically in directions to which the downstream policy is locally sensitive.
Values near $+1$ indicate aligned policy-relevant effects, values near $0$ indicate local Fisher orthogonality, and values near $-1$ indicate opposing policy-relevant effects.
We use this quantity later to characterize the geometry of cue competition in \textsc{Attribute Maze}.

\paragraph{Low-rank Fisher audit.}\label{app:fisher-rank-audit}
Because the activation-space Fisher matrix has rank at most $|\mathcal A|-1=14$ in this environment, we additionally test whether low Fisher rank alone can induce extreme pairwise cosine values.
Matched random $256$-dimensional vectors evaluated under rank-$14$ random Fisher metrics produce essentially no pairs with $|\cos_F|>0.9$, whereas $82$--$86\%$ of real state pairs exceed this threshold across the three GM policies on the black background.
The near-$\pm1$ Fisher structure therefore cannot be explained by Fisher rank alone.

\begin{definition}[Signed target-selection derivative]
\label{def:signed-target-selection-derivative}
Let
\[
A_\star(s),A_p(s)\subseteq\mathcal A
\]
denote the intended-goal- and proxy-consistent action sets for state $s$.
In \textsc{Maze} and \textsc{Attribute Maze}, both sets are taken from the base state:
$A_\star(s)$ is the first step toward the base cheese (respectively the line candidate)
and $A_p(s)$ the first step toward the top-right corner (respectively the yellow candidate).
In \textsc{Coin}, the base state holds the coin at a cell right of the agent, on the shortcut side, where the two sets would coincide,
and the edit moves it to the cell one step left of the agent.
So $A_\star(s)=A_{\mathrm{left}}$, the leftward actions toward the relocated coin, and $A_p(s)=A_{\mathrm{right}}$, the rightward shortcut.
At representation site $m$, define the target-selection contrast
\[
C_s^{(m)}(h)
=
\sum_{a\in A_\star(s)}
\pi^{(m)}(a\mid h)
-
\sum_{a\in A_p(s)}
\pi^{(m)}(a\mid h).
\]

For world edit $e$ with displacement
$\Delta_e^{(m)}(s)$,
the signed target-selection derivative is
\[
\partial_{\Delta_e}C_s^{(m)}
=
\nabla_h
C_s^{(m)}
\left(
h^{(m)}(s)
\right)^\top
\Delta_e^{(m)}(s).
\]
The action sets are determined from the environment state $s$ and are held fixed when differentiating with respect to $h$.
For a nonzero displacement, we additionally define the displacement-normalized derivative
\[
\widetilde{\partial_{\Delta_e}C_s^{(m)}}
=
\frac{
\partial_{\Delta_e}C_s^{(m)}
}{
\|\Delta_e^{(m)}(s)\|_2
}.
\]
\end{definition}

The raw derivative measures the first-order target-selective effect of the actual representation change induced by edit $e$.
Its sign indicates which side of the intended-versus-proxy contrast is favored, while its magnitude combines direction and edit size.
The normalized derivative removes the Euclidean magnitude of the displacement and therefore measures target-selective sensitivity per unit representation change.

Unlike the Fisher quadratic $Q$, the signed derivative is first-order and directional.
$Q$ asks how strongly the downstream action distribution changes along a world-induced representation direction.
$\partial_{\Delta_e}C_s^{(m)}$ asks whether that same direction moves the behaviorally relevant target-selection contrast toward the intended or proxy-consistent side. So, the two measurements are complementary: $Q$ measures whether the policy is sensitive to the world-induced displacement at all, while the signed derivative asks whether that sensitivity is organized along the behaviorally relevant target-selection contrast.

\paragraph{Input saliency.}
The same Fisher can be taken at the input instead of an internal site, which shows which parts of the observation the action distribution depends on.
Write $o$ for the observation, with $o_{c,x}$ the value of colour channel $c$ at pixel $x$, and treat $o$ as a site, so that $F^{(\mathrm{in})}(o)$ is the activation-space Fisher of Definition~\ref{def:activation-fisher} with $h=o$.

\begin{definition}[Policy-Fisher input saliency and enrichment]
\label{def:fisher-saliency}
The policy-Fisher input saliency of pixel $x$ is the diagonal of the input Fisher summed over the colour channels of $x$,
\[
S(x)
=
\sum_c F^{(\mathrm{in})}_{(c,x),(c,x)}(o)
=
\sum_c
\mathbb E_{a\sim\pi(\cdot\mid o)}
\left[
\left(
\frac{\partial\log\pi(a\mid o)}{\partial o_{c,x}}
\right)^{2}
\right].
\]
For a region $R$ of $|R|$ pixels in an image of $N$ pixels, the enrichment of $R$ is the saliency per pixel in $R$ relative to the saliency per pixel of the whole image,
\[
E_R(o)
=
\frac{N}{|R|}\,
\frac{\sum_{x\in R}S(x)}{\sum_{x}S(x)},
\]
and over states $o_1,\ldots,o_n$ we report its geometric mean
\[
E_R
=
\exp\!\Big(\frac1n\sum_{i=1}^{n}\log E_R(o_i)\Big).
\]
\end{definition}

By Proposition~\ref{prop:local-policy-geometry} at the input, changing channel $c$ of pixel $x$ by a small $\varepsilon$ moves the policy by $\tfrac12\varepsilon^2F^{(\mathrm{in})}_{(c,x),(c,x)}+O(\varepsilon^3)$ in KL, so $\tfrac12\varepsilon^2S(x)$ is this local KL summed over the channels of $x$.
For a softmax policy with logits $z$,
\[
\frac{\partial\log\pi(a\mid o)}{\partial o_{c,x}}
=
\frac{\partial z_a}{\partial o_{c,x}}
-\sum_b\pi(b\mid o)\,\frac{\partial z_b}{\partial o_{c,x}},
\]
and for a Gaussian policy Proposition~\ref{prop:fisher-gaussian} applies at the input with the body state held fixed, so $S$ has a closed form for both policy classes,
\[
S_{\mathrm{softmax}}(x)
=
\sum_c
\operatorname{Var}_{a\sim\pi(\cdot\mid o)}
\left[
\frac{\partial z_a}{\partial o_{c,x}}
\right],
\qquad
S_{\mathrm{Gauss}}(x)
=
\sum_c\sum_k
\frac{1}{\sigma_k^{2}}
\left(
\frac{\partial\mu_k}{\partial o_{c,x}}
\right)^{2}.
\]
The softmax saliency shrinks everywhere as the policy becomes nearly deterministic, and the Gaussian saliency grows everywhere as $\sigma$ shrinks (Corollary~\ref{cor:fisher-sharpness} and the remark after it).
Enrichment divides by the image mean of the same state and removes this global scale, so $\log_{10}E_R=0$ means that the region counts as much as an average pixel, whatever the sharpness of the policy.

\paragraph{How $S(x)$ differs from other saliency.}
\label{app:fisher-saliency}
Gradient saliency \citep{simonyan2014deep} differentiates a single output, one action logit or the value, and perturbation or occlusion saliency measures how that output changes when a patch of the image is replaced.
$S(x)$ is a Fisher-weighted variant of gradient saliency. It concerns the whole action distribution rather than one logit, it is invariant to a shift of all logits by a common amount, and it weights each action by its probability, so it measures what the policy actually reads out.
High-contrast sprites draw saliency in any network, so we read every region against randomly initialized networks, and since saliency decays with distance from the agent we compare cohorts at a fixed region.
Saliency maps suggest hypotheses rather than explain behavior \citep{atrey2020exploratory}. Our causal claims rest only on the world edits and the rollouts, never on the maps.

\subsection{Projections and Vector Analysis}
\label{app:projection-analysis}

The constructions below define projections relative to a chosen reference direction $u^{(m)}$.
They do not assume that one fixed reference direction provides a global description of the representation geometry across states.

\paragraph{Euclidean projection.}
Let $\Delta^{(m)}$ denote a representation displacement at site $m$, and let
$u^{(m)}\neq 0$ be a reference direction in the same representation space.
We write
\[
\hat u^{(m)}
=
\frac{u^{(m)}}{\|u^{(m)}\|_2}.
\]

The Euclidean projection of $\Delta^{(m)}$ onto the span of $u^{(m)}$ is
\[
\Delta_{u,E}^{(m)}
=
\left(
{\hat u^{(m)}}{}^{\top}
\Delta^{(m)}
\right)
\hat u^{(m)},
\]
with residual
\[
\Delta_{\perp,E}^{(m)}
=
\Delta^{(m)}
-
\Delta_{u,E}^{(m)}.
\]
By construction,
\[
{\Delta_{u,E}^{(m)}}^\top
\Delta_{\perp,E}^{(m)}
=
0.
\]

This decomposition measures how the representation displacement is oriented in the ordinary Euclidean geometry of activation space.

Although the Euclidean components are orthogonal in activation space, they are not generally orthogonal under the Fisher metric.
Consequently,
\[
Q^{(m)}\!\left(\Delta^{(m)}\right)
=
Q^{(m)}\!\left(\Delta_{u,E}^{(m)}\right)
+
Q^{(m)}\!\left(\Delta_{\perp,E}^{(m)}\right)
+
2
{\Delta_{u,E}^{(m)}}^\top
F^{(m)}(s)
\Delta_{\perp,E}^{(m)}.
\]
The cross term need not vanish because Euclidean orthogonality does not imply Fisher orthogonality.

\paragraph{Fisher-orthogonal projection.} To obtain an additive decomposition of the Fisher quadratic, we instead define the projection using the local Fisher geometry.
The downstream policy induces this local geometry through the activation-space Fisher matrix $F^{(m)}(s)$.
When
\[
{u^{(m)}}^\top
F^{(m)}(s)
u^{(m)}
>
0,
\]
we define the Fisher-orthogonal projection of $\Delta^{(m)}$ onto the span of $u^{(m)}$ as
\[
\Delta_{u,F}^{(m)}
=
\frac{
{u^{(m)}}^\top
F^{(m)}(s)
\Delta^{(m)}
}{
{u^{(m)}}^\top
F^{(m)}(s)
u^{(m)}
}
u^{(m)},
\]
with residual
\[
\Delta_{\perp,F}^{(m)}
=
\Delta^{(m)}
-
\Delta_{u,F}^{(m)}.
\]

The residual is orthogonal to the reference direction under the local Fisher metric,
\[
{u^{(m)}}^\top
F^{(m)}(s)
\Delta_{\perp,F}^{(m)}
=
0.
\]
Consequently, the Fisher quadratic decomposes exactly,
\[
Q^{(m)}
\left(
\Delta^{(m)}
\right)
=
Q^{(m)}
\left(
\Delta_{u,F}^{(m)}
\right)
+
Q^{(m)}
\left(
\Delta_{\perp,F}^{(m)}
\right),
\]
where, for any representation-space direction $v$,
\[
Q^{(m)}(v)
=
v^\top
F^{(m)}(s)
v.
\]

The Euclidean and Fisher projections answer different questions.
The Euclidean projection asks how much of the displacement lies along the reference direction in activation space.
The Fisher projection instead asks how the policy-sensitive component of that displacement is distributed relative to the same direction.

\paragraph{Signed-effect decomposition.}
At a fixed base state, the local signed target-selection derivative is linear in the representation displacement.
For the target-selection contrast $C_s^{(m)}(h)$,
\[
\partial_{\Delta}C_s^{(m)}
=
\nabla_h
C_s^{(m)}
\left(
h^{(m)}(s)
\right)^\top
\Delta^{(m)}.
\]
Therefore, for any additive decomposition
\[
\Delta^{(m)}
=
\Delta_1^{(m)}
+
\Delta_2^{(m)},
\]
we have
\[
\partial_{\Delta}C_s^{(m)}
=
\partial_{\Delta_1}C_s^{(m)}
+
\partial_{\Delta_2}C_s^{(m)}.
\]

This exact additivity applies to the local directional derivative.
It does not imply that the finite change
\[
C_s^{(m)}\!\left(h^{(m)}(s)+\Delta^{(m)}\right)
-
C_s^{(m)}\!\left(h^{(m)}(s)\right)
\]
decomposes additively for arbitrary finite displacements.
Unlike the quadratic Fisher decomposition, the derivative decomposition follows directly from linearity and therefore does not depend on how the additive split was constructed.

\paragraph{Local and shared-axis descriptions.}
The projection constructions above can be used to compare a fully state-specific
representation direction with a lower-dimensional shared description.

Let $d(s)$ denote a state-dependent representation direction of interest.
A fully local description retains this vector separately for every state.
A shared-axis description instead asks whether the state dependence can be
compressed to a fixed reference direction $u$ together with state-dependent
coefficients.

The simplest such approximation is
\[
d(s)
\approx
\alpha(s)u,
\]
where $\alpha(s)$ is a scalar that may vary across states.
If the orientation of the shared direction is itself determined by a discrete
state variable, we write
\[
d(s)
\approx
\sigma(s)\beta(s)u,
\qquad
\sigma(s)\in\{-1,+1\},
\qquad
\beta(s)\geq 0.
\]
Here, $\sigma(s)$ determines which orientation of the shared axis is active,
while $\beta(s)$ controls the local magnitude.

This form does not imply that the full representation geometry is
one-dimensional.
It only states that the component relevant to the particular displacement
family under study may be well described by one shared axis after accounting
for state-dependent sign and magnitude.
The residual
\[
r(s)
=
d(s)
-
\sigma(s)\beta(s)u
\]
contains the remaining state-specific structure.

\paragraph{Shared-axis coefficients under Euclidean and Fisher geometry.}
For a fixed shared direction $u$, the state-dependent coefficient can be
defined under different geometries.

In Euclidean activation space, the signed coordinate of $d(s)$ along the
unit direction
\[
\hat u
=
\frac{u}{\|u\|_2}
\]
is
\[
\alpha_E(s)
=
\hat u^\top d(s).
\]
The corresponding Euclidean projection is
\[
d_{u,E}(s)
=
\alpha_E(s)\hat u.
\]

Under the local Fisher geometry, the corresponding coefficient is
\[
\alpha_F(s)
=
\frac{
u^\top F(s)d(s)
}{
u^\top F(s)u
},
\]
giving the Fisher-orthogonal projection
\[
d_{u,F}(s)
=
\alpha_F(s)u.
\]

The two coefficients answer different questions.
$\alpha_E(s)$ measures how much of the displacement lies along $u$ in
ordinary activation space.
$\alpha_F(s)$ measures the coefficient of the same shared direction after
orthogonality is defined by local policy sensitivity.

The resulting residual
\[
d_{\perp,F}(s)
=
d(s)-d_{u,F}(s)
\]
satisfies
\[
u^\top F(s)d_{\perp,F}(s)=0,
\]
and therefore
\[
Q(d(s))
=
Q(d_{u,F}(s))
+
Q(d_{\perp,F}(s)).
\]

\paragraph{Interventions derived from these decompositions.}
The same distinction gives several intervention families.

A state-local intervention retains the complete direction,
\[
\delta_{\mathrm{local}}(s;\beta)
=
\beta\,d(s).
\]

A shared-axis intervention retains only a fixed direction and a
state-dependent sign and magnitude,
\[
\delta_{\mathrm{shared}}(s;\beta)
=
\beta\,
\sigma(s)\,
m(s)\,
\hat u,
\]
where $m(s)\geq 0$ specifies the chosen local magnitude.

Removing the sign dependence gives the corresponding unsigned intervention,
\[
\delta_{\mathrm{unsigned}}(s;\beta)
=
\beta\,
m(s)\,
\hat u.
\]

Finally, using the Fisher projection gives
\[
\delta_F(s;\beta)
=
\beta\,
\alpha_F(s)\,
u.
\]

These interventions separate different compression hypotheses.
The local intervention retains the full state-dependent representation
geometry.
The shared-axis intervention then asks whether its causal effect can be reproduced
using one common direction together with state-dependent gating.
The unsigned intervention tests whether that gating is necessary.
Finally, the Fisher-projected intervention retains only the component assigned to the
shared direction by the local policy-sensitive geometry.

\subsection{Environment-specific target geometry}
\label{sec:target-geometry}

On the training distribution the intended goal and the proxy are confounded.
At most states one and the same action moves the agent toward both, so the
policy's choice cannot tell us which of the two it is following. Behavior
separates goal from proxy only at states where the two require different first
actions, and finding these states requires knowing exactly which actions make
progress toward each target. Walls make the straight-line bearing of a target a
poor guide to this, so we define progress on the level's grid graph. The
resulting graph-exact action sets are used throughout the paper, both to choose
evaluation states (Section~\ref{app:banks}) and to score behavior on them.

This needs a local notion of progress toward a spatial target. The MDP state $s\in\mathcal S$ describes the complete
environment configuration, while navigation depends on the agent's location
within the spatial structure of the current level. We therefore introduce a
spatial projection of the environment state.

\paragraph{Spatial state projection.}
For a fixed level $\xi$, let
\[
\mathcal G_\xi=(V_\xi,E_\xi)
\]
denote its traversable grid graph. Each vertex $v\in V_\xi$ corresponds to a
traversable cell, and an edge $(u,v)\in E_\xi$ indicates that the agent can
move between the corresponding cells in one environment step.

We write
\[
x_\xi:\mathcal S\rightarrow V_\xi
\]
for the spatial projection that maps a full environment state $s$ to the
cell occupied by the agent. Thus $s$ retains the complete MDP state, while
$x_\xi(s)$ extracts only the agent's spatial location on level $\xi$.

For cells $u,v\in V_\xi$, we define the graph distance
\[
d_\xi(u,v)
=
\min
\left\{
k\in\mathbb N_0 :
\exists\, v_0,\ldots,v_k\in V_\xi,\;
v_0=u,\;
v_k=v,\;
(v_{i-1},v_i)\in E_\xi
\ \forall i=1,\ldots,k
\right\}.
\]
Hence $d_\xi(u,v)$ is the minimum number of valid grid transitions required
to move from $u$ to $v$. We compute this distance exactly using
breadth-first search.

\paragraph{Spatial state-transition projection.}
The MDP transition kernel acts on complete environment states,
\[
s' \sim P(\cdot\mid s,a).
\]
For the navigation environments considered here, the one-step spatial
transition is deterministic conditional on the full state $s$ and action
$a$. We therefore define
\[
x'_\xi(s,a)
=
x_\xi(s'),
\qquad
\text{where }
P(s'\mid s,a)=1,
\]
as the agent cell after applying action $a$ in state $s$ on level $\xi$.
Thus $x_\xi(s)$ and $x'_\xi(s,a)$ are respectively the spatial projections
before and after the MDP transition. This notation introduces no additional
transition model beyond $P$.

\begin{definition}[Graph-exact target-progress action set]
\label{def:target-progress-actions}
For a state $s$, level $\xi$, and candidate target cell $v\in V_\xi$, define
\[
A_v(s)
=
\left\{
a\in\mathcal A :
d_\xi\!\left(x'_\xi(s,a),v\right)
<
d_\xi\!\left(x_\xi(s),v\right)
\right\}
\subseteq \mathcal A.
\]
$A_v(s)$ is therefore the subset of actions whose induced spatial transition
strictly decreases the remaining shortest-path distance to $v$.
\end{definition}

Because the grid edges have unit cost, an action belongs to $A_v(s)$ exactly
when its induced movement lies on a shortest path from the current agent cell
toward $v$. An action from which $v$ remains eventually reachable need not
belong to $A_v(s)$ if it does not decrease the remaining shortest-path
distance on that step.

When multiple raw policy actions induce the same effective movement, we
aggregate their probability mass rather than treating action aliases as
distinct navigation choices.

Two candidate cells $v_1,v_2\in V_\xi$ are locally action-discriminating at
state $s$ when
\[
A_{v_1}(s)\neq A_{v_2}(s).
\]
At such states, the two spatial target hypotheses make different predictions
about immediate control and can therefore be distinguished from the native
policy's action distribution.

\paragraph{Environment-specific target action sets.}
In \textsc{Maze}, let $v_\star$ denote the cheese cell and $v_p$ the
proxy-associated top-right destination. We write
\[
A_\star(s)
\equiv
A_{v_\star}(s),
\qquad
A_p(s)
\equiv
A_{v_p}(s).
\]
Thus $A_\star(s)$ contains graph-exact progress actions toward the intended
goal location, while $A_p(s)$ contains progress actions toward the
proxy-associated destination.

In \textsc{Attribute Maze}, let $v_{\mathsf{line}}$ and $v_{\mathsf{yellow}}$ denote the locations of the
line-bearing and yellow-bearing objects. Depending on the world edit
under consideration, these locations instantiate the intended and proxy
candidates. We again use
\[
A_\star(s)
\qquad\text{and}\qquad
A_p(s)
\]
as shorthand for the corresponding graph-exact action sets, while retaining
$A_{v_{\mathsf{line}}}(s)$ and $A_{v_{\mathsf{yellow}}}(s)$ when the physical cue identity matters.

\textsc{Coin} does not use the grid-graph construction for its proxy.
The learned shortcut is represented by the right-moving action family, and
in every state of the \textsc{Coin} conflict bank the coin sits one cell to
the left of the agent in the state on which target selection is read (Appendix~\ref{app:banks}).
Progress toward the coin is therefore leftward motion, and we use the two
horizontal movement families
\[
A_{\mathrm{left}},\;A_{\mathrm{right}}
\subseteq
\mathcal A_{\mathrm{Coin}},
\]
each of which holds the three actions that move the agent in that direction.
For notational consistency in comparisons across environments, we write
\[
A_\star(s)
\equiv
A_{\mathrm{left}},
\qquad
A_p(s)
\equiv
A_{\mathrm{right}}.
\]
A state is locally action-discriminating whenever
\[
A_\star(s)\neq A_p(s).
\]

\paragraph{Action-discriminating evaluation states.}
For a level $\xi$, let $\mathcal S_\xi\subseteq\mathcal S$ denote the environment states belonging to that level.
For two candidate action sets $A_1(s)$ and $A_2(s)$, we call a state action-discriminating when
\[
A_1(s)\neq A_2(s).
\]
For the intended and proxy-associated targets, we therefore define
\[
\mathcal S_{\mathrm{conf}}(\xi)
=
\left\{
s\in\mathcal S_\xi:
A_\star(s)\neq A_p(s)
\right\}.
\]
These states are decision points at which progress toward the intended and proxy-associated targets implies different immediate action sets.
Restricting evaluation to $\mathcal S_{\mathrm{conf}}(\xi)$ therefore lets the observed policy action distinguish which target better organizes local behavior.
We call the members of $\mathcal S_{\mathrm{conf}}(\xi)$ \emph{conflict states}.

\paragraph{Strict conflict.}
A conflict state can still have progress sets that overlap, and an action in the overlap serves both targets and so reveals nothing about which one the policy follows.
We therefore build every evaluation bank from states in \emph{strict conflict},
\[
A_\star(s)\cap A_p(s)=\emptyset,
\]
at which every progress action serves exactly one of the two targets.
In \textsc{Maze} shortest paths are unique, so both sets contain a single action and every conflict state is a strict conflict.
In \textsc{Coin} the left-moving and right-moving families are disjoint by construction.
In \textsc{Attribute Maze} we additionally require the two candidates to be equidistant from the agent, so that neither target is favoured by being closer.

\paragraph{Continuous actions.}
The robot policies are Gaussian over actions $a\in\mathbb R^{m}$, $\pi(\cdot\mid s)=\mathcal N\big(\mu(s),\operatorname{diag}(\sigma^2)\big)$, with a state-independent standard deviation $\sigma\in\mathbb R^{m}_{>0}$. The first $k$ components of an action move the robot's effector, the gripper or the body, in the plane in which the task's objects lie.

\begin{definition}[Planned step and target direction]
\label{def:planned-step}
Let $M(s)\in\mathbb R^{2\times k}$ map the first $k$ action components at state $s$ to the motion of the effector in the task plane. The \emph{planned step} is the motion commanded by the mean action,
\[
\mathrm{step}(s)=M(s)\,\mu_{1:k}(s)\in\mathbb R^{2}.
\]
For an object $o$ with target point $y_o(s)$ in the task plane and the effector at $x(s)$, the \emph{target direction} is the unit vector $\hat t_o(s)=\big(y_o(s)-x(s)\big)/\|y_o(s)-x(s)\|_2$.
\end{definition}

The planned step is taken on the unclipped mean, so it can exceed what the robot executes in one control step.
The three kinds of robot task instantiate $M(s)$ and $y_o(s)$ as follows.

\paragraph{Quadruped.}
In \quadname{} the action is a walking command with $m=3$ components, a forward and a sideways velocity in the robot's own frame and a turn rate. The first $k=2$ components already describe the motion in the floor plane, so $M(s)=I_2$ and the planned step is $(\mu_1(s),\mu_2(s))$, in command units between $-1$ and $1$. The target point of the flag or the tower is the object's position in the robot frame.

\paragraph{Arm.}
In the arm tasks the action has $m=8$ components. The first $k=7$ change the seven joint angles of the arm by at most $0.1$\,rad each, and the eighth opens or closes the gripper.
How far the gripper moves when the joints turn depends on the arm's posture. The Jacobian $J(q)\in\mathbb R^{3\times7}$ at joint angles $q$ gives, for each joint, the velocity of the tool centre point per unit change of that joint. The tool centre point is the point between the two gripper fingers, the point that grasps or pushes and that the arm's controller positions. With $P\in\mathbb R^{2\times3}$ keeping the two components in the table plane,
\[
M(s)=0.1\,P\,J(q),
\]
so the planned step is the table-plane motion of the tool centre point that the mean joint change produces, in millimetres. In \textsc{LocationReach} and \textsc{AttributeTable} the target point of an object is its centre.

\paragraph{Push tasks.}
In \textsc{PushToMarker} and \textsc{SortZones} the gripper does not move toward the marker or pad $g$ but toward the place from which a push carries the cube onto it. With the cube centre at $c$, the target point of $g$ is the push point
\[
y_g=c-0.03\,\frac{g-c}{\|g-c\|_2},
\]
on the line from $g$ through the cube, $3$\,cm behind the cube centre. The cube's half-width is $2$\,cm, so this point lies $1$\,cm behind its rear face, where the gripper must stand to push the cube straight toward $g$. The dense reward of these tasks uses the same point (Appendix~\ref{app:env-arm}).

\clearpage
\section{Experimental design and extended results}
\label{sec:experiments-app}

This appendix extends every analysis of Section~\ref{sec:goal} to all eight environments of Appendix~\ref{sec:environments}.
Figure~\ref{fig:app-overview-all8} summarizes the result.
In each environment the true goal survives to a different depth of the agent, and this depth goes together with the intervention that recovers it.
The failures fall into the three types of the main text.
Target substitution appears in three environments, \textsc{Maze}, \textsc{LocationReach} and \textsc{PushToMarker}, where a location-based proxy dominates and fully substitutes the target, even though the goal object is present in the scene.
In \textsc{Maze} the substitute target is a spatial relation, the top-right extreme, and in \textsc{LocationReach} and \textsc{PushToMarker} a fixed place in the camera image.
The policy shortcut appears in \textsc{Coin}, where a very simple part of the policy, moving right, stands in for the goal object.
Cue competition appears, in different grades, in four environments.
In \textsc{Attribute Maze}, \textsc{AttributeTable} and \textsc{SortZones} it coincides with one object that carries two co-located characteristics, and \quadname{} replicates it with two different objects that stand closely together in space.
In all four we see a competition between the two characteristics, or between the two closely related objects.

To establish this, we walk through the agents with a ladder of questions at increasing depth.
We start in the representation and ask whether information about the goal is still there, move to the readout and ask whether the policy reacts to it and uses it, and then look at what organizes behavior instead and how the proxy is incorporated into the agent, from the outside and from the inside.
The ladder ends with recovery, where we ask whether goal-directed behavior can be restored without further reinforcement learning, by steering the existing policy or by learning a new readout from an earlier representation.

\begin{figure}[h]
\centering
\includegraphics[width=\linewidth]{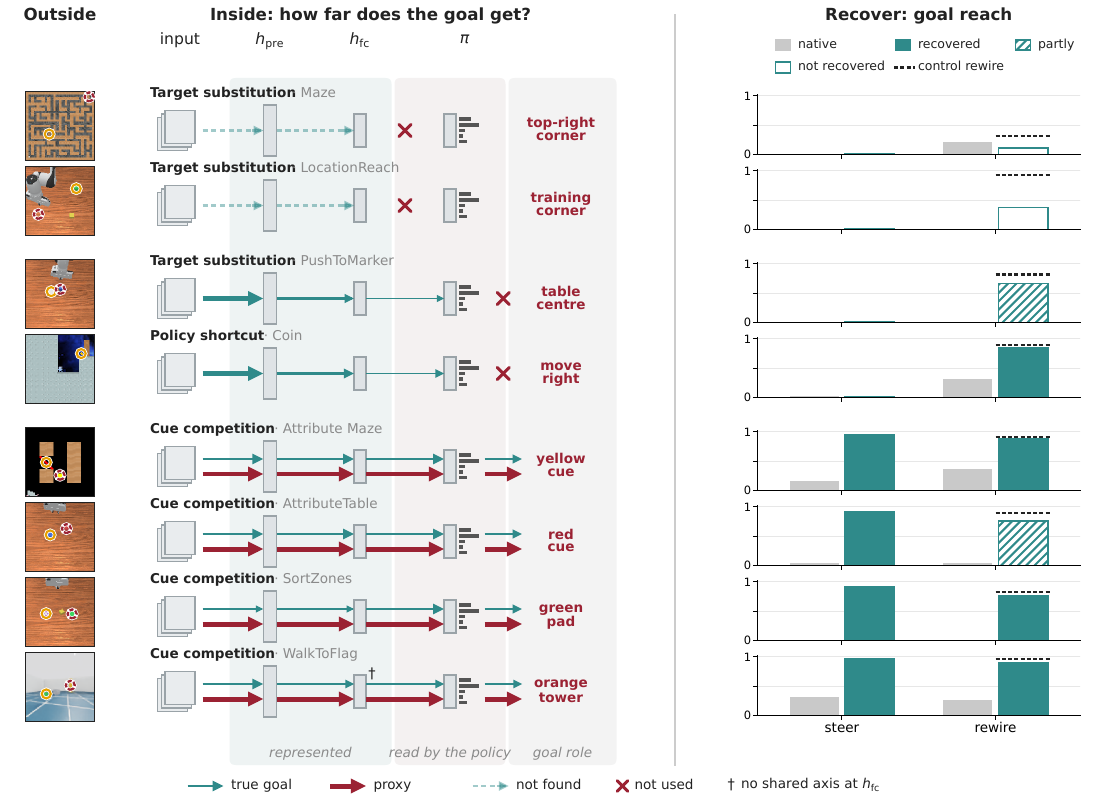}
\caption{
\textbf{How far does the goal get in each of the eight environments?}
The counterpart of Figure~\ref{fig:overview} for all environments, rows ordered from the shallowest to the deepest point the goal reaches. Arrow width is schematic.
Steering is the corrective push at $\beta=1$ (Table~\ref{tab:steering-local}) and rewiring the teacher-target readout from $h_{\mathrm{pre}}$ (Table~\ref{tab:rewire-main}), with oracle targets in \textsc{Maze} and \textsc{Attribute Maze} as in Figure~\ref{fig:overview} and for \textsc{PushToMarker} after three rounds of DAgger (Table~\ref{tab:maze-dagger}). Filled bars are recovered and hatched bars partial under the success criterion of Appendix~\ref{sec:rewiring}.
}
\label{fig:app-overview-all8}
\end{figure}

\subsection{Linear decodability: Is information about the intended goal still there?}
\label{app:probes}

\paragraph{Motivation.}
Goal misgeneralization tells us that an agent behaves according to something other than the intended goal, but this does not tell us what happened to information about the intended goal inside the agent.
One simple possibility is that the relevant goal information is no longer available in the learned representation.
We therefore begin with the most basic internal question: \emph{is information about the intended goal still there?}

We test this first through linear decodability.
A successful probe gives an existence result: the corresponding information is available in a form accessible to a linear readout.
We then ask where this information appears across the network, whether its accessibility differs between GM and Non-GM agents, and how much of that accessibility can be attributed to training rather than to structure already exposed by the input or architecture.

\paragraph{Design.}
For each environment, we define goal-related and, where meaningful, proxy-related target variables and fit regularized linear probes \citep{alain2017probes} to internal activations.
The probe targets are environment-specific because the information relevant to the goal differs across tasks.
For a binary target \(y\), the probe at representation \(m\) has score
\[
f_m(s)
=
w_m^\top h^{(m)}(s)+b_m,
\]
and we evaluate \(f_m\) using held-out ROC--AUC.
For continuous spatial targets, we fit the corresponding linear regressor
\[
\hat y_m(s)
=
w_m^\top h^{(m)}(s)+b_m
\]
and evaluate it using held-out \(R^2\).
Tables~\ref{tab:probe-banks} and~\ref{tab:robot-probe-banks} list every probe with its label, bank, split and sites.

Following the architectures in Appendix~\ref{sec:architecture}, we probe the same computational sites across the three Procgen environments,
\[
h^{(1)} = H^{(1)},
\qquad
h^{(2)} = H^{(2)},
\qquad
h^{(3)} = H^{(3)},
\]
where \(H^{(1)},H^{(2)},H^{(3)}\) denote the corresponding residual-block representations.
No additional ReLU is inserted after the first two blocks because no such operation occurs in the policy forward pass.
The subsequent shared representations are
\[
h_{\mathrm{pre}}
=
\operatorname{vec}
\left(
\operatorname{ReLU}(h^{(3)})
\right)
\]
and
\[
h_{\mathrm{fc}}
=
\operatorname{ReLU}
\left(
W_{\mathrm{fc}}h_{\mathrm{pre}}+b_{\mathrm{fc}}
\right).
\]
Thus \(h_{\mathrm{pre}}\) is the actual input to the fully connected layer and \(h_{\mathrm{fc}}\) is the final shared representation consumed by the policy and value readouts.
For \(h^{(1)},h^{(2)},h^{(3)}\), we globally average pool the convolutional feature maps before fitting the probe.
The later representations are used directly.
We exclude policy logits and value predictions from this sequence because they are separate readout branches rather than later shared representations.
In the four arm tasks we probe the unpooled feature maps \(H^{(1)}\) and \(H^{(2)}\) of the first two convolutions, then \(h_{\mathrm{pre}}\) and \(h_{\mathrm{fc}}\).
In \quadname{} we probe the three convolutional maps average-pooled to \(4{\times}4\) cells, \(\bar H^{(1)},\bar H^{(2)},\bar H^{(3)}\), then \(h_{\mathrm{pre}}\) and \(h_{\mathrm{fc}}\).
We do not pool the robot maps globally because pooling removes position: globally pooled \(H^{(1)}\) and \(H^{(2)}\) keep \(R^2<0.17\) of the marker position in \textsc{PushToMarker}, against \(0.74\) to \(0.90\) unpooled.

Every probe is fitted on a training partition and scored on held-out data: whole held-out levels in \textsc{Maze} and \textsc{Coin} over five split seeds, held-out object configurations in \textsc{Attribute Maze}, and held-out start frames in the robot tasks.
The Procgen probes are ridge regressions with fixed strength \(\alpha=1\) (logistic regression for \textsc{Coin} presence), and the robot probes are ridge regressions whose strength is selected by cross-validation on the training frames.
All probe preprocessing is fitted on the training partition only.

Our primary quantity is absolute held-out probe performance.
This answers the direct representational question: how linearly accessible is the target at each site?
We note that absolute decodability alone, however, does not establish how much accessibility was produced by training.
Simple visual or spatial structure can already be linearly exposed by an untrained convolutional network \citep{hermann2020shapes}.

\paragraph{Baselines.}
We compare every probe with three baselines.
First, randomly initialized networks of the same architecture (three draws in every environment), probed on the same frames.
Second, input-level or positional control probes, which probe the observation or a correlated state variable directly, for example the pixels of the cheese cell, the agent's position or block-averaged pixels.
They show how much information is trivially available in the input or carried by co-located objects and therefore need not be learned.
Third, in the Procgen environments, permuted labels give an empirical null with the same grouping and splits as the probe.
A score that an untrained network or an input control already reaches is not read as a training effect, so each result in the following first asks whether the probe separates from the baselines and only then compares GM with Non-GM agents.

\paragraph{\textsc{Maze} probe construction.}
In \textsc{Maze}, we first ask whether the cheese object itself is linearly detectable, using the probe bank of Appendix~\ref{app:banks-probe} (Table~\ref{tab:probe-banks}).
Standard \textsc{Maze} states always contain the cheese, so the negative class consists of the same states with the cheese removed.
We also probe the bearing of the cheese relative to the agent and, for the proxy, the bearing of the reachable top-right extreme of the maze.

\begin{figure*}[t]
    \centering
    \includegraphics[width=\textwidth]{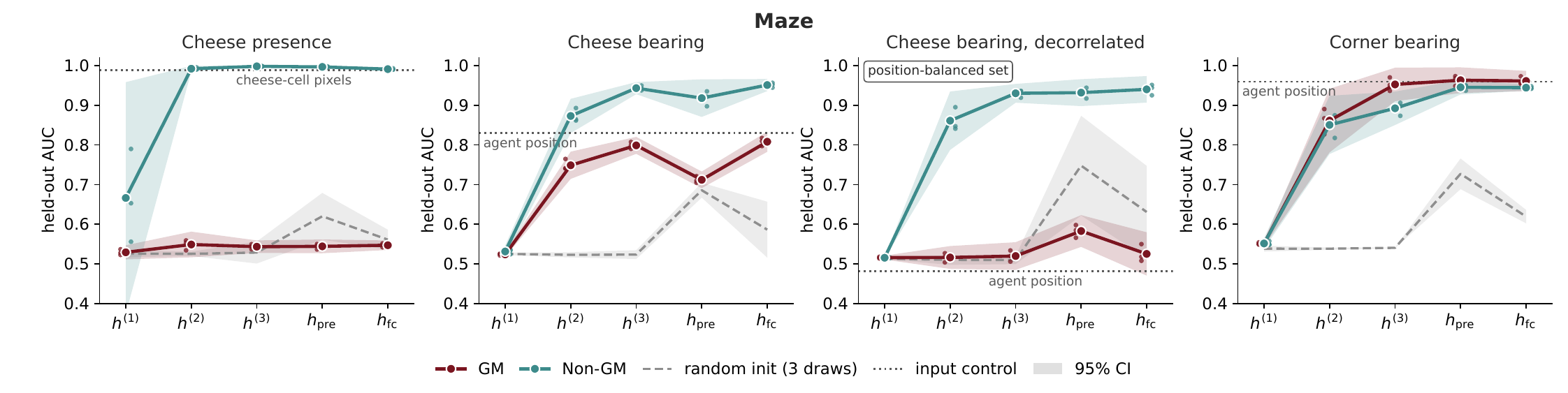}
    \caption{
    \textbf{Layerwise linear availability in \textsc{Maze}.}
    Absolute held-out ROC--AUC, from left to right, for cheese presence, cheese bearing relative to the agent, the same cheese bearing on the position-balanced set, in which agent position carries no information about the cheese bearing, and bearing to the reachable top-right corner.
    Points show the three checkpoint seeds, lines show cohort means, and shaded regions show 95\% $t$-intervals across the three checkpoints.
    Grey dashed: three randomly initialized networks, with a 95\% \(t\)-interval across draws. Dotted: input control, the cheese-cell pixels for presence and agent position for all bearings.
    }
    \label{fig:maze-layerwise-probes}
\end{figure*}

\paragraph{\textsc{Maze} results.}
Cheese presence is trivially decodable from the input, as a linear probe on the pixels of the cheese cell reaches an AUC of \(0.988\) (Figure~\ref{fig:maze-layerwise-probes}).
Moreover, Non-GM agents decode it just as well from \(h^{(2)}\) onward (\(0.991\) to \(0.998\)), while GM agents stay at the level of random networks in the convolutional blocks and even below them at the later layers \(h_{\mathrm{pre}}\) and \(h_{\mathrm{fc}}\) (\(0.529\) to \(0.549\) at every site, against \(0.620\) and \(0.561\) for random networks at the two later layers).
The true goal is therefore far more linearly decodable in Non-GM than in GM agents, a separation in the expected direction.
Because the presence of the cheese is already available in the input, however, a presence probe alone may not separate the cohorts cleanly, so we also probe the bearing of the cheese.

Cheese bearing separates from random networks in both cohorts from \(h^{(2)}\) onward, in GM only narrowly at \(h_{\mathrm{pre}}\) (\(0.712\) against \(0.686\)), and grows with depth, from chance at \(h^{(1)}\) to about \(0.8\) in GM and \(0.95\) in Non-GM agents.
The bearing probe thus separates, but we are unsure whether a subsidiary piece of information drives it, namely the agent's own position.
In GM agents the bearing stays close to and below what agent position alone gives (\(0.712\) to \(0.808\) from \(h^{(2)}\) onward, against \(0.830\)), while Non-GM agents exceed it (\(0.873\) to \(0.951\)).
With the corner bearing held fixed, so that only cheese-relative information helps, GM agents fall to \(0.598\) at \(h_{\mathrm{pre}}\), below both agent position (\(0.703\)) and random networks (\(0.729\)), while Non-GM agents keep \(0.930\) (Table~\ref{tab:probe-controls}).
We test this directly with a decorrelation check, on a set in which the agent's own position carries no information about the cheese bearing.
To build it, we keep inside every agent cell as many frames with the cheese to the left of the agent as to its right, which leaves \(902\) of the \(1{,}987\) cheese frames and brings the agent-position probe to chance (\(0.481\), Figure~\ref{fig:maze-layerwise-probes}, right).
On this set the cheese bearing in GM agents collapses at every site (\(0.516\) to \(0.583\)) and falls below random networks at \(h_{\mathrm{pre}}\) and \(h_{\mathrm{fc}}\) (\(0.748\) and \(0.631\)), while Non-GM agents keep it from \(h^{(2)}\) onward (\(0.861\) to \(0.940\)), and so does the pixel probe (\(0.895\)).
This rules out that GM agents carry cheese-bearing information beyond their own position.
The GM bearing probe only tracks the agent, while Non-GM agents and pixels keep the cheese bearing.

And lastly, corner bearing, the proxy probe, is decoded well by both cohorts and far above random networks from \(h^{(2)}\) onward, with GM agents slightly ahead (\(0.952\) against \(0.892\) at \(h^{(3)}\), \(0.963\) against \(0.945\) at \(h_{\mathrm{pre}}\)).
Agent position alone already predicts the corner bearing (AUC \(0.959\)), so neither cohort carries corner information beyond the geometry of the sampled states.
Across the two bearing probes, GM agents are somewhat worse than Non-GM agents for the true goal and slightly better for the proxy, a small effect that depends on the probe and the site.

Together we find that the cheese is trivially detectable from the observation, but GM training does not construct a strong cheese presence representation as seen in Non-GM agents.
Cheese bearing information in GM agents is also largely accounted for by the agent position, whereas Non-GM agents contain substantially stronger goal-related spatial information.
The \textsc{Maze} failure is therefore compatible with a learned representation that preserves generic spatial structure while failing to construct the goal-specific features that characterize the Non-GM policy.

\paragraph{\textsc{Coin} probe construction.}
In \textsc{Coin}, we separate object availability from spatial information about the goal, using the probe bank of Appendix~\ref{app:banks-probe} (Table~\ref{tab:probe-banks}).
Unlike cheese presence in \textsc{Maze}, coin presence varies naturally during ordinary \textsc{Coin} trajectories as the coin moves into and out of the visible frame.
Coin bearing is residualized on level progress because horizontal progress is strongly coupled to both observation structure and the training-time rightward trajectory.
For the continuous bearing target, the level-permutation null yields strongly negative held-out \(R^2\) because permuted labels are constant within levels, so we use \(R^2=0\) as the chance reference.

\begin{figure*}[t]
    \centering
    \includegraphics[width=\textwidth]{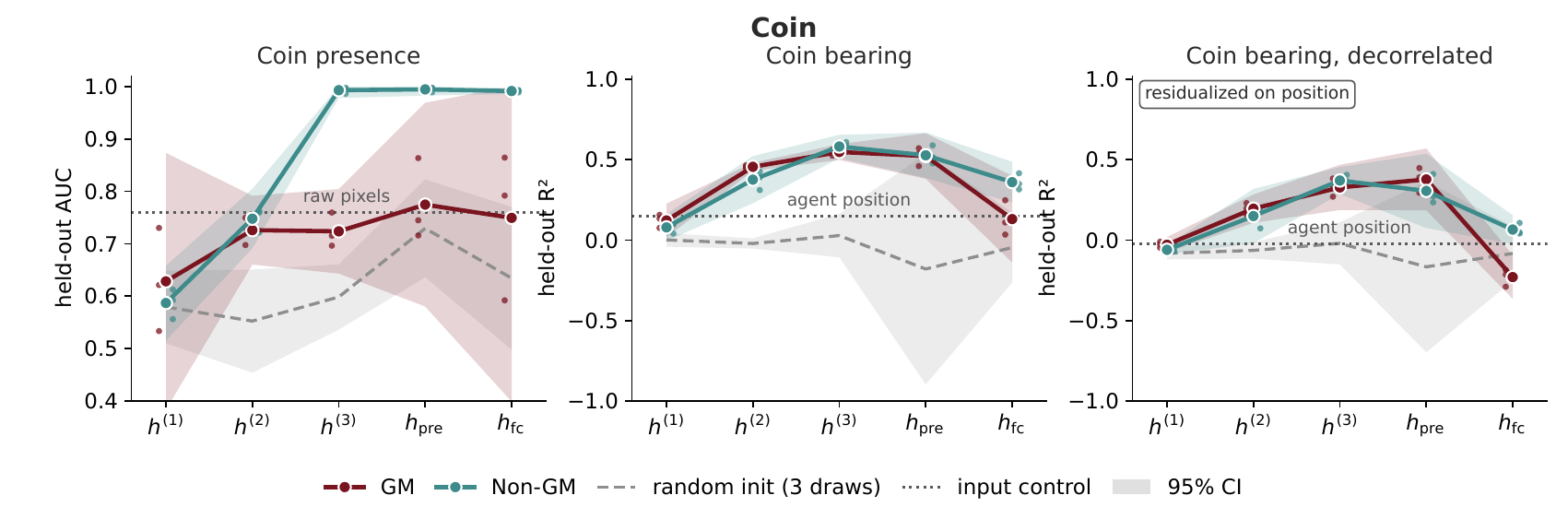}
    \caption{
    \textbf{Layerwise linear availability in \textsc{Coin}.}
    Absolute held-out probe performance for coin presence (ROC--AUC) and residualized coin bearing (\(R^2\)).
    Points show the three checkpoint seeds, lines show cohort means, and shaded regions show 95\% $t$-intervals across the three checkpoints.
    Grey dashed: three randomly initialized networks, with a 95\% \(t\)-interval across draws. Dotted: input control, raw pixels for presence and agent position for bearing (raw pixels give \(R^2=-0.552\) for bearing).
    Right: coin bearing residualized on agent position as well as on progress, so that agent position decodes none of it.
    }
    \label{fig:coin-layerwise-probes}
\end{figure*}

\paragraph{\textsc{Coin} results.}
The coin presence probe separates from the baselines only in Non-GM agents (Figure~\ref{fig:coin-layerwise-probes}).
A probe on the raw pixels of the coin already reaches \(0.760\), and GM agents stay at that level at every site from \(h^{(2)}\) onward (\(0.724\) to \(0.775\)).
They are above random networks in the convolutional blocks, but do not consistently exceed them at \(h_{\mathrm{pre}}\) and \(h_{\mathrm{fc}}\).
Non-GM agents start at the same level (\(0.748\) at \(h^{(2)}\)) and approach ceiling from \(h^{(3)}\) onward (\(0.991\) to \(0.995\)).
This separation holds on both frame sources of the probe bank.
The robust \textsc{Coin} result is therefore not that the coin becomes undecodable in GM agents, but that the goal is cleanly represented in Non-GM agents, whose training constructs a much stronger downstream coin presence representation than GM training does.

Coin bearing, by contrast, separates from every baseline in both cohorts.
From \(h^{(2)}\) to \(h_{\mathrm{pre}}\), held-out \(R^2\) is \(0.376\) to \(0.582\), against at most \(0.028\) for random networks, \(0.152\) for agent position and \(-0.552\) for raw pixels, and the two cohorts do not differ (\(0.523\) and \(0.527\) at \(h_{\mathrm{pre}}\)).
At \(h_{\mathrm{fc}}\) the cohort ordering depends on which half of the bank is evaluated (\(0.131\) for GM against \(0.360\) for Non-GM on the full bank, reversed on the later-stage half), so we do not interpret \textsc{Coin} at \(h_{\mathrm{fc}}\).
The bearing probe separates, but as in \textsc{Maze} we are unsure whether the agent's own position drives it, so we run the same decorrelation check.
With \(320\) frames we do not subsample but residualize the bearing on agent position as well as on progress, fitted on the training frames, so that a probe on agent position decodes none of it (\(R^2=-0.024\), Figure~\ref{fig:coin-layerwise-probes}, right).
Coin bearing stays decodable in both cohorts beyond agent position from \(h^{(2)}\) to \(h_{\mathrm{pre}}\) (\(0.150\) to \(0.378\)), against at most \(-0.020\) for random networks and \(-0.369\) for pixels, and the same holds with position entered as binned indicators (\(0.341\) and \(0.219\) at \(h_{\mathrm{pre}}\) for GM and Non-GM).
The values are lower than without the position term (\(0.523\) and \(0.527\) at \(h_{\mathrm{pre}}\)), so part of the original signal was position, but unlike in \textsc{Maze} the goal bearing in GM agents does not reduce to it.
This separates \textsc{Coin} from \textsc{Maze}.

\textsc{Coin} consequently provides a heterogeneous result across goal variables.
Goal bearing is strongly accessible in both cohorts through \(h_{\mathrm{pre}}\), with no stable cohort separation there.
Coin presence, however, becomes nearly perfectly accessible in Non-GM agents while remaining substantially weaker in GM agents at the downstream shared representations.
The robust representational distinction in \textsc{Coin} is therefore the organization of goal-object presence rather than a stable difference in goal bearing.
\citet{GMhelp} suggest that a CoinRun agent trained with the coin at the right end does not represent the coin, based on an anomaly detector and a learned help-request policy applied to its penultimate-layer features.
Their analysis differs from ours in method and site, since neither tool measures supervised linear decodability, and the penultimate layer corresponds to \(h_{\mathrm{fc}}\) in our architecture, where GM coin presence is near the raw-pixel level.

\paragraph{\textsc{Attribute Maze} probe construction.}
\textsc{Attribute Maze} requires a different decomposition because the intended and proxy cues are attributes of objects rather than separate spatial targets.
Simply detecting a feature does not establish whether the representation associates that feature with the rewarded object.
We therefore distinguish feature presence from goal role and probe four binary concepts,
\[
L_{\mathrm{pres}},
\qquad
Y_{\mathrm{pres}},
\qquad
L_{\mathrm{role}},
\qquad
Y_{\mathrm{role}},
\]
where \(L_{\mathrm{pres}}\) and \(Y_{\mathrm{pres}}\) indicate whether the line and yellow attributes are present, and \(L_{\mathrm{role}}\) and \(Y_{\mathrm{role}}\) indicate whether the corresponding attribute belongs to the rewarded object.
The first pair measures feature availability.
The second pair measures whether feature identity is available in relation to the rewarded role.

To separate presence from role, we use the configuration grid of the probe bank (Appendix~\ref{app:banks-probe}, Table~\ref{tab:probe-banks}) rather than sampling only from the ordinary task distribution.
The grid includes states in which an attribute is visible but belongs to the lure, allowing ``feature present'' and ``feature belongs to the rewarded object'' to disagree, and both candidate objects are independently placed on uniformly sampled free cells.
For the line attribute, presence and rewarded role disagree in \(4/12\) training configurations and \(2/6\) held-out configurations.
The yellow distinction has lower support, with \(Y_{\mathrm{pres}}\) and \(Y_{\mathrm{role}}\) differing in only one configuration in each split.
We therefore interpret the yellow-role probe more cautiously and do not assume that the two role probes provide equally strong separation of feature identity from rewarded role.

\begin{figure*}[t]
    \centering
    \includegraphics[width=\textwidth]{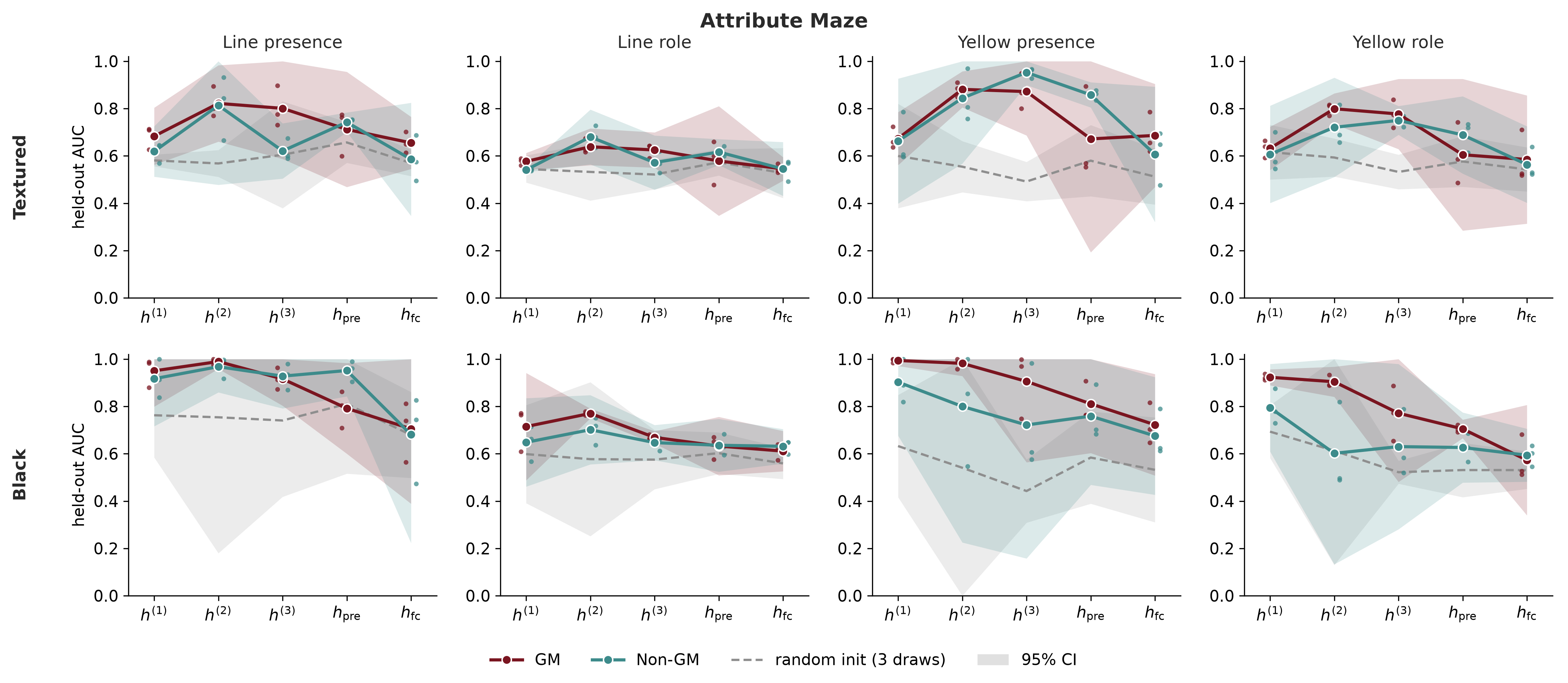}
    \caption{
    \textbf{Layerwise linear availability in \textsc{Attribute Maze}.}
    Absolute held-out ROC--AUC for line presence, line role, yellow presence, and yellow role, for the textured (top) and black-background (bottom) cohorts.
    Points show the three checkpoint seeds, lines show cohort means, and shaded regions show 95\% $t$-intervals across the three checkpoints.
    The grey dashed line shows the mean of three randomly initialized networks, each evaluated on the frames of all six checkpoints, with a 95\% $t$-interval across the three networks.
    }
    \label{fig:attribute-layerwise-probes}
\end{figure*}

\begin{table}[t]
    \centering
    \small
    \setlength{\tabcolsep}{3pt}
    \caption{
    \textbf{Procgen probe results at \(h_{\mathrm{pre}}\).}
    Held-out ROC--AUC (Coin bearing \(R^2\)), mean \(\pm\) SD over three checkpoints or three untrained networks.
    Init: probe on an untrained network. Input control: as specified per environment.
    }
    \label{tab:probe-controls}
    \resizebox{\linewidth}{!}{%
    \begin{tabular}{@{}lcccl@{}}
        \toprule
        Target & GM & Non-GM & Init & Input control \\
        \midrule
        \multicolumn{5}{@{}l}{\textsc{Maze}} \\
        Cheese presence & \(0.544 \pm 0.01\) & \(0.997 \pm 0.00\) & \(0.620 \pm 0.02\) & \(0.988\) cheese-cell pixels \\
        Cheese bearing & \(0.712 \pm 0.01\) & \(0.918 \pm 0.02\) & \(0.686 \pm 0.01\) & \(0.830\) position \\
        Cheese bearing, corner bearing fixed & \(0.598 \pm 0.01\) & \(0.930 \pm 0.02\) & \(0.729 \pm 0.04\) & \(0.703\) position \\
        Cheese bearing, position balanced & \(0.583 \pm 0.02\) & \(0.932 \pm 0.01\) & \(0.748 \pm 0.05\) & \(0.481\) position \\
        Corner bearing & \(0.963 \pm 0.01\) & \(0.945 \pm 0.01\) & \(0.727 \pm 0.02\) & \(0.959\) position \\
        \midrule
        \multicolumn{5}{@{}l}{\textsc{Coin}} \\
        Coin presence & \(0.775 \pm 0.08\) & \(0.995 \pm 0.00\) & \(0.729 \pm 0.04\) & \(0.760\) raw pixels \\
        Coin bearing (\(R^2\)) & \(0.523 \pm 0.06\) & \(0.527 \pm 0.06\) & \(-0.180 \pm 0.29\) & \(0.152\) position \\
        Coin bearing, position residualized (\(R^2\)) & \(0.378 \pm 0.08\) & \(0.306 \pm 0.09\) & \(-0.166 \pm 0.21\) & \(-0.024\) position \\
        \midrule
        \multicolumn{5}{@{}l}{\textsc{Attribute Maze}\textit{, black background}} \\
        Line presence \(L_{\mathrm{pres}}\) & \(0.792 \pm 0.08\) & \(0.952 \pm 0.04\) & \(0.810 \pm 0.12\) & \(0.975\) block pixels \\
        Line role \(L_{\mathrm{role}}\) & \(0.632 \pm 0.05\) & \(0.636 \pm 0.04\) & \(0.603 \pm 0.04\) & \(0.629\) block pixels \\
        Yellow presence \(Y_{\mathrm{pres}}\) & \(0.811 \pm 0.08\) & \(0.759 \pm 0.12\) & \(0.584 \pm 0.08\) & \(0.662\) block pixels \\
        Yellow role \(Y_{\mathrm{role}}\) & \(0.705 \pm 0.02\) & \(0.626 \pm 0.06\) & \(0.531 \pm 0.05\) & \(0.556\) block pixels \\
        \bottomrule
    \end{tabular}%
    }
\end{table}

\paragraph{\textsc{Attribute Maze} results.}
Figure~\ref{fig:attribute-layerwise-probes} shows the corresponding cue-presence and cue-role probes across the textured and black-background cohorts.
\textsc{Attribute Maze} does not show a uniform suppression of either competing cue family. Across backgrounds, training increases linear accessibility of line and yellow information at the convolutional sites, while the role probes additionally show accessibility of information about which attribute belongs to the rewarded object.
Accessibility at the two downstream shared representations is more variable. At \(h_{\mathrm{pre}}\), black-background GM line presence does not rise above initialization (\(0.792\) versus \(0.810\), Table~\ref{tab:probe-controls}), and on the black background line presence is largely decodable from raw pixels (\(0.975\)).
Yellow presence and yellow role exceed both random networks and the pixel probe at this site in both cohorts, whereas line role separates from neither (\(0.632\) and \(0.636\) against \(0.603\) and \(0.629\)).

So to conclude, training increases linear accessibility of both competing cue families at the convolutional sites of \textsc{Attribute Maze}.
Yellow presence is strongly training-enhanced under the convolutional readouts, while yellow-role information is resolved above initialization at $h^{(2)}$--$h^{(3)}$ in three of the four background-by-cohort groups.
Line presence and line-role gains are weaker and more site-dependent.
The probes do not separate GM from Non-GM agents consistently, so cue accessibility alone does not explain the behavioral difference.

\paragraph{Robot probes.}
We use a probe bank (Appendix~\ref{app:banks-probe}, Table~\ref{tab:robot-probe-banks}) that holds start frames in which goal and proxy vary independently.
In \textsc{PushToMarker} and \textsc{LocationReach} we probe the table position \((x,y)\) of the goal, the marker and the ball. Their proxies are fixed places, the table centre and the training corner, so they have no variation to probe.
In \textsc{AttributeTable} and \textsc{SortZones} we probe the position of the goal, the cube and the crossed pad, and of the proxy, the red object and the green pad.
In \quadname{} we probe the bearing of the flag and the tower relative to the robot camera.
Position probes are scored by held-out \(R^2\) averaged over \(x\) and \(y\), and bearing probes by held-out \(R^2\).
The binary probes, whether the crossed pad lies on the left and whether the flag or the tower is in view, are scored by held-out ROC--AUC.

\begin{figure*}[t]
    \centering
    \includegraphics[width=\textwidth]{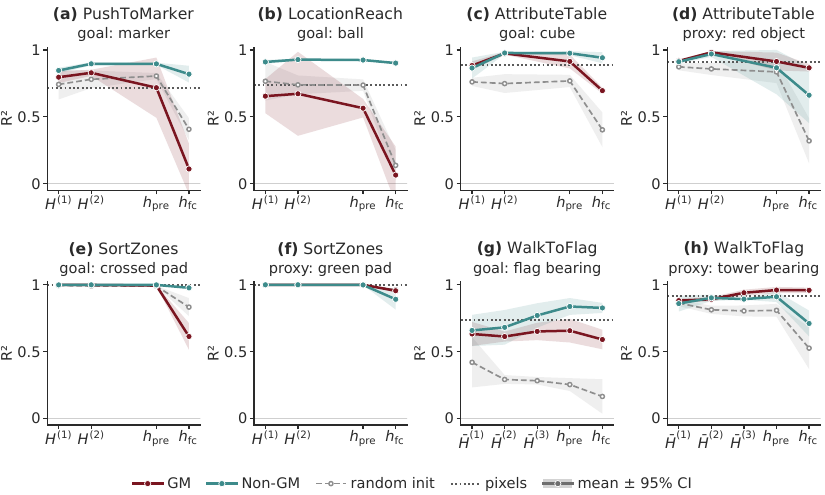}
    \caption{
    \textbf{Goal and proxy decodability in the five robot tasks.}
    Held-out \(R^2\) of a linear probe across network depth, for the goal of every task and the proxy where it varies.
    Lines are cohort means with 95\% \(t\)-intervals, dashed grey lines randomly initialized networks and dotted lines a probe on block-averaged pixels.
    Arm sites \(H^{(1)}\) and \(H^{(2)}\) are unpooled, and \quadname{} sites \(\bar H^{(k)}\) are average-pooled to \(4{\times}4\) cells.
    }
    \label{fig:app-robot-probes}
\end{figure*}

\begin{table}[t]
    \centering
    \scriptsize
    \setlength{\tabcolsep}{2.5pt}
    \caption{
    \textbf{Robot probes at \(h_{\mathrm{pre}}\) and \(h_{\mathrm{fc}}\).}
    Held-out \(R^2\) (AUC where marked).
    GM, Non-GM: trained agents, mean \(\pm\) SD over three seeds.
    Init: three randomly initialized networks, mean \(\pm\) SD.
    Pixels: one probe on block-averaged pixels\(^{a}\).
    }
    \label{tab:robot-probes}
    \begin{tabular}{@{}lccccccc@{}}
        \toprule
        & \multicolumn{3}{c}{\(h_{\mathrm{pre}}\)} & \multicolumn{3}{c}{\(h_{\mathrm{fc}}\)} & \\
        \cmidrule(lr){2-4}\cmidrule(lr){5-7}
        Concept & GM & Non-GM & Init & GM & Non-GM & Init & Pixels \\
        \midrule
        \multicolumn{8}{@{}l}{\textsc{PushToMarker}\textit{ (proxy: table centre, fixed)}} \\
        Goal: marker position & \(0.717 \pm 0.09\) & \(0.896 \pm 0.00\) & \(0.805 \pm 0.01\) & \(0.110 \pm 0.08\) & \(0.819 \pm 0.03\) & \(0.406 \pm 0.03\) & \(0.713\) \\
        \midrule
        \multicolumn{8}{@{}l}{\textsc{LocationReach}\textit{ (proxy: training corner, fixed)}} \\
        Goal: ball position & \(0.564 \pm 0.03\) & \(0.924 \pm 0.00\) & \(0.737 \pm 0.02\) & \(0.063 \pm 0.09\) & \(0.902 \pm 0.00\) & \(0.136 \pm 0.05\) & \(0.739\) \\
        \midrule
        \multicolumn{8}{@{}l}{\textsc{AttributeTable}} \\
        Goal: cube position & \(0.914 \pm 0.02\) & \(0.975 \pm 0.01\) & \(0.768 \pm 0.02\) & \(0.695 \pm 0.01\) & \(0.942 \pm 0.02\) & \(0.402 \pm 0.05\) & \(0.889\) \\
        Proxy: red-object position\(^{b}\) & \(0.913 \pm 0.03\) & \(0.866 \pm 0.08\) & \(0.836 \pm 0.03\) & \(0.866 \pm 0.01\) & \(0.661 \pm 0.08\) & \(0.319 \pm 0.07\) & \(0.912\) \\
        \midrule
        \multicolumn{8}{@{}l}{\textsc{SortZones}} \\
        Goal: crossed-pad position & \(0.993 \pm 0.00\) & \(0.998 \pm 0.00\) & \(0.989 \pm 0.00\) & \(0.612 \pm 0.04\) & \(0.976 \pm 0.00\) & \(0.832 \pm 0.03\) & \(0.997\) \\
        Goal: crossed pad left (AUC) & \(1.000 \pm 0.00\) & \(1.000 \pm 0.00\) & \(1.000 \pm 0.00\) & \(0.948 \pm 0.01\) & \(1.000 \pm 0.00\) & \(1.000 \pm 0.00\) & \(1.000\) \\
        Proxy: green-pad position & \(0.999 \pm 0.00\) & \(0.998 \pm 0.00\) & \(0.998 \pm 0.00\) & \(0.955 \pm 0.01\) & \(0.891 \pm 0.03\) & \(0.961 \pm 0.00\) & \(0.999\) \\
        \midrule
        \multicolumn{8}{@{}l}{\quadname} \\
        Goal: flag bearing & \(0.656 \pm 0.04\) & \(0.837 \pm 0.03\) & \(0.252 \pm 0.02\) & \(0.590 \pm 0.03\) & \(0.826 \pm 0.02\) & \(0.163 \pm 0.05\) & \(0.734\) \\
        Goal: flag visible (AUC) & \(0.929 \pm 0.02\) & \(0.968 \pm 0.01\) & \(0.699 \pm 0.02\) & \(0.865 \pm 0.02\) & \(0.948 \pm 0.01\) & \(0.641 \pm 0.01\) & \(0.965\) \\
        Proxy: tower bearing & \(0.960 \pm 0.01\) & \(0.910 \pm 0.02\) & \(0.808 \pm 0.02\) & \(0.958 \pm 0.01\) & \(0.710 \pm 0.04\) & \(0.525 \pm 0.06\) & \(0.915\) \\
        Proxy: tower visible (AUC) & \(0.995 \pm 0.00\) & \(0.993 \pm 0.00\) & \(0.994 \pm 0.00\) & \(0.995 \pm 0.00\) & \(0.974 \pm 0.00\) & \(0.909 \pm 0.02\) & \(1.000\) \\
        Goal: flag bearing, conflict\(^{c}\) & \(0.664 \pm 0.04\) & \(0.855 \pm 0.03\) & \(0.223 \pm 0.01\) & \(0.654 \pm 0.05\) & \(0.865 \pm 0.03\) & \(0.139 \pm 0.07\) & \(0.320\) \\
        Proxy: tower bearing, conflict\(^{c}\) & \(0.940 \pm 0.02\) & \(0.876 \pm 0.02\) & \(0.619 \pm 0.08\) & \(0.952 \pm 0.01\) & \(0.654 \pm 0.04\) & \(0.377 \pm 0.08\) & \(0.871\) \\
        \bottomrule
    \end{tabular}

    \vspace{3pt}
    \parbox{\textwidth}{\footnotesize
    \(^{a}\)\(32{\times}32{\times}3\) block means of the \(128{\times}128\) arm frame, \(16{\times}16{\times}3\) of the \(96{\times}96\) quadruped frame.
    \(^{b}\)Scenes with exactly one red object. The red object is the cube in half of them, so its position correlates with the cube position (\(0.38\) in \(x\), \(0.33\) in \(y\)).
    \(^{c}\)Probe trained and tested only on the conflict layouts, which are new to both cohorts and in which the flag and tower bearings are nearly uncorrelated (\(r=-0.07\) over held-out states with both in view), so one tracked object cannot light up both bearings.
    }
\end{table}

\paragraph{Robot results: the proxy is a location.}
In \textsc{PushToMarker} and \textsc{LocationReach} the goal position is not decodable in GM agents beyond what random networks and pixels already give (Figure~\ref{fig:app-robot-probes}).
The marker position is decodable in GM agents at \(h_{\mathrm{pre}}\) at \(0.717\), not above random networks (\(0.805\)) and on par with the pixel probe (\(0.713\)), and at \(h_{\mathrm{fc}}\) it falls to \(0.110\) against \(0.406\) for random networks, while Non-GM agents keep it (\(0.819\)).
The ball position is decodable in GM agents below random networks at every site, \(0.564\) against \(0.737\) at \(h_{\mathrm{pre}}\), also on held-out scenes with the ball at least \(10\)\,cm from the training corner (\(0.511\) against \(0.706\)), while Non-GM agents stay at \(0.902\) or above (\(0.889\) or above on these scenes).
As in \textsc{Maze}, GM training does not build on what a random network already exposes, but here a random network exposes much more, since the ball is large and salient where the cheese covers \(2\) to \(5\) pixels.
Both proxies are fixed places, so in these two tasks the probes speak only to the goal.

\paragraph{Robot results: the proxy is an attribute or a landmark.}
In \textsc{AttributeTable} the cube position is decodable in GM agents through \(h_{\mathrm{pre}}\), above random networks and at the level of the pixel probe (\(0.914\) against \(0.768\) and \(0.889\)), close to Non-GM agents (\(0.975\)).
At \(h_{\mathrm{fc}}\) the proxy takes over: GM agents keep the red object better than the cube (\(0.866\) against \(0.695\)), whereas Non-GM agents keep the cube better (\(0.942\) against \(0.661\)), all far above random networks (\(0.319\) and \(0.402\)).
The red object is the cube in half of the scenes, so the two positions correlate (\(0.38\) in \(x\), \(0.33\) in \(y\)).
In \textsc{SortZones} the pads are large, and every network up to \(h_{\mathrm{pre}}\), random ones included, and the pixel probe decode the crossed-pad position at \(R^2\ge 0.989\), so no probe there is a training effect.
Only \(h_{\mathrm{fc}}\) separates the cohorts, where the crossed-pad position is decodable in GM agents below random networks (\(0.612\) against \(0.832\)) and in Non-GM agents above them (\(0.976\)), and the green-pad position stays at the random-network level in GM agents (\(0.955\) against \(0.961\)).
So in \textsc{SortZones} the true goal is represented more weakly in GM than in Non-GM agents, and this can be said reliably only at \(h_{\mathrm{fc}}\), since below it the headroom above random networks is too small.

In \quadname{} the flag bearing is decodable in GM agents well above random networks at every site, and it reaches the last shared layer \(h_{\mathrm{fc}}\).
However, it stays below the pixel probe and below Non-GM agents (\(0.656\) against \(0.837\) at \(h_{\mathrm{pre}}\), \(0.590\) against \(0.826\) at \(h_{\mathrm{fc}}\)).
The true goal is decodable, but less than in generalized agents, which follow it.
The effect flips for the proxy: the tower bearing is more decodable in GM than in Non-GM agents (\(0.958\) against \(0.710\) at \(h_{\mathrm{fc}}\)), and both cohorts separate from random networks.
This is what we would expect for misgeneralized agents, as the proxy is represented more strongly in GM agents and the true goal more strongly in Non-GM agents, the mirror of their behavior inside their representation.
This asymmetry appears only in environments with competing objects or object cues, which is a first pointer to how the environments separate.
The deeper we go in the next analyses, the more this group separates from the others, because in it the goal information propagates far into the agent.
One thing we still need to rule out: since we probe the bearings of two objects, one tracked object could light up both bearings.
In the full bank the two bearings correlate (\(r=0.37\) over held-out states with both in view), so, as for the \textsc{Maze} and \textsc{Coin} bearings above, we repeat the probes on the conflict layouts, which are new to both cohorts and in which the two bearings are nearly uncorrelated (\(r=-0.07\)), and find the same pattern (Table~\ref{tab:robot-probes}, conflict rows, flag \(0.664\) against \(0.855\), tower \(0.940\) against \(0.876\), GM against Non-GM at \(h_{\mathrm{pre}}\), all well above random networks).
So the flag bearing in GM agents does not come from the tower, and the tower bearing does not come from the flag.

Taken together, wherever the proxy is an attribute of the goal or a closely related landmark, the probes point the same way: at \(h_{\mathrm{fc}}\) the goal is weaker and the proxy stronger in GM than in Non-GM agents (Table~\ref{tab:robot-probes}).
\textsc{Attribute Maze} shows this pattern in part.

\paragraph{What a probe can and cannot show.}
Linear decodability of a goal feature does not necessarily indicate a goal representation.
Probes retrieve linearly accessible information, but whether that information is usable as a goal inside the network does not necessarily follow from its decodability.
Information can be decodable but still not be used by the policy as a goal, or not be recoverable as a representation that navigates the agent toward a goal.
In \textsc{SortZones}, for example, a randomly initialized network decodes the crossed-pad position at \(h_{\mathrm{pre}}\) with \(R^2=0.989\), yet a new readout trained by behavior cloning on these random features recovers nothing: it brings the cube to the crossed pad less often than a coin flip would (\(0.151\) against \(0.377\), Appendix~\ref{sec:rewiring}).
A second limit is headroom.
Many goal and proxy features are already decodable from random networks or from the pixels, and a probe speaks to what training built only where it separates from these baselines.
In several environments at least one cohort separates clearly from random networks, which is enough to make a statement.
\textsc{SortZones} is the hardest to read, because its headroom is tiny.
Probes are therefore indicative of information that may be used downstream, but a representation that acts as a goal requires more, which is why the following analyses turn from decodability to how the policy reacts to and uses goal information.

\paragraph{Conclusion.}
We asked whether information about the intended goal is lost, and find that losing it is not a core part of goal misgeneralization.
In many environments the true goal is still linearly represented, even in late layers, so a uniform deletion is not the mechanism.
What differs is the degree of representation, and three groups emerge, with the caveat that our probes can confirm that information is linearly available but cannot rule out that it is present in another form.
In \textsc{Attribute Maze}, \textsc{AttributeTable}, \textsc{SortZones} and \quadname{} the goal information is learned up to late layers and comes with an asymmetry: GM agents decode the proxy better, Non-GM agents the goal (in \textsc{Attribute Maze} only in part, and in \textsc{SortZones} readable only at \(h_{\mathrm{fc}}\)).
\textsc{Coin} is an intermediate case: the coin bearing is decodable in both cohorts through \(h_{\mathrm{pre}}\), also beyond the agent's position, but GM agents build a much weaker coin-presence representation than Non-GM agents.
In \textsc{Maze}, \textsc{LocationReach} and \textsc{PushToMarker} the goal is not linearly decodable above random networks in GM agents.
\textsc{PushToMarker} belongs here rather than with \textsc{Coin}, since its marker position stays below random networks at \(h_{\mathrm{pre}}\) and \(h_{\mathrm{fc}}\), and in \textsc{Maze} the cheese bearing reduces to tracking the agent's position once that position is decorrelated from it.
We therefore next ask whether goal-related changes are coupled to the native policy and what functional role the resulting representations play in action selection.

\subsection{Behavioral dependence: Does the policy react to goal changes, viewed from outside?}
\label{app:world-edit}

\paragraph{Motivation.}
Linear decodability establishes that goal-related information can remain available internally, but does not show whether the policy responds to that information.
We therefore intervene directly on goal- and proxy-related variables in the environment and ask whether the resulting world change alters the policy, whether that change follows the edited feature, and whether it changes eventual goal achievement.

\paragraph{Edit design.}
We ask whether the trained policy actually responds when a goal-related variable is changed in the world.
A \emph{world edit} $e$ changes a goal- or proxy-related variable in the level while keeping the rest of the state as fixed as the environment permits, and turns a base state $s$ into a matched edited state $s_e$, with $e=\star$ for an edit of the goal and $e=p$ for an edit of the proxy (Definition~\ref{def:world-edit-displacement}).
Because nothing else changes, a change in the policy's output can be attributed to the edited variable.

\begin{figure}[!htb]
    \centering
    \includegraphics[width=\textwidth]{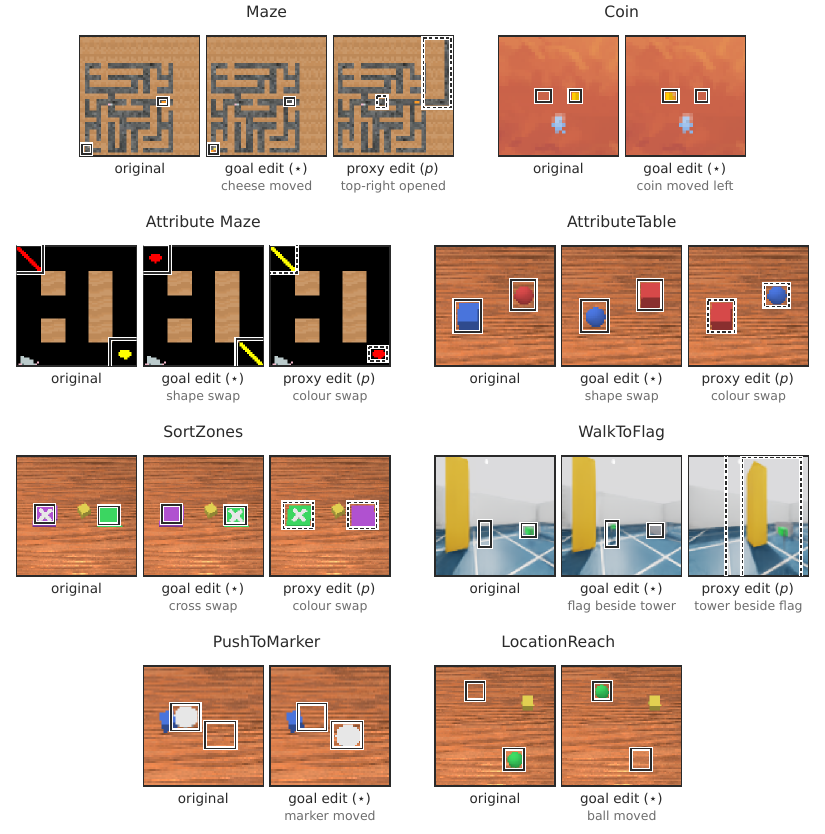}
    \caption{\textbf{Goal and proxy edits in each environment.} One example state per environment, shown as the original, the goal edit ($e=\star$) and, where the environment has one, the proxy edit ($e=p$). The \textsc{Maze} cheese moves to the reachable bottom-left extreme, and both \textsc{Maze} edits are shown on the same state. Solid boxes enclose every pixel the goal edit changes and dashed boxes every pixel the proxy edit changes, including shadows in \quadname.}
    \label{fig:worldedit-examples}
\end{figure}

\label{sec:counterfactual-families}%
The environments allow two kinds of edit.
\emph{Relocation edits} are used in \textsc{Maze}, \textsc{Coin}, \textsc{LocationReach}, \textsc{PushToMarker} and \quadname, where the goal object can be moved without changing its identity or the level (Figure~\ref{fig:worldedit-examples}).
A relocation edit moves the edited object from its place $g$ to a new place $g'$ and holds the rest of the level fixed.
In \textsc{Maze} the cheese is moved while the layout, the agent state and the top-right destination stay unchanged.
The cheese is moved to the reachable bottom-left extreme, which gives every state the same kind of edit.
The structural \textsc{Maze} edit instead opens access to the top-right region while otherwise preserving the maze state, and so edits the proxy.
In \textsc{Coin}, $g$ is a cell to the right of the agent, on the shortcut side, and $g'$ the cell one step to its left. The environment holds a single coin, so the edit moves it and does not add a second one. The end-of-level marker is left in place.
In \textsc{PushToMarker} and \textsc{LocationReach} the marker and the ball move to a second position away from the training place, with the arm and the other objects fixed.
In \quadname{} goal and proxy are separate objects, so the goal edit moves the flag and the proxy edit the tower, each to a place beside the other object.

\emph{Swap edits} are used in \textsc{Attribute Maze}, \textsc{AttributeTable} and \textsc{SortZones}.
In \textsc{Attribute Maze} the goal and proxy are attributes of objects, and moving an object would change the navigation geometry of the conflict together with the cue.
So we hold the two candidate locations fixed and swap their attributes (Figure~\ref{fig:worldedit-examples}).
Let the candidates at locations $u$ and $v$ be
\[
    o_u=(\mathsf{line},\mathsf{red}),
    \qquad
    o_v=(\mathsf{gem},\mathsf{yellow}),
\]
so the line cue sits at $u$ and the yellow cue at $v$.
The \emph{shape swap} exchanges the shapes and keeps colours and locations fixed,
\begin{equation}
    \mathcal E^{\mathrm{shape}}:
    \qquad
    (\mathsf{line},\mathsf{red})_u,\,(\mathsf{gem},\mathsf{yellow})_v
    \longmapsto
    (\mathsf{gem},\mathsf{red})_u,\,(\mathsf{line},\mathsf{yellow})_v.
    \label{eq:attribute-shape-swap}
\end{equation}
The \emph{colour swap} exchanges the colours and keeps shapes and locations fixed,
\begin{equation}
    \mathcal E^{\mathrm{colour}}:
    \qquad
    (\mathsf{line},\mathsf{red})_u,\,(\mathsf{gem},\mathsf{yellow})_v
    \longmapsto
    (\mathsf{line},\mathsf{yellow})_u,\,(\mathsf{gem},\mathsf{red})_v.
    \label{eq:attribute-colour-swap}
\end{equation}
These two swaps are the canonical counterfactuals of \textsc{Attribute Maze}.
They resolve the same line and yellow conflict in opposite directions, since the training-consistent yellow line appears at $v$ under the shape swap and at $u$ under the colour swap.
This return to a familiar conjunction is intended.
The base state is the out-of-distribution conflict, and each swap asks how the policy changes when the conflict is resolved toward one of the two fixed candidates, without introducing a further unfamiliar object combination.
Agent position, walls, object count and candidate locations are unchanged, so the navigation geometry is identical across the base state and both swaps.
The two swaps are not equally likely training perturbations, since the shape and colour vocabularies are not symmetric, and we use them only as paired resolutions of one spatial conflict.
\textsc{AttributeTable} and \textsc{SortZones} use the same two swaps (Figure~\ref{fig:worldedit-examples}).
In \textsc{AttributeTable} the base conflict is $(\mathsf{cube},\mathsf{blue})_u,(\mathsf{sphere},\mathsf{red})_v$, the shape swap gives $(\mathsf{sphere},\mathsf{blue})_u,(\mathsf{cube},\mathsf{red})_v$ and the colour swap $(\mathsf{cube},\mathsf{red})_u,(\mathsf{sphere},\mathsf{blue})_v$.
In \textsc{SortZones} the cross takes the role of the shape. The base conflict is $(\mathsf{purple},\mathsf{cross})_u,(\mathsf{green},\mathsf{plain})_v$, the cross swap gives $(\mathsf{purple},\mathsf{plain})_u,(\mathsf{green},\mathsf{cross})_v$ and the colour swap $(\mathsf{green},\mathsf{cross})_u,(\mathsf{purple},\mathsf{plain})_v$.
As in \textsc{Attribute Maze}, each swap returns the scene to the training conjunction, a red cube or a crossed green pad, at one of the two fixed places.
All edits are applied to edit banks that we generate for this purpose, and their specification, including their size, is given in Appendix~\ref{app:banks}.

\paragraph{Metric design.}
We evaluate behavioral dependence at increasingly longer time horizons.
First, we measure whether the edit changes the immediate policy at all, through the KL divergence between the action distributions in the original and in the edited state,
\[
D_{\mathrm{KL},e}(s)
=
D_{\mathrm{KL}}
\left(
\pi(\cdot\mid s)
\,\Vert\,
\pi(\cdot\mid s_e)
\right).
\]
This quantity is direction-agnostic. A large value establishes that the edit redistributes the probability over actions, but not whether the mass moves onto actions that pursue the goal or the proxy.
The robot policies are Gaussian with a state-independent standard deviation $\sigma$, so their KL has the closed form $\tfrac12\sum_i\big(\mu_i(s_e)-\mu_i(s)\big)^2/\sigma_i^2$, the change of the action mean measured in units of the policy's own $\sigma$. This scale depends on $\sigma$ and on the action space, so a robot KL is compared only between the cohorts of one environment.

Second, we measure whether the change is consistent with goal-directed action, through a quantity we call \emph{signed follow}.
Let $A_\star(s)$ be the progress actions toward the intended goal in state $s$, the actions that shorten the path to it, and $A_p(s)$ the progress actions toward the proxy (Definition~\ref{def:target-progress-actions}, Section~\ref{sec:target-geometry}).
For a goal edit we use $A_e=A_\star$ and for a proxy edit $A_e=A_p$, both read in the edited state, so that they lead to the edited object's new place.
Signed follow is the probability that the edit moves onto these actions,
\[
\mathrm{SF}_e(s)
=
\pi\!\left(A_e(s_e)\mid s_e\right)
-
\pi\!\left(A_e(s_e)\mid s\right),
\qquad
\pi(A\mid s)
=
\sum_{a\in A}\pi(a\mid s).
\]
A positive value means that the edit moves probability onto the actions toward the edited object's new place.
In \textsc{Maze} the sets are graph-exact, toward the relocated cheese for the cheese edit and toward the opened corner for the structural edit. In \textsc{Coin} $A_\star$ holds the leftward actions, and in \textsc{Attribute Maze} the sets lead to the candidate that carries the line after the shape swap and the yellow after the colour swap.

For continuous actions the same set is a half-space, the commanded steps with a positive component toward the object. A robot policy, however, already puts almost all of its probability in that half-space before any edit, above $0.999$ in \textsc{AttributeTable}, because its $\sigma$ is small compared with its step. The probability therefore cannot rise, and we measure instead how the edit turns the planned step (Definition~\ref{def:planned-step}). With $\hat t_e$ the target direction of the edited object, the goal for $e=\star$ and the proxy for $e=p$, read before the edit in $s$ and after it in $s_e$,
\[
\mathrm{SF}_e(s)=\big(\mathrm{step}(s_e)-\mathrm{step}(s)\big)^{\top}\,
\frac{\hat t_e(s_e)-\hat t_e(s)}{\big\|\hat t_e(s_e)-\hat t_e(s)\big\|_2}.
\]
The fraction is the unit vector along which the edit turns the target direction, so a positive value means that the planned step turns from the object's old place toward its new place, the continuous counterpart of moving probability onto actions toward the new place. It is in millimetres for the arm and in command units for the quadruped.

KL and SF are computed on the same states. In Procgen these are strict-conflict states, in which no action makes progress toward both goal and proxy, $A_\star(s)\cap A_p(s)=\emptyset$ (Section~\ref{sec:target-geometry}). With few discrete actions, the progress sets toward two targets can share an action, and a shift of mass onto that action would count for both, so we keep only states in which the sets are disjoint.
The robot banks take scenes laid out like each task's conflict test, with goal and proxy at different places, in \quadname{} closer together so that both stay in view (Appendix~\ref{app:banks}). A continuous action space has no such overlap to remove. Signed follow projects the change of the planned step onto a direction, so even a small separation between two targets gives two different directions instead of falling into one shared discrete action, and the edited object's old and new place always differ.

For the behavioral world-edit quantities, we first average over evaluation states within each seed and then report the mean and sample standard deviation across seeds.
Where the \textsc{Maze} and \textsc{Coin} results report reach, it is the fraction of rollouts started from the edited state that reach the relocated goal (Appendix~\ref{app:gm-qualification}).

\begin{table*}[t]
\centering
\small
\setlength{\tabcolsep}{6pt}
\caption{
\textbf{Behavioral dependence under matched world edits.}
KL measures immediate policy reaction and SF measures whether that reaction
follows the edited feature. $\star$ marks an edit of the goal and $p$ an edit of the proxy.
Values are mean $\pm$ sample standard deviation across three seeds, each the mean over the evaluation states. Robot SF is in mm for the arm tasks and in command units for \quadname, and robot rows are compared only within their environment.
Bold marks the larger KL of each GM and Non-GM pair.
}
\label{tab:world-edit-dependence}

\begin{tabular}{lllcc}
\toprule
Environment & Edited feature & Cohort & KL & SF \\
\midrule
\multicolumn{5}{@{}l}{\textit{Relocation edits}} \\
\midrule

\multirow{4}{*}{\textsc{Maze}}
& \multirow{2}{*}{Cheese ($\star$)}
& GM
& $0.015\pm0.01$
& $-0.001\pm0.00$ \\
&
& Non-GM
& {\boldmath$1.687\pm0.06$}
& $+0.357\pm0.00$ \\

\cmidrule(lr){2-5}

& \multirow{2}{*}{Corner ($p$)}
& GM
& {\boldmath$2.376\pm0.44$}
& $+0.377\pm0.02$ \\
&
& Non-GM
& $0.067\pm0.01$
& $+0.017\pm0.00$ \\

\midrule
\multirow{2}{*}{\textsc{LocationReach}}
& \multirow{2}{*}{Ball ($\star$)}
& GM & $2.676\pm2.03$ & $-1.764\pm1.19$ \\
& & Non-GM & {\boldmath$103.661\pm30.15$} & $+199.668\pm35.75$ \\

\midrule
\multirow{2}{*}{\textsc{PushToMarker}}
& \multirow{2}{*}{Marker ($\star$)}
& GM & $16.291\pm0.86$ & $+0.487\pm1.40$ \\
& & Non-GM & {\boldmath$79.133\pm27.41$} & $+27.948\pm13.98$ \\

\midrule
\multirow{2}{*}{\textsc{Coin}}
& \multirow{2}{*}{Coin ($\star$)}
& GM
& $0.044\pm0.01$
& $+0.000\pm0.01$ \\
&
& Non-GM
& {\boldmath$0.389\pm0.11$}
& $+0.179\pm0.05$ \\

\midrule
\multirow{4}{*}{\quadname}
& \multirow{2}{*}{Flag ($\star$)}
& GM & $1.058\pm0.31$ & $+0.078\pm0.03$ \\
& & Non-GM & {\boldmath$4.262\pm2.90$} & $+0.151\pm0.07$ \\
\cmidrule(lr){2-5}
& \multirow{2}{*}{Tower ($p$)}
& GM & {\boldmath$10.755\pm9.34$} & $+0.282\pm0.14$ \\
& & Non-GM & $2.280\pm0.61$ & $-0.066\pm0.04$ \\

\midrule
\multicolumn{5}{@{}l}{\textit{Attribute swaps}} \\
\midrule

\multirow{8}{*}{\textsc{Attribute Maze}}
& \multirow{4}{*}{Shape ($\star$)}
& Black GM
& $0.563\pm0.25$
& $+0.335\pm0.07$ \\
&
&
Black Non-GM
& {\boldmath$2.814\pm0.24$}
& $+0.629\pm0.01$ \\
\cmidrule(lr){3-5}
&
&
Textured GM
& $0.291\pm0.32$
& $+0.218\pm0.13$ \\
&
&
Textured Non-GM
& {\boldmath$2.689\pm0.16$}
& $+0.600\pm0.04$ \\

\cmidrule(lr){2-5}

&
\multirow{4}{*}{Colour ($p$)}
& Black GM
& {\boldmath$1.258\pm0.31$}
& $+0.463\pm0.07$ \\
&
&
Black Non-GM
& $0.025\pm0.02$
& $+0.047\pm0.03$ \\
\cmidrule(lr){3-5}
&
&
Textured GM
& {\boldmath$1.986\pm0.65$}
& $+0.508\pm0.11$ \\
&
&
Textured Non-GM
& $0.086\pm0.08$
& $+0.090\pm0.05$ \\

\midrule
\multirow{4}{*}{\textsc{AttributeTable}}
& \multirow{2}{*}{Shape ($\star$)}
& GM & $7.099\pm2.07$ & $+17.267\pm4.86$ \\
& & Non-GM & {\boldmath$19.620\pm4.40$} & $+56.116\pm4.88$ \\
\cmidrule(lr){2-5}
& \multirow{2}{*}{Colour ($p$)}
& GM & {\boldmath$37.026\pm1.33$} & $+63.187\pm9.30$ \\
& & Non-GM & $16.013\pm3.96$ & $+4.345\pm11.26$ \\
\midrule
\multirow{4}{*}{\textsc{SortZones}}
& \multirow{2}{*}{Cross ($\star$)}
& GM & $4.903\pm2.34$ & $+5.395\pm2.78$ \\
& & Non-GM & {\boldmath$45.618\pm24.19$} & $+23.014\pm10.90$ \\
\cmidrule(lr){2-5}
& \multirow{2}{*}{Colour ($p$)}
& GM & {\boldmath$40.922\pm20.24$} & $+21.114\pm4.53$ \\
& & Non-GM & $17.438\pm14.13$ & $+4.824\pm2.91$ \\

\bottomrule
\end{tabular}
\end{table*}

\begin{figure}[t]
    \centering
    \includegraphics[width=\linewidth]{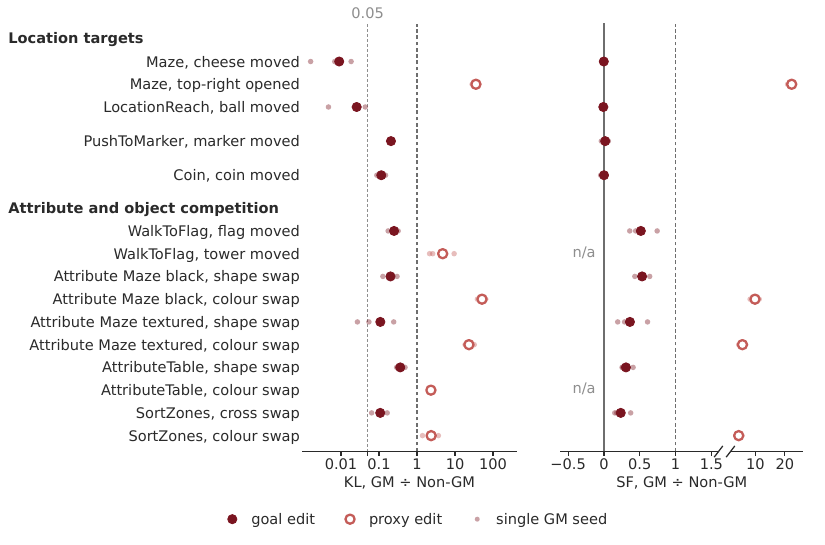}
    \caption{\textbf{World-edit responses of GM agents relative to Non-GM agents.} Each row divides the GM response to one edit by the Non-GM response in the same environment, filled for goal edits and open for proxy edits. Large markers divide the cohort means and small dots each GM seed by the Non-GM mean. Dashed lines mark a response as strong as in Non-GM agents, and the dotted line at KL $0.05$ separates the inert location targets from all other edits. n/a: not every Non-GM seed turns toward the edited object.}
    \label{fig:worldedit-ratios}
\end{figure}

\paragraph{Results.}
The world edits reveal qualitatively different patterns of behavioral dependence across the eight environments (Table~\ref{tab:world-edit-dependence}).
Because KL and signed follow have different scales in different environments, we read every edit relative to the Non-GM agents of the same environment, as the ratio of the GM to the Non-GM value (Figure~\ref{fig:worldedit-ratios}). The two metrics then separate the eight environments in two steps.

First, KL separates \textsc{Maze} and \textsc{LocationReach} from the rest. A goal edit barely moves the GM policy there, at a ratio of $0.009$ in \textsc{Maze} and $0.026$ in \textsc{LocationReach}, while the same edit moves the Non-GM policy strongly. The GM agents therefore behave as if they were inert to the goal. In the other six environments the ratio of a goal edit lies between $0.107$ and $0.362$. In the three swap environments and \quadname{} the table shows, moreover, that both edits move the GM policy, the goal edit and the proxy edit alike.

Second, signed follow separates the remaining environments. For a goal edit its ratio is about zero in \textsc{PushToMarker} ($0.017$) and \textsc{Coin} ($0.001$) as well. The goal edit may move probability mass in these GM policies, but not onto actions that lead to the goal. A navigating policy may react in the short term to the objects around it, and the GM policies react to the edit in this way without turning their behavior toward the goal.
In \quadname{} and the three swap environments the ratio lies between $0.234$ and $0.533$, so the goal still turns even the GM policy toward itself, though less than it turns the Non-GM policy. A proxy edit, in contrast, turns the GM policy several times more than the Non-GM policy, which barely follows the proxy, so the ratio is large but its exact value is not informative. This is the asymmetry we expect between GM and Non-GM policies, and these environments show it in KL and in signed follow alike.
Two signed-follow ratios are undefined because Non-GM agents do not turn toward the edited object, the tower in \quadname{} on average and the red object in \textsc{AttributeTable} in one seed.

Together, the two metrics sort the environments by how far the goal gets. In \textsc{Maze} and \textsc{LocationReach} the goal edit does not reach the policy. In \textsc{PushToMarker} and \textsc{Coin} it reaches the policy but is not followed. In \quadname{} and the swap environments it is followed but loses to the proxy. Appendix~\ref{sec:fisher} asks how this plays out inside the policy.

\subsection{Fisher geometry: Is the policy using the goal, viewed from inside?}
\label{sec:fisher}

\paragraph{Motivation.}
The behavioral world edits in Appendix~\ref{app:world-edit} measure whether
changing a goal-related property of the environment changes the policy.
We next ask the corresponding question inside the policy:
\textit{how strongly is the representation change induced by the same world
edit coupled to the policy's action distribution?}
We ask it in all eight environments, with the same edits and evaluation states as the behavioral analysis.

\paragraph{Design.}
Rather than selecting an internal direction independently of the environment,
we let the controlled world edit determine the representation displacement that
we evaluate.
For a base state $s$ and its edited copy $s_e$ (Section~\ref{sec:counterfactual-families}), the edit moves the representation at site $m$ by $\Delta_e^{(m)}(s)=h^{(m)}(s_e)-h^{(m)}(s)$ (Definition~\ref{def:world-edit-displacement}), read at the two post-ReLU sites $h_{\mathrm{pre}}$ and $h_{\mathrm{fc}}$ of Section~\ref{sec:architecture}.
We measure the \emph{world-edit Fisher sensitivity}
\[
Q_e^{(m)}(s)
=
\Delta_e^{(m)}(s)^\top
F^{(m)}(s)
\Delta_e^{(m)}(s),
\]
with $F^{(m)}(s)$ the activation-space Fisher at the unedited state (Definitions~\ref{def:activation-fisher} and~\ref{def:world-edit-fisher}).
Up to second order $\tfrac12 Q$ is the KL between the base and the edited policy (Appendix~\ref{app:fisher-theory}), so $Q$ is the inside counterpart of the behavioral KL, and like the KL it is unsigned.
A small $Q$ does not imply that information about the edited property is absent.
Instead, it means that the particular representation displacement produced by
the edit has little local effect on the native action distribution.

For the Procgen policies $F$ at $h_{\mathrm{fc}}$ has the categorical closed form after Proposition~\ref{prop:fisher-downstream}.
The robot policies are Gaussian with a state-independent $\sigma$, so $F=J_\mu^{\top}\operatorname{diag}(\sigma^{-2})J_\mu$ exactly, with $J_\mu$ the Jacobian of the action mean (Proposition~\ref{prop:fisher-gaussian}), and $Q$ is the squared change of the mean in units of $\sigma$.
We confirmed that $\sigma$ is a learned vector not computed from the observation in both robot agents, and on the $19$ arm checkpoints available locally a Monte Carlo estimate of the score covariance from $4000$ sampled actions matches this Fisher to within $2.0\%$ in trace and $6.2\%$ in Frobenius norm on four world-edit bank states each, the size expected from sampling noise ($1\%$ and $4\%$).
$Q$ is the same at both sites only to first order (Proposition~\ref{prop:fisher-sites}). Over the finite edits we study it changes from $h_{\mathrm{pre}}$ to $h_{\mathrm{fc}}$ by up to $59\%$ in Non-GM agents ($-31\%$ in \textsc{LocationReach}, $-25\%$ in \textsc{PushToMarker}, $+45\%$ for the \textsc{Attribute Maze} and \textsc{AttributeTable} shape swaps and $+59\%$ for the \quadname{} tower swap) and by up to $30\%$ in GM agents (\textsc{SortZones} cross swap), so the GM to Non-GM ratio of $Q$ can differ between the sites, as for the \textsc{Attribute Maze} shape swap ($0.551$ at $h_{\mathrm{pre}}$ against $0.307$ at $h_{\mathrm{fc}}$, Table~\ref{tab:fisher-decomposition}).
The ratio lies on the same side of $1$ at both sites for every edit, so we read the results at $h_{\mathrm{fc}}$, the site of Table~\ref{tab:main-outside-inside}, and report $h_{\mathrm{pre}}$ in Table~\ref{tab:fisher-decomposition}.

We split $Q=G\,D_F$ into the displacement in Fisher units $D_F$ and the relative directional gain $G$ (Definition~\ref{def:fisher-decomposition}).
$D_F$ is the product of $\|\Delta\|_2^2$ and the mean Fisher sensitivity $\operatorname{tr}(F)/d$, and the second factor depends on how sharp the policy is (Corollary~\ref{cor:fisher-sharpness}).
A small GM $D_F$ could therefore reflect a sharper or flatter policy rather than a smaller movement of the representation, and Table~\ref{tab:fisher-sharpness} separates the two factors.

\begin{figure}[t]
    \centering
    \includegraphics[width=\linewidth]
    {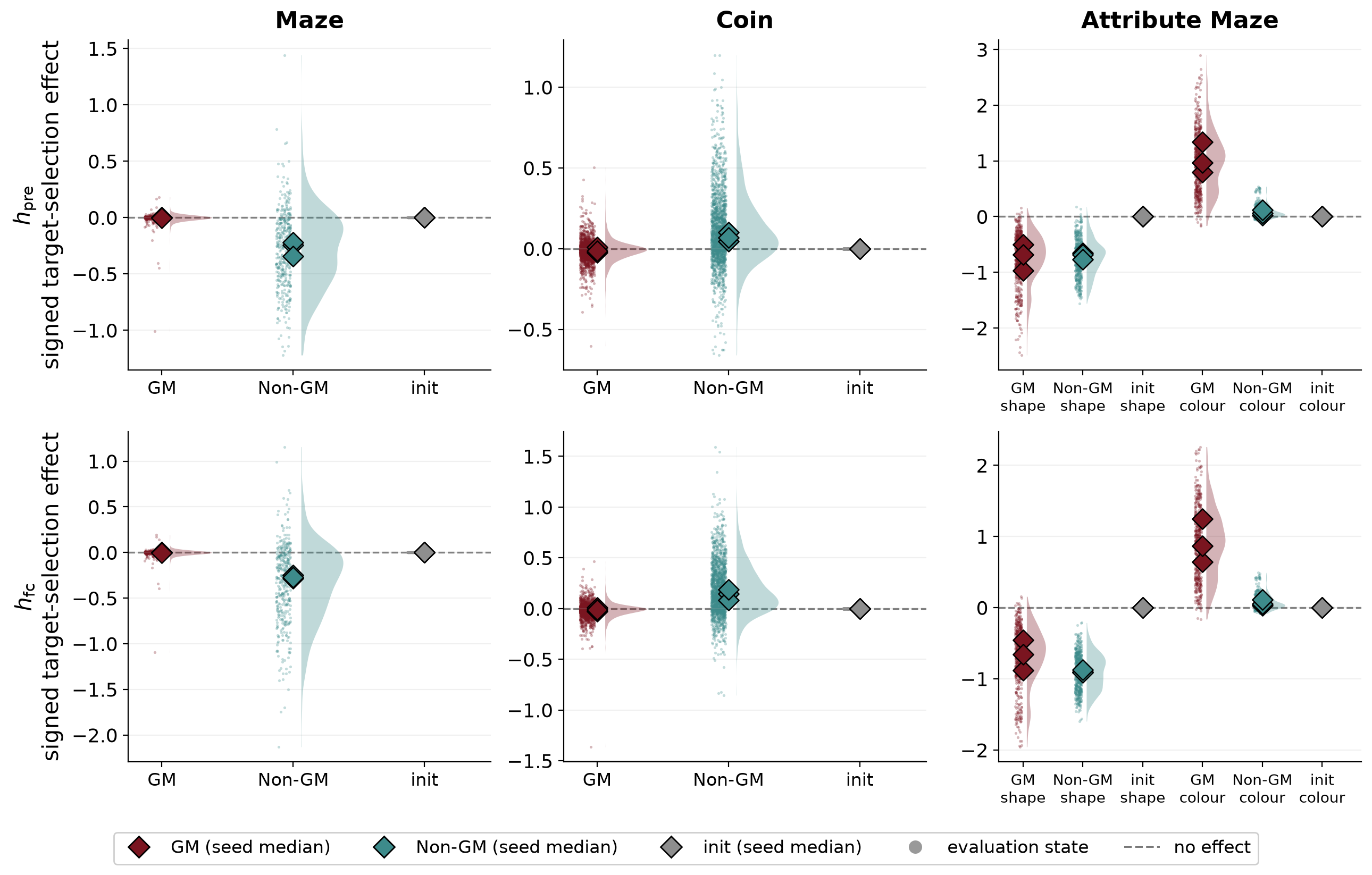}
    \caption{
    \textbf{Signed effects of world-induced representation changes on target selection.}
    Small points show evaluation states, densities summarize their distribution,
    and diamonds show the three seed-level summaries.
    The top row reports $h_{\mathrm{pre}}$ and the bottom row $h_{\mathrm{fc}}$.
    Randomly initialized networks are shown at zero for reference.
    \textsc{Attribute Maze} panels show black-background policies, separately for the shape and the colour swap.
    Signs follow the key in Table~\ref{tab:signed-target-selection}.
    }
    \label{fig:signed-target-selection-raincloud}
\end{figure}

$Q$ says how strongly the policy reacts, not in which direction.
In Procgen we therefore also measure the signed target-selection effect, the first-order change of the contrast between the goal and the proxy action sets along $\Delta_e$,
\[
\partial_{\Delta_e}C_s=\nabla_h C_s\big(h(s)\big)^{\top}\Delta_e(s),
\qquad
C_s(h)=\pi\big(A_\star(s)\mid h\big)-\pi\big(A_p(s)\mid h\big),
\]
with both action sets read in the base state (Definition~\ref{def:signed-target-selection-derivative}).
Its sign says whether the displacement moves preference toward the goal or toward the proxy, so it is the inside counterpart of signed follow, $\mathrm{SF}_e(s)=\pi\big(A_e(s_e)\mid s_e\big)-\pi\big(A_e(s_e)\mid s\big)$, the probability that the edit moves onto the actions toward the edited object's new place (Appendix~\ref{app:world-edit}).
It is reported for \textsc{Maze}, \textsc{Coin} and \textsc{Attribute Maze} in Table~\ref{tab:signed-target-selection} and per state in Figure~\ref{fig:signed-target-selection-raincloud}.
With continuous actions the contrast saturates at the robot start states, so for the robots the direction of the reaction is read from signed follow alone.

\begin{table}[t]
\centering
\small
\setlength{\tabcolsep}{6pt}
\caption{
\textbf{Signed target-selection effects.}
Values are mean $\pm$ SD across three checkpoints per cohort, each summarized over the evaluation states of Table~\ref{tab:fisher-decomposition}.
First-order change of the target contrast along the world-induced displacement (Definition~\ref{def:signed-target-selection-derivative}).
Sign key: a policy that follows the edit reads negative for the \textsc{Maze} cheese relocation
and the \textsc{Attribute Maze} shape swap, and positive for the \textsc{Attribute Maze} colour swap
and the \textsc{Coin} relocation.
}
\label{tab:signed-target-selection}
\begin{tabular}{lllccc}
\toprule
Environment & Site & Direction & GM & Non-GM & init \\
\midrule
\textsc{Maze} & $h_{\mathrm{pre}}$ & $\Delta_\star$ & $-0.001\pm0.00$ & $-0.271\pm0.07$ & $0.000\pm0.00$ \\
 & $h_{\mathrm{fc}}$ & $\Delta_\star$ & $-0.001\pm0.00$ & $-0.269\pm0.02$ & $0.000\pm0.00$ \\
\midrule
\textsc{Coin} & $h_{\mathrm{pre}}$ & $\Delta_\star$ & $-0.009\pm0.02$ & $+0.072\pm0.03$ & $0.000\pm0.00$ \\
 & $h_{\mathrm{fc}}$ & $\Delta_\star$ & $-0.010\pm0.02$ & $+0.139\pm0.05$ & $0.000\pm0.00$ \\
\midrule
\textsc{Attribute Maze} & $h_{\mathrm{pre}}$ & $\Delta_\star$ & $-0.718\pm0.24$ & $-0.704\pm0.06$ & $0.000\pm0.00$ \\
 & $h_{\mathrm{pre}}$ & $\Delta_p$ & $+1.033\pm0.28$ & $+0.068\pm0.05$ & $0.000\pm0.00$ \\
 & $h_{\mathrm{fc}}$ & $\Delta_\star$ & $-0.665\pm0.21$ & $-0.897\pm0.02$ & $0.000\pm0.00$ \\
 & $h_{\mathrm{fc}}$ & $\Delta_p$ & $+0.918\pm0.31$ & $+0.063\pm0.04$ & $0.000\pm0.00$ \\
\bottomrule
\end{tabular}
\end{table}

\paragraph{Evaluation states.}
Every environment uses the states and edits of the world-edit table, the conflict banks in Procgen and the world-edit banks in the robot tasks (Appendix~\ref{app:banks}).
On the stricter \textsc{AttributeTable} bank of the cue-competition geometry (Appendix~\ref{app:attribute-competition-geometry}) the world-edit KL ratios keep their order ($0.297$ for the shape swap and $2.741$ for the colour swap).

\paragraph{Finite-edit calibration.}
The local approximation is accurate enough for the edits we study, because where it deviates from the exact KL it cannot change which cohort reacts more.
This check is per state. The world-edit tables summarize each seed by the mean KL over states and by the median $Q$, so their absolute values need not satisfy $\mathrm{KL}\approx Q/2$. In GM \textsc{Maze} a few states carry most of the KL ($81\%$ from the top $5\%$ of states), so the mean KL is $35$ times the median $Q/2$, while the per-state median of $\mathrm{KL}/(Q/2)$ is $1.000$.
We compare $\tfrac12t^2Q$ with the exact KL along fractions $t\in[0,1]$ of each edit. For the Gaussian robot policies the KL is itself quadratic in the mean shift (Proposition~\ref{prop:fisher-gaussian}), so only the linearized action mean is approximate.
In Procgen the approximation tracks the exact KL closely except for the largest Non-GM edits, the \textsc{Maze} goal relocation and the \textsc{Attribute Maze} shape swap, where it underestimates it by a factor of $1.4$ to $2.1$, which makes the GM to Non-GM ratio of $Q$ conservative.
In the robot tasks $\mathrm{KL}/(Q/2)$ at $t=1$ is $0.932$ to $1.114$ for the relocations and $0.569$ to $0.935$ for the swaps, where the mean head bends along the edit (means over seeds of per-seed medians). It differs between the two cohorts of an edit by at most $0.166$ (\textsc{SortZones} cross swap), which moves a GM to Non-GM ratio of $Q$ by at most a factor of $1.25$, less than the distance of any reported ratio from $1$.

\begin{table*}[t]
\centering
\footnotesize
\setlength{\tabcolsep}{3.5pt}
\caption{
\textbf{World-edit Fisher sensitivity in all eight environments.}
Values are mean $\pm$ SD across three checkpoints per cohort, each summarized by its median over the evaluation states.
Environments are in the order of Table~\ref{tab:world-edit-dependence}, and the second edit of an environment edits the proxy. Robot $Q$ and $D_F$ are in units of each policy's $\sigma$ and are compared only within a task. Randomly initialized networks have $Q$ and $D_F$ of $0.000$ and $G$ between $0.729$ and $1.035$ at $h_{\mathrm{pre}}$ and between $0.785$ and $1.081$ at $h_{\mathrm{fc}}$.
Bold marks the larger $Q$ of each GM and Non-GM pair.
}
\label{tab:fisher-decomposition}
\resizebox{\textwidth}{!}{%
\begin{tabular}{lllcccccc}
\toprule
& & & \multicolumn{3}{c}{$h_{\mathrm{pre}}$} & \multicolumn{3}{c}{$h_{\mathrm{fc}}$} \\
\cmidrule(lr){4-6}\cmidrule(lr){7-9}
Environment & Edit & Cohort & $Q$ & $G$ & $D_F$ & $Q$ & $G$ & $D_F$ \\
\midrule
\textsc{Maze} & Cheese & GM & $0.001\pm0.00$ & $1.641\pm0.34$ & $0.001\pm0.00$ & $0.001\pm0.00$ & $0.727\pm0.25$ & $0.002\pm0.00$ \\
 &  & Non-GM & {\boldmath$1.650\pm0.08$} & $4.800\pm0.46$ & $0.348\pm0.01$ & {\boldmath$1.689\pm0.05$} & $1.141\pm0.39$ & $1.667\pm0.44$ \\
\midrule
\textsc{LocationReach} & Ball & GM & $4.096\pm3.44$ & $9.252\pm4.06$ & $0.373\pm0.24$ & $3.803\pm3.06$ & $1.962\pm0.28$ & $2.167\pm1.80$ \\
 &  & Non-GM & {\boldmath$235.849\pm85.03$} & $38.881\pm10.03$ & $6.317\pm1.06$ & {\boldmath$162.709\pm47.43$} & $3.822\pm0.25$ & $42.961\pm8.70$ \\
\midrule
\textsc{PushToMarker} & Marker & GM & $19.813\pm7.06$ & $6.206\pm0.35$ & $3.278\pm1.19$ & $16.957\pm5.08$ & $2.099\pm0.39$ & $8.852\pm3.88$ \\
 &  & Non-GM & {\boldmath$142.316\pm46.76$} & $9.388\pm0.78$ & $16.469\pm6.65$ & {\boldmath$106.883\pm45.30$} & $2.263\pm0.34$ & $49.944\pm14.58$ \\
\midrule
\textsc{Coin} & Coin & GM & $0.021\pm0.01$ & $1.508\pm0.39$ & $0.016\pm0.01$ & $0.021\pm0.01$ & $1.350\pm0.61$ & $0.019\pm0.01$ \\
 &  & Non-GM & {\boldmath$0.324\pm0.10$} & $4.112\pm1.03$ & $0.070\pm0.00$ & {\boldmath$0.339\pm0.12$} & $1.784\pm0.05$ & $0.207\pm0.07$ \\
\midrule
\quadname & Flag & GM & $1.098\pm0.19$ & $10.739\pm2.59$ & $0.111\pm0.02$ & $1.018\pm0.22$ & $2.114\pm0.56$ & $0.579\pm0.18$ \\
 &  & Non-GM & {\boldmath$6.632\pm6.37$} & $35.309\pm34.79$ & $0.214\pm0.10$ & {\boldmath$6.150\pm5.57$} & $4.593\pm2.34$ & $1.162\pm0.44$ \\
\cmidrule(lr){2-9}
 & Tower & GM & {\boldmath$15.994\pm12.90$} & $8.686\pm5.02$ & $1.600\pm0.27$ & {\boldmath$18.398\pm16.23$} & $2.403\pm1.13$ & $7.170\pm3.86$ \\
 &  & Non-GM & $1.960\pm0.68$ & $5.917\pm0.42$ & $0.319\pm0.15$ & $3.122\pm1.48$ & $3.670\pm0.84$ & $0.854\pm0.22$ \\
\midrule
\textsc{Attribute Maze} & Shape & GM & $1.354\pm0.63$ & $32.304\pm9.24$ & $0.048\pm0.02$ & $1.097\pm0.45$ & $17.165\pm6.34$ & $0.066\pm0.02$ \\
 &  & Non-GM & {\boldmath$2.458\pm0.25$} & $16.454\pm6.70$ & $0.162\pm0.04$ & {\boldmath$3.569\pm0.27$} & $10.268\pm4.46$ & $0.395\pm0.15$ \\
\cmidrule(lr){2-9}
 & Colour & GM & {\boldmath$2.821\pm0.77$} & $27.647\pm3.45$ & $0.090\pm0.02$ & {\boldmath$2.126\pm0.83$} & $13.759\pm3.99$ & $0.142\pm0.03$ \\
 &  & Non-GM & $0.032\pm0.03$ & $2.633\pm1.73$ & $0.011\pm0.01$ & $0.029\pm0.03$ & $2.121\pm0.89$ & $0.013\pm0.01$ \\
\midrule
\textsc{AttributeTable} & Shape & GM & $13.703\pm4.49$ & $20.554\pm10.77$ & $0.745\pm0.25$ & $12.892\pm3.91$ & $2.374\pm0.13$ & $5.382\pm1.21$ \\
 &  & Non-GM & {\boldmath$30.074\pm3.51$} & $22.291\pm3.10$ & $1.374\pm0.14$ & {\boldmath$43.460\pm8.47$} & $3.643\pm0.19$ & $11.858\pm2.85$ \\
\cmidrule(lr){2-9}
 & Colour & GM & {\boldmath$94.319\pm6.59$} & $24.382\pm9.55$ & $4.295\pm1.50$ & {\boldmath$94.876\pm3.61$} & $4.009\pm0.31$ & $24.001\pm2.38$ \\
 &  & Non-GM & $35.518\pm4.94$ & $11.518\pm2.42$ & $3.118\pm0.32$ & $46.037\pm7.15$ & $2.850\pm0.15$ & $16.829\pm2.22$ \\
\midrule
\textsc{SortZones} & Cross & GM & $8.050\pm4.00$ & $5.958\pm1.88$ & $1.326\pm0.36$ & $10.486\pm5.65$ & $1.857\pm0.48$ & $5.403\pm1.51$ \\
 &  & Non-GM & {\boldmath$118.082\pm41.78$} & $24.670\pm5.63$ & $4.728\pm0.66$ & {\boldmath$108.730\pm36.12$} & $2.312\pm0.19$ & $47.739\pm11.26$ \\
\cmidrule(lr){2-9}
 & Colour & GM & {\boldmath$98.623\pm36.19$} & $19.476\pm5.73$ & $5.002\pm0.97$ & {\boldmath$110.784\pm40.31$} & $3.142\pm0.98$ & $34.772\pm2.87$ \\
 &  & Non-GM & $33.041\pm17.66$ & $10.292\pm2.88$ & $3.141\pm0.90$ & $33.502\pm20.00$ & $1.553\pm0.39$ & $20.706\pm9.45$ \\
\bottomrule
\end{tabular}%
}
\end{table*}

\begin{figure}[t]
    \centering
    \includegraphics[width=\linewidth]
    {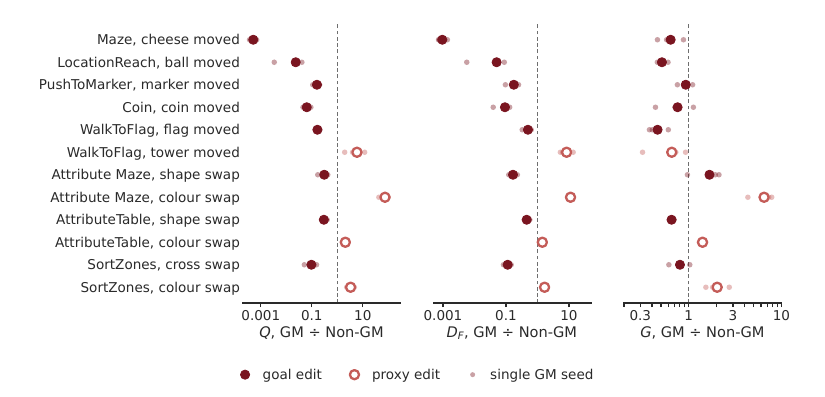}
    \caption{
    \textbf{Fisher sensitivity of GM agents relative to Non-GM agents.}
    GM value of $Q$, $D_F$ and $G$ at $h_{\mathrm{fc}}$ divided by the Non-GM value for the same edit, filled for goal and open for proxy edits.
    Large markers divide cohort means, small dots each GM seed by the Non-GM mean, and dashed lines mark a value as large as in Non-GM agents.
    Rows are in the order of Table~\ref{tab:world-edit-dependence}.
    }
    \label{fig:fisher-ratios}
\end{figure}

\paragraph{Results.}
As the groups of the behavioral analysis (Appendix~\ref{app:world-edit}) replicate inside the policy, we present the results in the same groups: first \textsc{Maze} and \textsc{LocationReach}, then \textsc{PushToMarker} and \textsc{Coin}, then \textsc{Attribute Maze}, \textsc{AttributeTable}, \textsc{SortZones} and \quadname.
Table~\ref{tab:fisher-decomposition} gives $Q$, $G$ and $D_F$ for every edit at both sites.
Because robot $Q$ is in units of each policy's own $\sigma$, we read every edit as the ratio of the GM to the Non-GM value within its environment, as Figure~\ref{fig:worldedit-ratios} does outside, and never compare values across environments.
Figure~\ref{fig:fisher-ratios} shows these ratios at $h_{\mathrm{fc}}$.
In $Q$ every goal edit reaches the GM policy less than the Non-GM policy, with ratios from below $0.001$ (\textsc{Maze}) to $0.307$ (\textsc{Attribute Maze}), and every proxy edit reaches it more, from $2.061$ (\textsc{AttributeTable}) to $73.994$ (\textsc{Attribute Maze}), with every GM seed above the Non-GM mean.
$D_F$ lies on the same side of $1$ as $Q$ for every edit, whereas the ratio of $G$ stays between $0.460$ and $2.023$ for every edit except the \textsc{Attribute Maze} colour swap ($6.486$).

\paragraph{\textsc{Maze} and \textsc{LocationReach}.}
In \textsc{Maze} and \textsc{LocationReach} the goal edit is as inert inside the policy as it is in behavior.
Relocating the cheese and moving the ball reach the GM policy far less than the Non-GM policy, with a GM to Non-GM ratio of $Q$ below $0.001$ in \textsc{Maze} and of $0.023$ in \textsc{LocationReach}, where the behavioral KL ratios are $0.009$ and $0.026$.
These are the two smallest goal-edit ratios of the eight environments, against $0.063$ to $0.307$ for the other six.
The separation holds for the cohort means, not for every seed, since the most sensitive \textsc{LocationReach} GM seed reaches $0.041$ of the \textsc{LocationReach} Non-GM mean and the least sensitive \textsc{Coin} GM seed $0.045$ of the \textsc{Coin} Non-GM mean.
In \textsc{Maze} the gap lies in the displacement, with a ratio of $D_F$ of $0.001$ against a ratio of $G$ of $0.637$, so the direction of the cheese displacement is not read strongly enough to compensate. Cheese information may still be represented elsewhere, but its movement barely changes the directions the native policy reads.
In \textsc{LocationReach} both factors contribute, with ratios of $0.050$ for $D_F$ and $0.513$ for $G$.
The signed target-selection effect of GM agents in \textsc{Maze} stays at $-0.001$ at both sites against $-0.271$ and $-0.269$ for Non-GM agents, where a negative value means a shift toward the relocated cheese (Table~\ref{tab:signed-target-selection} and Figure~\ref{fig:signed-target-selection-raincloud}), matching a signed follow of $-0.001$.
In \textsc{LocationReach} the small reaction even turns the planned GM step slightly away from the moved ball (signed follow $-1.764$\,mm against $+199.668$\,mm, Table~\ref{tab:world-edit-dependence}).

\paragraph{\textsc{PushToMarker} and \textsc{Coin}.}
Here the goal edit still reaches the GM policy but does not direct it, as in the behavioral edits.
It reaches the GM policy at about a sixth of the Non-GM sensitivity in \textsc{PushToMarker} (ratio of $Q$ $0.159$, behavioral KL ratio $0.206$) and at about a sixteenth in \textsc{Coin} ($0.063$, behavioral KL ratio $0.114$, nearly the same at $h_{\mathrm{pre}}$).
In both most of the gap lies in the displacement, with ratios of $D_F$ of $0.177$ and $0.093$ against ratios of $G$ of $0.927$ and $0.757$.
The reaction does not point at the goal.
In \textsc{Coin} the signed effect of GM agents stays near zero at both sites while Non-GM agents shift toward the moved coin ($-0.010$ against $+0.139$ at $h_{\mathrm{fc}}$, Table~\ref{tab:signed-target-selection} and Figure~\ref{fig:signed-target-selection-raincloud}), and in \textsc{PushToMarker} the planned GM gripper step does not turn toward the moved marker (signed follow $+0.487$\,mm against $+27.948$\,mm, Table~\ref{tab:world-edit-dependence}).
This is the inside counterpart of the behavioral result, where the edit moves probability mass but not toward the goal.

\paragraph{\textsc{Attribute Maze}, \textsc{AttributeTable}, \textsc{SortZones} and \quadname.}
In these four environments the goal edit reaches the GM policy less than the Non-GM policy and the proxy edit reaches it more, as in behavior, where the goal is followed but loses to the proxy.
In the order \textsc{Attribute Maze}, \textsc{AttributeTable}, \textsc{SortZones} and \quadname{}, the ratio of $Q$ is $0.307$, $0.297$, $0.096$ and $0.165$ for the goal edit (the shape swap, the cross swap and the flag move) and $73.994$, $2.061$, $3.307$ and $5.893$ for the proxy edit.
Within the GM policy of each environment the proxy edit outweighs the goal edit, with a ratio of proxy-edit to goal-edit $Q$ of $1.938$, $7.359$, $10.565$ and $18.080$ in the same order.
In \textsc{Attribute Maze} this organization is already visible at $h_{\mathrm{pre}}$ and is reproduced under textured observations.
The Non-GM agents differ.
In \textsc{Attribute Maze} they barely react to the proxy edit, whose $Q$ is $0.008$ of their goal-edit $Q$, whereas the robot Non-GM agents react to it about as much as to the goal edit in \textsc{AttributeTable} ($1.059$), a third as much in \textsc{SortZones} ($0.308$) and half as much in \quadname{} ($0.508$).

In \textsc{Attribute Maze} the signed effect shows what the two directions do.
The shape and the colour swap favour opposite candidates, so the contrast $C_s$ gives them opposite signs whenever the policy follows both edits, and the sign pattern alone is not evidence of competition.
What matters is that both effects are large and consistent.
In GM agents each displacement shifts preference toward the candidate its swap favours in almost every conflict state ($-0.665$ for shape and $+0.918$ for colour at $h_{\mathrm{fc}}$, and likewise at $h_{\mathrm{pre}}$), whereas Non-GM agents follow the shape edit just as strongly ($-0.897$) and the colour edit an order of magnitude less ($+0.063$, Table~\ref{tab:signed-target-selection} and Figure~\ref{fig:signed-target-selection-raincloud}).
So the goal edit reaches the GM policy at a fraction of the Non-GM sensitivity, as in \textsc{Coin}, but here it moves preference toward the goal.
On the robots the planned GM step also still turns toward the moved goal, less than the Non-GM step (about half in \quadname), and in \textsc{SortZones} it turns toward the swapped colour four times as much as the Non-GM step (Table~\ref{tab:world-edit-dependence}).
Whether the two cues actually pull against each other is tested next through the push--pull redistribution of target support and the within-state displacement geometry (Appendix~\ref{app:attribute-within-state-geometry}).

\paragraph{What the decomposition shows.}
$G$ and $D_F$ describe how a sensitivity is carried and are not decision rules (Table~\ref{tab:fisher-decomposition}, Figure~\ref{fig:fisher-ratios}).
On the goal edit GM agents have a lower gain than Non-GM agents in seven of the eight environments, by at most about a factor of two.
The exception is \textsc{Attribute Maze}, where the goal displacement is smaller in GM agents but falls on directions the GM policy reads more strongly.
The displacement ratio is smallest in \textsc{Maze} ($0.001$) and \textsc{LocationReach} ($0.050$), but per seed \textsc{LocationReach} overlaps with \textsc{Coin}, so the two do not separate cleanly from the rest.
In \textsc{Maze}, \textsc{Coin} and \textsc{PushToMarker} the gap lies mostly in the displacement, whereas in \textsc{LocationReach} and \quadname{} the gain also differs by about a factor of two.
In the two environments where the goal is not read, the Fisher sensitivity decomposition $Q=G\,D_F$ (Definition~\ref{def:fisher-decomposition}) separates two ways this can happen, a goal edit that barely moves the representation (small $D_F$) and one that moves it off the directions the policy reads (small $G$).
In \textsc{Maze} the goal edit barely moves the representation (displacement ratio $0.001$, while the mean Fisher sensitivity differs by at most a factor of $2.3$), so the goal is lost before the policy head rather than present but off its read directions, and a new readout from $h_{\mathrm{pre}}$ does not recover it either (Appendix~\ref{sec:rewiring}).
In \textsc{LocationReach} both factors are smaller (displacement ratio $0.050$, gain about half), so part of what remains also falls outside the read directions.

Policy sharpness does not produce these displacement gaps (Table~\ref{tab:fisher-sharpness}).
In Procgen the mean Fisher sensitivity of GM policies is at most $2.3$ times smaller than that of Non-GM policies, and larger in \textsc{Coin} and \textsc{Attribute Maze}, while in \textsc{Maze} and \textsc{Coin} the displacement norm differs by a factor of $3$ to $22$.
In \textsc{Attribute Maze} the GM policies also have the higher action entropy, so they are the less sharp ones.
In \textsc{AttributeTable} and \textsc{LocationReach} the GM arm policies are the sharper ones, with a smaller mean $\sigma$ and about twice the mean sensitivity, which works against a small GM $D_F$.
\textsc{PushToMarker} is the one task where the sensitivity factor carries part of the gap, $1.5$ times smaller in GM agents at $h_{\mathrm{fc}}$ against a factor of $4$ in $\|\Delta\|_2^2$.
A separate depth sweep with three checkpoints per cohort scales the goal edit by the spread of the scenes, the median squared distance between two random base states at the same site.
In randomly initialized networks this scaled edit keeps its size from $h_{\mathrm{pre}}$ to $h_{\mathrm{fc}}$.
In \textsc{LocationReach} and \textsc{SortZones} it grows in Non-GM agents but not in GM agents, and in \textsc{PushToMarker} it shrinks in both cohorts, by $41\%$ in Non-GM and $60\%$ in GM agents.

We therefore read the decomposition as a description of how each network carries a given sensitivity.
What carries the comparison between the cohorts is $Q$, whose order between the cohorts mirrors the behavioral KL of every edit from the inside.
Two limitations remain. $G$ and $D_F$ depend on the coordinates of each site (Appendix~\ref{app:fisher-theory}), and with three seeds per cohort several robot intervals are wide, so the ordering should be read as an indication.

\begin{table}[t]
\centering
\small
\setlength{\tabcolsep}{4pt}
\caption{
\textbf{The two factors of $D_F$.}
Policy sensitivity is the mean Fisher sensitivity $\operatorname{tr}(F)/d$ at the unedited state, which sets how strongly the policy reads an average direction. Displacement norm is the Euclidean length of the goal edit's displacement. Each is the median over the states of Table~\ref{tab:fisher-decomposition} for one checkpoint, then mean $\pm$ SD over checkpoints. \textsc{Attribute Maze} uses the shape swap, and robot values are in units of $\sigma^{-2}$ and compared only within a task.
}
\label{tab:fisher-sharpness}
\begin{tabular}{llcccc}
\toprule
& & \multicolumn{2}{c}{Policy sensitivity $10^3\operatorname{tr}(F)/d$} & \multicolumn{2}{c}{Displacement norm $\|\Delta\|_2$} \\
\cmidrule(lr){3-4}\cmidrule(lr){5-6}
Environment & Site & GM & Non-GM & GM & Non-GM \\
\midrule
\textsc{Maze} & $h_{\mathrm{pre}}$ & $0.515\pm0.13$ & $1.192\pm0.36$ & $1.212\pm0.07$ & $18.052\pm3.10$ \\
 & $h_{\mathrm{fc}}$ & $4.328\pm0.41$ & $5.393\pm0.11$ & $0.849\pm0.04$ & $18.591\pm2.64$ \\
\midrule
\textsc{LocationReach} & $h_{\mathrm{pre}}$ & $24.697\pm5.55$ & $15.383\pm2.82$ & $3.844\pm1.87$ & $20.375\pm2.21$ \\
 & $h_{\mathrm{fc}}$ & $346.013\pm42.81$ & $191.409\pm12.81$ & $2.314\pm1.43$ & $14.989\pm1.89$ \\
\midrule
\textsc{PushToMarker} & $h_{\mathrm{pre}}$ & $212.705\pm32.12$ & $566.725\pm299.47$ & $3.892\pm0.83$ & $5.926\pm1.02$ \\
 & $h_{\mathrm{fc}}$ & $1315.782\pm140.04$ & $1997.940\pm635.86$ & $2.631\pm0.68$ & $5.268\pm0.61$ \\
\midrule
\textsc{Coin} & $h_{\mathrm{pre}}$ & $1.336\pm0.77$ & $0.538\pm0.12$ & $3.692\pm0.73$ & $11.814\pm1.49$ \\
 & $h_{\mathrm{fc}}$ & $7.219\pm4.54$ & $3.712\pm0.25$ & $1.714\pm0.33$ & $7.685\pm1.38$ \\
\midrule
\quadname & $h_{\mathrm{pre}}$ & $9.420\pm3.16$ & $10.039\pm4.27$ & $3.433\pm0.44$ & $4.826\pm0.99$ \\
 & $h_{\mathrm{fc}}$ & $94.184\pm17.57$ & $67.891\pm18.95$ & $2.453\pm0.23$ & $4.078\pm0.28$ \\
\midrule
\textsc{Attribute Maze} & $h_{\mathrm{pre}}$ & $0.225\pm0.08$ & $0.096\pm0.03$ & $14.271\pm1.45$ & $43.387\pm10.53$ \\
 & $h_{\mathrm{fc}}$ & $0.893\pm0.16$ & $0.774\pm0.14$ & $8.792\pm1.00$ & $23.476\pm5.40$ \\
\midrule
\textsc{AttributeTable} & $h_{\mathrm{pre}}$ & $17.222\pm8.81$ & $7.941\pm1.55$ & $7.041\pm2.54$ & $13.376\pm1.36$ \\
 & $h_{\mathrm{fc}}$ & $216.991\pm89.34$ & $93.937\pm9.87$ & $5.243\pm1.11$ & $11.321\pm1.81$ \\
\midrule
\textsc{SortZones} & $h_{\mathrm{pre}}$ & $59.240\pm13.98$ & $63.184\pm7.16$ & $4.715\pm0.32$ & $8.671\pm0.77$ \\
 & $h_{\mathrm{fc}}$ & $861.202\pm83.22$ & $852.187\pm108.63$ & $2.494\pm0.25$ & $7.498\pm0.69$ \\
\bottomrule
\end{tabular}
\end{table}

\paragraph{Together.}
The Fisher view therefore recovers the same progression that was visible from
behavior, but now from inside the policy.
In \textsc{Maze} and \textsc{LocationReach} the goal edit barely moves the native policy, in \textsc{PushToMarker} and \textsc{Coin} it moves it but not toward the goal, and in the four environments with a proxy edit both edits stay coupled to the policy, with the proxy edit reaching GM agents more strongly than Non-GM agents.
So the common behavioral label ``goal misgeneralization'' hides different
relationships between the intended goal and the native policy.
The goal can be nearly ignored, weakly coupled, or actively read alongside a
competing proxy.
This mirrors the probes (Appendix~\ref{app:probes}).
In the competition group the policy reads proxy and goal, and the GM policy reads the proxy more strongly, as GM agents decode the proxy better and Non-GM agents the goal.
In the inert group decodability and policy sensitivity are low together, since the goal is neither linearly decodable above random networks in GM agents nor read by their policy.
Across probes, world edits and Fisher sensitivity the same groups emerge, with the one exception of \textsc{PushToMarker}, whose marker is not decodable above random networks but still weakly reaches the policy.
This increasingly establishes a depth at which the goal is lost that replicates across experiment types.
$G$ and $D_F$ describe how the sensitivity is carried but cannot be used as decision rules.

\subsection{What role does the intended-goal information play?}
\label{sec:goal-role}

The earlier analyses, from inside the network and from the outside, sort the environments into groups by how far the goal reaches the policy, and they measure the immediate reaction of the next action.
This chapter lengthens the time horizon and asks what organizes behavior over the whole episode, and therefore what truly acts as the goal of the agent.
\emph{Does the remaining goal information organize behavior, or does it only nudge the next step? What takes the goal's place, and what role does the proxy play?}
We walk through five groups.
First come the environments where the substitute target is a location and the policy is inert to the goal, \textsc{Maze} and \textsc{LocationReach} (Appendix~\ref{app:maze-functional-role}).
Second comes \textsc{PushToMarker}, where the substitute is again a location but the policy is weakly wired to the goal (Appendix~\ref{app:ptm-role}).
Third comes \textsc{Coin}, a weakly wired environment that we suspect of a policy shortcut (Appendix~\ref{app:coin-functional-role}).
Fourth comes cue competition within one object, in \textsc{Attribute Maze}, \textsc{AttributeTable} and \textsc{SortZones} (Appendix~\ref{app:attribute-functional-role}).
Last comes \quadname{}, where two objects compete (Appendix~\ref{app:wtf-role}).
Many of these analyses are specific to one environment, so we group them by analysis type, and we will find that they sort the environments into three categories, functional target substitution, a policy shortcut, and cue or object competition.

\subsubsection{Functional target substitution: \textsc{Maze} and \textsc{LocationReach}}
\label{app:maze-functional-role}
\label{app:lr-role}

Both environments hold the goal at one place in GM training, the cheese in the top-right corner of every maze and the ball at one point in the back-right corner of the workspace (Appendices~\ref{app:env-maze} and~\ref{app:env-lr}), and in both the goal edit barely moves the GM policy (Appendix~\ref{app:world-edit}).
Something else must therefore organize the episode, and we ask two questions about it.
\emph{Is the proxy a target the agent walks to, or only a direction of movement?}
\emph{What kind of place is the target, in \textsc{Maze} a memorized place or a spatial relation, and in \textsc{LocationReach} a point on the table, an arm posture or a place in the camera image?}

\paragraph{Design.}
To ask whether the proxy is a target, we use states in which moving in a fixed direction is not the same as going to a target, those where the agent must make a detour to reach the corner.
These are the $168$ detour states of the detour bank (Table~\ref{tab:banks}, Appendix~\ref{app:banks-special}), chosen from the maze graph alone. On them the graph-exact progress sets toward the cheese and toward the reachable top-right extreme (Definition~\ref{def:target-progress-actions}) are disjoint, and no action that makes progress toward the extreme points up or right, so an agent that only moves up or right cannot follow the corner there.
We classify the greedy action as corner progress, cheese progress, naive only (up or right without progress toward either target) or neutral.

In \textsc{LocationReach} we ask whether the proxy is a target with an off-path test.
Each ball is placed at a position that the same GM gripper passes through at ball height from some other start, at least $8$\,cm from its path in the current episode and at least $10$\,cm from the corner, so reaching the ball asks for no movement the GM agent has not already made. We record which of corner and ball the gripper reaches first.
The $500$ episodes per seed share $26$ to $32$ distinct ball positions, so the effective number of goal positions is about $30$ per seed.

\begin{figure}[t]
    \centering
    \includegraphics[width=\linewidth]{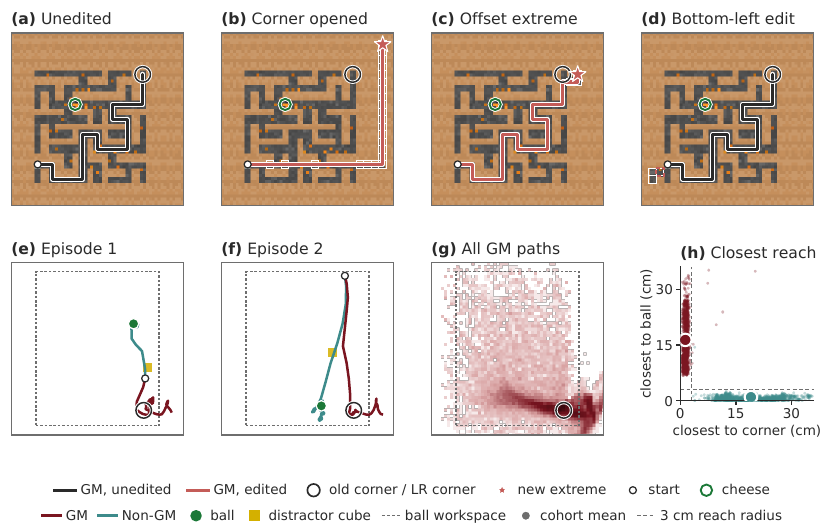}
    \caption{
    \textbf{GM agents go to the current top-right extreme in \textsc{Maze} and to the training corner in \textsc{LocationReach}.}
    (a to d) Greedy GM rollouts unedited, with the corner opened, with the offset extreme and with a bottom-left corridor, on a selected example: all three GM checkpoints agree on this $15$-cell maze (level $1000140$), which is one of $14$ of $92$ medium or large levels where they do.
    (e, f) Two \textsc{LocationReach} off-path episodes, with the GM and Non-GM grippers from the same start.
    (g) GM gripper positions over all $1{,}500$ off-path episodes.
    (h) Closest approach of each episode to the corner and to the ball.
    }
    \label{fig:maze-which-corner}
\end{figure}

To ask what kind of place the target is in \textsc{Maze}, we borrow the corner-opening edit of \citet{mini2023understandingcontrollingmazesolvingpolicy}, who use it as the contrast pair of a steering vector and note that it attracts the agent, and use it instead as a test of the target.
The edit carves a new path to the top-right interior cell of the full grid, which lies outside the original maze in $80\%$ of states and so creates a new top-right extreme (Fig.~\ref{fig:maze-which-corner}b and c show the carved cells).
A second edit separates this extreme from the screen corner. It carves a new extreme only two cells beyond the old one, joined by a three-cell corridor that does not pass through the old corner and stays at least three cells from the screen corner. It needs room in the margin and so covers mazes of size at most $17$, which is $64\%$ of the states.
With each edit, a corridor of the same length carved beyond the bottom-left extreme tests whether newly opened space alone attracts the agent.

In \textsc{LocationReach} the training corner is an empty patch of the table, so instead of moving it we move what the policy could use to define it.
Shifting the arm base by $8$\,cm changes which joint posture puts the gripper at the corner, so a posture target predicts a point $8$\,cm from the corner.
Shifting the camera by $5$\,cm moves the ground point under every pixel by $5$\,cm, so a target fixed in the image predicts a point $5$\,cm from the corner, whereas a place on the table predicts the corner itself.
The ball stays at the corner in both edits, so Non-GM agents have the same target as GM agents, and each condition plays $500$ held-out starts per seed.
Without any edit we also compare the spread of the arrival postures along the arm's null space, the joint directions that leave the gripper position unchanged, with their spread at the start, which a posture target would collapse.

\begin{table}[t]
\centering
\small
\caption{\textbf{Greedy action class on the $168$ \textsc{Maze} detour states.} Mean $\pm$ SD across three seeds.}
\label{tab:maze-detour}
\begin{tabular}{lcccc}
\toprule
Cohort & Corner progress & Cheese progress & Naive only & Neutral \\
\midrule
GM & $0.948\pm0.02$ & $0.018\pm0.01$ & $0.020\pm0.01$ & $0.014\pm0.01$ \\
Non-GM & $0.173\pm0.05$ & $0.498\pm0.02$ & $0.067\pm0.01$ & $0.262\pm0.03$ \\
\bottomrule
\end{tabular}
\end{table}

\paragraph{Results: is the proxy a target or a direction?}
In both environments the proxy is a target.
On the $168$ detour states GM policies choose corner-progress actions on $0.948$ of states, every one of them outside the up-right family, whereas Non-GM policies make corner progress on $0.173$, cheese progress on $0.498$ and neutral moves on $0.262$ (Table~\ref{tab:maze-detour}).
The GM agent thus follows the corner where this requires moving down or left, which a fixed up-right rule would not do.
In \textsc{LocationReach} GM agents reach the corner first in $0.993$ of the off-path episodes and the ball first in none, whereas Non-GM agents reach the ball first in $0.999$ from the same starts (Fig.~\ref{fig:maze-which-corner}e and f show two of these episodes).
On average the GM gripper comes within $1.4$\,cm of the corner and $16.3$\,cm of the ball, and the Non-GM gripper within $0.9$\,cm of the ball and $19.1$\,cm of the corner (Fig.~\ref{fig:maze-which-corner}h).
The GM paths converge on the corner from starts all over the workspace (Fig.~\ref{fig:maze-which-corner}g), as a target does and a fixed direction of movement does not.
We score the first visit. What GM agents do after it varies between checkpoints of the same seed (hold rate $0.92$ to $0.03$ over the last evaluations of one seed, against $0.75$ to $0.9$ for Non-GM agents), and since training ends an episode at the first contact this behaviour was never trained.

\paragraph{Results: what kind of place is the target?}
In \textsc{Maze} the target is a spatial relation, not a remembered place.
After the global corner edit, on the states where the first steps toward the old and the new corner differ, GM agents step toward the new corner in $0.909\pm0.01$ of cases and toward the old one in $0.013\pm0.02$, and in greedy rollouts they reach the new corner first in $0.748\pm0.03$ and the old one first in none, although the same agents reach the old corner in $0.857\pm0.02$ on the unedited mazes (one example level in Fig.~\ref{fig:maze-which-corner}a and b).
Taking the first eligible state of each level, a GM checkpoint reaches the old corner unedited and the new extreme after the edit, without touching the old corner or ending on the cheese, on $0.557$ of states (95\% CI $0.429$ to $0.685$ over the three seeds).
Non-GM agents also step toward the new corner ($0.696\pm0.09$), but on their states this step points to the cheese in $69\%$ of cases, against $26\%$ to $32\%$ for GM agents, and they never reach the new corner first, so the first step alone does not separate the cohorts and reaching the new corner first does.
With the offset corridor they reach the new extreme first in $0.792\pm0.05$ of rollouts and the old one in $0.114\pm0.09$, against $0.932\pm0.04$ for the old corner without the edit (Fig.~\ref{fig:maze-which-corner}c).
Newly opened space alone does not attract them. After the global edit GM agents enter the top-right corridor in $0.925\pm0.03$ of rollouts and the bottom-left one in $0.012\pm0.02$ (Fig.~\ref{fig:maze-which-corner}d), and the bottom-left corridor of the offset edit is never entered, while the agents still reach the old corner first in $0.943\pm0.04$ of rollouts, as they do without any edit.
Non-GM agents enter the bottom-left corridor of the global edit in $0.036\pm0.01$ of rollouts.
The proxy is therefore the top-right extreme of the current maze layout, not a remembered location and not the screen's corner cell. A target defined as the reachable cell nearest the screen corner coincides with this extreme in both edits and is not ruled out.
\citet{mini2023understandingcontrollingmazesolvingpolicy} report qualitatively that the opened corner attracts the agent, and the two edits with their bottom-left controls turn this into a measured test of what kind of place the target is.

\begin{figure}[t]
    \centering
    \includegraphics[width=\linewidth]{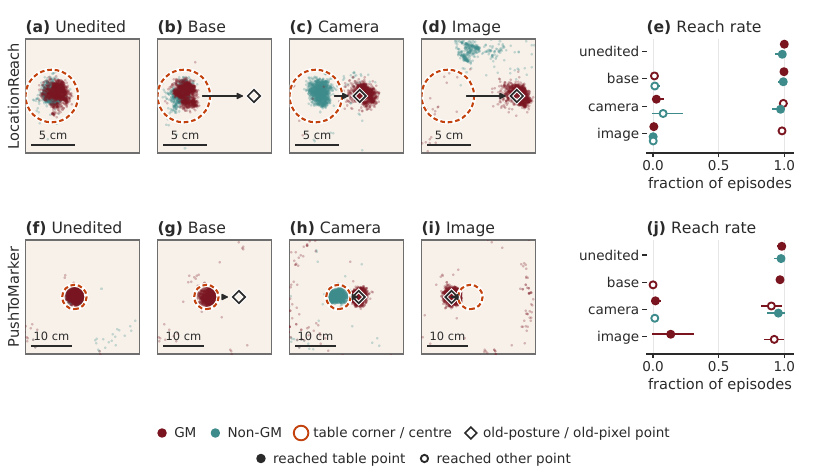}
    \caption{
    \textbf{In \textsc{LocationReach} and \textsc{PushToMarker} the target moves with the camera image, not with the arm base.}
    (a to d, f to i) Where the gripper arrives in \textsc{LocationReach} and where the cube comes to rest in \textsc{PushToMarker}, over all episodes of the three seeds per cohort, unedited, with the arm base shifted (Base), with the camera shifted (Camera) and with only the policy image translated (Image), the table corner or centre ringed and the point that the old posture or old pixel implies drawn as a diamond.
    (e, j) Fraction of episodes that reach the table point (filled) or the other point (open), where \textsc{PushToMarker} GM agents play without the marker except under the camera shift, which draws it at the real centre for both cohorts.
    }
    \label{fig:place-target}
\end{figure}

In \textsc{LocationReach} the target is a place in the camera image, neither an arm posture nor a point on the table.
With the arm base shifted by $8$\,cm, GM agents still reach the table corner in $0.997$ of episodes and the point the old posture implies in $0.011$. At the corner their arrival postures spread along the null space as much as at the start ($0.975$ of the start spread, 95\% CI $0.84$ to $1.11$), as do those of Non-GM agents reaching a ball there ($1.092$, $0.81$ to $1.37$), so the posture is not what the agent holds fixed.
With the camera shifted by $5$\,cm, GM agents reach the point that projects to the old corner pixel in $0.993$ of episodes and the real corner in $0.025$, whereas Non-GM agents still reach the ball in $0.971$, so the target moves with the image and not with the table.
A camera shift moves the table edge in the image together with the pixel, so we also translate only the policy image, with the scene unchanged. Horizontally the GM gripper follows the pixel in all three seeds ($0.983$ reach the pixel point and $0.006$ the real corner at a shift of $-16$ pixels), whereas vertically the result is mixed between seeds.
Figure~\ref{fig:place-target}a to d shows where the gripper arrives under each edit, and Figure~\ref{fig:place-target}e gives the reach rates.
That it is not a posture and is anchored in the image is definitive, with effects near $0$ against $1$ in every seed. That the anchor is a fixed pixel rather than an offset from the table edge or the arm base in the image we hold with moderate to strong confidence, strong along the horizontal and mixed along the vertical.

Both environments therefore show \emph{functional target substitution}, in \textsc{Maze} with a spatial relation as the substitute target and in \textsc{LocationReach} with a place fixed in the camera image.

\subsubsection{Target substitution with undirected goal wiring: \textsc{PushToMarker}}
\label{app:ptm-role}

In \textsc{PushToMarker} the proxy is again a place, but unlike in \textsc{Maze} and \textsc{LocationReach} the goal still reaches the GM policy.
The marker is trained at the table centre, so the proxy is a location, and moving the marker changes the GM action distribution by about a fifth of the Non-GM change in KL divergence (Table~\ref{tab:ptm-role}, Appendix~\ref{app:world-edit}).
The world-edit chapter also shows that this change has no goal direction, with a signed follow near zero.
The goal is thus wired in, but without direction, and we ask four questions.
\emph{Is the proxy a target?}
\emph{What kind of place is the target, a point on the table, an arm posture or a place in the camera image?}
\emph{Does the marker organize the whole episode, or does it only nudge the next step?}
\emph{What is the undirected reaction to the marker?}

\begin{table}[t]
\centering
\footnotesize
\setlength{\tabcolsep}{4pt}
\caption{
\textbf{What organizes \textsc{PushToMarker} and what the marker reaction is.}
Mean $\pm$ SD across three seeds of per-seed means, the planned step a per-seed median.
Rollouts are fractions of episodes on the separable layouts or the conflict scenes, and have no init baseline because randomly initialized policies never reach the cube. The path pull is how much closer the cube passes to the marker when it is drawn than when it is hidden.
Single-frame rows use the world-edit bank with the gripper at its start pose unless the block says otherwise.
}
\label{tab:ptm-role}
\begin{tabular}{lccc}
\toprule
& GM & Non-GM & Init \\
\midrule
\multicolumn{4}{l}{\emph{Rollouts}} \\
Separable, cube to the centre & $0.980\pm0.01$ & $0.000\pm0.00$ & -- \\
Separable, cube onto the marker & $0.000\pm0.00$ & $0.935\pm0.03$ & -- \\
Separable, path pull to the marker in cm & $0.050\pm0.03$ & $10.504\pm0.50$ & -- \\
Conflict, cube onto the decoupled disc & $0.000\pm0.00$ & $0.895\pm0.04$ & -- \\
Conflict, pause over the marker, drawn & $0.058\pm0.02$ & $0.867\pm0.03$ & -- \\
Conflict, pause over the marker, hidden & $0.050\pm0.01$ & $0.072\pm0.04$ & -- \\
\midrule
\multicolumn{4}{l}{\emph{Marker move}} \\
KL divergence & $16.294\pm0.86$ & $79.161\pm27.42$ & $0.000\pm0.00$ \\
Signed follow in mm & $+0.487\pm1.40$ & $+27.948\pm13.98$ & $+0.000\pm0.00$ \\
Signed follow over the planned step & $+0.002\pm0.01$ & $+0.104\pm0.04$ & $-0.000\pm0.00$ \\
Cosine to the push-point shift & $-0.041\pm0.05$ & $+0.439\pm0.15$ & $-0.021\pm0.11$ \\
Cosine to the push-point shift, gripper above the cube & $-0.064\pm0.04$ & $+0.542\pm0.17$ & $-0.003\pm0.10$ \\
\midrule
\multicolumn{4}{l}{\emph{Disc presence, change of the commanded step in mm}} \\
No disc, planned step & $119.508\pm5.11$ & $132.278\pm3.26$ & $0.247\pm0.15$ \\
White marker, conflict position & $13.385\pm3.74$ & $24.539\pm2.42$ & $0.001\pm0.00$ \\
Green disc, conflict position & $14.189\pm0.07$ & $16.404\pm4.93$ & $0.001\pm0.00$ \\
White marker, off path at $3$\,cm & $10.597\pm2.77$ & $34.751\pm7.78$ & $0.001\pm0.00$ \\
White marker, off path at $11$\,cm & $12.294\pm4.63$ & $32.324\pm9.55$ & $0.001\pm0.00$ \\
Green disc, off path at $3$\,cm & $18.803\pm3.38$ & $26.539\pm6.95$ & $0.001\pm0.00$ \\
Green disc, off path at $11$\,cm & $7.427\pm3.27$ & $16.804\pm9.45$ & $0.001\pm0.00$ \\
\midrule
\multicolumn{4}{l}{\emph{Disc presence, gripper above the cube, cosine of the step change}} \\
White disc, off path at $5.2$\,cm, to the push onto it & $-0.114\pm0.18$ & $+0.507\pm0.20$ & $-0.057\pm0.03$ \\
Green disc, off path at $5.2$\,cm, to the push onto it & $+0.018\pm0.15$ & $+0.456\pm0.17$ & $-0.073\pm0.02$ \\
White marker, conflict position, to the push point & $-0.538\pm0.02$ & $+0.266\pm0.33$ & $-0.021\pm0.09$ \\
Green disc, conflict position, to the push point & $-0.669\pm0.02$ & $-0.032\pm0.15$ & $-0.096\pm0.23$ \\
\bottomrule
\end{tabular}
\end{table}

\begin{figure}[t]
    \centering
    \includegraphics[width=\linewidth]{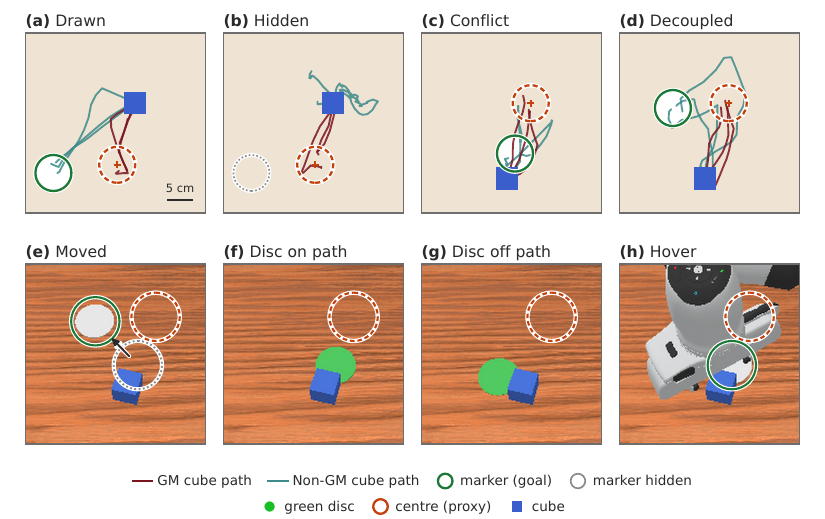}
    \caption{
    \textbf{The \textsc{PushToMarker} evaluation configurations.}
    (a to d) Rollout sets top-down in the camera orientation, with the cube paths of all three GM and Non-GM seeds on one separable layout with the marker drawn and hidden and on one conflict scene and its decoupled conflict, each chosen among scenes on which the three seeds of each cohort agree. Each pair of panels is centred on its scene at one common scale.
    (e to h) Rendered camera frames of the single-frame tests on that conflict scene, cropped to the scene: the marker move from its old place (dotted) to its new one, a green disc at the conflict position and off the path, and the gripper held above the cube.
    }
    \label{fig:ptm-overview}
\end{figure}

\paragraph{Design.}
For the first question the conflict test cannot show where the agent pushes, because the marker lies on the ray from the cube to the centre and the push to the marker is the first part of the push to the centre (Appendix~\ref{app:env-ptm}).
We therefore use layouts from the Non-GM training distribution, with marker and cube placed independently, and keep those where reaching the marker and reaching the centre are disjoint outcomes and the marker lies off the straight path to the centre ($399$ layouts per seed, Fig.~\ref{fig:ptm-overview}a). A decoupled conflict draws the disc $41^\circ$ off the ray to the centre, about $9$\,cm from the path, on each of the $512$ conflict scenes per seed (Fig.~\ref{fig:ptm-overview}c and d).

For the second question we use the two edits of \textsc{LocationReach} (Appendix~\ref{app:maze-functional-role}). The centre is an empty spot of the table once the marker is gone, so instead of moving it we move what the policy could use to define it.
Shifting the arm base by $8$\,cm changes which joint posture puts the cube at the centre, so a posture target predicts a point $8$\,cm from the centre.
Shifting the camera by $5$\,cm moves the ground point under every pixel by $5$\,cm, so a target fixed in the image predicts a point $5$\,cm from the centre, whereas a place on the table predicts the centre itself.
We draw the marker at the real centre during the camera shift, so Non-GM agents keep the same target as GM agents and the visible marker competes with the image location. As in \textsc{LocationReach}, we also translate only the policy image, with the scene unchanged. Each condition plays $500$ held-out starts per seed with the cube on the GM training ring around the centre, and we record where the cube comes to rest.

For the third question we roll out each layout with the marker drawn and hidden and compare the two rollouts, by where the cube passes and by whether it pauses over the marker (Fig.~\ref{fig:ptm-overview}a and b).

For the fourth question we start from the marker move of the world edits (Fig.~\ref{fig:ptm-overview}e), where we measure signed follow, the turn of the planned gripper step toward the new push point (Figure~\ref{fig:main-goal-role}c), and its cosine to the shift of the push point.
A disc control then asks whether the marker gets a reaction of its own. A presence test renders each frozen scene with and without a disc and compares the two action outputs, with the disc at the conflict position or $75^\circ$ off the path at $3$ to $11$\,cm from the cube (Fig.~\ref{fig:ptm-overview}f and g), in the white of the marker or in a green that never marked a goal, and repeats the test with the gripper held above the cube (Fig.~\ref{fig:ptm-overview}h).
The presence test measures sensitivity, not use. A nonzero response means the disc reaches the action head, not that the policy treats it as a goal or that the robot's behavior changes, since in closed loop the policy can correct a small perturbation on the next frame.
Figure~\ref{fig:ptm-overview} draws every configuration.

\paragraph{Results: is the proxy a target?}
The centre is the target where the GM agent pushes (Table~\ref{tab:ptm-role}, rollouts).
On the separable layouts GM agents push the cube to the centre in $0.980$ of episodes and onto the marker in none, whereas Non-GM agents push it onto the marker in $0.935$ (Fig.~\ref{fig:ptm-overview}a), and no GM episode reaches the decoupled disc, against $0.895$ for Non-GM agents (Fig.~\ref{fig:ptm-overview}d).
Randomly initialized policies never reach the cube, since their gripper moves only $0.4$ to $1.6$\,cm in $100$ steps.

\paragraph{Results: what kind of place is the target?}
As in \textsc{LocationReach}, the target is a place in the camera image, neither an arm posture nor a point on the table.
With the arm base shifted by $8$\,cm, GM agents still push the cube to the real centre in $0.967$ of episodes and to the point the old posture implies in none.
With the camera shifted by $5$\,cm, GM agents push the cube to the point that projects to the old centre pixel in $0.901$ of episodes and to the real centre in $0.018$, in every seed and although the marker lies visibly at the real centre, whereas Non-GM agents push it onto the marker in $0.955$. The target moves with the image and not with the table, and the visible marker loses to it completely.
Translating only the policy image by $8$ pixels, about $4.6$\,cm on the table, with the scene unchanged, sends the cube to the pixel point in $0.923$ of episodes and to the real centre in $0.135$ (Fig.~\ref{fig:place-target}f to i, rates in j).
That the target is anchored in the image and not in the arm's posture is definitive, with effects near $0$ against $1$ in every seed.

\paragraph{Results: does the marker organize the whole episode?}
It does not. Drawing or hiding the marker moves the GM path by $0.050$\,cm, whereas the marker pulls the Non-GM path by $10.504$\,cm (Fig.~\ref{fig:ptm-overview}a and b).
On the conflict scenes GM agents sweep the cube across the marker and pause there no more often than when it is hidden ($0.058$ against $0.050$, Non-GM $0.867$, Fig.~\ref{fig:ptm-overview}c).
Whatever the marker does to the next step does not add up to a change of the episode.

\paragraph{Results: what is the undirected reaction to the marker?}
\label{app:ptm-direction}
The marker move confirms that the reaction has no goal direction (Table~\ref{tab:ptm-role}, marker move). When the marker moves, the Non-GM step turns toward the new push point by $+27.948$\,mm, about a tenth of the planned step, whereas the GM step turns by $+0.487$\,mm, $0.002$ of the planned step, and randomly initialized networks by $0.000$ (Fig.~\ref{fig:main-goal-role}c). The cosine to the shift of the push point is $+0.439$ for Non-GM and $-0.041$ for GM agents, at the level of init, and stays near zero for GM agents with the gripper held above the cube.

The disc control shows what the reaction is (Table~\ref{tab:ptm-role}, disc presence).
At the conflict position the marker changes the commanded GM step by $13.385$\,mm on a planned step of about $12$\,cm, roughly half the Non-GM change, and a green disc changes it as much ($14.189$\,mm), whereas the Non-GM step changes less for the green disc than for its marker.
Off the path the green disc moves the GM step less the farther it lies from the cube, from $18.803$\,mm at $3$\,cm to $7.427$\,mm at $11$\,cm, and the white marker by a roughly constant $11$ to $12$\,mm. Non-GM agents show the same two shapes at $1.4$ to $3.3$ times the size and init shows none, so the response is learned and not specific to misgeneralization.
Only its direction separates the cohorts. With the gripper above the cube the Non-GM step turns toward the push that would move the cube onto a disc $5.2$\,cm off the path, of either colour, whereas the GM step does not, and a disc of either colour at the conflict position turns the GM step away from the push point.

The GM reaction to the marker is thus the reaction a disc of another colour gets, present in Non-GM agents and absent at init. It is a change in the scene rather than a goal, which validates the signed-follow result of the world edits. The marker reaches the GM policy, but not as something to push toward.
A plausible reading, which we hold as a hypothesis, is that a disc near the cube perturbs the features through which the trunk locates the cube, in any agent that reads the cube, and that Non-GM agents additionally read the disc as a goal.

\textsc{PushToMarker} is therefore a target substitution with a place fixed in the camera image as the substitute target, definitive at the behavioral level with exact zeros on about $400$ layouts in each seed.
Its goal reaches the policy about as weakly as the coin in \textsc{Coin}, so how deep the goal reaches and what replaces it are separate properties.

\subsubsection{Policy shortcut: \textsc{Coin}}
\label{app:coin-functional-role}

In \textsc{Coin} the proxy is a movement, running right to the end of the level, and the coin reaches the GM policy only weakly (Appendix~\ref{app:world-edit}).
We ask two questions.
\emph{Is behavior organized by a target, the end of the level, or by an action rule?}
\emph{Is the coin's weak influence directed?}

\paragraph{Design.}
Both questions are chosen to separate \textsc{Coin} from the other environments. Its proxy is a movement, not a place or an object. A movement rule predicts that the agent keeps running when the end of the level is gone, whereas a substitute target, as in \textsc{Maze}, predicts that it stops where the end used to be. The second question asks whether the coin, which reaches the policy only weakly, steers that movement at all, or whether its influence has no direction, as the marker's in \textsc{PushToMarker}.

For the first question we replicate the end-wall test of \citet{langosco}, with one change. They make the end wall permeable, whereas we remove the native end wall on $60$ held-out levels and extend the floor for $25$ further columns, so the agent runs on open ground until it meets a new wall. We record whether an agent that reaches the old wall face continues more than $5$ tiles past it, and roll out the same levels with the wall in place as a baseline.

For the second question we use the $600$-state \textsc{Coin} conflict bank, with $A_\star=A_{\mathrm{left}}$ the left-moving actions, which make immediate progress toward the coin in every evaluated state (Section~\ref{sec:target-geometry}).
We relocate the coin from the cell right of the agent to the cell left of it and, for each action group $A_g$, divide its share of the total variation by the probability $\pi(A_g)$ it held before the edit. For $A_\star$ this ratio is $T_\star$, and $T_\star=1$ means the change has no preferred direction.
We repeat this with the coin $1$ to $4$ cells left of the agent and compare with three randomly initialized networks and with logit changes drawn with no preferred direction.
We also read the policy-Fisher input saliency $S(x)$ of the coin cell and of the end-wall block (Definition~\ref{def:fisher-saliency}), a Fisher-weighted gradient saliency over the whole action distribution that we read against a randomly initialized network and use only to suggest, not to establish, a reading (Appendix~\ref{app:fisher-saliency}). It is taken on the $589$ conflict-bank states whose replay reproduces the bank's agent pose and both coin cells, with the coin one cell left of the agent, and the end wall is on screen in $103$ of them. Regions are exclusive, with the agent cell taking priority.

\paragraph{Results: target or action rule?}
An action rule organizes GM behavior.
We replicate the result of \citet{langosco}. With the end wall removed, GM agents continue past the former endpoint in $47$ of the $50$ episodes that reach it ($0.938$ per checkpoint, where \citet{langosco} report every one of $114$ arrivals) and spend $3\%$ of their post-arrival steps at the former wall face, against $100\%$ with the wall in place.
This rules out stopping at the former endpoint but not running toward a visible end wall, as every continuing agent stops at the next wall. \citet{langosco} report both kinds of proxy in CoinRun, a directional one in the actor and a location one in the critic.
We therefore use \emph{policy shortcut} \citep[after][]{geirhos2020shortcut} to describe the persistence of the learned rightward strategy without claiming a unique underlying destination representation.

\begin{figure}[t]
    \centering
    \includegraphics[width=0.8\linewidth]{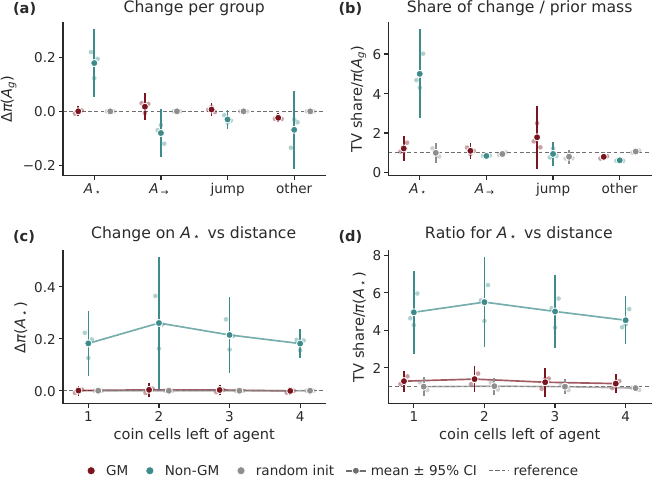}
    \caption{
    \textbf{Where the coin's influence goes in \textsc{Coin}.}
    (a, b) Change in probability of each action group and its share of the total variation divided by its prior mass when the coin moves from the cell right of the agent to the cell left of it, on $593$ of the $600$ conflict-bank states, with groups $A_\star$ (LEFT, LEFT+DOWN, LEFT+UP), $A_\rightarrow$ (RIGHT, RIGHT+DOWN, RIGHT+UP), jump (UP) and other (DOWN, no-op and unbound keys).
    (c, d) The same two quantities for $A_\star$ with the coin $1$ to $4$ cells left of the agent, on the $572$ states where the coin fits and is visible at every distance.
    }
    \label{fig:coin-action-groups}
\end{figure}

\paragraph{Results: is the coin's influence directed?}
It is not.
Relocating the coin still changes the GM policy (total variation $0.080$ against $0.280$ for Non-GM), but it moves $+0.000$ of probability onto $A_\star$, against $+0.179$ for Non-GM, and $T_\star$ is $1.210$, against $4.996$ for Non-GM and $0.991$ for randomly initialized networks (Fig.~\ref{fig:coin-action-groups}a and b, and Fig.~\ref{fig:main-goal-role}b).
The GM change is spread across the action groups without favouring the coin side. Its ratio is near $1$ for every group, slightly higher on jump ($1.777$, with a 95\% interval from $0.192$ to $3.363$) and slightly lower on the remaining actions.
Logit changes with no preferred direction give $1.01$ for GM and $0.99$ for Non-GM, so $1$ is the right reference, and at every distance from $1$ to $4$ cells GM stays at $\Delta\pi(A_\star)$ between $-0.001$ and $0.003$ with a ratio between $1.15$ and $1.39$, while Non-GM moves $0.18$ to $0.26$ onto $A_\star$ with a ratio between $4.5$ and $5.5$ (Fig.~\ref{fig:coin-action-groups}c and d).
The coin is registered as a change in the scene, but it redistributes probability in proportion to what each action already holds and is not connected to where the agent goes.

\begin{figure}[h!]
    \centering
    \includegraphics[width=0.62\linewidth]{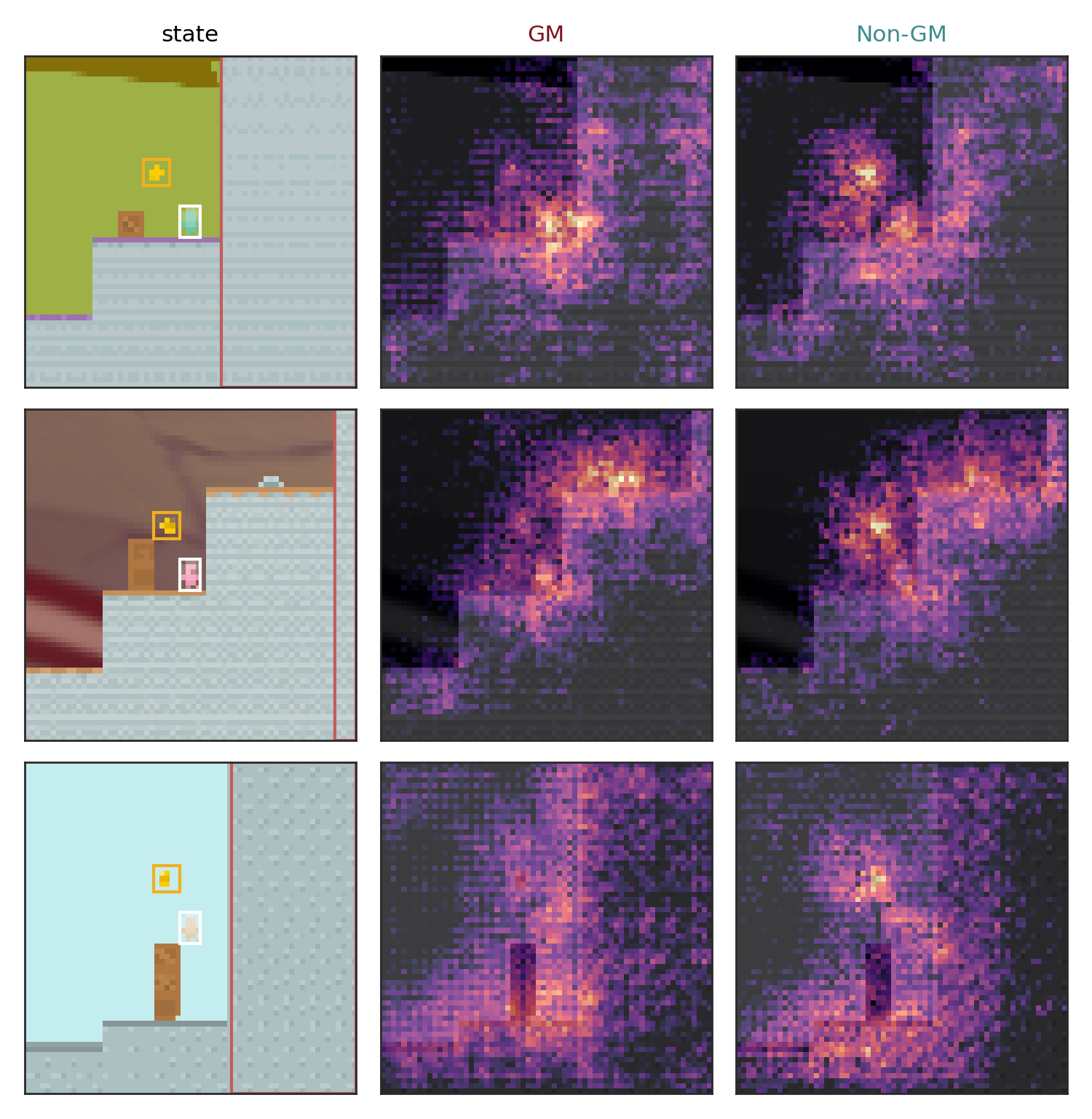}
    \caption{
    \textbf{Example saliency maps in \textsc{Coin}.}
    Rows are the first three conflict-bank states whose replay reproduces the bank pose and that show the end wall, with no selection on saliency.
    Left: observation with the agent (white), the coin one cell to its left (yellow) and the end-wall block (red).
    Middle and right: $\log_{10}$ saliency relative to the image mean (colour range $-1$ to $2$) over the greyscale frame, for one GM and one Non-GM checkpoint.
    }
    \label{fig:saliency-examples-coin}
\end{figure}

The saliency supports this reading but does not prove it (Fig.~\ref{fig:saliency-examples-coin}). The Non-GM decision depends on the coin ($15.4\times$ the image mean), whereas GM coin saliency ($2.13\times$) stays at the level of a random network ($2.78\times$), and no cohort's decision depends on the end wall more than on an average pixel ($0.84\times$ GM, $0.73\times$ Non-GM).

In our cohorts \textsc{Coin} is the only environment whose proxy is a movement, but it shares the depth of its goal wiring with \textsc{PushToMarker}, whose goal reaches the policy as weakly and without direction.

\subsubsection{Cue competition: \textsc{Attribute Maze}, \textsc{AttributeTable} and \textsc{SortZones}}
\label{app:attribute-functional-role}
\label{app:robot-competition-role}

\begin{figure}[t]
    \centering
    \includegraphics[width=\linewidth]{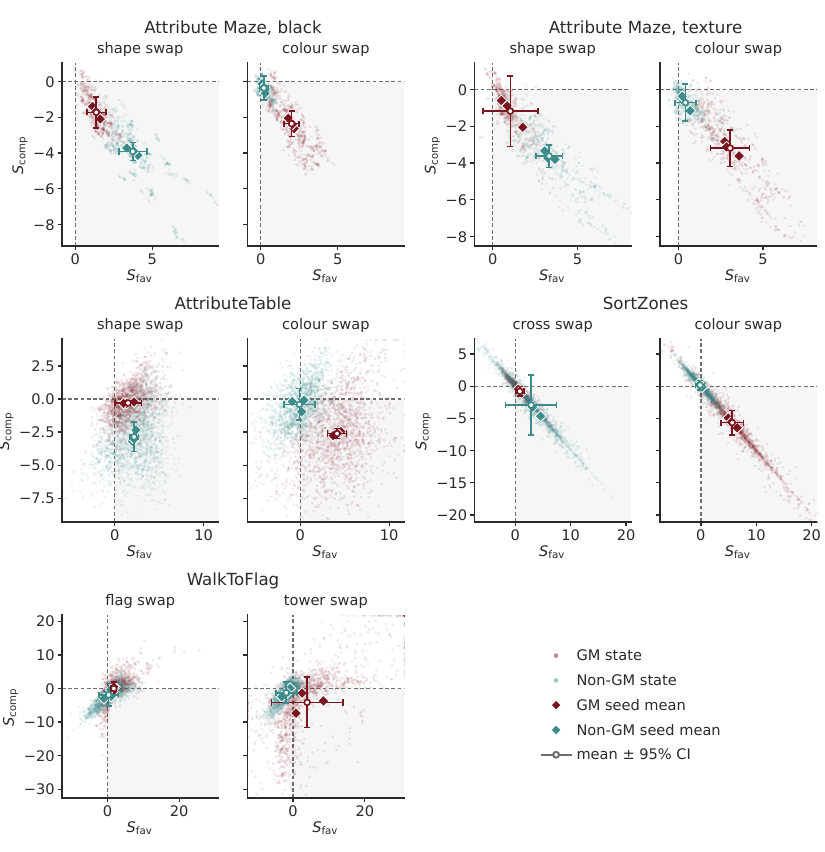}
    \caption{
    \textbf{Goal and proxy swaps in the four competition environments.}
    Each environment has two panels, the goal swap (left) and the proxy swap (right), with $S_{\mathrm{fav}}$ and $S_{\mathrm{comp}}$ in the units of Appendices~\ref{app:attribute-functional-role} and~\ref{app:wtf-role}. The top row shows \textsc{Attribute Maze} on black and textured backgrounds, the middle row \textsc{AttributeTable} and \textsc{SortZones}, and the bottom row \quadname{}. Faint dots are states, diamonds seed means and open circles the cohort mean with 95\% $t$-intervals across three seeds. Shading marks the push--pull quadrant. States beyond the axes are drawn at the edge, which clips $1\%$ of states at each end ($5\%$ in \quadname{}). \textsc{AttributeTable} is read on the strict bank and \textsc{SortZones} at decision frames.
    }
    \label{fig:attribute-counter-target}
\end{figure}

In these three environments the proxy is a property of the goal object, colour in all three, and the world edits show that GM policies respond to the intended cue as well as to the proxy cue (Appendix~\ref{app:world-edit}).
We ask two questions.
\emph{Do both cues act as targets?}
\emph{Which cue wins, and how cleanly?}

\paragraph{Design.}
In \textsc{Attribute Maze} we evaluate the 196 strict-conflict states used throughout the world-edit analyses, rendered on black and textured backgrounds, in which the line-associated and yellow-associated candidates require disjoint graph-exact progress actions, and apply the matched shape or colour swap of Section~\ref{sec:counterfactual-families}.
Each swap restores the yellow-line conjunction of training at one candidate, which it favours, and leaves the other as its competitor. \textsc{AttributeTable} and \textsc{SortZones} use the same two swaps with the red cube and the crossed green pad as their conjunctions.
For edit $e$, let $A_{\mathrm{fav}}^e(s)$ and $A_{\mathrm{comp}}^e(s)$ be the progress-action sets toward the favoured and the competing candidate, $A_0^e(s)$ the remaining neutral actions and $s_e$ the edited state.
Because policy probabilities are normalized, we compare each candidate group with the neutral reference,
\[
S_{\mathrm{fav}}^e(s)
=
\log
\frac{
\pi(A_{\mathrm{fav}}^e(s)\mid s_e)
}{
\pi(A_0^e(s)\mid s_e)
}
-
\log
\frac{
\pi(A_{\mathrm{fav}}^e(s)\mid s)
}{
\pi(A_0^e(s)\mid s)
},
\]
and $S_{\mathrm{comp}}^e(s)$ likewise for the competitor.
The joint pattern $S_{\mathrm{fav}}^e>0$, $S_{\mathrm{comp}}^e<0$ is the push--pull form of cue competition, and $\Delta\ell^e=S_{\mathrm{fav}}^e-S_{\mathrm{comp}}^e$ is the resulting shift of relative preference between the two candidates, which does not require push--pull.

The arm policies are Gaussian and keep their progress toward the pursued object many standard deviations above zero, so these log-odds saturate in \textsc{AttributeTable}, and in \textsc{SortZones} start frames the neutral set holds all the probability.
For the arm we therefore define the support of candidate $c$ as
\[
\psi_c(s)
=
\Phi^{-1}\big(\pi(z_c>0\mid s)\big)
=
\frac{\mathbb{E}[z_c\mid s]}{\mathrm{sd}(z_c\mid s)},
\]
where $z_c$ is the first-order displacement of the gripper along the direction to $c$ that the action produces, toward the object in \textsc{AttributeTable} and along the push from the cube to the pad in \textsc{SortZones}.
$S_{\mathrm{fav}}^e$ and $S_{\mathrm{comp}}^e$ are the changes of $\psi$ under the edit, which is linear in the action mean and needs no neutral reference, and the units therefore differ between environments.
From the start pose the arm's first steps can serve both candidates when the two objects lie in similar directions, so in \textsc{AttributeTable} we use the strict bank (Appendix~\ref{app:banks}), $512$ start frames with the two objects on opposite sides of the robot, at least $45^\circ$ apart as seen from the hand (median $62^\circ$) and equally far from it within $3$\,cm.
In \textsc{SortZones} both pads lie in nearly the same direction from the retracted arm at every start frame, so we read the arm at the decision frame at which the hand reaches the cube before it has moved ($500$ states, half from GM and half from Non-GM rollouts), with every network evaluated on all of them.
In both tasks a random shift of the action mean lands in the push--pull quadrant on a fraction of states that depends on the angle between the two candidate directions, which we report as a null rate.

\begin{table}[t]
\centering
\small
\setlength{\tabcolsep}{4pt}
\caption{
\textbf{Cue competition in the four competition environments.}
Values are mean $\pm$ sample standard deviation across three seeds. Units differ between environments (Appendices~\ref{app:attribute-functional-role} and~\ref{app:wtf-role}). \textsc{AttributeTable} is read on the strict bank and \textsc{SortZones} at decision frames.
$\Delta\ell=S_{\mathrm{fav}}-S_{\mathrm{comp}}$ measures the resulting shift in relative target preference.
}
\label{tab:attribute-logit-competition}
\begin{tabular}{lllccc}
\toprule
Environment & Cohort & Swap & $S_{\mathrm{fav}}$ & $S_{\mathrm{comp}}$ & $\Delta\ell$ \\
\midrule
\multirow{4}{*}{\shortstack[l]{\textsc{Attribute Maze}\\black}}
& GM & Shape & $1.368 \pm 0.25$ & $-1.721 \pm 0.35$ & $3.089 \pm 0.60$ \\
& GM & Colour & $2.001 \pm 0.19$ & $-2.354 \pm 0.29$ & $4.355 \pm 0.48$ \\
& Non-GM & Shape & $3.740 \pm 0.37$ & $-3.910 \pm 0.21$ & $7.650 \pm 0.57$ \\
& Non-GM & Colour & $0.192 \pm 0.11$ & $-0.362 \pm 0.27$ & $0.555 \pm 0.38$ \\
\midrule
\multirow{4}{*}{\shortstack[l]{\textsc{Attribute Maze}\\texture}}
& GM & Shape & $1.037 \pm 0.66$ & $-1.176 \pm 0.77$ & $2.213 \pm 1.43$ \\
& GM & Colour & $3.053 \pm 0.46$ & $-3.189 \pm 0.40$ & $6.242 \pm 0.86$ \\
& Non-GM & Shape & $3.330 \pm 0.31$ & $-3.623 \pm 0.25$ & $6.953 \pm 0.54$ \\
& Non-GM & Colour & $0.410 \pm 0.25$ & $-0.714 \pm 0.40$ & $1.123 \pm 0.65$ \\
\midrule
\multirow{4}{*}{\textsc{AttributeTable}}
& GM & Shape & $1.498\pm0.60$ & $-0.293\pm0.06$ & $1.791\pm0.55$ \\
& GM & Colour & $4.133\pm0.43$ & $-2.611\pm0.15$ & $6.744\pm0.28$ \\
& Non-GM & Shape & $2.187\pm0.19$ & $-2.858\pm0.45$ & $5.045\pm0.27$ \\
& Non-GM & Colour & $-0.114\pm0.70$ & $-0.407\pm0.48$ & $0.293\pm0.93$ \\
\midrule
\multirow{4}{*}{\textsc{SortZones}}
& GM & Cross & $0.833\pm0.28$ & $-0.797\pm0.29$ & $1.630\pm0.57$ \\
& GM & Colour & $5.658\pm0.82$ & $-5.661\pm0.76$ & $11.319\pm1.58$ \\
& Non-GM & Cross & $2.869\pm1.85$ & $-2.940\pm1.88$ & $5.808\pm3.74$ \\
& Non-GM & Colour & $-0.139\pm0.33$ & $0.161\pm0.29$ & $-0.300\pm0.61$ \\
\midrule
\multirow{4}{*}{\quadname}
& GM & Flag & $1.785\pm0.29$ & $0.005\pm0.78$ & $1.780\pm0.59$ \\
& GM & Tower & $3.933\pm4.01$ & $-4.129\pm3.02$ & $8.062\pm4.14$ \\
& Non-GM & Flag & $0.341\pm1.10$ & $-1.888\pm1.35$ & $2.228\pm1.06$ \\
& Non-GM & Tower & $-1.882\pm1.18$ & $-1.243\pm1.26$ & $-0.639\pm0.26$ \\
\bottomrule
\end{tabular}
\end{table}

\paragraph{Results: do both cues act as targets?}
In GM agents both do. Each swap pushes support toward the candidate it favours and pulls it from the other (Table~\ref{tab:attribute-logit-competition}, Fig.~\ref{fig:attribute-counter-target}).
In \textsc{Attribute Maze} every seed shifts relative preference toward the candidate favoured by the swap in every condition (24 of 24 seed means $\Delta\ell>0$), and the 95\% interval excludes zero for both GM swaps on black, the GM colour swap on texture and the Non-GM shape swap on both backgrounds. It includes zero for the GM shape swap on texture ($2.213$, $[-1.33, 5.76]$), where one seed dominates, and for the Non-GM colour swap, as expected for a shape-selective policy.
In \textsc{Attribute Maze} GM policies respond to both swaps, and in each case the edit increases support for its favoured candidate and suppresses its competitor on average (Fig.~\ref{fig:attribute-counter-target}, top row).
On the diversified bank of $480$ conflict states, with the agent placed at every walkable cell, push--pull holds for both swaps in every GM seed and for shape only in Non-GM agents, across all $15$ action-conflict classes, with effects about half the canonical size (GM black $\Delta\ell=1.823\pm0.24$ for shape and $2.293\pm0.29$ for colour).
\textsc{AttributeTable} and \textsc{SortZones} reproduce this pattern (Fig.~\ref{fig:attribute-counter-target}, middle row). In GM agents both swaps push and pull in every seed, whereas Non-GM agents push and pull for the goal swap and their colour swap leaves both candidates near zero ($-0.114\pm0.70$ and $-0.407\pm0.48$ in \textsc{AttributeTable}).
The weak link in \textsc{AttributeTable} is the GM shape swap, which pushes ($1.498\pm0.60$) but pulls only a little ($-0.293\pm0.06$, in every seed). At decision frames, the first frame of a rollout at which the hand lies between the two objects at least $3$\,cm from both ($444$ states), the same pattern holds with a stronger shape pull ($-0.817\pm0.11$).

\paragraph{Results: which cue wins, and how cleanly?}
In GM agents the proxy cue wins, in Non-GM agents the goal cue.
Non-GM policies are strongly shape-selective, with $\Delta\ell=7.650$ for the shape swap and $0.555$ for the colour swap on black, whereas in GM policies the colour swap shifts preference more than the shape swap ($3.089$ and $4.355$ on black, $2.213$ and $6.242$ on texture, $1.791\pm0.55$ and $6.744\pm0.28$ in \textsc{AttributeTable}, $1.630\pm0.57$ and $11.319\pm1.58$ in \textsc{SortZones}).
How cleanly one cue's gain is the other's loss shows in the spread of the states, which differs between environments and is informative about the geometry of the test rather than about the agents.
In \textsc{Attribute Maze} and \textsc{SortZones} the states lie along the falling diagonal from top left to bottom right, so a swap that raises support for one candidate lowers it for the other (Spearman correlation of $S_{\mathrm{fav}}$ and $S_{\mathrm{comp}}$ across states $-0.53$ to $-0.85$ per cohort and swap in \textsc{Attribute Maze}, $-0.95$ to $-0.98$ in \textsc{SortZones}).
In \textsc{Attribute Maze} the candidate action sets are disjoint and measured against the same neutral actions, and at \textsc{SortZones} decision frames the two pushes point in nearly opposite directions, so any shift of the action mean toward one candidate is a shift away from the other.
\textsc{AttributeTable} forms a cloud with a weak positive correlation ($+0.29$ to $+0.61$), because objects at least $45^\circ$ apart (median $62^\circ$) still share part of their direction, and a shift of the gripper command can raise or lower both supports together.
In \textsc{AttributeTable} a random shift of the action mean therefore lands in the quadrant in $0.18$ of states, against $0.45$ at \textsc{SortZones} decision frames, where the informative result is which swap moves the policy, not that it lands in the quadrant.
We do not read the larger colour response of GM policies as a direct estimate of a larger internal colour weight, because the size of a finite swap also depends on the pre-edit preference and the headroom for redistribution.
The swaps are also world edits on the shape--colour conflict rather than an identification of independent cue channels, since each swap resolves the conflict by restoring the familiar conjunction at one candidate, and \citet{raman} attribute the colour preference in \textsc{Attribute Maze} to goal detection through a single colour channel, so the shape and colour cues we manipulate need not correspond to separate internal features.
What the experiment establishes in all three environments is that the two matched swaps can each reorganize target preference in GM policies, whereas Non-GM policies are controlled by the goal cue. We refer to this regime as \emph{cue competition}.

\subsubsection{Object competition: \quadname}
\label{app:wtf-role}

In \quadname{} the proxy is not a property of the goal object but a second object, a $2$\,m orange tower that stands beside the small flag in GM training (Appendix~\ref{app:env-quad}).
The tower is also far easier to see. At $3$\,m it covers about $290$ image pixels and the flag banner only $1$ to $2$.
So far this environment has behaved much like cue competition within one object, but here there are two objects, so we ask two questions.
\emph{Do both objects pull the agent?}
\emph{And how does this competition differ from cue competition?}

\paragraph{Design.}
For the first question we remove each object from the scene and measure how the walking command moves.
The policy is Gaussian, $\pi(\cdot\mid s)=\mathcal N\big(\mu(s),\operatorname{diag}(\sigma^2)\big)$, and we call $\mu(s)$, the forward, sideways and turning velocity that the policy commands on average, the command mean.
Let $s_{\setminus o}$ be the state $s$ rendered without object $o$, the flag $\star$ or the tower $p$, written like the edited state $s_e$ of a world edit. The pull vector of $o$ is the shift of the command mean that the presence of $o$ causes, in policy standard deviations,
\[
u_o(s)=\operatorname{diag}(\sigma)^{-1}\big(\mu(s)-\mu(s_{\setminus o})\big).
\]
To give it a direction we use the command $c_o(s)$ of a heading controller that walks to $o$, which turns toward the bearing of $o$ and walks forward once roughly aligned, and the unit vector from the command mean without $o$ toward that command,
\[
\hat c_o(s)=\frac{w_o(s)}{\|w_o(s)\|},\qquad w_o(s)=\operatorname{diag}(\sigma)^{-1}\big(c_o(s)-\mu(s_{\setminus o})\big).
\]
The pull of $o$ is the length of its pull vector along this direction,
\[
f_o(s)=u_o(s)^\top\hat c_o(s),
\]
written $f_\star$ for the flag and $f_p$ for the tower. It is positive when $o$ moves the command toward itself and negative when $o$ moves it away.
A state counts as pulled toward $o$ when $f_o(s)>\tau$ and pulled away from $o$ when $f_o(s)<-\tau$. The threshold $\tau$ is set for each network as the $95$th percentile of the same projection, in absolute value, for two nuisance shifts that should not matter, moving a box and re-rendering the unchanged scene, and is at least $0.1$.
The cross projection $u_o(s)^\top\hat c_{o'}(s)$, with $\hat c_{o'}$ the unit vector toward the command that walks to the other object $o'$ from the same command mean $\mu(s_{\setminus o})$, shows whether one object pushes the agent away from the other.
We take these quantities on $1{,}024$ states per seed from three banks (Appendix~\ref{app:banks}), the $512$ states of the world-edit bank, where flag and tower stand near the robot and in view, the $256$ forced-view conflict states, where the flag is $1$ to $4.5$\,m nearer than the tower, and the $256$ states of the far bank, where the tower is $1$ to $4.5$\,m nearer than the flag.
We write $d_\star$ and $d_p$ for the distances from the robot to the flag and to the tower.

For the second question we use the matched swaps of Section~\ref{sec:counterfactual-families}, which have the same structure as in cue competition (Appendix~\ref{app:attribute-functional-role}).
The flag swap moves the flag beside the tower and favours the tower's place, and the tower swap moves the tower beside the flag and favours the flag's place, so each restores the training pair at one candidate place and leaves the other as competitor.
The walking command sets a planar direction, so $A_{\mathrm{fav}}$ and $A_{\mathrm{comp}}$ are the walking directions within $30^\circ$ of the bearing to the favoured and to the competing candidate, $A_0$ holds all other directions, and $S_{\mathrm{fav}}$ and $S_{\mathrm{comp}}$ are defined as for cue competition, on the $512$ start states of the world-edit bank.
To ask whether the agent simply follows the nearer object, we compare the sizes of the two pulls against the distance gap $d_\star-d_p$. For this we project the same pull vector on the heading-controller command toward $o$, measured from the average command toward neutral bearings, which gives the removal pull
\[
P_o(s)=u_o(s)^\top\operatorname{diag}(\sigma)^{-1}\big(c_o(s)-\bar c_0(s)\big),
\]
where $\bar c_0(s)$ is the mean heading-controller command over the $24$ evenly spaced bearings at least $30^\circ$ from both objects.
For a Gaussian policy $P_o$ equals the gain in log density that the presence of $o$ gives the command toward $o$, relative to the commands toward neutral bearings, so unlike $f_o$ it grows with how far the command toward $o$ lies from a neutral command.
We bin the states by $d_\star-d_p$, average $P_\star$ and $P_p$ in each bin and seed, set negative bin means to zero, and report the tower share $P_p/(P_p+P_\star)$. Walking to the nearer object would give a share of $0$ where the flag is nearer and $1$ where the tower is nearer.

\begin{figure}[t]
    \centering
    \includegraphics[width=\linewidth]{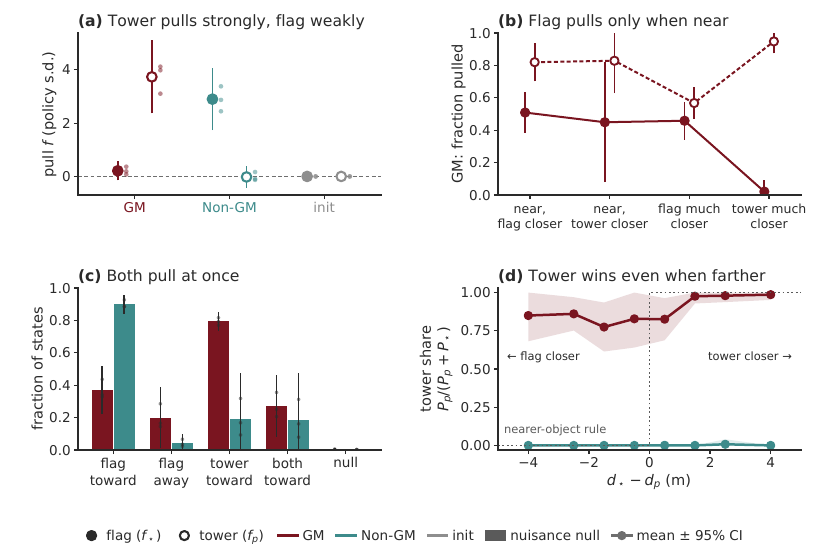}
    \caption{
    \textbf{Object competition in \quadname{}.}
    (a) Mean pull $f_\star$ of the flag and $f_p$ of the tower, the shift of the command mean caused by each object along the direction to the command that walks to it, in policy standard deviations, with dots for seeds.
    (b) Fraction of GM states pulled toward each object ($f_o>\tau$) in four groups, the world-edit bank split into states with the flag or the tower more than $0.3$\,m nearer (near), the forced-view conflict bank (flag much nearer) and the far bank (tower much nearer).
    (c) Fraction of states pulled toward the flag, away from it, toward the tower and toward both at once, and the fraction on which both nuisance shifts pass the threshold together (null).
    (d) Tower share $P_p/(P_p+P_\star)$ of the removal pull against the distance gap $d_\star-d_p$, negative where the flag is nearer. The dotted step is the share that walking to the nearer object would give.
    Means with $95\%$ $t$-intervals across three seeds.
    }
    \label{fig:wtf-two-pulls}
\end{figure}

\paragraph{Results: do both objects pull the agent?}
Both objects pull GM agents, the tower strongly and the flag weakly and with mixed sign.
In GM agents the tower pull is $f_p=3.736$ policy standard deviations, whereas the flag pull is $f_\star=0.211$ (Fig.~\ref{fig:wtf-two-pulls}a). The tower pulls toward itself on $0.793$ of states, and the flag pulls toward itself on $0.368$ and away from itself on $0.198$ (Fig.~\ref{fig:wtf-two-pulls}c).
Non-GM agents show the reverse, $f_\star=2.899$ with a pull toward the flag on $0.899$ of states and $f_p=-0.019$, and randomly initialized networks show no pull.
The GM flag pull lives near the flag (Fig.~\ref{fig:wtf-two-pulls}b). It points toward the flag on $0.449$ to $0.509$ of states where both objects are near and on $0.458$ where the flag is much nearer, but on $0.021$ where the tower is much nearer, while the tower pulls toward itself on $0.568$ to $0.948$ of states in every group.
The two pulls also coexist on the same states. Both point toward their own object on $0.271$ of GM states, against $0.003$ for the two nuisance shifts (Fig.~\ref{fig:wtf-two-pulls}c).
Neither object pushes the agent away from the other. The cross projection of the GM flag pull on the direction to the tower is $-0.293$ (95\% CI $-0.730$ to $0.144$), and that of the tower pull on the direction to the flag is $+2.177$ ($0.342$ to $4.012$), positive because the two objects often lie in similar directions.

\paragraph{Results: how does this differ from cue competition?}
Unlike cue competition, the two objects do not trade support against each other. They pull at the same time, and the tower wins at every distance.
The swaps show no push and pull. The states lie along the rising diagonal, the inverse of the other environments (Spearman correlation $+0.65$ to $+0.81$ per cohort and swap, Fig.~\ref{fig:attribute-counter-target}, bottom row), because flag and tower stand close together in the world-edit bank, so a turn toward one candidate is mostly a turn toward the other and a swap raises or lowers both supports together.
In the seed means (Table~\ref{tab:attribute-logit-competition}) the GM tower swap lies in the push--pull quadrant ($S_{\mathrm{fav}}=3.933$, $S_{\mathrm{comp}}=-4.129$), with a spread across seeds as large as the effect, and the GM flag swap raises support for the place where the flag joins the tower ($1.785$) without taking support from the place it leaves ($0.005$). In Non-GM agents the flag swap mainly takes support from the place the flag leaves ($-1.888$), and the tower swap lowers both candidates ($-1.882$ and $-1.243$).
The tower swap result holds with cones of $15^\circ$ and $45^\circ$ and when the cones are replaced by the heading-controller commands toward the two candidates, whereas the GM flag swap lies in the quadrant in all three seeds at $15^\circ$ but in one seed at $30^\circ$, at $45^\circ$ and with the heading-controller commands.
The tower's dominance does not come from being nearer (Fig.~\ref{fig:wtf-two-pulls}d). It takes $0.775$ to $0.861$ of the GM removal pull where the flag is $1$ to $5$\,m nearer and $0.976$ to $0.986$ where the tower is more than $1$\,m nearer, against at most $0.008$ in Non-GM agents.
This agrees with panel b, which shows the flag pulling on about half of the states where it is nearer. Panel b counts states on which a pull passes the nuisance threshold, whereas panel d compares the sizes of the two pulls, and where the flag is much nearer the GM flag pull passes the threshold on $0.458$ of states but averages $f_\star=0.241$ against $f_p=1.750$ for the tower. The flag pulls often but weakly, and the tower's larger pull dominates the share.

GM behavior in \quadname{} is therefore an \emph{object competition}, in which both objects pull at once without pushing the agent away from each other, the orange tower dominates at every distance and the flag keeps a weak pull only when it is near.

The next section asks how this target-level interaction is organized in
representation space.

\subsection{Geometry of Cue Competition}
\label{app:attribute-competition-geometry}

We found that the goal and proxy cues both remain active in \textsc{Attribute Maze}, \textsc{AttributeTable} and \textsc{SortZones}, and so do the two competing objects in \quadname{}, and that changing either cue shifts the target preference toward the candidate that cue supports.
The preceding sections show different variants of this competition.
What remains open is what the competition looks like inside the representation.

We therefore ask two questions.
First, \textit{when the intended and proxy cues are changed from the same base state, how are the resulting representation displacements related?}
Second, \textit{once we define their contrast, is the resulting competition direction reused across states?}

\subsubsection{How are the two cue directions organised within a state?}
\label{app:attribute-within-state-geometry}

The edits are those of the world-edit analysis (Appendix~\ref{app:world-edit}).
Because the analysis compares edits across states, it needs states in which it is clear on which side the goal candidate lies and the choice between the candidates is still open, so \textsc{AttributeTable} and \textsc{SortZones} use stricter conflict banks than their world edits, in which the two candidates lie on clearly different sides at equal distance from the hand (\textsc{AttributeTable}) or the cube (\textsc{SortZones}), both conflict orientations occur equally often, and in \textsc{SortZones} the frame is taken when the hand has reached the cube but not yet pushed it (Appendix~\ref{app:banks-conflict}).
For each state $s\in S_{\mathrm{conf}}$, the intended-cue swap induces the representation displacement $\Delta_\star(s)$ and the proxy-cue swap induces $\Delta_p(s)$.
The swaps are shape and colour in \textsc{Attribute Maze} and \textsc{AttributeTable}, cross mark and colour in \textsc{SortZones}, and flag and tower in \quadname{}.
Both displacements are measured relative to the same unedited representation.
Everything except the swapped cue stays fixed, so the comparison shows how the two cue edits move the representation from the same starting point.

To ask whether the two cue edits induce related representation changes, we first compare the orientation of their displacement vectors.
In ordinary activation space, their alignment is measured by the Euclidean cosine
\[
\cos_E\!\left(
\Delta_\star(s),
\Delta_p(s)
\right)
=
\frac{
\Delta_\star(s)^\top \Delta_p(s)
}{
\|\Delta_\star(s)\|_2
\|\Delta_p(s)\|_2
}.
\]
Positive values indicate aligned representation changes, values near zero indicate weak alignment, and negative values indicate opposing directions.

The Euclidean cosine treats every activation-space direction equally.
We are additionally interested in whether the two edits oppose each other specifically along directions to which the downstream policy is sensitive.
Because both displacements originate from the same base state, they share the same local Fisher metric.
We therefore also compute the Fisher cosine from Definition~\ref{def:fisher-cosine},
\[
\cos_F\!\left(
\Delta_\star(s),
\Delta_p(s)
\right).
\]
So $\cos_E$ measures how similar the two displacements are as raw activation changes, and $\cos_F$ measures how similar the changes in the action distribution are that they cause.
In \textsc{Attribute Maze} the metric is that of a choice among $15$ discrete actions, and in the robot tasks that of a Gaussian action head, which reads all eight action dimensions of the arm and the three of \quadname{}.

\begin{figure}[t]
    \centering
    \includegraphics[width=\linewidth]
    {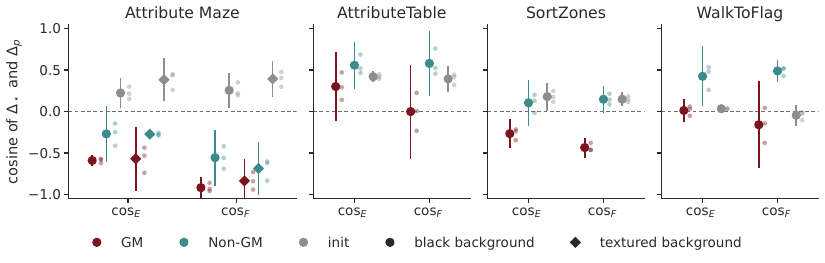}
    \caption{
    \textbf{Within-state cue geometry in the four competition environments} ($h_{\mathrm{fc}}$).
    Euclidean cosine $\cos_E$ and Fisher cosine $\cos_F$ between $\Delta_\star(s)$ and $\Delta_p(s)$, averaged over states.
    In \textsc{Attribute Maze} circles are the black-background policies and diamonds the textured-background policies, each on its own background.
    Mean and 95\% $t$-interval over three seeds, small points are seeds.
    }
    \label{fig:attribute-within-state-geometry}
\end{figure}

We report both quantities at $h_{\mathrm{fc}}$, the representation the policy and value heads read, with \textsc{Attribute Maze} also at $h_{\mathrm{pre}}$ in Table~\ref{tab:attribute-cross-state-geometry} and Figure~\ref{fig:attribute-dcomp-heatmap}.
\textsc{Attribute Maze} refers to the black-background policies, and the textured-background policies are shown as diamonds in Figure~\ref{fig:attribute-within-state-geometry}, in the right half of Figure~\ref{fig:attribute-dcomp-heatmap} and in their own rows of Table~\ref{tab:attribute-cross-state-geometry}.
Freshly initialized networks serve as the baseline.

\paragraph{Results.}
In GM agents the two cues are opposed within a state to very different degrees (Figure~\ref{fig:attribute-within-state-geometry}).
The Fisher cosine $\cos_F$, which compares the changes in the action distribution, is $-0.919$ in \textsc{Attribute Maze}, $-0.435$ in \textsc{SortZones}, $-0.160$ in \quadname{} and $-0.002$ in \textsc{AttributeTable}.
So only \textsc{Attribute Maze} is close to antipodal, and the robot tasks form a weaker gradient below it.
The \textsc{SortZones} opposition holds in every GM policy, while the \textsc{AttributeTable} mean near zero averages policies that differ in sign.
The textured \textsc{Attribute Maze} policies are opposed almost as strongly as the black ones ($\cos_F=-0.838$ in GM and $-0.687$ in Non-GM agents).

Initialized networks may explain part of the gradient (grey points in Figure~\ref{fig:attribute-within-state-geometry}).
Both swaps redraw the same two objects, so in an untrained network the two displacements already share a component, with $\cos_F=0.254$ in \textsc{Attribute Maze} and $0.393$ in \textsc{AttributeTable}.
Training in GM agents moves $\cos_F$ away from this starting point by $1.173$ in \textsc{Attribute Maze}, $0.583$ in \textsc{SortZones}, $0.395$ in \textsc{AttributeTable} and $0.113$ in \quadname{}.
So \textsc{AttributeTable} starts from the largest shared component and GM training removes only a third of what it removes in \textsc{Attribute Maze}.

The action space explains most of the rest of the \textsc{AttributeTable} result.
The first-order gripper displacements that the two edits cause point in opposite directions in every GM agent (cosine of the two 3-d gripper motions $-0.590\pm0.11$, against $0.393$ in Non-GM agents and $0.263$ at initialization), so the two cues do push the gripper toward opposite objects.
But the arm has seven joints, and a large part of the Fisher energy of each edit lies in joint directions that leave the gripper in place ($0.470$ for the shape swap and $0.297$ for the colour swap).
In those directions the two edits move the policy the same way ($\cos_F=0.402\pm0.29$), and the full $\cos_F$ averages the opposed gripper component with this aligned one.
\textsc{Attribute Maze} has no such directions, since every action changes where the agent goes.
In \textsc{SortZones} the gripper displacements are opposed as well ($-0.730\pm0.06$), and so are the two edits in the directions that leave the gripper in place ($\cos_F=-0.506\pm0.16$), so nothing cancels the opposition.

In the robot tasks, within-state $\cos_F$ is lower in GM than in Non-GM agents in every task, but clearly negative only in \textsc{SortZones}.
In \textsc{Attribute Maze} Non-GM agents are opposed as well ($\cos_F=-0.557$), and single-cue ablations, which remove one cue instead of exchanging it and share no edited sprite, leave GM agents no more opposed than Non-GM agents ($-0.70$ at $h_{\mathrm{fc}}$ for GM).
In the robot tasks Non-GM agents are aligned on average ($\cos_F=0.579$ in \textsc{AttributeTable}, $0.147$ in \textsc{SortZones} and $0.488$ in \quadname{}).
A Non-GM agent barely reads the proxy, so its proxy displacement keeps mostly the shared component that initialization already has and never acquires an opposed part.
Within-state opposition is therefore a property of trained policies in \textsc{Attribute Maze}, and in the robot tasks it is clearly GM-specific only in \textsc{SortZones}.

Within a state, then, the two cue edits move the representation in clearly different directions from the same starting point, opposed in \textsc{Attribute Maze} and in the GM agents of \textsc{SortZones}, and their contrast is what the policy has to resolve.
This motivates a competition direction defined separately for every state,
\[
d_{\mathrm{comp}}(s)
=
\Delta_p(s)
-
\Delta_\star(s)
=
h(s_p)-h(s_\star).
\]
Because both displacements are measured from the same base representation, the base representation cancels in the difference.
The resulting vector is therefore the direct representation-space contrast between the two competing cue assignments at state $s$.
Its positive orientation points from the swap under which both cues support the proxy's candidate toward the swap under which both cues support the goal's candidate.
So, pushing along $+d_{\mathrm{comp}}(s)$ moves the representation toward the intended goal.
What remains open is whether this direction is specific to each state or shared across states.

\subsubsection{Is the competition direction shared across states?}
\label{app:attribute-cross-state-geometry}

The within-state analysis gives us one competition direction $d_{\mathrm{comp}}(s)$ for every conflict state.
We now change the unit of comparison and ask how these directions relate across states.

\paragraph{Two possible organizations.}
The simplest possibility is that the same orientation is reused everywhere,
\[
d_{\mathrm{comp}}(s)
\approx
\beta(s)u_{\mathrm{comp}},
\qquad
\beta(s)>0.
\]
A slightly richer possibility is that the same underlying axis is reused while its orientation changes with the state,
\[
d_{\mathrm{comp}}(s)
\approx
\sigma(s)\beta(s)u_{\mathrm{comp}},
\qquad
\sigma(s)\in\{-1,+1\}.
\]
Under this account, different states can express opposite orientations of the same competition axis.
The sign $\sigma(s)$ is the conflict orientation of the state, and it is set by the scene, not read from the network.
Every conflict has a mirrored version in which the goal candidate lies on the other side.
In \textsc{Attribute Maze} the agent always starts at the same cell and the line candidate lies either along UP or along RIGHT ($98$ states each), in \textsc{AttributeTable} and \textsc{SortZones} the goal candidate lies left or right of the robot, and in \quadname{} the flag appears left or right in the camera image.
We set $\sigma(s)=+1$ for one side and $\sigma(s)=-1$ for the other.
The sign matters because $d_{\mathrm{comp}}(s)$ points toward the goal's candidate, so mirrored states call for opposite actions, for example UP instead of RIGHT.
Averaged without the sign, the two orientations cancel, whereas $\sigma(s)\,d_{\mathrm{comp}}(s)$ folds them onto one direction.
Since $\sigma$ is known by design, the question is whether independently generated cue edits collapse onto one signed axis in the part of the representation the policy reads.

\paragraph{Comparing competition directions across states.}
For two states $s_i,s_j\in S_{\mathrm{conf}}$, we first compare their competition directions in ordinary activation space,
\[
\kappa_E(s_i,s_j)
=
\cos_E\!\left(
d_{\mathrm{comp}}(s_i),
d_{\mathrm{comp}}(s_j)
\right).
\]
For the policy-sensitive comparison, the two states generally have different local Fisher metrics.
We therefore use their pair-average metric $F_{ij}=\frac{1}{2}(F_i+F_j)$ and define
\[
\kappa_F(s_i,s_j)
=
\frac{
d_i^\top F_{ij}d_j
}{
\sqrt{
(d_i^\top F_{ij}d_i)
(d_j^\top F_{ij}d_j)
}
},
\]
where $d_i=d_{\mathrm{comp}}(s_i)$ and $d_j=d_{\mathrm{comp}}(s_j)$.

We summarize both similarities over pairs of states with the same orientation.
We fit $u_{\mathrm{comp}}$ as the mean of $\sigma(s)\,d_{\mathrm{comp}}(s)$ over the fit states ($146$ in \textsc{Attribute Maze}, $308$ in \textsc{AttributeTable}, $302$ in \textsc{SortZones} and $256$ in \quadname{}) and evaluate its alignment with $\sigma(s)\,d_{\mathrm{comp}}(s)$ on the held-out states ($50$, $204$, $198$ and $200$).

\begin{table}[t]
\centering
\small
\setlength{\tabcolsep}{4pt}
\caption{
\textbf{Cross-state competition geometry.}
The first two columns give mean similarity over pairs of states with the same sign $\sigma$.
The last two give the alignment of held-out competition directions with the shared axis $u_{\mathrm{comp}}$ fit on the remaining states.
Entries are cohort mean $\pm$ SD across three policies. $\kappa_E$ and $\cos_E$ compare raw activations, $\kappa_F$ and $\cos_F$ compare them in the policy's Fisher metric.
\textsc{Attribute Maze} has two blocks with the same rows, the black-background and the textured-background policies, each on its own background at $h_{\mathrm{pre}}$ and $h_{\mathrm{fc}}$ (initialized networks shown on the same background). Robots at $h_{\mathrm{fc}}$ on the banks of Appendix~\ref{app:banks-conflict}.
The held-out $\cos_F$ uses the Fisher metric at each held-out state.
}
\label{tab:attribute-cross-state-geometry}
\begin{tabular}{lllcccc}
\toprule
Environment & Cohort & Position
& $\kappa_E$ same
& $\kappa_F$ same
& \begin{tabular}[b]{@{}c@{}}$\cos_E(\sigma d_{\mathrm{comp}},$\\$u_{\mathrm{comp}})$\end{tabular}
& \begin{tabular}[b]{@{}c@{}}$\cos_F(\sigma d_{\mathrm{comp}},$\\$u_{\mathrm{comp}})$\end{tabular} \\
\midrule
\textsc{Attribute Maze} & GM & $h_{\mathrm{pre}}$ & $0.354\pm0.02$ & $0.938\pm0.04$ & $0.521\pm0.02$ & $0.961\pm0.03$ \\
\quad black & GM & $h_{\mathrm{fc}}$ & $0.589\pm0.02$ & $0.942\pm0.05$ & $0.709\pm0.00$ & $0.959\pm0.03$ \\
 & Non-GM & $h_{\mathrm{pre}}$ & $0.284\pm0.02$ & $0.844\pm0.05$ & $0.468\pm0.02$ & $0.914\pm0.05$ \\
 & Non-GM & $h_{\mathrm{fc}}$ & $0.532\pm0.04$ & $0.952\pm0.02$ & $0.679\pm0.05$ & $0.970\pm0.02$ \\
 & Init & $h_{\mathrm{pre}}$ & $0.173\pm0.02$ & $0.160\pm0.03$ & $0.357\pm0.03$ & $0.330\pm0.06$ \\
 & Init & $h_{\mathrm{fc}}$ & $0.177\pm0.03$ & $0.148\pm0.03$ & $0.364\pm0.03$ & $0.301\pm0.06$ \\
\midrule
\textsc{Attribute Maze} & GM & $h_{\mathrm{pre}}$ & $0.265\pm0.03$ & $0.913\pm0.06$ & $0.447\pm0.03$ & $0.943\pm0.05$ \\
\quad textured & GM & $h_{\mathrm{fc}}$ & $0.588\pm0.06$ & $0.918\pm0.07$ & $0.709\pm0.05$ & $0.946\pm0.05$ \\
 & Non-GM & $h_{\mathrm{pre}}$ & $0.243\pm0.04$ & $0.820\pm0.04$ & $0.426\pm0.04$ & $0.931\pm0.05$ \\
 & Non-GM & $h_{\mathrm{fc}}$ & $0.458\pm0.02$ & $0.900\pm0.04$ & $0.626\pm0.02$ & $0.932\pm0.04$ \\
 & Init & $h_{\mathrm{pre}}$ & $0.105\pm0.05$ & $0.073\pm0.02$ & $0.263\pm0.06$ & $0.205\pm0.00$ \\
 & Init & $h_{\mathrm{fc}}$ & $0.100\pm0.05$ & $0.061\pm0.01$ & $0.249\pm0.07$ & $0.192\pm0.01$ \\
\midrule
\textsc{AttributeTable} & GM & $h_{\mathrm{fc}}$ & $0.187\pm0.01$ & $0.667\pm0.02$ & $0.432\pm0.02$ & $0.824\pm0.01$ \\
 & Non-GM & $h_{\mathrm{fc}}$ & $0.175\pm0.04$ & $0.561\pm0.11$ & $0.419\pm0.05$ & $0.727\pm0.08$ \\
 & Init & $h_{\mathrm{fc}}$ & $0.052\pm0.02$ & $0.069\pm0.04$ & $0.221\pm0.04$ & $0.249\pm0.06$ \\
\midrule
\textsc{SortZones} & GM & $h_{\mathrm{fc}}$ & $0.271\pm0.04$ & $0.541\pm0.08$ & $0.502\pm0.04$ & $0.707\pm0.05$ \\
 & Non-GM & $h_{\mathrm{fc}}$ & $0.283\pm0.02$ & $0.446\pm0.01$ & $0.524\pm0.02$ & $0.701\pm0.03$ \\
 & Init & $h_{\mathrm{fc}}$ & $0.332\pm0.01$ & $0.292\pm0.04$ & $0.582\pm0.02$ & $0.560\pm0.02$ \\
\midrule
\quadname{} & GM & $h_{\mathrm{fc}}$ & $0.525\pm0.02$ & $0.443\pm0.16$ & $0.680\pm0.01$ & $0.625\pm0.14$ \\
 & Non-GM & $h_{\mathrm{fc}}$ & $0.498\pm0.09$ & $0.478\pm0.19$ & $0.539\pm0.06$ & $0.505\pm0.28$ \\
 & Init & $h_{\mathrm{fc}}$ & $0.280\pm0.02$ & $0.229\pm0.07$ & $0.427\pm0.01$ & $0.307\pm0.11$ \\
\bottomrule
\end{tabular}
\end{table}

\begin{figure}[t]
    \centering
    \includegraphics[width=\linewidth]
    {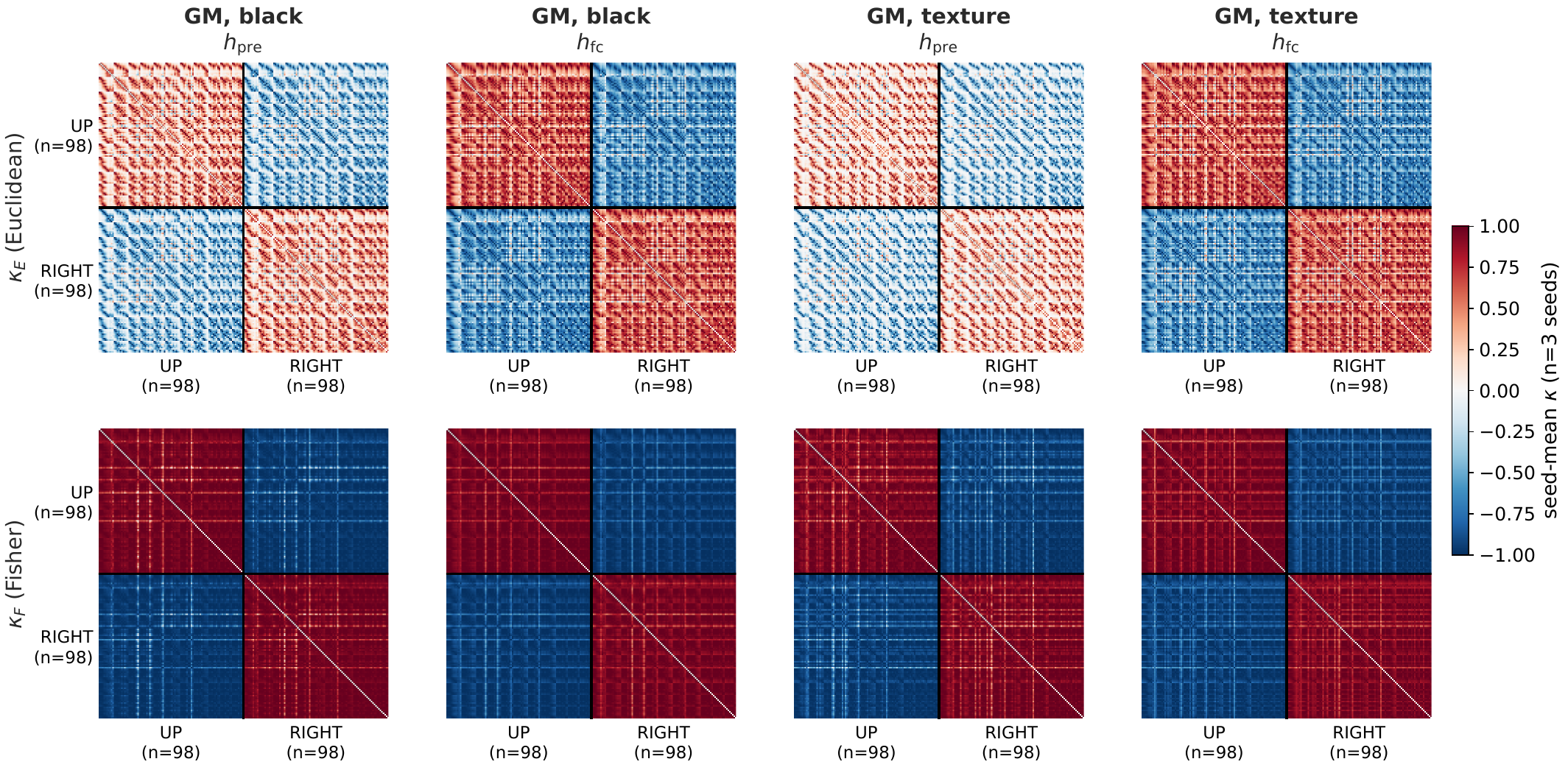}
    \caption{
    \textbf{The policy compresses a rich competition geometry onto one signed axis} (\textsc{Attribute Maze}).
    Pairwise similarity between state-local competition directions across the $196$ conflict-bank states, averaged over the three GM policies of each background, black (left four panels) and textured (right four panels), each on its own background.
    Columns show $h_{\mathrm{pre}}$ and $h_{\mathrm{fc}}$ for each background, rows show $\kappa_E$ and $\kappa_F$.
    States are grouped by whether the line candidate lies along UP or along RIGHT.
    Under the swap construction the cross-sign blocks are antisymmetric by construction, so the informative structure is the within-sign alignment.
    }
    \label{fig:attribute-dcomp-heatmap}
\end{figure}

\begin{figure}[t]
    \centering
    \includegraphics[width=\linewidth]
    {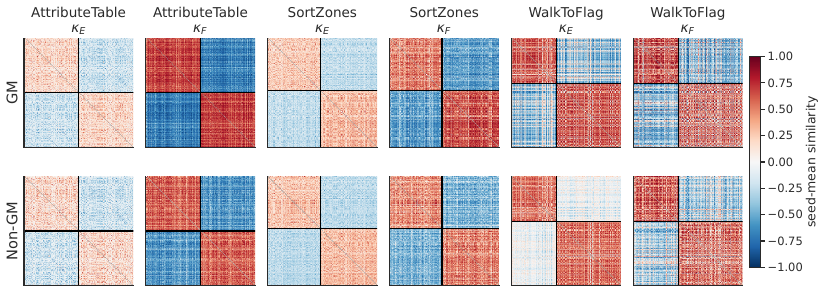}
    \caption{
    \textbf{Competition directions across states in the robot tasks} ($h_{\mathrm{fc}}$, fit states of the strict \textsc{AttributeTable} bank, the \textsc{SortZones} decision-frame bank and the \quadname{} world-edit bank).
    Pairwise $\kappa_E$ and $\kappa_F$ of $d_{\mathrm{comp}}$, averaged over three seeds, for GM (top) and Non-GM (bottom) agents.
    States are grouped by conflict orientation.
    Unlike in \textsc{Attribute Maze}, mirrored states are separate scenes, so the cross-sign blocks are not antisymmetric by construction.
    }
    \label{fig:robot-dcomp-heatmaps}
\end{figure}

\paragraph{Results.}
Every environment has a shared competition axis, but the policy compresses the competition onto it to very different degrees (Table~\ref{tab:attribute-cross-state-geometry}).
In \textsc{Attribute Maze} the within-sign blocks are rich in raw activations ($\kappa_E=0.589$ at $h_{\mathrm{fc}}$ for GM), and the Fisher metric removes most of that texture ($\kappa_F=0.942$, against $0.148$ at initialization, where the metric still has full rank, Figure~\ref{fig:attribute-dcomp-heatmap}).
An axis fit on $146$ states aligns with held-out competition directions at $\cos_F=0.959$, and between $74\%$ and $86\%$ of the Fisher energy of $d_{\mathrm{comp}}$ lies on a single logit contrast, UP against the RIGHT action group.
The state mainly chooses the sign and the magnitude of one shared action contrast, $d_{\mathrm{comp}}(s)\approx\sigma(s)\beta(s)u_{\mathrm{comp}}$.
The textured policies compress the same way ($\kappa_F=0.918$ and held-out $\cos_F=0.946$ in GM agents, right half of Figure~\ref{fig:attribute-dcomp-heatmap}).

The robot tasks show the same signed block structure, but weaker (Figure~\ref{fig:robot-dcomp-heatmaps}).
Since initialized networks already share part of the direction through the input, we read $\kappa_F$ against its initial value.
Above initialization, GM agents gain $0.793$ in \textsc{Attribute Maze}, $0.599$ in \textsc{AttributeTable}, $0.249$ in \textsc{SortZones} and $0.213$ in \quadname{}.
In \textsc{AttributeTable} the left-right gripper contrast carries $0.570$ of the Fisher energy of $d_{\mathrm{comp}}$, against $0.053$ at initialization, and in \textsc{SortZones} $0.277$, against $0.064$, so \textsc{AttributeTable} comes closest to reducing its competition to one action contrast, as \textsc{Attribute Maze} does with $74\%$ to $86\%$.
The axis is not specific to goal misgeneralization in any environment, as Non-GM agents share it almost as much as GM agents ($\kappa_F=0.952$, $0.561$, $0.446$ and $0.478$, lower row of Figure~\ref{fig:robot-dcomp-heatmaps}).

The robot results depend on the bank in the expected direction.
On the \textsc{AttributeTable} world-edit bank, where the conflict orientation is often ambiguous, same-sign $\kappa_F$ is $0.281$ and both cues pull toward their own candidates in $0.597$ of GM states, against $0.667$ and $0.884$ on the strict bank, while the within-state $\cos_F$ is near zero on both ($0.098$ and $-0.002$).
At the \textsc{SortZones} start frame, before a pad is chosen, the within-state $\cos_F$ is less opposed than at the decision frame ($-0.293$ against $-0.435$), while $\kappa_F$ ($0.612$ against $0.541$) and the fraction of states in which both cues pull their own way ($0.908$ against $0.820$) change little.

\paragraph{Controls in \textsc{Attribute Maze}.}
Some concerns could come up about the \textsc{Attribute Maze} axis.
First, the local Fisher metric has rank at most $14$, and in trained policies its effective rank is only about $2.6$ (participation ratio), which makes cosines more extreme.
Under norm-matched random displacements and the same metric, $\cos_F$ and $\kappa_F$ average zero, with mean absolute values of $0.43$ to $0.82$ depending on how the directions are sampled, whereas real GM displacements are opposed in $97\%$ to $100\%$ of states and same-bearing pairs always align. The size of these values is therefore partly the metric, and the evidence for the axis is their sign.
Second, one might consider the axis an artefact of the swap construction, which makes mirrored states exact negatives of each other.
However, competition directions built from single-cue ablations, which have no such symmetry, put $92\%$ of their Fisher energy on the swap axis in GM policies, against $9\%$ to $10\%$ at initialization, and recover the sign in $92\%$ to $100\%$ of held-out states, so the axis is a property of trained policies, GM and Non-GM alike, and not of the edit.
Third, the within-sign blocks are rich in raw activations, and one might ask whether this fine structure is part of the competition.
However, it tracks the spatial similarity of the inputs (correlation $0.735$ at $h_{\mathrm{fc}}$, against $0.733$ at initialization) and lies mostly in directions the policy does not read, so it reflects how similar the scenes look rather than how the policy decides.
Finally, every state of the conflict bank starts the agent at the same cell, so the single axis might only hold for that one situation.
However, on the diversified bank, where the agent starts at every walkable cell, the axis splits into roughly one shared axis per action-conflict class ($\kappa_F=0.889$ for pairs that share the action contrast and the bearing, $-0.213$ for pairs that share neither), so the single axis of the conflict bank is the special case of one conflict class.

\subsubsection{Why does the proxy win on the shared axis?}
\label{app:attribute-tug-of-war}

If both cues act on one axis, the competition is a tug of war, and we can ask which cue pulls harder and whether this decides the agent's choice.

\paragraph{Measuring each cue's pull.}
Each swap moves one cue to the other candidate, so the displacement it causes is exactly the influence that cue had on the conflict state.
To express this influence in the currency the policy uses, we project each swap displacement onto $u_{\mathrm{comp}}$ in the Fisher metric of the unedited state, as in Appendix~\ref{app:projection-analysis},
\[
\alpha_F(\Delta)
=
\frac{u_{\mathrm{comp}}^\top F(s)\,\Delta}{u_{\mathrm{comp}}^\top F(s)\,u_{\mathrm{comp}}},
\qquad
\Delta\in\{\Delta_\star(s),\Delta_p(s)\}.
\]
So $\alpha_F(\Delta_\star)$ measures how far the goal cue pulls the state along the axis, and $\alpha_F(\Delta_p)$ how far the proxy cue pulls it, both oriented so that a positive value points toward the candidate the swap favours.
All pulls are reported on held-out states at $h_{\mathrm{fc}}$, $50$ in \textsc{Attribute Maze}, $204$ in \textsc{AttributeTable}, $198$ in \textsc{SortZones} and $200$ in \quadname{}.

\begin{figure}[t]
    \centering
    \includegraphics[width=\linewidth]
    {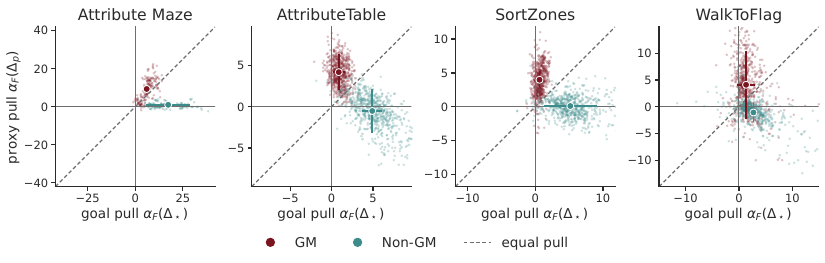}
    \caption{
    \textbf{Tug of war on the shared axis in four environments} ($h_{\mathrm{fc}}$, held-out states).
    Each small point is one state of one policy, placed by how far the goal-cue swap moves it along $u_{\mathrm{comp}}$ toward the goal cue's candidate (goal pull, $x$) and how far the proxy-cue swap moves it toward the proxy cue's candidate (proxy pull, $y$), both in the Fisher metric.
    GM and Non-GM agents are overlaid, large markers are the mean and 95\% $t$-interval over three seeds.
    Above the dashed identity line the proxy pulls harder, in the upper-right quadrant both cues pull toward their own candidates, and left of the vertical axis the goal cue pushes toward the proxy's candidate.
    Axes are linear and scaled per environment.
    }
    \label{fig:tug-of-war-four-envs}
\end{figure}

\paragraph{The proxy pulls harder in every GM cohort.}
In \textsc{Attribute Maze} the colour swap moves GM policies along the axis by $9.321$ and the shape swap by $5.974$, a ratio of means of $1.564$ (95\% CI $[1.34,1.79]$), and Non-GM policies show the mirror image, with a ratio of $0.067$.
To compare environments with one estimator we use the median over states of the ratio of the two pulls (Figure~\ref{fig:competition-four-envs}, bottom right).
It is $1.527$ for GM and $0.082$ for Non-GM in \textsc{Attribute Maze}, and in the robot tasks $5.106$, $6.216$ and $4.228$ for GM against $0.267$, $0.211$ and $0.449$ for Non-GM (\textsc{AttributeTable}, \textsc{SortZones}, \quadname{}).
So on the axis the policy reads, the proxy outweighs the goal in GM agents and the goal outweighs the proxy in Non-GM agents in all four environments.

Randomly initialized networks favour the proxy on average in every one of these environments, with a median ratio of $1.644$ in \textsc{Attribute Maze} and $2.077$, $2.901$ and $6.028$ in the robot tasks, although the 95\% CI includes $1$ in \textsc{Attribute Maze} ($0.66$ to $2.63$) and \textsc{SortZones} ($0.97$ to $4.83$). %
Their pulls are small, $30$ to $50$ times smaller than those of trained policies in \textsc{Attribute Maze} and $100$ to $1000$ times smaller in the robot tasks, so the initial ratio is a direction rather than a precise value.
GM training keeps this preference (the log ratio changes by $-0.07$ in \textsc{Attribute Maze}) and Non-GM training reverses it ($-3.0$).
Per unit of pixel change, initialized networks do not favour the proxy in the robot tasks (Fisher $Q$ ratio $0.76$ to $0.85$, every 95\% CI including $1$). There the preference comes from the input, as the tower fills about $14$ times more pixels than the flag and the \textsc{AttributeTable} colour swap carries $2.69$ times the pixel energy of the shape swap.
In \textsc{Attribute Maze} the colour swap is the smaller edit but moves initialized networks as much as the shape swap ($Q$ ratio $1.17$, 95\% CI $0.71$ to $1.94$), possibly because its pixel changes all point one way and survive the pooling down to $h_{\mathrm{pre}}$, while those of the shape swap cancel within a cell.
The proxy's advantage at initialization is therefore mostly an input effect, in line with networks preferring the more easily extracted feature \citep{hermann2020shapes}, that GM training does not overturn, and the comparison with Non-GM agents carries the result.
A prior of this kind may explain why misgeneralization is favoured early in training, which we leave as an exciting open direction.

\paragraph{How clean the tug of war is.}
The environments differ in how balanced the tug of war is and in whether the two cues pull in opposite directions state by state (Figure~\ref{fig:tug-of-war-four-envs}).
In \textsc{Attribute Maze} both cues pull toward their own candidates in $0.967$ of GM states, so the cloud sits in the upper-right quadrant, slightly above the identity line.
\textsc{AttributeTable} ($0.884$) and \textsc{SortZones} ($0.820$) look the same, but their clouds sit far above the identity line, because the goal pulls with only a fifth to a sixth of the proxy's strength.
In \quadname{} ($0.650$) the goal pull changes sign from state to state, so the cloud spreads across the vertical axis, and one seed is pulled only weakly by the tower (ratio $1.521$).
This matches the object competition found for \quadname{} in Appendix~\ref{app:wtf-role}, where flag and tower each pull the agent toward themselves and the tower wins, without the flag being pushed away.
Two independent pulls do not form the two ends of one contrast, so the flag's pull on $u_{\mathrm{comp}}$ is weak and its sign is not tied to the flag's side.
The tug of war is therefore balanced in \textsc{Attribute Maze}, consistent but lopsided in \textsc{AttributeTable} and \textsc{SortZones}, and inconsistent in \quadname{}.

\begin{figure}[t]
    \centering
    \includegraphics[width=0.9\linewidth]
    {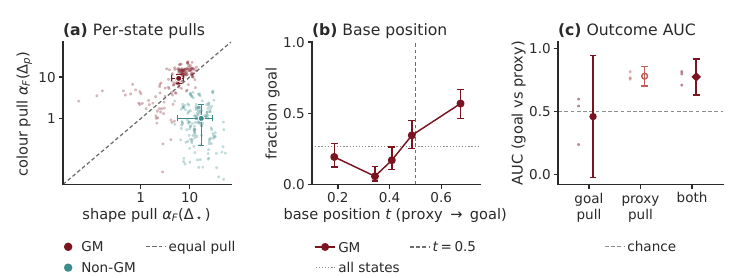}
    \caption{
    \textbf{GM policies weight the proxy cue more strongly on the shared competition axis} (\textsc{Attribute Maze}, $h_{\mathrm{fc}}$, black background).
    (a) Goal pull against proxy pull per held-out state (small points) with cohort means and 95\% $t$-intervals. States above the identity line are pulled harder by the proxy. Axes are symmetric-logarithmic.
    (b) Fraction of GM rollouts reaching the goal in five quantile bins of the base position $t$, with Wilson 95\% intervals (all $196$ states, timeouts excluded).
    (c) AUC for predicting a goal versus a proxy outcome from the goal pull, the proxy pull, or both.
    }
    \label{fig:attribute-tug-of-war}
\end{figure}

\paragraph{Where the unedited state sits.}
In \textsc{Attribute Maze} the two pulls also predict what the agent does.
The shape swap puts both cues on the proxy's candidate (the proxy consensus), and the colour swap puts both cues on the goal's candidate (the goal consensus).
The unedited state lies between these two consensus states, and its position is
\[
t(s)=\frac{\alpha_F(\Delta_\star)}{\alpha_F(\Delta_\star)+\alpha_F(\Delta_p)},
\]
with $t=0$ at the proxy consensus and $t=1$ at the goal consensus.
The median held-out state sits at $t=0.396$ in GM policies and at $t=0.933$ in control policies.
In Figure~\ref{fig:attribute-tug-of-war}b, the fraction of GM rollouts that reach the intended goal stays below $0.2$ in the bins on the proxy side and reaches $0.57$ in the bin closest to the goal side.
The base position predicts the GM outcome.
The signed distance of the base state from the midpoint separates rollouts that reach the goal from rollouts that reach the proxy with an AUC of $0.774$ over all $196$ states (95\% CI $[0.723,0.822]$, bootstrap over states within each seed) and $0.846$ within each conflict orientation ($[0.800,0.891]$), timeouts excluded, and every seed lies above $0.70$.
The interval across the three seeds is wider ($[0.631,0.916]$), because each seed's AUC rests on few goal outcomes, as GM agents rarely reach the goal ($27$ to $60$ of $196$ states per seed, and $5$ to $11$ of the $50$ held-out states, where the AUC is $0.736$, $[0.614,0.839]$).
The proxy pull alone predicts the GM outcome as well as both pulls together ($0.779$), while the goal pull alone is at chance ($0.459$).
We did not run this outcome analysis in the robot tasks, where $t$ is undefined in the states in which the goal pull changes sign.

\subsubsection{What does the geometry suggest for recovering the goal?}
\label{app:competition-four-envs}

\begin{figure}[t]
    \centering
    \includegraphics[width=\linewidth]
    {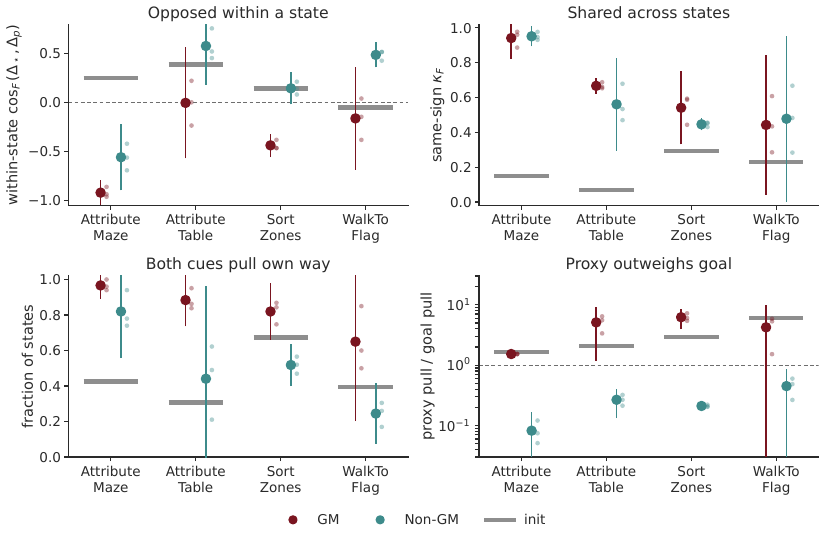}
    \caption{
    \textbf{In every environment the proxy outweighs the goal in GM agents and not in Non-GM agents, while the rest of the geometry differs between environments} ($h_{\mathrm{fc}}$).
    Each panel is one component of the competition.
    Top left, Fisher cosine of the two cue edits within a state (negative means opposed).
    Top right, same-sign $\kappa_F$ (how much the competition lies on one axis shared across states).
    Bottom left, fraction of states in which both cues pull toward their own candidates.
    Bottom right, median per-state ratio of proxy pull to goal pull (log scale, above $1$ the proxy pulls harder).
    Grey bars are initialized networks. Mean and 95\% $t$-interval over three seeds, small points are seeds.
    }
    \label{fig:competition-four-envs}
\end{figure}

Figure~\ref{fig:competition-four-envs} puts the components of the competition side by side.
All four environments share one property, a proxy that outweighs the goal on the axis the policy reads in GM agents and not in Non-GM agents (bottom right).
Initialized networks favour the proxy on average (grey bars, 95\% CI including $1$ in \textsc{Attribute Maze} and \textsc{SortZones}), and GM training keeps this preference while Non-GM training reverses it (Appendix~\ref{app:attribute-tug-of-war}).
The environments differ in the rest of the geometry (other three panels), and this difference suggests what to expect when we try to recover the goal next.

A global shared axis makes a simple prediction.
If the competition in every state is mostly one signed contrast, $d_{\mathrm{comp}}(s)\approx\sigma(s)\beta(s)u_{\mathrm{comp}}$, then one fixed push along $u_{\mathrm{comp}}$, applied with the sign $\sigma(s)$ of the state, should move the policy toward the goal in every state, without rebuilding the direction for each state.

\textsc{Attribute Maze}, with the cleanest push and pull on one shared global axis, should be the easiest to recover.
Its two cues are near antipodal within a state, the competition collapses onto one axis shared across states ($\kappa_F$ $0.793$ above initialization), both cues pull their own way in almost every state, and the tug of war is close to balanced.
A single push along $u_{\mathrm{comp}}$ is therefore a push from the proxy toward the goal in almost every state.

\quadname{} sits at the other end.
Its competition geometry is largely individual to each state, with within-state opposition close to its initial value, a shared axis that rises only $0.213$ above initialization and both cues pulling their own way in only $0.650$ of states.
Without a global axis there is no single direction along which to steer it.

\textsc{SortZones} has a push and pull within each state but a weaker shared axis.
It is the only robot task whose cues are clearly opposed in GM agents ($\cos_F=-0.435$, against $0.147$ in Non-GM agents), in the gripper motion and in the joint directions that leave the gripper in place alike, but its axis rises only $0.249$ above initialization.
Its decision is where to push the cube, so the competition direction depends on where the cube and the gripper are at the decision frame, and the policy reads a broad direction rather than one contrast.
Steering along a shared axis may therefore be less efficient there, while steering along the direction of each state can still work.

\textsc{AttributeTable} has the opposite profile.
Its shared axis is strong, with $\kappa_F$ $0.599$ above initialization and more than half of the Fisher energy of its competition direction on the left-right gripper contrast, but its two cues are not opposed within a state on average, because the arm's redundant joints carry an aligned component that cancels the opposed gripper motion.
A push along the axis moves the gripper contrast but leaves this aligned part in place, so moving the policy toward the goal may need a different or larger effort, similar to \quadname{}.

These expectations rest on three seeds per cohort, and the robot tasks add caveats of their own.
Their policies act in a continuous space whose Fisher metric has rank at most eight, their pull ratios start above one at initialization, and the \textsc{SortZones} decision frames come from GM and Non-GM rollouts in equal parts and so lie partly off each cohort's own state distribution.
The next section tests them by steering the existing wiring (Appendix~\ref{app:shared-axis-steering}).

\clearpage
\section{Recoverability: Can Goal-Directed Behavior Be Recovered?}
\label{sec:recoverability-app}

We have located where the goal is lost and which role the proxy takes in each environment.
We now ask whether the goal can be recovered from the representations the agents already have, either through the existing wiring of the policy, by steering (Appendix~\ref{app:steering}), or by relearning the wiring downstream of a representation, by rewiring (Appendix~\ref{sec:rewiring}).
The two methods act at different depths of the network and answer two questions.
First, can a representation found inside the agent be used as a goal at all?
Information present in an agent need not be usable to orient its behavior.
Second, which recovery method is needed at which depth?
Our hypothesis is that the depth at which the goal is lost indicates which method works.
If the goal is lost only at the decision, a goal representation is present but too weak, and steering the existing policy should suffice.
If goal information reaches the policy but is not used to navigate, the existing wiring is not enough to change behavior, but a new readout trained on the existing representation can be taught to use it.
If the goal is already lost in the representation, neither method should work, and recovery would need new training.
Both methods therefore ask whether the information inside the agent can be reused to move it to the right target, not whether they restore a goal-directed agent.

\subsection{Steering the Existing Wiring}
\label{app:steering}

The preceding analyses identify internal goal-related directions, but geometry alone does not establish that these directions control behavior.
A representation displacement can be decodable, policy-sensitive, or geometrically well organized without providing a useful intervention direction.
We therefore ask whether the internal structure identified above can be used to redirect the existing policy without retraining its weights.

The geometry suggests two increasingly specific causal questions.
\emph{First, can the state-specific counterfactual direction available at the current state causally redirect behavior?}
\emph{Second, in the environments where this local intervention succeeds, how much of that causal effect survives compression into the shared competition axis identified across states?}

All interventions are applied at $h_{\mathrm{fc}}$, after the final ReLU.
This is the representation the policy and value heads read directly, so a push there changes the input to the head without passing through any further nonlinearity.
The counterfactual direction used for steering is reconstructed at the same representation at which it is injected.
In \textsc{Attribute Maze}, pushing at $h_{\mathrm{pre}}$ instead gives GM policies a reach within $0.06$ of the $h_{\mathrm{fc}}$ push at every dose up to $\beta=1$ ($0.933$ against $0.953$ at $\beta=1$), so the choice of site does not drive the results below.

\subsubsection{Can the state-local counterfactual direction redirect the policy?}
\label{app:local-steering}

\paragraph{The intervention.}
The most direct causal use of the counterfactual analyses is to retain the complete state-dependent direction.
This makes no assumption that the relevant geometry is shared across states or low-dimensional.
At each rollout state $s_t$, we reconstruct the counterfactual representations associated with the intended and proxy-supported alternatives and use their contrast as the intervention direction $d_{\mathrm{local}}(s_t)$.
Its positive orientation always points toward the counterfactual in which the evidence supports the intended target.
Where goal and proxy cues sit on the same candidates (\textsc{Attribute Maze}, \textsc{AttributeTable}, \textsc{SortZones}) and in \quadname{}, this is exactly the competition direction defined above,
\[
d_{\mathrm{local}}(s_t)
=
d_{\mathrm{comp}}(s_t)
=
\Delta_p(s_t)-\Delta_\star(s_t).
\]
Its swaps put the proxy cue on the goal candidate and the goal cue on the proxy candidate (colour and shape in \textsc{Attribute Maze} and \textsc{AttributeTable}, colour and cross in \textsc{SortZones}, flag and tower in \quadname{}).
In \textsc{Maze}, the two counterfactuals place the cheese at its own cell and in the top-right corner, $d_{\mathrm{local}}(s_t)=h_{\mathrm{fc}}(s_t^{\star})-h_{\mathrm{fc}}(s_t^{p})$, in \textsc{LocationReach} the ball at its own position and in the training corner, and in \textsc{PushToMarker} the marker at its own position and at the table centre.
In \textsc{Coin}, they place the coin at its position and on the shortcut side, ahead of the agent, and rollouts start from the edited conflict state in which the coin lies behind the agent, off the shortcut path.
The important commonality is that the intervention is reconstructed from the current state at every step rather than replaced by one fixed direction learned across states.
In the robot tasks, the edited copies of the scene are re-synchronised with the acting copy after every step, so that $d_{\mathrm{local}}(s_t)$ always contrasts the same arm, cube or body state.

For steering dose $\beta$, the corrective intervention, a state-local form of activation addition \citep{turner2023actadd}, is
\[
\delta_{\mathrm{local}}^{+}(s_t;\beta)
=
\beta\,d_{\mathrm{local}}(s_t),
\]
giving
\[
h_{\mathrm{fc}}'(s_t)
=
h_{\mathrm{fc}}(s_t)
+
\delta_{\mathrm{local}}^{+}(s_t;\beta).
\]
The anticorrective intervention reverses the same direction,
\[
\delta_{\mathrm{local}}^{-}(s_t;\beta)
=
-\beta\,d_{\mathrm{local}}(s_t).
\]

This sign reversal is important.
A direction that merely perturbs the policy may occasionally improve goal reach without being a meaningful causal axis.
A genuine target-selection direction should instead have an ordered effect.
Moving along the corrective orientation should favour intended-goal behavior, while reversing the same intervention should move behavior in the opposite direction.

For sufficiently small interventions, this follows the local first-order prediction
\[
C_s(h+\delta)-C_s(h)
\approx
\nabla_h C_s(h)^\top \delta,
\]
where $C_s$ is the target-selection contrast defined above.
The rollout experiment asks the stronger question of whether this local effect remains useful after repeatedly intervening as the state changes.

\paragraph{Controls and dose response.}
We compare four conditions,
\[
\text{baseline},
\qquad
\text{matched random},
\qquad
\text{corrective},
\qquad
\text{anticorrective}.
\]
The baseline leaves the representation unchanged.
The random control injects three random directions with norm matched to the corresponding treatment intervention.
It therefore controls for the possibility that sufficiently large perturbations alter behavior regardless of their orientation.
The corrective and anticorrective conditions use opposite signs of the same state-local direction.
Together with the random control, this separates directional causal control from generic intervention effects.

We evaluate the dose grid $\beta\in\{0.125,\ 0.25,\ 0.5,\ 1.0\}$ in \textsc{Maze} and \textsc{Attribute Maze}, extend it to $\beta\in\{0.75,\ 1.25,\ 1.5,\ 2.0\}$ in \textsc{Attribute Maze}, and use $\beta\in\{0.25,\ 0.5,\ 1,\ 2,\ 4\}$ in \textsc{Coin}, where the direction is smallest in norm.
The robot tasks use $\beta\in\{0.125,\ 0.25,\ 0.5,\ 1,\ 2\}$, extended to $\beta=4$ in \textsc{PushToMarker} and \quadname{}.
This lets us distinguish a stable dose-response relationship from an effect that occurs only at one arbitrary intervention magnitude.
The policies act greedily, with the most likely action in Procgen and the mean action in the robot tasks.

\paragraph{Evaluation states.}
Every steering state is a conflict state in which goal and proxy require different behavior, taken from the conflict banks or from the steering samples of Appendix~\ref{app:banks}, and in \quadname{} from the conflict-test bank with the native start pose, as in rewiring.
Steering fits nothing to these states, and the shared axis below is fit on states disjoint from them.

\paragraph{Outcomes.}
The primary outcome is intended-target reach.
Where an episode can also end at the proxy target, reach factorizes exactly into decisiveness and conditional preference,
\[
\mathrm{reach}
=
D\cdot \Pr(\mathrm{goal}\mid\mathrm{dec}),
\qquad
D=\frac{N_{\mathrm{goal}}+N_{\mathrm{proxy}}}{N}.
\]
$D$ asks whether the policy still completes the task at all, and $\Pr(\mathrm{goal}\mid\mathrm{dec})$ asks which target it chooses when it does.
This lets us tell a change in target selection apart from a policy that has simply been pushed into timeouts.
The factorization applies in \textsc{Attribute Maze} and \textsc{Coin}, and in every robot task.
In \textsc{Maze}, the episode ends only at the cheese, so reach is the only outcome.
In \textsc{LocationReach}, \textsc{AttributeTable} and \textsc{SortZones} the outcome is the first candidate the gripper touches or the cube enters, in \textsc{PushToMarker} it is where the cube rests at the end of the episode, and in \quadname{} it is the first object the robot reaches.

\begin{figure}[t]
    \centering
    \includegraphics[width=\linewidth]
    {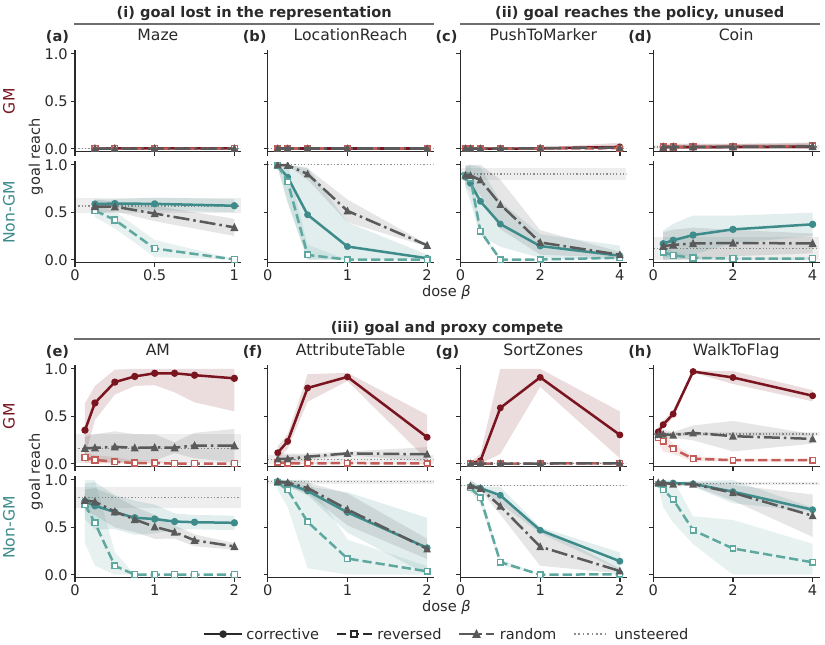}
    \caption{
    \textbf{State-local steering, ordered by where the goal is lost.}
    Goal reach under corrective, reversed and norm-matched random pushes at $h_{\mathrm{fc}}$ against the dose $\beta$, GM (upper row of each block) and Non-GM (lower row).
    AM is \textsc{Attribute Maze} on the black background, the textured background is in Table~\ref{tab:steering-local}, and \quadname{} uses the native set.
    Grey bands show the unsteered baseline.
    Mean with 95\% $t$-interval over three checkpoints.
    }
    \label{fig:steering-local}
\end{figure}

\begin{table}[t]
\centering
\small
\setlength{\tabcolsep}{4pt}
\caption{
\textbf{State-local steering at $\beta=1$.}
Intended-target reach, mean $\pm$ SD across three checkpoints.
Each entry is read against the unsteered baseline in its own row.
Blocks follow Figure~\ref{fig:steering-local}. \quadname{} rows use the native set.
Bold marks corrective values whose 95\% $t$-interval lies above those of both the unsteered baseline and the random push.
}
\label{tab:steering-local}
\begin{tabular}{llcccc}
\toprule
Environment & Cohort
& Baseline
& Corrective
& Anticorrective
& Random \\
\midrule
\textsc{Maze} & GM
& $0.001\pm0.00$ & $0.002\pm0.00$ & $0.000\pm0.00$ & $0.001\pm0.00$ \\
 & Non-GM
& $0.567\pm0.03$ & $0.569\pm0.03$ & $0.002\pm0.00$ & $0.340\pm0.04$ \\
\textsc{LocationReach} & GM
& $0.000\pm0.00$ & $0.000\pm0.00$ & $0.000\pm0.00$ & $0.000\pm0.00$ \\
 & Non-GM
& $1.000\pm0.00$ & $0.138\pm0.10$ & $0.000\pm0.00$ & $0.514\pm0.05$ \\
\midrule
\textsc{PushToMarker} & GM
& $0.000\pm0.00$ & $0.000\pm0.00$ & $0.000\pm0.00$ & $0.000\pm0.00$ \\
 & Non-GM
& $0.903\pm0.02$ & $0.375\pm0.10$ & $0.000\pm0.00$ & $0.582\pm0.10$ \\
\textsc{Coin} & GM
& $0.018\pm0.01$ & $0.016\pm0.01$ & $0.020\pm0.01$ & $0.020\pm0.01$ \\
 & Non-GM
& $0.121\pm0.05$ & $0.259\pm0.08$ & $0.019\pm0.01$ & $0.171\pm0.07$ \\
\midrule
\textsc{Attribute Maze} (black) & GM
& $0.160\pm0.06$ & $\mathbf{0.953\pm0.05}$ & $0.007\pm0.01$ & $0.171\pm0.06$ \\
 & Non-GM
& $0.813\pm0.05$ & $0.587\pm0.06$ & $0.000\pm0.00$ & $0.504\pm0.05$ \\
\textsc{Attribute Maze} (texture) & GM
& $0.040\pm0.07$ & $\mathbf{0.840\pm0.18}$ & $0.000\pm0.00$ & $0.049\pm0.07$ \\
 & Non-GM
& $0.567\pm0.13$ & $0.507\pm0.05$ & $0.000\pm0.00$ & $0.418\pm0.07$ \\
\textsc{AttributeTable} & GM
& $0.040\pm0.02$ & $\mathbf{0.915\pm0.02}$ & $0.005\pm0.00$ & $0.108\pm0.01$ \\
 & Non-GM
& $0.983\pm0.01$ & $0.657\pm0.09$ & $0.170\pm0.08$ & $0.688\pm0.07$ \\
\textsc{SortZones} & GM
& $0.000\pm0.00$ & $\mathbf{0.910\pm0.04}$ & $0.002\pm0.00$ & $0.001\pm0.00$ \\
 & Non-GM
& $0.943\pm0.00$ & $0.468\pm0.01$ & $0.000\pm0.00$ & $0.296\pm0.08$ \\
\quadname{} & GM
& $0.312\pm0.01$ & $\mathbf{0.971\pm0.01}$ & $0.052\pm0.01$ & $0.325\pm0.03$ \\
 & Non-GM
& $0.961\pm0.01$ & $0.958\pm0.01$ & $0.469\pm0.06$ & $0.955\pm0.01$ \\
\bottomrule
\end{tabular}
\end{table}

\paragraph{Results.}
Steering works exactly following the depth at which the goal is lost, confirming our hypothesis for the environments we tested (Figure~\ref{fig:steering-local}, Table~\ref{tab:steering-local}).
Where the goal is lost in the representation or reaches the policy without being used, no push makes the GM policy pursue it.
Where goal and proxy compete at the decision, the state-local push raises GM goal reach to at least $0.84$.

\emph{(i) Goal lost in the representation: \textsc{Maze} and \textsc{LocationReach}.}
No condition moves the GM policy in either environment.
It reaches the cheese in $0.001$ of the $320$ states without steering and in at most $0.002$ under any push at any dose, and it never touches the ball first in \textsc{LocationReach}.
The representation and policy results leave nothing to steer here.
Cheese bearing is not decodable beyond the agent's own position, the cheese edit barely reaches the GM policy ($Q=0.001$ against $1.689$, Section~\ref{sec:goal-read}), and in \textsc{LocationReach} a moved ball even turns the planned GM step slightly away from it (Appendix~\ref{app:world-edit}).
The steering direction is built from these edits, so it is small, with a median $\|d_{\mathrm{local}}(s_0)\|_2$ of $1.40$ against $17.97$ for Non-GM policies in \textsc{Maze} and $2.536$ against $20.015$ in \textsc{LocationReach}.
Large pushes sometimes reach the goal.
Pushed at fixed norms of $16$ to $32$, about the Non-GM norm at $\beta=1$, GM policies in \textsc{Maze} reach the cheese in up to $7$ of $48$ states for one seed ($0.090$ averaged over seeds at norm $32$), above random pushes of equal norm (at most $0.049$), while the reversed push reaches it in up to $4$ of $48$.
We read this as a large perturbation in the direction of the cheese rather than goal pursuit, since it stays far below Non-GM cheese pursuit and is built from an edit the GM policy barely registers.
In \textsc{LocationReach} the grid stops where the GM push is a quarter of the Non-GM push at $\beta=1$, so there we cannot separate a missing direction from a small one.
In Non-GM policies, the same construction is clearly causal in the other direction.
The anticorrective push removes cheese pursuit almost completely, from $0.567$ to $0.002$ at $\beta=1$, while random pushes of equal norm only reduce it to $0.340$, and in \textsc{LocationReach} it sends the gripper to the corner in $0.983$ of states.
So, removing the cheese-associated direction from a policy that already uses the cheese destroys goal pursuit, but adding the same kind of direction to GM does not create cheese pursuit.
The intervention can suppress an existing computation, but it cannot instantiate one that the native GM policy does not already use.
This matches the \textsc{Maze} Fisher analysis, computed on a separate set of decision states, where the policy-sensitive energy of the GM cheese edit is roughly three orders of magnitude below that of Non-GM policies and its signed effect on target selection is indistinguishable from zero.
This asymmetry is consistent with earlier interventions on \textsc{Maze} policies, where subtracting a cheese vector suppressed cheese-seeking while adding it left cheese-seeking unchanged, and editing goal-tracking channels retargeted the policy only partially~\citep{mini2023understandingcontrollingmazesolvingpolicy}.
Those policies were trained with the cheese anywhere in the top-right $5\times5$ region and pursued it at baseline, whereas our GM agents are trained with the cheese on the exact corner cell (\texttt{rand\_region}$=0$) and our Non-GM agents with a randomized position (\texttt{rand\_region}$=25$), so the agreement is in the direction of the effect rather than a replication of our setting.

\emph{(ii) Goal reaches the policy but is not used: \textsc{PushToMarker} and \textsc{Coin}.}
Here the goal edit still moves the GM policy, if weakly, but no push creates goal pursuit.
In \textsc{PushToMarker}, GM goal reach is $0.000$ at $\beta=1$ and at most $0.018$ at $\beta=4$.
Push size does not explain this, since at $\beta=2$ the GM push has the norm of the Non-GM push at $\beta=1$ (median $2.516$ against $5.033$) and GM goal reach is still $0.003$.
What the corrective push does is weaken the substitute target, with the cube resting at the centre in $0.628$ of scenes at $\beta=1$ against $0.925$ unsteered and $0.896$ under random pushes, without moving the cube to the marker.
So the push breaks the pursuit of the centre but gives the head no marker location to pursue instead.
In \textsc{Coin}, GM coin reach is $0.018$ without steering and stays between $0.016$ and $0.027$ under corrective, anticorrective and random pushes at every dose up to $\beta=4$.
At $\beta=4$, the corrective push exceeds the random push by $-0.001$ ($95\%$ CI $\pm0.013$ across checkpoints), and the conditional preference $\Pr(\mathrm{goal}\mid\mathrm{dec})$ moves only from $0.028$ to $0.035$.
Because the GM baseline is close to zero, the anticorrective push cannot show a decrease, so the evidence for GM rests on the corrective push matching the random push.
Here too push size is not the reason, since at $\beta=4$ the GM push has about the norm of the Non-GM push at $\beta=1$ (mean $1.93$ against $7.85$), where Non-GM policies already respond.
Both nulls follow from how the goal reaches these policies.
The goal edit is registered as a change of the scene but not as a direction.
It reaches the GM policy at about a sixth of the Non-GM sensitivity in \textsc{PushToMarker} and a sixteenth in \textsc{Coin} (ratios of $Q$ of $0.159$ and $0.063$, Appendix~\ref{sec:fisher}), but its signed follow is close to zero in both (ratios $0.017$ and $0.001$, Appendix~\ref{app:world-edit}), and the role analysis finds the GM reaction to be the one any disc gets in \textsc{PushToMarker} and spread over the actions in proportion to their prior probability in \textsc{Coin} (Appendices~\ref{app:ptm-role} and~\ref{app:coin-functional-role}).
The steering direction is built from this edit, so it carries a change of the scene without an orientation, and even at the norm that moves Non-GM policies the push is too weak to be registered as an orientation toward the goal.
In Non-GM policies, the same construction is causal in both directions and grows with dose.
In \textsc{Coin}, the corrective push raises reach from $0.121$ to $0.259$ at $\beta=1$ and $0.373$ at $\beta=4$, against $0.171$ for random pushes, and the anticorrective push lowers it to $0.019$ and $0.012$, and in \textsc{PushToMarker} the reversed push sends the cube to the centre in $0.943$ of scenes.
So, in both environments the direction is causally live where the policy reads it, and inert in GM policies.

\emph{(iii) Goal and proxy compete: \textsc{Attribute Maze}, \textsc{AttributeTable}, \textsc{SortZones} and \quadname{}.}
The local competition direction is a strong causal handle on the GM policy in every competition environment.
On the black background, intended-goal reach rises from $0.160$ without steering to $0.953$ at $\beta=1$, on the textured background from $0.040$ to $0.840$, in \textsc{AttributeTable} from $0.040$ to $0.915$ and in \textsc{SortZones} from $0.000$ to $0.910$.
In \quadname{} flag-first reach rises from $0.312$ to $0.971$, against $0.325$ under random pushes, and the reversed push lowers it to $0.052$.
As a robustness check, the forced-view set, in which both objects are in view from the start, gives the same picture ($0.378$ to $0.992$, random $0.389$, reversed $0.000$).
This was not guaranteed by the construction. A direction built from counterfactual edits at one state could have shifted the first action and then washed out, broken competent navigation, or worked only on a narrow subset of states. Instead it survives closed-loop rollout across the evaluation states.
Norm-matched random pushes leave reach close to its baseline across the full dose range, and the anticorrective push drives it to zero.
So, the effect is directional and ordered, not a consequence of perturbing the representation.
Unlike in (i) and (ii), the GM direction is about as large as the Non-GM one, with a median norm of $10.710$ against $12.033$ in \textsc{AttributeTable} and $7.429$ against $8.244$ in \textsc{SortZones}.

\paragraph{The decomposition shows what the push changes.}
For GM policies on the black background, the conditional preference $\Pr(\mathrm{goal}\mid\mathrm{dec})$ rises from $0.235$ to $0.993$ at $\beta=1$, while decisiveness also rises, from $0.687$ to $0.960$.
It rises from $0.069$ to $1.000$ on the textured background, from $0.048$ to $0.996$ in \textsc{AttributeTable}, from $0.000$ to $1.000$ in \textsc{SortZones} and from $0.315$ to $0.995$ in \quadname{}.
Steering therefore does not merely make the agent more active.
It changes which competing target wins.
The texture policies need a larger dose before this happens, with reach of only $0.167\pm0.24$ at $\beta=0.25$ before rising to $0.740$ at $\beta=0.5$, and so do the robot policies, with $0.235$ (\textsc{AttributeTable}) and $0.030$ (\textsc{SortZones}) at $\beta=0.25$.
At $\beta=2$ robot reach falls back to $0.278$ and $0.303$, as the push becomes large enough to disrupt the reach itself.

Non-GM policies behave differently, and we treat them as a positive control for the direction rather than as a second recovery target.
They already choose the intended target whenever they complete the task, with $\Pr(\mathrm{goal}\mid\mathrm{dec})=1.00$ at baseline on both backgrounds, and this stays at $1.00$ under every corrective dose.
Their reach nevertheless falls, from $0.813$ to $0.587$ on the black background, $0.983$ to $0.657$ in \textsc{AttributeTable} and $0.943$ to $0.468$ in \textsc{SortZones}, because a large push into an already working policy costs decisiveness and produces timeouts.
\quadname{} is the exception, with Non-GM reach $0.958$ at $\beta=1$ against $0.961$ unsteered, falling only at $\beta=4$, to $0.685$, about as far as under random pushes of equal norm ($0.622$), while the reversed push lowers it to $0.130$.
The anticorrective push, by contrast, lowers intended-goal reach to $0.000$ in \textsc{Attribute Maze} and \textsc{SortZones}, $0.170$ in \textsc{AttributeTable} and $0.469$ in \quadname{}, so the same direction still carries target selection in these policies.

\paragraph{Steering redirects the policy toward a location.}
Every competition direction is built from swaps that exchange what lies at the goal location $u$ and the lure location $v$, so the push may tell the policy which cue to follow or only where to go.
We separate the two by acting in a scene where the goal cue is removed or moved to $v$, while the push is still built from the swap copies of the original scene.
The GM push at $\beta=1$ reaches $u$ either way (Table~\ref{tab:steering-location}).
With the goal cue removed, it reaches $u$ in $0.913$ of \textsc{Attribute Maze} states, $0.900$ of \textsc{AttributeTable} and $0.970$ of \textsc{SortZones} scenes, and with the cue moved to the lure, it goes to the now empty $u$ in $0.880$, $0.993$ and $0.985$.
Unsteered GM policies reach $u$ in $0.160$, $0.052$ and $0.000$ of \textsc{Attribute Maze}, \textsc{AttributeTable} and \textsc{SortZones} scenes with the cue in place, at most $0.023$ with it removed and never with it moved.
In \textsc{Attribute Maze} the control scenes hold one object, which no policy saw in training, but the decoupled scene is still informative, since unsteered agents there reach the lone goal object at $v$.
So the push redirects the policy using information inside the agent, and we describe all steering below, local and shared-axis, in this way and not as recovered goal pursuit.
This follows from what steering does.
It does not change the wiring, the weights with which the policy reads out its inputs.
It injects information, changing the size of the arrows that point into the policy rather than how they are read, a kind of overwriting, so it does not amplify the agent's perception of the goal.
This is consistent with the environments where the goal is inert, where the information that steering would need to reorient the agent is not there.
Steering still shows that a goal-relevant representation can be extracted from the internals of a misgeneralized agent and used to make it goal-seeking again, without any diversified training set and with only a permutation of the scene at test time.
This is not ready for deployment, since a scene cannot always be edited, but it is a form of recovery, though not a recovery of navigation.

\begin{table}[t]
\centering
\small
\setlength{\tabcolsep}{4pt}
\caption{
\textbf{Steering redirects the policy toward a location.}
Reach of the goal location $u$ by GM policies, mean $\pm$ SD across three checkpoints, with the goal cue in place, removed, or moved to the lure location $v$.
The push is built from the original scene in all three.
Bold marks the local push, which keeps reach of $u$ high with the goal cue removed or moved.
The robot rows are a separate set of rollouts on the same $200$ scenes as Table~\ref{tab:steering-local}, so their cue-in-place values differ from it by at most $0.012$, and their shared axis is fit on the world-edit bank (cosine $0.935$ with the strict-bank axis of Table~\ref{tab:steering-compression}).
}
\label{tab:steering-location}
\begin{tabular}{llccc}
\toprule
Environment & Push & Goal cue at $u$ & Removed & Moved to $v$ \\
\midrule
\textsc{Attribute Maze} & none
& $0.160\pm0.06$ & $0.007\pm0.01$ & $0.000\pm0.00$ \\
(black) & local, $\beta=1$
& $\mathbf{0.953\pm0.05}$ & $\mathbf{0.913\pm0.03}$ & $\mathbf{0.880\pm0.07}$ \\
 & reversed, $\beta=1$
& $0.007\pm0.01$ & $0.000\pm0.00$ & $0.000\pm0.00$ \\
 & random, $\beta=1$
& $0.171\pm0.06$ & $0.024\pm0.01$ & $0.031\pm0.01$ \\
 & shared axis, $\beta=0.25$
& $0.587\pm0.11$ & $0.093\pm0.04$ & $0.147\pm0.06$ \\
\midrule
\textsc{AttributeTable} & none
& $0.052\pm0.02$ & $0.023\pm0.02$ & $0.000\pm0.00$ \\
 & local, $\beta=1$
& $\mathbf{0.915\pm0.02}$ & $\mathbf{0.900\pm0.04}$ & $\mathbf{0.993\pm0.01}$ \\
 & reversed, $\beta=1$
& $0.002\pm0.00$ & $0.005\pm0.01$ & $0.000\pm0.00$ \\
 & random, $\beta=1$
& $0.106\pm0.01$ & $0.081\pm0.01$ & $0.023\pm0.01$ \\
 & shared axis, $\beta=0.25$
& $0.310\pm0.04$ & $0.247\pm0.04$ & $0.198\pm0.06$ \\
\midrule
\textsc{SortZones} & none
& $0.000\pm0.00$ & $0.000\pm0.00$ & $0.000\pm0.00$ \\
 & local, $\beta=1$
& $\mathbf{0.907\pm0.05}$ & $\mathbf{0.970\pm0.01}$ & $\mathbf{0.985\pm0.01}$ \\
 & reversed, $\beta=1$
& $0.000\pm0.00$ & $0.002\pm0.00$ & $0.002\pm0.00$ \\
 & random, $\beta=1$
& $0.002\pm0.00$ & $0.000\pm0.00$ & $0.000\pm0.00$ \\
\bottomrule
\end{tabular}
\end{table}

\subsubsection{How much control survives compression onto a shared axis?}
\label{app:shared-axis-steering}

\paragraph{Design.}
State-local steering retains the full counterfactual direction $d_{\mathrm{comp}}(s)$ at every state.
It therefore provides the strongest test of causal accessibility, but it does not tell us how much of the intervention is actually necessary.
The geometry analysis gives us a more specific hypothesis.
In \textsc{Attribute Maze}, within the canonical UP-versus-RIGHT conflict class, the policy-relevant component of $d_{\mathrm{comp}}(s)$ is well described by one signed axis, $d_{\mathrm{comp}}(s)\approx\sigma(s)\beta(s)u_{\mathrm{comp}}$ (Section~\ref{app:attribute-cross-state-geometry}), and \textsc{AttributeTable} is the robot task with the strongest shared axis, while \textsc{SortZones} has the weakest (Appendix~\ref{app:attribute-competition-geometry}).
If the successful local intervention acts primarily through this shared axis, then much of its causal effect should survive after the remaining state-specific structure is removed.

We fit the axis only on the $146$ fit states of the conflict bank, which are disjoint from the $50$ evaluation states used for steering, with mirrored states always kept on the same side of the split.
Within the fit states, $118$ determine the axis and $28$ serve as a held-out check before any steering.
In \textsc{AttributeTable} the axis is fit on all $512$ scenes of the strict bank, in which the two objects lie on opposite sides of the robot, and in \textsc{SortZones} on all $500$ frames of the decision-frame bank, at which the hand first reaches the cube (Appendix~\ref{app:banks}).
Neither bank shares scenes with the world-edit scenes on which the robots are steered.
In \quadname{} the axis is fit on the states of the world-edit bank, and the fixed pushes were run in GM and Non-GM policies on the forced-view set and on the native set.
The axis is the sign-folded mean of the competition directions,
\[
u_{\mathrm{comp}}
=
\frac{
\bar d_{+}-\bar d_{-}
}{
\|\bar d_{+}-\bar d_{-}\|_2
},
\qquad
\bar d_{\pm}
=
\frac{1}{|S_{\pm}|}
\sum_{s\in S_{\pm}}
d_{\mathrm{comp}}(s),
\]
where $S_{+}$ contains the fit states in which the line candidate lies along UP and $S_{-}$ those in which it lies along RIGHT, and in the robot tasks the states in which the goal candidate lies on one side of the robot and on the other.
The sign $\sigma(s)\in\{-1,+1\}$ used at a steering state is determined only by this orientation of its conflict.
The rollout outcomes used to evaluate steering do not enter this construction.

We compare the complete local intervention with progressively more compressed versions of it.
Every compressed intervention is a fixed push, computed once at the start state $s_0$ and held constant during the rollout.
The fair reference for compression is therefore the local direction fixed in the same way, $\delta_{\mathrm{fixed}}(\beta)=\beta\,d_{\mathrm{comp}}(s_0)$.
The fully local intervention, which rebuilds $d_{\mathrm{comp}}(s_t)$ at every step, remains the positive control.
The sign-gated shared-axis intervention removes the state-specific direction while preserving its local magnitude and geometry-derived orientation,
\[
\delta_{\mathrm{shared}}(\beta)
=
\beta\,
\sigma(s_0)\,
\|d_{\mathrm{comp}}(s_0)\|_2\,
\hat u_{\mathrm{comp}}.
\]
This is the central compression hypothesis.
If it approaches the effect of $\delta_{\mathrm{fixed}}$, then the causal component of $d_{\mathrm{comp}}(s_0)$ is largely captured by one shared direction together with its sign.
We then remove the sign gate, $\delta_{\mathrm{unsigned}}(\beta)=\beta\,\|d_{\mathrm{comp}}(s_0)\|_2\,\hat u_{\mathrm{comp}}$.
This preserves the shared axis and the intervention magnitude but uses the same orientation in every state.
It therefore tests whether the sign identified by the geometry is causally necessary.
Finally, we retain only the component of the local direction that the policy-sensitive geometry assigns to the shared axis.
Using the Fisher-orthogonal coefficient of Appendix~\ref{app:projection-analysis},
\[
\alpha_F(s_0)
=
\frac{
u_{\mathrm{comp}}^\top
F(s_0)
d_{\mathrm{comp}}(s_0)
}{
u_{\mathrm{comp}}^\top
F(s_0)
u_{\mathrm{comp}}
},
\qquad
\delta_F(\beta)
=
\beta\,
\alpha_F(s_0)\,
u_{\mathrm{comp}}.
\]
Each intervention is also applied with the opposite sign, and a norm-matched random direction again provides the nonspecific perturbation control.

At large doses the fixed pushes stop testing the axis.
At $\beta=1$, the sign-gated and unsigned pushes collapse every rollout into one repeated movement, and reach then equals a ceiling set by the maze walls, $0.48$ on the evaluation states.
In \textsc{AttributeTable} the sign-gated push already falls from $0.300$ at $\beta=0.25$ to $0.053$ at $\beta=0.5$.
We therefore compare the compressed interventions at $\beta\le0.25$, where none of them has reached this ceiling.
The fully local intervention does not have this problem, because its direction is rebuilt as the agent moves.

\begin{figure}[t]
    \centering
    \includegraphics[width=\linewidth]
    {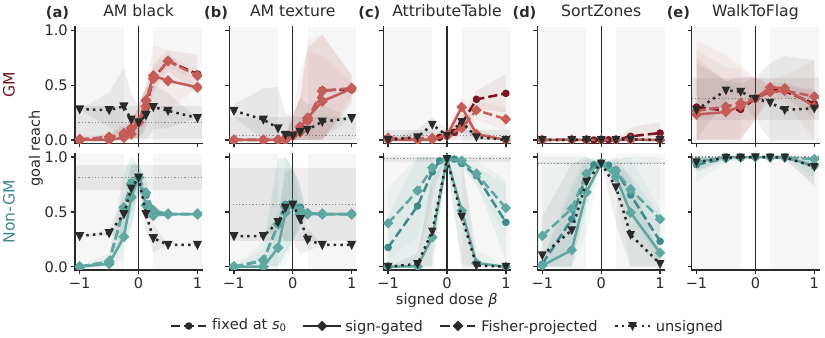}
    \caption{
    \textbf{Compressing local steering onto the shared axis.}
    Goal reach of GM (top) and Non-GM (bottom) policies against the signed dose, negative doses reversed.
    All arms are fixed pushes computed at the start state.
    The \textsc{AttributeTable} axis is fit on the strict bank and the \textsc{SortZones} axis on the decision-frame bank.
    \quadname{} is shown on the forced-view set. On the native set no fixed push moves reach in either cohort (see text).
    Shaded regions mark $|\beta|>0.25$.
    Mean with 95\% $t$-interval over three checkpoints.
    }
    \label{fig:steering-compression}
\end{figure}

\begin{table}[t]
\centering
\small
\setlength{\tabcolsep}{3pt}
\caption{
\textbf{Compression test at $\beta=0.25$.}
Intended-goal reach of GM policies, mean $\pm$ SD across three checkpoints.
AM, AT, SZ and WTF are \textsc{Attribute Maze}, \textsc{AttributeTable}, \textsc{SortZones} and \quadname{} (forced-view set).
In the texture row, the robustness variant of \textsc{Attribute Maze}, one of the three checkpoints carries the mean and the variation across checkpoints is as large as the effects.
Bold marks compressed pushes toward the goal (sign-gated, Fisher-projected and unsigned) whose mean $\pm$ SD range lies entirely above that of the unsteered value.
}
\label{tab:steering-compression}
\resizebox{\linewidth}{!}{%
\begin{tabular}{lccccccc}
\toprule
Environment
& Unsteered
& \begin{tabular}[b]{@{}c@{}}Local,\\fixed at $s_0$\end{tabular}
& Sign-gated
& \begin{tabular}[b]{@{}c@{}}Fisher-\\projected\end{tabular}
& Unsigned
& \begin{tabular}[b]{@{}c@{}}Unsigned,\\reversed\end{tabular}
& \begin{tabular}[b]{@{}c@{}}Local,\\rebuilt\end{tabular} \\
\midrule
AM, black
& $0.160\pm0.06$ & $0.560\pm0.05$ & $\mathbf{0.587\pm0.11}$ & $\mathbf{0.560\pm0.04}$ & $0.300\pm0.08$ & $0.307\pm0.14$ & $0.640\pm0.07$ \\
AM, texture
& $0.040\pm0.07$ & $0.173\pm0.27$ & $0.200\pm0.25$ & $0.180\pm0.29$ & $0.093\pm0.09$ & $0.107\pm0.15$ & $0.167\pm0.24$ \\
AT
& $0.040\pm0.02$ & $0.147\pm0.02$ & $\mathbf{0.300\pm0.04}$ & $\mathbf{0.108\pm0.02}$ & $\mathbf{0.165\pm0.03}$ & $0.140\pm0.03$ & $0.235\pm0.03$ \\
SZ
& $0.000\pm0.00$ & $0.000\pm0.00$ & $0.000\pm0.00$ & $0.000\pm0.00$ & $0.000\pm0.00$ & $0.000\pm0.00$ & $0.030\pm0.03$ \\
WTF
& $0.378\pm0.08$ & $0.466\pm0.01$ & $0.479\pm0.03$ & $0.435\pm0.03$ & $0.341\pm0.15$ & $0.440\pm0.03$ & $0.544\pm0.01$ \\
\bottomrule
\end{tabular}}
\end{table}

\paragraph{Results.}
One shared axis carries the push in \textsc{Attribute Maze} on the black background, part of it in \textsc{AttributeTable}, and none of it in \textsc{SortZones} (Figure~\ref{fig:steering-compression}, Table~\ref{tab:steering-compression}).
On the black background at $\beta=0.25$, the sign-gated shared axis reaches $0.587$ and the Fisher-projected axis $0.560$, compared with $0.560$ for the local direction fixed at the start state.
One shared direction, together with the sign of the conflict, therefore carries essentially all of the causal effect of the state-specific direction when both are held fixed from the start state.
Reversing either compressed intervention suppresses intended-goal reach just as reversing the local direction does.
Without the sign, the same shared direction reaches only $0.300$, and reversing it gives almost the same value, $0.307$.
The remaining gap to the fully local intervention, $0.640$ at the same dose and $0.953$ at $\beta=1$, comes from rebuilding the direction as the state changes during the rollout, not from compressing it.
On the textured background the sign-gated push reaches $0.200$ on average, but $0.100$, $0.020$ and $0.480$ in the three checkpoints, so one checkpoint carries the mean, and we base the compression conclusion on the black-background policies.
Unlike the local push, the sign-gated push depends on the goal cue here. With the cue removed or moved to the lure, it reaches $u$ in $0.093$ and $0.147$ of states at $\beta=0.25$, against $0.587$ with the cue in place (Table~\ref{tab:steering-location}).

In \textsc{AttributeTable}, the sign-gated push raises reach from $0.040$ to $0.300$, against $0.050$ for random pushes of equal norm, and the reversed push lowers it to $0.005$.
It even exceeds the local direction fixed at the start state ($0.147$), partly because the two are not the same dose.
At equal $\beta$ and equal Euclidean norm, the sign-gated push moves the GM policy about twice as far in the Fisher norm (median $19.33$ against $10.13$ per unit $\beta$), since the axis is more policy-sensitive per unit norm than $d_{\mathrm{comp}}(s_0)$ itself.
Here the sign carries only about half of the gain, since the unsigned push also raises reach, by $0.125$ to $0.165$, and so does the reversed unsigned push ($0.140$).
Part of the effect is therefore a push off the proxy that does not depend on the orientation of the conflict, a part that is smaller in \textsc{Attribute Maze} (a gain of $0.140$ unsigned against $0.427$ signed).
Unlike in \textsc{Attribute Maze}, the sign-gated push still reaches the goal location with the goal cue removed ($0.247$) or moved to the lure ($0.198$), against $0.310$ with the cue in place, for an axis fit on the world-edit bank whose cosine with the strict-bank axis is $0.935$ (Table~\ref{tab:steering-location}).
So in \textsc{AttributeTable} the fixed push sets where the policy goes, as the local push does, rather than how strongly it reads the goal cue.
In \textsc{SortZones}, every fixed push stays at $0.000$ at $\beta=0.25$, including the local direction held fixed, which reaches only $0.063$ at $\beta=1$, while the same direction rebuilt at every step reaches $0.910$.
We read this as a difference in what is being chosen.
In \textsc{AttributeTable} the competition is mostly a left or right hand movement, the same contrast in every scene, so one push along it carries part of the effect.
In \textsc{SortZones} the decision is where to push the cube, so the useful direction depends on where the cube and the hand are, which changes as the hand moves.
In \quadname{} the sign-gated push and the local direction held fixed raise reach by about $0.1$ at $\beta=0.25$, to $0.479$ and $0.466$ against $0.378$ unsteered and $0.375$ for random pushes, while the direction rebuilt at every step reaches $0.544$ at the same dose and $0.992$ at $\beta=1$.
On the native set, where the agent starts in its training pose and the two objects need not be in view, no fixed push moves GM reach (all between $0.273$ and $0.339$ up to $|\beta|=1$, against $0.312$ unsteered), which we read as the start state carrying too little of the conflict for a push computed there to steer the rest of the rollout.
The start-state direction is consistent with this, with a median norm of $0.43$ against $8.11$ on the forced-view set, and its Fisher projection agrees with the conflict sign in only about half of the native start states.

Non-GM policies show the same axis from the other side (Figure~\ref{fig:steering-compression}, bottom).
The reversed sign-gated push removes goal pursuit, lowering reach to $0.273$ at $\beta=0.25$ and $0.027$ at $\beta=0.5$ on the black background, and from $0.983$ to $0.268$ in \textsc{AttributeTable} and from $0.943$ to $0.680$ in \textsc{SortZones} at $\beta=0.25$.
\quadname{} is the exception, with Non-GM policies at the ceiling under every fixed push, on the forced-view set (reversed sign-gated push $1.000$ at $\beta=0.25$ and $0.953$ at $\beta=1$) and on the native set (between $0.919$ and $0.971$ up to $|\beta|=1$, against $0.961$ unsteered), while the reversed direction rebuilt at every step lowers forced-view reach to $0.542$ at $\beta=1$.
So in \quadname{} a push held fixed from the start state cannot remove goal pursuit, consistent with the weak shared axis.
In the corrective direction, every fixed push lowers \textsc{Attribute Maze} reach to about the wall-set ceiling of $0.48$, because a policy that already chooses the goal has nothing to gain, and a fixed push collapses its rollouts into one repeated movement.
The shared axis is scoped to the canonical UP-versus-RIGHT conflict class.
As the diversified geometry analysis shows, it should not be read as one universal goal-versus-proxy direction, and steering was not evaluated beyond this class.

\paragraph{How much control survives compression onto a shared axis?}
Control survives only where the geometry found a global axis, and how much survives follows how clean that geometry is (Appendix~\ref{app:attribute-competition-geometry}).
In \textsc{Attribute Maze}, where one action contrast carries $74\%$ to $86\%$ of the Fisher energy of $d_{\mathrm{comp}}$, the fixed push along the axis does as well as the local direction held fixed.
In \textsc{AttributeTable}, where the left-right gripper contrast carries $57\%$, it carries part of the effect, and in \textsc{SortZones} and \quadname{}, whose axes are weak, the fixed pushes do little.
The sign-gated push works and the unsigned push fails because conflicts of opposite orientation point in opposite directions on the axis, as the geometry showed.
Without the sign, one fixed push helps the states of one orientation and hurts the mirrored ones, so it barely moves average reach in either direction.
The Fisher-projected push keeps the axis and the sign but is slightly weaker than the sign-gated push ($0.560$ against $0.587$ in \textsc{Attribute Maze}, $0.435$ against $0.479$ in \quadname{}), and clearly weaker in \textsc{AttributeTable} ($0.108$ against $0.300$).
It retains only the part of $d_{\mathrm{comp}}(s_0)$ that the policy's sensitivity at $s_0$ assigns to the axis and drops the rest, whereas the sign-gated push puts the full norm of $d_{\mathrm{comp}}(s_0)$ on the axis.
At the start states of the steered GM policies, the Fisher-projected push is $0.73$ times the sign-gated push in \textsc{Attribute Maze}, $0.68$ times in \quadname{} and $0.30$ times in \textsc{AttributeTable} (median over states, mean over checkpoints, and $0.67$ in \textsc{SortZones}, where no fixed push moves reach).
Both pushes lie along $u_{\mathrm{comp}}$, so the ratio is the same in the Euclidean and in the Fisher norm.
In \textsc{AttributeTable} and \quadname{} the Fisher push also points against the conflict sign at about a tenth of the start states, and in \textsc{Attribute Maze} at none.
So the Fisher push is the smaller push along the same direction, about a third of the sign-gated push where it loses the most, and it loses the most where the axis carries the least of the competition.

\paragraph{Implication for recoverability.}
Steering asks whether the native policy can be redirected using structure it already reads.
It can wherever goal and proxy compete at the decision, and there the competition contrast is a direct causal handle on which location the policy goes to.
It cannot where the goal is lost in the representation (\textsc{Maze}, \textsc{LocationReach}) or reaches the policy without being used (\textsc{PushToMarker}, \textsc{Coin}), even though the same construction remains causal in matched Non-GM policies.
These nulls are partly expected from Section~\ref{sec:fisher}: the steering directions are built from world edits to which the GM policies barely respond ($Q=0.001$ in \textsc{Maze}, $0.021$ in \textsc{Coin}). What steering adds is the closed-loop test, together with the Non-GM positive control showing that the same construction does redirect a policy that reads the goal.
Where it works, steering redirects the policy using information inside the agent, a form of recovery but not a recovery of navigation.
This leaves open whether the relevant goal information is absent or merely unused by the existing downstream computation.
We therefore next freeze the representation and retrain the readout.

\subsection{Recoverability by rewiring}
\label{sec:rewiring}

\subsubsection{Design}

\paragraph{Motivation.}
The preceding steering experiments ask whether intended-goal behavior can be recovered through the agent's existing downstream policy. Rewiring asks a complementary question. Even when the native downstream computation does not use the goal structure successfully, the goal may still be represented in a form from which a new downstream policy can recover goal-directed behavior. We therefore freeze the representation and progressively rebuild the downstream computation, from replacing only the policy head \citep{kirichenko2023lastlayer} to relearning the transformation from the final spatial representation to actions.

\textit{If the frozen representation still contains information in a form useful for goal-directed control, can rebuilding the downstream wiring recover the intended behavior?}

\begin{figure*}[t]
    \centering
    \includegraphics[width=0.98\textwidth]{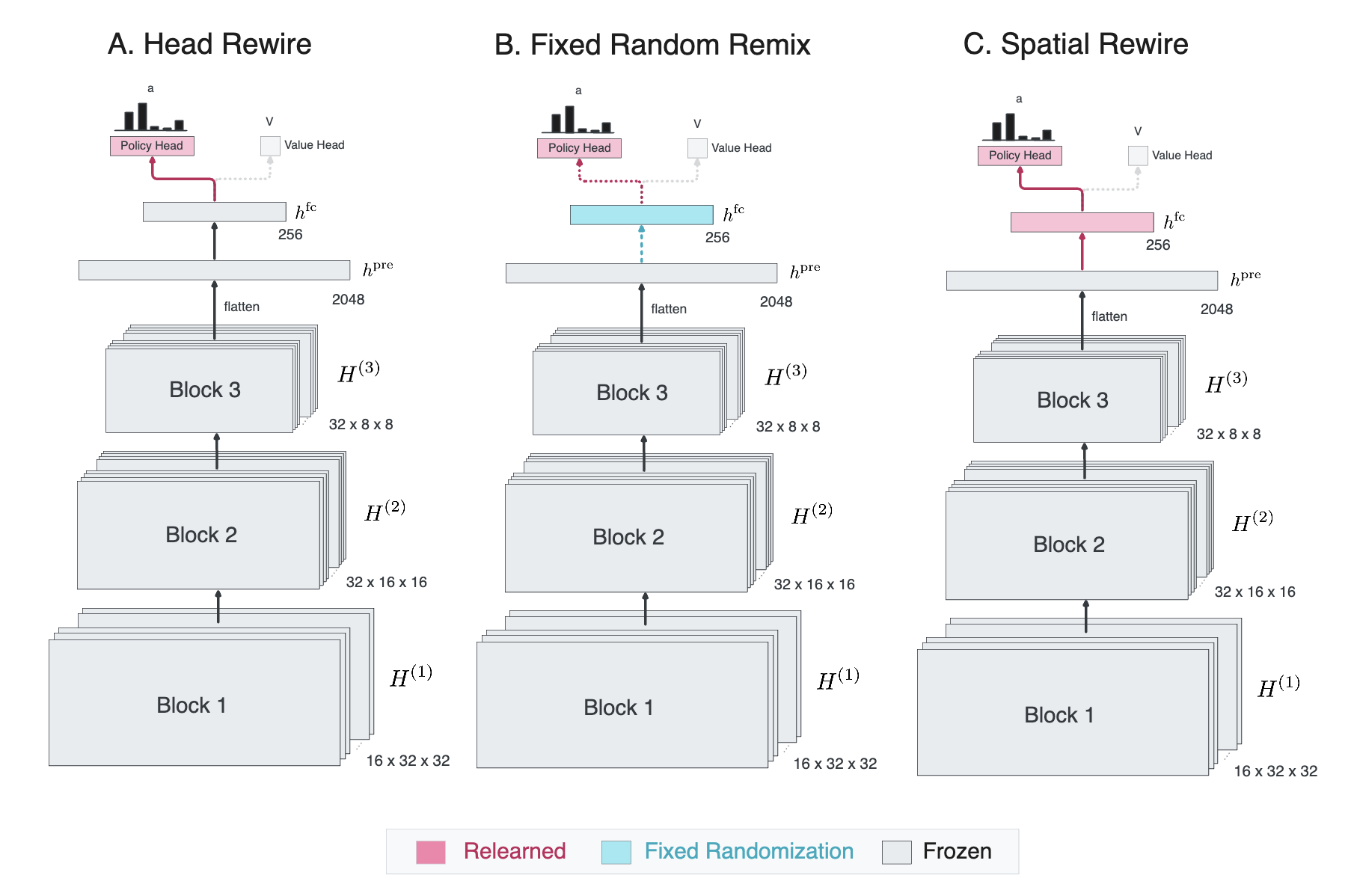}
    \caption{
    \textbf{Frozen-representation rewiring.}
    The diagram shows the Procgen IMPALA network. The robot networks follow the same scheme with a Nature CNN trunk instead (Appendix~\ref{sec:architecture}). What the figure shows is which downstream part each rewire relearns, not the trunk, which is frozen in every arm.
    Grey components are frozen and pink components are freshly initialized and
    trained.
    \textbf{A}, the head rewire retains the learned $h_{\mathrm{fc}}$
    representation and learns only a new policy head.
    \textbf{B}, the fixed random remix replaces the learned
    $h_{\mathrm{pre}}\!\to h_{\mathrm{fc}}$ transformation by a fixed random
    $2048\!\to256$ map and learns only the policy head.
    \textbf{C}, the spatial rewire starts from $h_{\mathrm{pre}}$ and relearns
    both the intermediate transformation and policy head.
    The value head is not used for behavior cloning.
    }
    \label{fig:rewire-architecture}
\end{figure*}

\paragraph{Frozen-representation rewiring.}
For representation site $m$, we freeze the trained network up to $h^{(m)}$ and replace the learned policy computation downstream of that site.
We denote the freshly initialized downstream policy by
\[
\widetilde{\pi}^{(m)}_{\phi}(\cdot\mid h),
\]
where $\phi$ contains the newly optimized downstream policy parameters.
The rewired policy is
\[
\widetilde{\pi}^{(m)}_{\phi}(\cdot\mid s)
=
\widetilde{\pi}^{(m)}_{\phi}
\!\left(
\cdot\mid h^{(m)}(s)
\right),
\]
where $h^{(m)}(s)$ is always produced by the frozen representation of the original GM or Non-GM checkpoint.
Figure~\ref{fig:rewire-architecture} shows the three arms, the head rewire from $h_{\mathrm{fc}}$, the fixed random remix and the spatial rewire from $h_{\mathrm{pre}}$.

Our primary comparison uses the two representation sites closest to action selection.
The first is the shared fully connected representation
\[
h_{\mathrm{fc}}(s),
\]
which is read directly by the policy head.
Rewiring at this site therefore keeps the entire learned trunk, including the final shared transformation, fixed and learns only a new policy readout.

We additionally rewire one step earlier, from the final spatial representation
\[
h_{\mathrm{pre}}(s)
=
\operatorname{vec}
\!\left(
\operatorname{ReLU}\!\left(H^{(3)}(s)\right)
\right).
\]
This is the post-ReLU representation immediately before the learned fully connected transformation.
Rewiring from $h_{\mathrm{pre}}$ therefore replaces both that transformation and the policy head.

The comparison between these sites is especially informative because $h_{\mathrm{pre}}$ still retains the spatially indexed structure produced by the convolutional trunk.
Flattening only changes the indexing of the tensor, so every component of $h_{\mathrm{pre}}$ still corresponds to a particular channel and spatial location in the final feature map.
By contrast, the learned map from $h_{\mathrm{pre}}$ to $h_{\mathrm{fc}}$ mixes and compresses these spatial features before action selection.
Comparing recovery at the two sites therefore asks whether goal-relevant structure that remains usable in the final spatial representation remains similarly accessible after this learned late transformation.

We use the post-ReLU spatial representation because it is the representation actually presented to the downstream computation.
The native network passes
\[
\operatorname{ReLU}\!\left(H^{(3)}(s)\right)
\]
to the subsequent fully connected layer rather than the pre-activation tensor $H^{(3)}(s)$ itself.
A pre-ReLU intervention would therefore expose variation that the trained network normally removes before downstream action selection and would answer a different question.

The two architectures share this late structure (Appendix~\ref{sec:architecture}), and the two sites are defined in the same way in both.
In the Procgen agents the trunk is the IMPALA network \citep{espeholt2018impala}. Its third block gives $H^{(3)}(s)\in\mathbb R^{32\times8\times8}$, so $h_{\mathrm{pre}}\in\mathbb R^{2048}$, a learned linear layer with ReLU maps $h_{\mathrm{pre}}$ to $h_{\mathrm{fc}}\in\mathbb R^{256}$, and a linear policy head maps $h_{\mathrm{fc}}$ to logits over $|\mathcal A|=15$ actions.
In the robot agents the trunk is the Nature CNN \citep{mnih2015human}. Its third convolution gives $H^{(3)}(s)\in\mathbb R^{64\times12\times12}$ in the arm tasks and $\mathbb R^{64\times8\times8}$ in \quadname, so $h_{\mathrm{pre}}\in\mathbb R^{9216}$ and $\mathbb R^{4096}$, a learned linear layer with ReLU maps $h_{\mathrm{pre}}$ to $h_{\mathrm{fc}}\in\mathbb R^{256}$, and a two-layer policy head with $512$ hidden units maps $h_{\mathrm{fc}}$ to the mean of a Gaussian over the $8$-dimensional arm action or the $3$-dimensional walking command of \quadname.
The robot head also reads the body state, which passes through its own linear layer to $256$ dimensions and is concatenated with $h_{\mathrm{fc}}$. This state layer stays frozen in every rewire, so the rewires differ from the Procgen ones only in the head that is retrained.

\paragraph{Behavior-cloning objective.}
We train the new downstream computation by behavior cloning \citep{pomerleau1991efficient} rather than by continuing policy optimization with PPO \citep{schulman2017ppo}.
This isolates whether the frozen representation can support the desired action mapping from the additional exploration, credit assignment, and representation learning introduced by another round of reinforcement learning.
It also lets us provide the same target behavior directly to GM and Non-GM representations.

For each rewiring arm, we construct
\[
\mathcal D_{\mathrm{BC}}
=
\left\{
(s_i,q_i)
\right\}_{i=1}^{N},
\]
where
\[
q_i(a)\geq 0,
\qquad
\sum_{a\in\mathcal A}q_i(a)=1
\]
defines the target action distribution at state $s_i$.
The frozen representation presented to the new downstream policy is
\[
z_i^{(m)}
=
h^{(m)}(s_i).
\]
We optimize
\[
\phi^\star
=
\arg\min_{\phi}
\frac{1}{N}
\sum_{i=1}^{N}
\mathcal L_{\mathrm{BC}}
\left(
\widetilde{\pi}^{(m)}_{\phi}
(\cdot\mid z_i^{(m)}),
q_i
\right),
\]
with
\[
\mathcal L_{\mathrm{BC}}
\left(
\widetilde{\pi}^{(m)}_{\phi},
q
\right)
=
-
\sum_{a\in\mathcal A}
q(a)
\log
\widetilde{\pi}^{(m)}_{\phi}(a).
\]
The robot actions are continuous. There the new head outputs the action mean $\mu_\phi(z_i^{(m)})\in\mathbb R^{d_{\mathcal A}}$, the target is the teacher's mean action $\mu_{\mathrm T}(s_i)$, and the loss is their squared distance,
\[
\mathcal L_{\mathrm{BC}}
\left(
\mu_\phi(z_i^{(m)}),
\mu_{\mathrm T}(s_i)
\right)
=
\frac{1}{d_{\mathcal A}}
\sum_{j=1}^{d_{\mathcal A}}
\left(
\frac{
\mu_{\phi,j}(z_i^{(m)})-\mu_{\mathrm T,j}(s_i)
}{
\sigma_{\mathrm T,j}
}
\right)^{2},
\]
where $\sigma_{\mathrm T,j}$ is the teacher's state-independent action standard deviation in dimension $j$, so the loss is proportional to the Gaussian negative log-likelihood of the teacher's mean action at the teacher's variance, up to an additive constant.

We use two forms of supervision where available.
Under \emph{teacher supervision}, $q_i$ places all mass on the greedy action of the competent Non-GM checkpoint used to generate the corresponding behavior-cloning bank.
This favours Non-GM rewires at $h_{\mathrm{fc}}$, where the representation already encodes the teacher's own actions. The oracle rows avoid this and show the same ordering.
The robot tasks use teacher supervision only, with no oracle targets. The teacher of each robot rewire is a Non-GM agent other than the one paired with it, so no Non-GM rewire clones its own policy, and its target is the teacher's mean action $\mu_{\mathrm T}(s_i)$.
Under \emph{oracle supervision}, the maze graph defines
\[
A^\star(s_i),
\]
the set of graph-exact first actions on a shortest path toward the intended goal, and
\[
q_i(a)
=
\frac{\mathbf 1\{a\in A^\star(s_i)\}}
{|A^\star(s_i)|}.
\]
In the \textsc{Maze} bank, $|A^\star(s)|=1$ for every state because the generated mazes have unique shortest paths, so this reduces to ordinary hard cross-entropy.
In \textsc{Attribute Maze}, ties can occur and the set-valued target is substantive.
Shortest paths in \textsc{Attribute Maze} never pass through the other candidate object, so an oracle path cannot end the episode at the proxy.
We use teacher targets in all eight environments and oracle targets in \textsc{Maze} and \textsc{Attribute Maze} as an additional variant. Figure~\ref{fig:rewire-main} shows the oracle variant for these two, since its targets follow the intended goal exactly.

The value head is not part of the behavior-cloning objective and does not affect action selection.
Implementations differ slightly in whether its parameters are included in the optimizer, but because no value loss is applied they receive no behavior-cloning gradient.

\paragraph{State distributions and training.}
For teacher supervision, \textsc{Maze}, \textsc{Coin}, and \textsc{Attribute Maze} use balanced behavior-cloning banks with equal contributions from GM and competent Non-GM rollouts.
Within each assay, the same bank is shared across matched GM and Non-GM rewires and representation sites.
The robot banks are described in Appendix~\ref{app:banks-cloning}. In \textsc{AttributeTable} and \textsc{SortZones} the rewires clone only the episodes drawn from the Non-GM training distribution, so no cloned state shows the training co-occurrence of goal and proxy.

Where graph-exact supervision is available, states are additionally labelled with optimal shortest-path actions toward the intended goal.
In \textsc{Maze}, teacher and oracle labels are attached to the same state bank, so the two supervision schemes can be compared on an identical state distribution.
In \textsc{Attribute Maze}, oracle-labelled states must be generated separately because the graph information required for the oracle is not retained in the teacher bank.
We therefore do not interpret teacher--oracle differences in \textsc{Attribute Maze} as isolated effects of the supervision source.

We train all downstream policies for $20$ epochs with Adam \citep{kingma2015adam} at learning rate $3\times10^{-4}$, batch size $256$, gradient clipping at norm $0.5$, and an $80/20$ train--validation split, which in the robot tasks is drawn by episode.
Because the behavior-cloning banks differ in size across assays, the resulting number of optimizer steps also differs.
All GM--Non-GM comparisons are therefore made within a matched assay, where the training budget is identical.
Training and validation cross-entropy together with one-step validation accuracy are recorded as optimization diagnostics rather than measures of behavioral recovery.
For teacher targets, validation accuracy denotes exact action agreement.
For set-valued oracle targets, it denotes whether the predicted action belongs to $A^\star(s)$, so the two accuracy measures are not compared directly.

\paragraph{Closed-loop recovery.}
One-step validation performance does not establish that the rewired policy reaches the intended goal under its own rollout distribution.
We therefore evaluate recovery using the task-level reach quantity introduced above.

We use the behavioral objective of Section~\ref{sec:formal-setup}. $\Gamma^\star(\tau)=1$ when trajectory $\tau$ reaches the intended goal, and the intended-goal success of the rewired policy on the held-out evaluation distribution $\Lambda_{\mathrm{eval}}$ is $J^\star(\widetilde{\pi}^{(m)}_{\phi};\Lambda_{\mathrm{eval}})$.
We estimate it by the goal reach
\[
\hat J^\star\big(\widetilde{\pi}^{(m)}_{\phi}\big)
=
\frac{1}{N}
\sum_{i=1}^{N}
\Gamma^\star(\tau_i),
\qquad
\tau_i\sim P_{\widetilde{\pi}^{(m)}_{\phi}}\big(\tau\mid s_0=s_i\big),
\]
the fraction of $N$ rollouts from held-out initial states $s_1,\ldots,s_N$ that reach the goal, as for the native policies (Appendix~\ref{app:gm-qualification}).
Every rewired policy acts greedily (deterministically in the robot tasks).
For \textsc{Maze}, $\Gamma^\star(\tau)=1$ when the trajectory reaches the cheese.
For \textsc{Coin}, it indicates reaching the goal coin.
For \textsc{Attribute Maze}, it indicates reaching the intended object.
In \textsc{LocationReach} it is the first gripper contact with the ball, within $3$\,cm and before the training corner, in $50$ steps, on the $500$ off-path episodes of each seed pair. In \textsc{PushToMarker} it is the cube resting within $2.5$\,cm of the marker after $100$ steps, on the $512$ conflict scenes. In \textsc{AttributeTable} it is the first touch of the goal object before the proxy, and in \textsc{SortZones} the first entry of the cube into the crossed pad before the plain one, both in $100$ steps on the $512$ world-edit scenes (Appendix~\ref{app:banks}).
In \quadname{} it is the first object reached, flag before tower, within $200$ commands on the first $256$ layouts of the conflict-test bank, with the native start pose (Appendix~\ref{app:banks}).

Where the task has two candidate objects (\textsc{Attribute Maze}, \textsc{AttributeTable}, \textsc{SortZones}, \quadname), intended-goal reach admits an exact
decomposition into whether the policy reaches either candidate and which
candidate it selects conditional on doing so.
Let
\[
D
=
\frac{
N_{\mathrm{goal}}+N_{\mathrm{proxy}}
}{
N
}
\]
denote the decisive fraction, and let
\[
\Pr(\mathrm{goal}\mid\mathrm{dec})
=
\frac{
N_{\mathrm{goal}}
}{
N_{\mathrm{goal}}+N_{\mathrm{proxy}}
}
\]
denote intended-goal preference conditional on reaching either candidate.
Then
\[
\hat J^\star
=
D
\Pr(\mathrm{goal}\mid\mathrm{dec})
\]
holds exactly, as in Equation~\eqref{eq:gm-factorization}.
So the two factors answer different questions.
$D$ measures whether the policy successfully executes a trajectory to either
candidate, whereas $\Pr(\mathrm{goal}\mid\mathrm{dec})$ measures target
selection once the policy becomes decisive.
The decomposition is what separates a rewire that selects the goal from one that only executes better, and we use it in every comparison against the random trunk.

Evaluation is performed on held-out levels disjoint from the
behavior-cloning range, and in the robot tasks on scenes and start poses disjoint from the cloning banks.
We use 300 episodes with a 1000-step cap in \textsc{Maze},
1000 episodes with a 1000-step cap in \textsc{Coin},
and 1000 episodes with a 256-step cap in \textsc{Attribute Maze}.
In \textsc{Attribute Maze} level separation does not imply state separation, since the compact state space puts $98\%$ of evaluation start frames into the oracle cloning bank.

\paragraph{Baselines.}
A successful rewire reestablishes the use of the goal. To decide when this happens, we compare the rewired policies against five complementary baselines.
(1) The untouched native GM and Non-GM checkpoints give the behavior before the intervention and the competence of a policy trained toward the intended goal.
(2) The same rewiring applied to the frozen representation of a Non-GM agent is our primary method control. It shows whether the method can reestablish the goal use that the Non-GM agent originally had.
(3) The random trunk is a randomly initialized network of the same architecture whose trunk up to $h_{\mathrm{pre}}$ is frozen at initialization and never trained. Only the downstream computation, the $h_{\mathrm{pre}}\!\to h_{\mathrm{fc}}$ layer and the policy head, is trained, with the same bank, targets, loss and training as the rewire. We use three such networks, one per GM checkpoint. The random trunk measures whether the trained trunk provides value or whether any trunk lets the readout relearn the task.
(4) A goal-blind chance policy tells what a policy that ignores the goal achieves at the rewire's own competence. In the four two-object tasks it is a coin flip between the two candidates at the decisiveness of the GM rewire from $h_{\mathrm{pre}}$, $0.5\,D$. In \textsc{LocationReach} and \textsc{PushToMarker}, where the goal position varies continuously, it is phantom-goal reach, the rate at which the GM rewire's own paths hit $20$ goal positions drawn from the same distribution as the real one. In \textsc{Coin} it is the untrained head with sampled actions, and in \textsc{Maze} a uniform random walk over the $15$ actions, since no competent goal-blind policy is defined without a second target.
(5) Parameter count is a confounder, since in some environments a certain capacity suffices to relearn the whole structure. A fixed random remix therefore freezes a random \(2048\!\to\!256\) map ($9216\!\to\!256$ in the arm tasks and $4096\!\to\!256$ in \quadname) in place of the learned layer and trains only the downstream head. It is exactly parameter-matched to the \(h_{\mathrm{fc}}\) rewire while removing the learned \(h_{\mathrm{pre}}\!\to h_{\mathrm{fc}}\) transformation.

\paragraph{Parameter counts.}
The \(h_{\mathrm{fc}}\) rewire trains \(257|\mathcal A|\) policy parameters, whereas the \(h_{\mathrm{pre}}\) rewire additionally learns the \(2048\!\to\!256\) transformation, adding \(524{,}544\) trainable parameters. With \(|\mathcal A|=15\), this gives \(3{,}855\) versus \(528{,}399\) policy parameters. In \textsc{Maze} and \textsc{Attribute Maze}, the reinitialized value head is also passed to the optimizer, adding \(257\) parameters to the recorded counts (\(4{,}112\) and \(528{,}656\), respectively). Because the behavior-cloning loss depends only on policy logits, these parameters receive no gradient and cannot affect action selection. The fixed-remix control has exactly the same recorded trainable count as the corresponding \(h_{\mathrm{fc}}\) rewire in every environment.
In the robot agents the $h_{\mathrm{fc}}$ rewire trains the two-layer head, $266{,}760$ parameters in the arm tasks and $264{,}195$ in \quadname, and the $h_{\mathrm{pre}}$ rewire adds the $9216\!\to\!256$ or $4096\!\to\!256$ transformation ($2{,}359{,}552$ and $1{,}048{,}832$ parameters), for $2{,}626{,}312$ and $1{,}313{,}027$. No value parameters are passed to the optimizer, and the fixed remix again matches the $h_{\mathrm{fc}}$ rewire.

\paragraph{Success criterion.}
We need to define when a rewire recovers the goal, recovers it partly or does not recover it. We compare reaches at the same site and with the same targets. The reference is the lower of the native Non-GM reach (1) and the Non-GM rewire reach (2), since a new rewire can overshoot the native policy. A GM rewire is \emph{recovered} when it reaches at least $90\%$ of this reference. It is \emph{partial} when it stays below that but its excess over the floor, the higher of the random trunk (3) and the goal-blind policy (4), has a $95\%$ interval above zero, and \emph{none} otherwise. The intervals are Welch intervals across the three GM checkpoints and the three random trunks or chance estimates.
The Non-GM rewire is needed to rule out a failed method only when recovery fails. We therefore count the method as working when the Non-GM rewire reaches at least $90\%$ of its native policy or the GM rewire is recovered. If recovery works, the method evidently works, so the control is informative only when the experiment fails. Where the method does not work, a GM rewire that is not recovered must be read with care.
\label{sec:rewiring-floor-above-reference}
When a random-trunk rewire reaches the native Non-GM reach, a readout trained on a frozen trunk has enough capacity to solve the environment from any trunk. Goal information in the GM or the Non-GM trunk is then not necessary for recovery. This does not rule out that a recovery from these trunks uses the goal, but it makes it much harder to tell where the recovery comes from, as the readout can sidestep the trunk by relearning the whole environment, with goal direction, in its trained layers. A readout with less capacity separates the trunks again. In that case we therefore also use the strict test of the additional ablations below, a linear readout cloned on a bank without the evaluation start configurations and on $10\%$ and $1\%$ of it, and ask whether the GM trunk exceeds the random trunk there.

\paragraph{Additional ablations.}
In \textsc{Maze} and \textsc{PushToMarker} offline behavior cloning leaves the Non-GM rewire below $90\%$ of its native policy, and cloning can fail through distribution shift \citep{ross2011dagger}. Both therefore receive three rounds of DAgger at both sites.
In \textsc{Maze}, each round rolls out the current rewired policy on training levels, labels $50$k visited states with oracle actions, adds them to the training set and fine-tunes the downstream computation. A matched control adds the same number of frames resampled from the base set, so the two arms take identical optimizer steps and differ only in whether the added states are on-policy.
In \textsc{PushToMarker}, with teacher targets, each round rolls out the current rewired policy from the training half of the start poses on $288$ fresh scenes of the Non-GM training distribution and labels all $28{,}800$ visited states with the teacher's actions. That is $25\%$ of the $115{,}200$-frame cloning bank, the same share as $50$k states of the $200$k \textsc{Maze} bank. The matched control again adds as many frames resampled from the base set, and the random trunk receives both treatments too.
We judge these arms, like all others, by closed-loop reach rather than by cloning accuracy, since cloning accuracy only measures whether the rewire replicates its targets, not whether it pursues the goal.

A rewire can raise goal reach without using the goal, by sending the policy to a fixed position or to the proxy. As in the location controls of steering (Table~\ref{tab:steering-location}), we therefore keep the GM trunk and take the goal away (Section~\ref{sec:rewiring-goal-directed}). We clone a rewire on frames in which the goal cue is masked, and we evaluate the standard rewire with the goal removed, by deleting the cheese (\textsc{Maze}), hiding the coin (\textsc{Coin}), drawing the goal line as a red gem (\textsc{Attribute Maze}), removing the ball, marker or flag (\textsc{LocationReach}, \textsc{PushToMarker}, \quadname) or giving the goal the proxy's cue (two spheres in \textsc{AttributeTable}, no cross on either pad in \textsc{SortZones}). We then count arrivals at the former goal location. A rewire that pursues the goal drops to chance or below, while one that goes to a fixed position or to the proxy keeps arriving there.
A few further tests serve single environments. In \textsc{Maze} a random trunk freezes under greedy actions, so all \textsc{Maze} arms are also evaluated with sampled actions. In \textsc{PushToMarker} the marker of a conflict scene lies on the path to the table centre, so every arm is also evaluated on the $512$ scenes of the control-layout set, in which marker and cube are placed independently and a push towards the centre cannot score (Appendix~\ref{app:banks}).
In \textsc{Coin} we split every arm's episodes by outcome, coin, level end, death or step cap, to separate choosing the coin from surviving the platforming.
In \textsc{Attribute Maze} a strict test clones on a bank from which every evaluation start configuration is removed, and on $10\%$ and $1\%$ of it by level, with the relearned head and with a linear readout from $h_{\mathrm{pre}}$ or $h_{\mathrm{fc}}$, as a weaker readout has less room to compensate for the features it reads. All \textsc{Attribute Maze} arms are repeated with the textured-background checkpoints, on the textured version of the banks, to test whether the ordering of the two sites survives a change of background.

\paragraph{What a rewire can show.}
A successful rewire shows that the frozen representation suffices for goal-directed control under the downstream function class, supervision and state distribution that we test. A failed rewire is weaker evidence, since it can also reflect limits of the downstream architecture, supervision, optimization or state distribution, which is why we read it together with the Non-GM rewire and DAgger. A successful rewire in turn says something about the agent only if it exceeds what a new readout extracts from an untrained trunk, which already passes the position and appearance of a visible goal object.

\subsubsection{Does new wiring recover the goal?}

\begin{table}[t]
\centering
\footnotesize
\setlength{\tabcolsep}{5pt}
\caption{\textbf{Rewiring at a glance.}
The method works when the Non-GM rewire reaches at least $90\%$ of its native policy (its fraction in brackets) or the GM rewire is recovered, since a recovery shows that the method works. Recovery is that of the GM rewire under the success criterion, with teacher targets, and is in bold where the method works. Oracle targets in \textsc{Maze} and \textsc{Attribute Maze} give the same verdicts. The DAgger rows use three rounds of on-policy relabeling. Values in Tables~\ref{tab:rewire-main} and~\ref{tab:maze-dagger}.}
\label{tab:rewire-summary}
\begin{tabular}{l cc cc}
\toprule
& \multicolumn{2}{c}{Method works?} & \multicolumn{2}{c}{Recovery successful?} \\
\cmidrule(lr){2-3}\cmidrule(lr){4-5}
Environment & $h_{\mathrm{pre}}$ & $h_{\mathrm{fc}}$ & $h_{\mathrm{pre}}$ & $h_{\mathrm{fc}}$ \\
\midrule
\multicolumn{5}{l}{\textit{(i) Goal lost in the representation}} \\
\textsc{Maze} & no ($0.80$) & no ($0.64$) & none & none \\
\quad with DAgger & \textbf{yes} ($1.01$) & no ($0.78$) & none & none \\
\textsc{LocationReach} & \textbf{yes} ($0.92$) & \textbf{yes} ($0.97$) & none & none \\
\midrule
\multicolumn{5}{l}{\textit{(ii) Goal reaches the policy but is not used}} \\
\textsc{PushToMarker} & no ($0.79$) & no ($0.77$) & partial & partial \\
\quad with DAgger & \textbf{yes} ($0.92$) & \textbf{yes} ($0.92$) & \textbf{partial} & \textbf{partial} \\
\textsc{Coin} & \textbf{yes} ($0.99$) & \textbf{yes} ($0.98$) & \textbf{recovered} & none \\
\midrule
\multicolumn{5}{l}{\textit{(iii) Goal and proxy compete}} \\
\textsc{Attribute Maze} & \textbf{yes} ($0.98$) & \textbf{yes} ($0.94$) & \textbf{recovered} & none \\
\textsc{AttributeTable} & \textbf{yes} ($0.91$) & no ($0.87$) & \textbf{partial} & none \\
\textsc{SortZones} & \textbf{yes} ($0.87$) & no ($0.71$) & \textbf{recovered} & none \\
\quadname & \textbf{yes} ($1.00$) & \textbf{yes} ($0.94$) & \textbf{recovered} & \textbf{partial} \\
\bottomrule
\end{tabular}
\end{table}

\begin{table}[t]
\centering
\scriptsize
\setlength{\tabcolsep}{3pt}
\caption{\textbf{Closed-loop goal reach after rewiring, ordered by where the goal is lost.}
Goal reach $\hat J^\star$, mean $\pm$ SD across three checkpoints or random trunks, with teacher targets unless marked oracle or DAgger. Native is the untouched policy, remix the fixed random remix (5), and the random trunk (3) is rewired from the same site as the row. Chance is the goal-blind policy (4). Bold GM rewires are recovered and underlined ones partial under the success criterion.
$^a$Uniform random walk, sampled actions.
$^b$Freezes under greedy actions.
$^c$Does not fit the cloning targets.
$^d$Two of three checkpoints.
$^e$Cloned on a rebuilt copy of the bank, on which the rewires reproduce within $0.02$.
$^f$$0.5\,D$ at $D=0.690$, one estimate.
$^g$Non-GM rewire below $90\%$ of its native policy.
$^h$Three rounds of DAgger (Table~\ref{tab:maze-dagger}), oracle targets in \textsc{Maze}.}
\label{tab:rewire-main}
\label{tab:rewire-remix}
\begin{tabular}{ll cc cc cc}
\toprule
& & \multicolumn{2}{c}{Native} & \multicolumn{2}{c}{Rewire} & \multicolumn{2}{c}{Baselines} \\
\cmidrule(lr){3-4}\cmidrule(lr){5-6}\cmidrule(lr){7-8}
Environment & Site & GM & Non-GM & GM & Non-GM & Random trunk & Chance \\
\midrule
\multicolumn{8}{l}{\textit{(i) Goal lost in the representation}} \\
\textsc{Maze} & $h_{\mathrm{pre}}$ & $0.219 \pm 0.02$ & $0.347 \pm 0.02$ & $0.089 \pm 0.01$ & $0.277 \pm 0.02^g$ & $0.016 \pm 0.01^{b,e}$ & $0.226^a$ \\
 & remix & & & $0.038 \pm 0.02$ & $0.167 \pm 0.03$ & & \\
 & $h_{\mathrm{fc}}$ & & & $0.048 \pm 0.00$ & $0.222 \pm 0.07^g$ & $0.008 \pm 0.00^{b,e}$ & \\
\quad oracle & $h_{\mathrm{pre}}$ & & & $0.116 \pm 0.02$ & $0.308 \pm 0.04^g$ & $0.018 \pm 0.00^b$ & $0.226^a$ \\
 & remix & & & $0.092 \pm 0.02$ & $0.173 \pm 0.01$ & & \\
 & $h_{\mathrm{fc}}$ & & & $0.119 \pm 0.02$ & $0.236 \pm 0.03^g$ & $0.009 \pm 0.00^{b,e}$ & \\
\quad DAgger$^h$ & $h_{\mathrm{pre}}$ & & & $0.128 \pm 0.02$ & $0.351 \pm 0.03$ & & $0.226^a$ \\
 & $h_{\mathrm{fc}}$ & & & $0.122 \pm 0.02$ & $0.271 \pm 0.03^g$ & & \\
\addlinespace
\textsc{LocationReach} & $h_{\mathrm{pre}}$ & $0.000 \pm 0.00$ & $0.999 \pm 0.00$ & $0.373 \pm 0.08$ & $0.921 \pm 0.05$ & $0.179 \pm 0.09$ & $0.096 \pm 0.03$ \\
 & remix & & & $0.117 \pm 0.04$ & $0.468 \pm 0.17$ & & \\
 & $h_{\mathrm{fc}}$ & & & $0.137 \pm 0.09$ & $0.973 \pm 0.02$ & $0.146 \pm 0.05$ & \\
\midrule
\multicolumn{8}{l}{\textit{(ii) Goal reaches the policy but is not used}} \\
\textsc{PushToMarker} & $h_{\mathrm{pre}}$ & $0.000 \pm 0.00$ & $0.889 \pm 0.04$ & \underline{$0.314 \pm 0.07$} & $0.702 \pm 0.10^g$ & $0.014 \pm 0.00^c$ & $0.000 \pm 0.00$ \\
 & remix & & & $0.068 \pm 0.03$ & $0.154 \pm 0.07$ & & \\
 & $h_{\mathrm{fc}}$ & & & \underline{$0.132 \pm 0.00$} & $0.682 \pm 0.15^g$ & $0.004 \pm 0.00^c$ & \\
\quad DAgger$^h$ & $h_{\mathrm{pre}}$ & & & \underline{$0.661 \pm 0.16$} & $0.816 \pm 0.08$ & $0.135 \pm 0.07$ & \\
 & $h_{\mathrm{fc}}$ & & & \underline{$0.326 \pm 0.09$} & $0.819 \pm 0.08$ & $0.009 \pm 0.01$ & \\
\addlinespace
\textsc{Coin} & $h_{\mathrm{pre}}$ & $0.316 \pm 0.01$ & $0.896 \pm 0.00$ & {\boldmath$0.843 \pm 0.01$} & $0.888 \pm 0.01$ & $0.646 \pm 0.02$ & $0.625 \pm 0.00^d$ \\
 & remix & & & $0.676 \pm 0.03$ & $0.825 \pm 0.03$ & & \\
 & $h_{\mathrm{fc}}$ & & & $0.666 \pm 0.07$ & $0.881 \pm 0.02$ & $0.511 \pm 0.00^e$ & \\
\midrule
\multicolumn{8}{l}{\textit{(iii) Goal and proxy compete}} \\
\textsc{Attribute Maze} & $h_{\mathrm{pre}}$ & $0.358 \pm 0.02$ & $0.696 \pm 0.02$ & {\boldmath$0.689 \pm 0.03$} & $0.684 \pm 0.02$ & $0.631 \pm 0.03^e$ & $0.345^f$ \\
 & remix & & & $0.262 \pm 0.02$ & $0.440 \pm 0.05$ & & \\
 & $h_{\mathrm{fc}}$ & & & $0.273 \pm 0.02$ & $0.653 \pm 0.04$ & $0.224 \pm 0.04^e$ & \\
\quad oracle & $h_{\mathrm{pre}}$ & & & {\boldmath$0.871 \pm 0.02$} & $0.908 \pm 0.01$ & $0.840 \pm 0.05$ & $0.450 \pm 0.01$ \\
 & remix & & & $0.481 \pm 0.07$ & $0.626 \pm 0.01$ & & \\
 & $h_{\mathrm{fc}}$ & & & $0.478 \pm 0.06$ & $0.793 \pm 0.01$ & $0.296 \pm 0.03^e$ & \\
\addlinespace
\textsc{AttributeTable} & $h_{\mathrm{pre}}$ & $0.037 \pm 0.02$ & $0.986 \pm 0.01$ & \underline{$0.757 \pm 0.03$} & $0.898 \pm 0.04$ & $0.462 \pm 0.12$ & $0.411 \pm 0.02$ \\
 & remix & & & $0.361 \pm 0.10$ & $0.391 \pm 0.16$ & & \\
 & $h_{\mathrm{fc}}$ & & & $0.545 \pm 0.12$ & $0.854 \pm 0.06^g$ & $0.173 \pm 0.03$ & \\
\addlinespace
\textsc{SortZones} & $h_{\mathrm{pre}}$ & $0.000 \pm 0.00$ & $0.946 \pm 0.01$ & {\boldmath$0.753 \pm 0.02$} & $0.821 \pm 0.01^g$ & $0.151 \pm 0.04$ & $0.377 \pm 0.01$ \\
 & remix & & & $0.270 \pm 0.07$ & $0.387 \pm 0.06$ & & \\
 & $h_{\mathrm{fc}}$ & & & $0.348 \pm 0.07$ & $0.669 \pm 0.12^g$ & $0.036 \pm 0.05$ & \\
\addlinespace
\quadname & $h_{\mathrm{pre}}$ & $0.270 \pm 0.01$ & $0.964 \pm 0.01$ & {\boldmath$0.892 \pm 0.05$} & $0.965 \pm 0.01$ & $0.122 \pm 0.01$ & $0.464 \pm 0.01$ \\
 & remix & & & $0.462 \pm 0.12$ & $0.888 \pm 0.03$ & & \\
 & $h_{\mathrm{fc}}$ & & & \underline{$0.697 \pm 0.03$} & $0.905 \pm 0.12$ & $0.049 \pm 0.01$ & \\
\bottomrule
\end{tabular}
\end{table}

\begin{figure}[t]
    \centering
    \includegraphics[width=\linewidth]{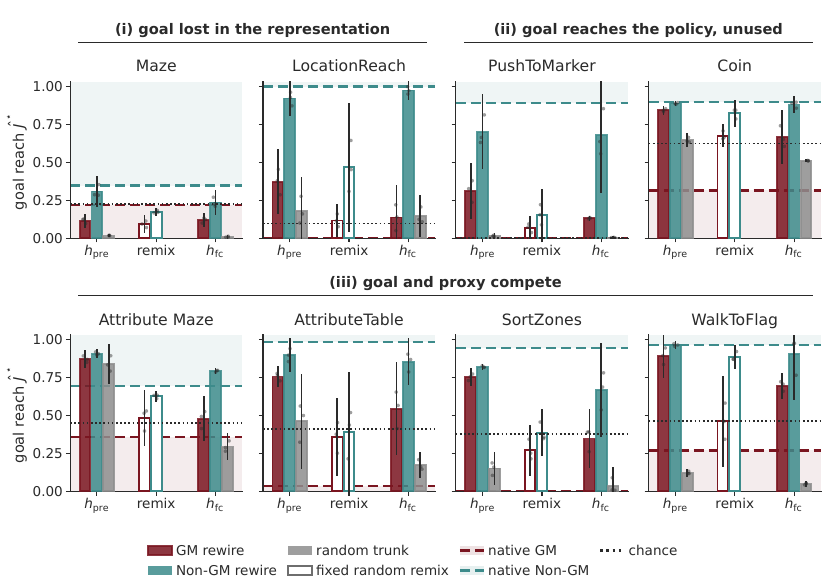}
    \caption{\textbf{Closed-loop goal reach after rewiring, ordered by where the goal is lost.}
    Goal reach $\hat J^\star$ of the GM (red) and Non-GM (teal) rewires from $h_{\mathrm{pre}}$ and $h_{\mathrm{fc}}$, of the fixed random remix (open bars) and of the random trunk (grey), with oracle targets in \textsc{Maze} and \textsc{Attribute Maze} and teacher targets in the other six environments (Table~\ref{tab:rewire-main}).
    Bars are means, whiskers $95\%$ $t$-intervals across the three checkpoints or random trunks, and dots individual checkpoints where they are available.
    Dashed lines mark the native GM policy, with the band below it, and the native Non-GM policy, with the band above it. The dotted line is chance.
    The random trunks of \textsc{Maze}, \textsc{Attribute Maze} and \textsc{Coin} at $h_{\mathrm{fc}}$ are cloned on the rebuilt bank (Table~\ref{tab:rewire-main}, note $e$).}
    \label{fig:rewire-main}
\end{figure}

From $h_{\mathrm{pre}}$, the GM rewire recovers the goal in \textsc{Coin}, \textsc{Attribute Maze}, \textsc{SortZones} and \quadname, partly in \textsc{PushToMarker} and \textsc{AttributeTable}, and not in \textsc{Maze} or \textsc{LocationReach} (Tables~\ref{tab:rewire-summary} and~\ref{tab:rewire-main}, Figure~\ref{fig:rewire-main}). Recovery thus lines up with depth. Where goal and proxy compete or the goal reaches the policy without being used, it succeeds fully or partly, and where the goal is lost in the representation it does not. From the later layer $h_{\mathrm{fc}}$, it recovers the goal nowhere and only partly in \textsc{PushToMarker} and \quadname.
The drop at $h_{\mathrm{fc}}$ follows from what this layer is trained for. It compresses thousands of spatial features into $256$ and keeps what the policy acts on, the goal in a Non-GM agent and the proxy in a GM agent, so in a GM agent nothing in training carries the goal through (Section~\ref{sec:rewiring-layers}).

\begin{figure}[t]
    \centering
    \includegraphics[width=\linewidth]{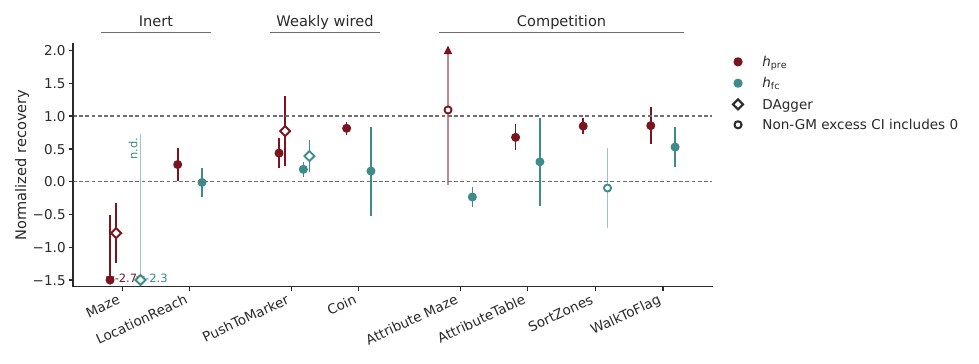}
    \caption{\textbf{Normalized rewire recovery.} $(\text{GM}-\text{floor})/(\text{Non-GM}-\text{floor})$ for the rewires from $h_{\mathrm{pre}}$ (red) and $h_{\mathrm{fc}}$ (teal), with the floor the higher of random trunk and chance, for teacher-target behavior cloning (circles) and three rounds of DAgger (diamonds, oracle targets in \textsc{Maze}, teacher targets in \textsc{PushToMarker}), with $95\%$ delta-method intervals across three checkpoints per arm. Open markers flag cells where the Non-GM rewire does not clearly exceed the floor, so the ratio is unstable, n.d.\ marks a Non-GM rewire at or below the floor, and values beyond the axis are printed at its edge.}
    \label{fig:rewire-normalized}
\end{figure}

In \textsc{Maze}, \textsc{PushToMarker} and \textsc{SortZones}, offline behavior cloning leaves the Non-GM rewire below $90\%$ of its native policy ($0.80$, $0.79$ and $0.87$ from $h_{\mathrm{pre}}$). In \textsc{SortZones} the GM rewire is recovered, so the method works there by the success criterion. In \textsc{Maze} and \textsc{PushToMarker} it is not, so there the method itself has to be questioned. Three rounds of DAgger give a working method in \textsc{PushToMarker} at both sites and in \textsc{Maze} from $h_{\mathrm{pre}}$ (Table~\ref{tab:maze-dagger}). \textsc{Maze} still does not recover, while \textsc{PushToMarker} recovers partly, with the GM rewire from $h_{\mathrm{pre}}$ at $0.81$ of the Non-GM rewire. \textsc{SortZones} needs no DAgger, since its method already works. Figure~\ref{fig:rewire-normalized} puts every GM rewire on one scale, its reach above the floor as a fraction of the Non-GM rewire's.

In some environments the random trunk comes close to the GM rewire, most clearly in \textsc{Attribute Maze}, where a readout with enough capacity relearns the simple structure of the environment from any trunk. Behavior cloning hands the readout the goal through its targets, and a visible goal object is passed on by random convolutional features too (grey bars). GM agents keep an advantage ($0.684$ against $0.631$ on the same bank), and with less capacity, a linear readout, the two separate.

In the following we go through each environment in depth.

\begin{table}[t]
\centering
\footnotesize
\caption{\textbf{Rewiring with DAgger.} Goal reach after three rounds of on-policy relabeling (DAgger) or of matched extra training, with oracle targets in \textsc{Maze} ($300$ held-out levels) and teacher targets in \textsc{PushToMarker} ($512$ conflict scenes). Mean $\pm$ SD across three checkpoints or random trunks. Round 0 is the rewire before the added rounds. In \textsc{Maze} it is trained on states visited by the freshly initialized downstream policy, so it differs from Table~\ref{tab:rewire-main}, and both arms then share a base set along the native agent's trajectories. Underlined GM cells are partial under the success criterion, and none is recovered.}
\label{tab:maze-dagger}
\label{tab:rewire-dagger}
\begin{tabular}{llcccc}
\toprule
Site & Cohort & Native & Round 0 & DAgger & Matched \\
\midrule
\multicolumn{6}{l}{\textit{\textsc{Maze}, oracle targets}} \\
$h_{\mathrm{pre}}$ & GM & $0.219 \pm 0.02$ & $0.112 \pm 0.02$ & $0.128 \pm 0.02$ & $0.118 \pm 0.01$ \\
 & Non-GM & $0.347 \pm 0.02$ & $0.187 \pm 0.03$ & $0.351 \pm 0.03$ & $0.357 \pm 0.03$ \\
\addlinespace
$h_{\mathrm{fc}}$ & GM & $0.219 \pm 0.02$ & $0.114 \pm 0.01$ & $0.122 \pm 0.02$ & $0.098 \pm 0.01$ \\
 & Non-GM & $0.347 \pm 0.02$ & $0.246 \pm 0.02$ & $0.271 \pm 0.03$ & $0.111 \pm 0.03$ \\
\midrule
\multicolumn{6}{l}{\textit{\textsc{PushToMarker}, teacher targets}} \\
$h_{\mathrm{pre}}$ & GM & $0.000 \pm 0.00$ & \underline{$0.296 \pm 0.06$} & \underline{$0.661 \pm 0.16$} & \underline{$0.311 \pm 0.08$} \\
 & Non-GM & $0.889 \pm 0.04$ & $0.686 \pm 0.09$ & $0.816 \pm 0.08$ & $0.715 \pm 0.08$ \\
 & Random trunk & & $0.007 \pm 0.00$ & $0.135 \pm 0.07$ & $0.055 \pm 0.04$ \\
\addlinespace
$h_{\mathrm{fc}}$ & GM & $0.000 \pm 0.00$ & \underline{$0.135 \pm 0.02$} & \underline{$0.326 \pm 0.09$} & \underline{$0.126 \pm 0.01$} \\
 & Non-GM & $0.889 \pm 0.04$ & $0.677 \pm 0.11$ & $0.819 \pm 0.08$ & $0.726 \pm 0.09$ \\
 & Random trunk & & $0.006 \pm 0.01$ & $0.009 \pm 0.01$ & $0.005 \pm 0.00$ \\
\bottomrule
\end{tabular}
\end{table}

\emph{\textsc{Maze}: no recovery.}
In \textsc{Maze} the GM rewire fails. With teacher targets it reaches the cheese in $0.089$ of levels from $h_{\mathrm{pre}}$ and $0.048$ from $h_{\mathrm{fc}}$, below its own native policy ($0.219$) and a uniform random walk ($0.226$), while the Non-GM rewire reaches $0.277$ and $0.222$, against its native $0.347$ (Table~\ref{tab:rewire-main}, first rows).
With oracle targets the GM rewire does not recover the goal either ($0.116$ at $h_{\mathrm{pre}}$ and $0.119$ at $h_{\mathrm{fc}}$, against $0.308$ and $0.236$ for the Non-GM rewires, Table~\ref{tab:rewire-main}). Sampled instead of greedy actions bring the GM oracle rewire from $h_{\mathrm{pre}}$ only to the level of the random walk ($0.235$), against $0.582$ for Non-GM.
The method itself underperforms here, since the Non-GM rewire stays below $90\%$ of its native performance, closest with oracle targets at $h_{\mathrm{pre}}$ ($0.89$) and furthest at $h_{\mathrm{fc}}$.
To rule out the method, three rounds of DAgger on states along the agents' own trajectories bring the Non-GM rewire from $h_{\mathrm{pre}}$ back to its native performance ($0.351$ against $0.347$, Table~\ref{tab:maze-dagger}), which makes the method viable there. The matched extra training alone does the same ($0.357$). From $h_{\mathrm{fc}}$ the Non-GM rewire still falls short ($0.271$). The GM rewires stay below even the native GM reach ($0.128$ and $0.122$ against $0.219$), so the goal information in the \textsc{Maze} agent cannot be turned into goal use by a new wiring.
Cloning accuracy cannot show this failure, since in $95\%$ of \textsc{Maze} states the step toward the cheese is also the step toward the top-right corner, and only the few forks where the two differ decide whether a rollout reaches the cheese.

\emph{\textsc{LocationReach}: no more than random features.}
The Non-GM rewire nearly matches the native policy, but the GM rewire reaches the ball in only $0.373$ of episodes from $h_{\mathrm{pre}}$ and $0.137$ from $h_{\mathrm{fc}}$. That is about the random trunk at $h_{\mathrm{fc}}$, and above it at $h_{\mathrm{pre}}$ by a difference whose interval includes zero, because a random convolution passes the position of a single salient ball almost as well (Appendix~\ref{app:probes}).

\emph{\textsc{PushToMarker}: partial recovery.}
\textsc{PushToMarker} differs from \textsc{LocationReach}. The GM rewire from $h_{\mathrm{pre}}$ pushes the cube onto the marker in $0.314$ of conflict scenes, far above the random trunk ($0.014$), and as often in the control-layout scenes, where a push toward the table centre cannot score (Figure~\ref{fig:rewire-main}, PushToMarker). So the new readout uses goal information that only the GM trunk provides.
It reaches about $45\%$ of the Non-GM rewire ($0.702$), a partial recovery, consistent with a goal that is weakly wired into the GM policy. The Non-GM rewire itself reaches only $0.79$ of its native policy, so this comparison must be read with care.
Behavior cloning fits the rewire only on the states of the bank, and on the GM rewire's own trajectories its errors compound. Three rounds of on-policy relabeling (Table~\ref{tab:maze-dagger}) lift the GM rewire from $h_{\mathrm{pre}}$ from $0.296$ to $0.661$, while the matched control stays at $0.311$ and the random trunk gains only to $0.135$. So most of the gain uses goal information in the GM $h_{\mathrm{pre}}$ that cloning could not exploit. On the control-layout set the GM rewire rises from $0.333$ to $0.793$, so the gain is not a push toward the table centre.
The Non-GM rewire reaches $0.816$ under DAgger, $0.92$ of its native policy, so the method no longer underperforms, and the GM rewire reaches $0.81$ of it, still a partial recovery. Normalized, it moves from $0.44$ under cloning to $0.77$ under DAgger, with an interval from $0.25$ to $1.29$, against $0.39$ for the matched control (Figure~\ref{fig:rewire-normalized}).
The three GM checkpoints spread widely, $0.848$, $0.592$ and $0.543$, and only the first reaches the mean Non-GM rewire.
At $h_{\mathrm{fc}}$ the gain is small, from $0.135$ to $0.326$ against $0.819$ for the Non-GM rewire, again partial.
The goal-removed controls of Table~\ref{tab:rewire-goal-directed} were not repeated after DAgger.

\emph{\textsc{Coin} and \textsc{Attribute Maze}: recovered.}
In \textsc{Coin} the GM rewire from $h_{\mathrm{pre}}$ reaches $0.843$, $0.95$ of the Non-GM rewire ($0.888$). The random trunk reaches $0.646$, and its gap to the GM rewire is mostly survival, not coin choice. Among episodes that end at the coin or at the level end, the random trunk takes the coin almost as often as the GM rewire ($0.962$ against $0.987$), and with the GM rewire's survival it would reach $0.822$. The coin is a visible sprite that a head on random features already steers toward, and the GM trunk adds the platforming that keeps the agent alive.
In \textsc{Attribute Maze} the GM rewire matches the Non-GM rewire with teacher targets ($0.689$ against $0.684$), and with oracle targets it reaches $0.871$, above the native Non-GM policy ($0.696$, Table~\ref{tab:rewire-main}). On the rebuilt bank the random trunk reaches $0.631$ against $0.684$ for the GM rewire, and both go to the goal in $0.999$ of the episodes in which they reach an object.
The $5\times5$ maze is small enough for a readout with $528{,}656$ parameters to relearn the task from random features. With oracle targets the random trunk ($0.840$) lies above the native Non-GM reach ($0.696$), so at full capacity the rewire cannot tell where the recovery comes from (Section~\ref{sec:rewiring-floor-above-reference}). The strict test with a linear readout separates the trunks. Cloned on the bank without the evaluation start configurations, the GM trunk exceeds the random trunk by $0.136$ from $h_{\mathrm{pre}}$ ($0.797$ against $0.661$, $95\%$ Welch interval $0.05$ to $0.22$) and by $0.185$ from $h_{\mathrm{fc}}$ ($0.487$ against $0.303$, $0.06$ to $0.31$), and it stays above it with $10\%$ and $1\%$ of that bank. So the GM trunk carries goal-relevant structure that a random trunk lacks. Most of this advantage is execution, reaching an object at all. Once decisive, the GM rewire picks the goal only slightly more often than the random trunk from $h_{\mathrm{pre}}$ ($0.96$ against $0.92$).

\emph{\quadname{} and \textsc{SortZones} recovered, \textsc{AttributeTable} partial.}
The GM rewire from $h_{\mathrm{pre}}$ reaches $0.93$ of the reference in \quadname, $0.92$ in \textsc{SortZones} and $0.84$ in \textsc{AttributeTable}, and stays well above the random trunk (Figure~\ref{fig:rewire-main}, bottom row). In \textsc{SortZones} the Non-GM rewire reaches only $0.87$ of its native policy, but the GM rewire is recovered, so the method works (Table~\ref{tab:rewire-summary}). What the rewire recovers differs.
In \quadname{} the rewire changes which object the agent selects. It reaches the flag first in $0.892$ of the conflict layouts, against $0.464$ for chance and $0.965$ for the Non-GM rewire, while the native GM policy is just as decisive but picks the flag in only $0.272$ of decisive episodes. The random trunk cannot walk, so it cannot show whether random features would also select the flag.
In \textsc{SortZones} and \textsc{AttributeTable} the rewire mainly improves execution. In \textsc{SortZones} every trunk, the random one included, brings the cube to the crossed pad whenever it brings it to a pad at all, so the gain over the random trunk is better pushing. In \textsc{AttributeTable} the GM rewire reaches an object far more often than the random trunk ($D=0.822$ against $0.530$) but prefers the goal only slightly more often once it does ($0.922$ against $0.867$).

\subsubsection{From which layer can the goal be recovered?}
\label{sec:rewiring-layers}

\begin{figure}[t]
    \centering
    \includegraphics[width=\linewidth]{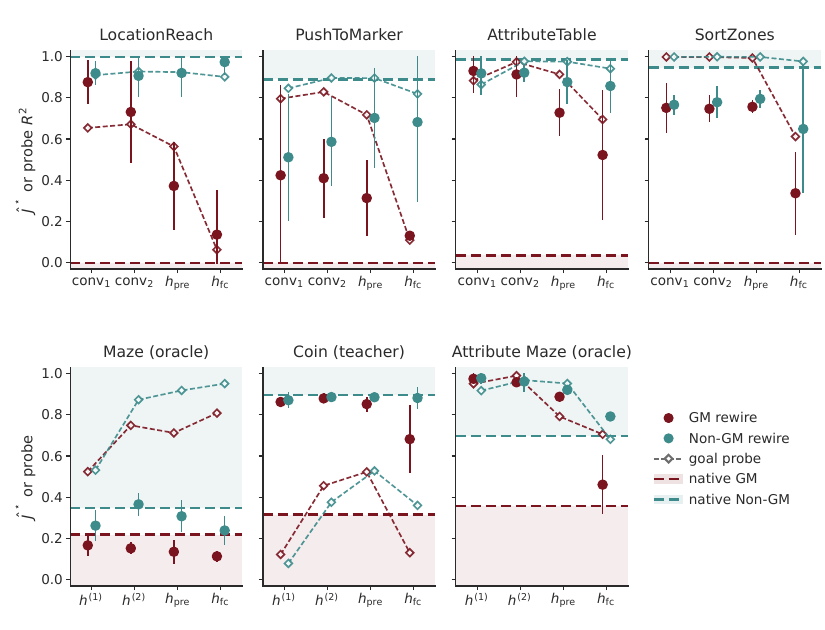}
    \caption{\textbf{Rewiring from every layer.}
    Goal reach $\hat J^\star$ of the GM (red) and Non-GM (teal) rewires from each layer of the frozen trunk, means with $95\%$ $t$-intervals across three checkpoints, with the native policies as dashed lines and the goal probe of Appendix~\ref{app:probes} at the same layer as open diamonds (goal-position $R^2$ in the arm tasks, cheese-bearing AUC, coin-bearing $R^2$ and line-presence AUC in Procgen).}
    \label{fig:rewire-layers}
\end{figure}

\emph{The goal is lost in the last learned layer.}
In every environment where the GM rewire from $h_{\mathrm{pre}}$ is recovered or partial, the rewire from $h_{\mathrm{fc}}$ recovers clearly less, in \textsc{PushToMarker}, \textsc{Coin}, \textsc{Attribute Maze}, \textsc{AttributeTable}, \textsc{SortZones} and \quadname{} (Table~\ref{tab:rewire-main}), while the Non-GM rewires lose much less. No GM rewire from $h_{\mathrm{fc}}$ is recovered, and only those of \textsc{PushToMarker} and \quadname{} are partial.
So in GM agents the goal survives the convolutional trunk and is lost in the map from $h_{\mathrm{pre}}$ to $h_{\mathrm{fc}}$. This layer compresses thousands of spatial features into $256$ and is trained only to keep what the policy acts on. For a Non-GM agent that is the goal. For a GM agent, whose policy acts on the proxy, nothing in training carries the goal through.
On the textured background of \textsc{Attribute Maze}, with oracle targets, the ordering of the two sites survives in both cohorts, although there the Non-GM rewire loses almost as much ($0.786$ to $0.530$, against $0.753$ to $0.419$ for GM).

\emph{The learned layer keeps what the policy uses.}
The fixed random remix separates this from compression alone (Figure~\ref{fig:rewire-main}, open bars). It replaces the learned layer by a random map of the same size, so it keeps what any compression keeps.
For Non-GM agents the learned $h_{\mathrm{fc}}$ beats the remix in every environment, as expected if training shaped it to keep the goal. A random compression alone already keeps most of the Non-GM goal reach in \quadname{} and \textsc{Coin}.
For GM agents the learned $h_{\mathrm{fc}}$ is no better than the remix in \textsc{Coin} and \textsc{Attribute Maze} and only slightly better in \textsc{LocationReach}, \textsc{PushToMarker} and \textsc{SortZones}, so there their fc layer keeps no more of the goal than a random one would. Only in \textsc{AttributeTable} and \quadname{} does it keep clearly more.
The robot probes show the same pattern (Table~\ref{tab:robot-probes}). A random fc layer already loses much of the goal position, so the compression itself is not specific to GM training. What training changes is which cue survives it. Non-GM agents keep the goal and drop the proxy, GM agents keep the proxy and let the goal go.
In \textsc{Attribute Maze} and \textsc{Maze} the probes do not follow the rewires. In \textsc{Attribute Maze} a linear readout separates GM from the random trunk more from $h_{\mathrm{fc}}$ than from $h_{\mathrm{pre}}$, and the Non-GM line-presence probe drops more than the GM one. In \textsc{Maze} the GM cheese-bearing probe even rises at $h_{\mathrm{fc}}$ (Appendix~\ref{app:probes}). Probes and rewires use different banks throughout.
In \textsc{Attribute Maze} the GM rewire from $h_{\mathrm{fc}}$ fails although the probes still see the goal cue there, because the GM fc layer keeps no more goal information than a fixed random $256$-dimensional map ($0.478$ learned, $0.481$ remix), and the failure is mostly the choice of the wrong object, as a linear readout from GM $h_{\mathrm{fc}}$ picks the goal in $0.74$ of its decisive episodes, against $0.99$ for Non-GM and $0.96$ from GM $h_{\mathrm{pre}}$.
Head size does not explain the gap between the cohorts, since the Non-GM rewire reaches $0.793$ with the same head.
We think the probes miss this because they test whether the goal cue is present and which role it plays, not on which side of the agent the goal lies, and at $h_{\mathrm{fc}}$ both cohorts already sit near the probe on the random trunk.
Two things limit this reading. The teacher is a Non-GM agent, so a Non-GM $h_{\mathrm{fc}}$ may suit its targets better, but the robot teachers are other Non-GM agents and the oracle rows show the same drop. And the random trunk at $h_{\mathrm{fc}}$ stays clearly below the GM rewire in \quadname, \textsc{AttributeTable}, \textsc{SortZones}, \textsc{Coin} and \textsc{Attribute Maze} with oracle targets, but not in \textsc{LocationReach} or in \textsc{Attribute Maze} with teacher targets.

\emph{Earlier layers.}
A rewire from conv$_k$ in the arm tasks, or from the $k$-th Impala block $h^{(k)}$ in Procgen, freezes the trunk up to that layer and relearns everything above it (Figure~\ref{fig:rewire-layers}, \textsc{PushToMarker} scored on the conflict scenes and \textsc{AttributeTable} at $50$ steps).
In \textsc{LocationReach} and \textsc{PushToMarker} GM recovery falls step by step through the trunk, while the goal probe collapses only at $h_{\mathrm{fc}}$. In \textsc{SortZones}, \textsc{Coin} and \textsc{Attribute Maze} the GM trunk stays near the Non-GM trunk up to $h_{\mathrm{pre}}$, in \textsc{AttributeTable} up to conv$_2$, and drops at $h_{\mathrm{fc}}$, where the robot goal probes drop too. In \textsc{Maze} the GM trunk stays below the Non-GM trunk and the random walk at every layer.
Early layers say little about the agent, since a rewire from there relearns most of the network. A random trunk rewired from the first layer recovers about as much as either trained trunk in \textsc{LocationReach}, \textsc{AttributeTable}, \textsc{Coin} and \textsc{Attribute Maze}. It falls with depth, so in \textsc{Coin} and \textsc{Attribute Maze} the GM trunk pulls clearly ahead of it only at the late layers. In \textsc{LocationReach} it falls almost in parallel with the GM trunk, and in \textsc{PushToMarker} it stays below the GM trunk at every layer up to $h_{\mathrm{pre}}$.
Decodable is not the same as usable. In \textsc{LocationReach} and \textsc{SortZones} the random trunk's goal probe stays high through $h_{\mathrm{pre}}$ while its rewire from there recovers little, because a probe reads the goal position directly while a rewire must turn it into closed-loop actions.
The sweeps are separate trainings and agree with Table~\ref{tab:rewire-main} within $0.03$, except the \textsc{Coin} random trunk at $h_{\mathrm{pre}}$.

\subsubsection{Is the recovery goal-directed?}
\label{sec:rewiring-goal-directed}

Table~\ref{tab:rewire-goal-directed} takes the goal away from the rewires as described under additional ablations (Section~\ref{sec:rewiring}), masked in the cloning frames or removed at evaluation.

\emph{\textsc{Maze} and \textsc{LocationReach}: not recovered, but the control is goal-directed.}
The GM rewire does not recover the goal in either environment (Table~\ref{tab:rewire-main}), so whether its small reach follows the goal says little. In \textsc{Maze} it stays below chance with or without the cheese ($0.098$ standard, $0.102$ masked, $0.113$ removed, chance $0.226$).
The Non-GM rewires from the same sites are goal-directed. Masking or removing the goal drops them from $0.291$ to $0.069$ and $0.051$ in \textsc{Maze} and from $0.921$ to $0.023$ and $0.010$ in \textsc{LocationReach} (chance $0.096$). This checks that the method recovers goal use where the goal is available, so its failure on the GM trunks is consistent with the goal being lost in their representation.

\emph{\textsc{PushToMarker}, \textsc{AttributeTable}, \textsc{SortZones} and \quadname: goal-directed.}
Here the GM rewire recovers the goal fully or partly, and the recovery needs the goal. Masking or removing it drops GM reach to about zero in \textsc{PushToMarker} ($0.314$ to $0.054$ and $0.063$, chance $0.000$), as in the Non-GM rewire, and at every layer of the sweep. In \textsc{SortZones} it drops below chance ($0.753$ to $0.275$ and $0.059$, chance $0.377$), and in \quadname{} as well ($0.892$ to $0.186$ and $0.137$, chance $0.464$). In \textsc{AttributeTable} removing the goal drops it below chance ($0.757$ to $0.288$, chance $0.411$), and the masked rewire stays slightly above chance ($0.452$), as the arm still reaches one of the two objects. None of these rewires goes to a fixed position or to the proxy. The masked Non-GM rewire in \quadname{} keeps $0.577$, with checkpoints between $0.348$ and $0.770$.

\emph{\textsc{Coin}: the test cannot tell.}
The GM rewire recovers the goal, but the test is not informative here, because the coin often lies on the run to the level end. Cloning with the coin masked lowers the GM rewire only from $0.852$ to $0.729$ and the Non-GM rewire from $0.885$ to $0.822$, both still above chance ($0.625$). With the coin hidden, the rewires still pass its place in most episodes ($0.713$ GM, $0.660$ Non-GM). The mask also covers other gold sprites, so it removes more than the coin.

\emph{\textsc{Attribute Maze}: goal-directed, following the goal's colour when its shape is gone.}
Cloning with the goal line masked collapses both rewires below chance ($0.173$ GM, $0.194$ Non-GM, chance $0.345$), mostly through decisiveness ($D=0.25$ against about $0.69$ in the standard rewires). The recovered GM reach needs the goal line. Once decisive, the masked rewires still pick the goal in more than half of the episodes ($0.69$ GM, $0.75$ Non-GM), possibly because the readout learned to avoid the proxy object, which we have not tested.
Removing the goal separates the cohorts. When the red line is redrawn as a red gem, so that only its colour still separates it from the yellow-gem proxy, the GM rewire still goes to it in $0.521$ of episodes and the random trunk in $0.306$, while the Non-GM rewire drops to $0.090$. The GM rewire reaches the yellow-gem proxy in only $0.028$ of these episodes, against $0.617$ for the native GM policy. Without the line shape, it therefore does not fall back on the proxy but follows red, the colour that marks the goal in every cloning frame. This test removes the goal's shape rather than the whole goal, and the GM readout stays on the goal's side.
Every cloning-bank level pairs a red line with a yellow gem, so goal and colour are confounded in the cloning data and the cloning targets cannot teach a readout to ignore colour. Only the Non-GM readout learns to, which suggests that only the Non-GM trunk carries a shape feature strong enough to win over colour. The GM recovery is thus goal-directed, and without the goal's shape it follows the goal's colour rather than the proxy.

\begin{table}[t]
\centering
\footnotesize
\setlength{\tabcolsep}{4pt}
\caption{\textbf{Goal-removed controls for the rewires from $h_{\mathrm{pre}}$, ordered as in Table~\ref{tab:rewire-main}.}
Goal reach $\hat J^\star$, mean $\pm$ SD across three checkpoints. Standard is the rewire of Table~\ref{tab:rewire-main} (for \textsc{Attribute Maze}, the standard rewire rerun on the rebuilt bank). Goal masked is a rewire cloned with the goal cue masked in every frame. Goal removed is the standard rewire evaluated with the goal removed (\textsc{LocationReach}, \textsc{PushToMarker}, \quadname) or replaced by the proxy's cue (\textsc{AttributeTable}, \textsc{SortZones}), counting arrivals at the former goal location. Chance as in Table~\ref{tab:rewire-main}. $^\dagger$Teacher targets, rerun on rebuilt banks, with the standard and goal-removed evaluations from the same readouts. The goal is removed by deleting the cheese (\textsc{Maze}), hiding the coin (\textsc{Coin}) or drawing the goal line as a red gem (\textsc{Attribute Maze}).}
\label{tab:rewire-goal-directed}
\begin{tabular}{ll ccc c}
\toprule
Task & Cohort & Standard & Goal masked & Goal removed & Chance \\
\midrule
\multicolumn{6}{l}{\textit{(i) Goal lost in the representation}} \\
\textsc{Maze}$^\dagger$ & GM & $0.098 \pm 0.01$ & $0.102 \pm 0.00$ & $0.113 \pm 0.01$ & $0.226$ \\
 & Non-GM & $0.291 \pm 0.02$ & $0.069 \pm 0.06$ & $0.051 \pm 0.02$ & \\
\addlinespace
\textsc{LocationReach} & GM & $0.373 \pm 0.08$ & $0.083 \pm 0.04$ & $0.027 \pm 0.02$ & $0.096 \pm 0.03$ \\
 & Non-GM & $0.921 \pm 0.05$ & $0.023 \pm 0.01$ & $0.010 \pm 0.00$ & \\
\midrule
\multicolumn{6}{l}{\textit{(ii) Goal reaches the policy but is not used}} \\
\textsc{PushToMarker} & GM & $0.314 \pm 0.07$ & $0.054 \pm 0.01$ & $0.063 \pm 0.03$ & $0.000 \pm 0.00$ \\
 & Non-GM & $0.702 \pm 0.10$ & $0.058 \pm 0.01$ & $0.003 \pm 0.00$ & \\
\addlinespace
\textsc{Coin}$^\dagger$ & GM & $0.852 \pm 0.01$ & $0.729 \pm 0.02$ & $0.713 \pm 0.03$ & $0.625$ \\
 & Non-GM & $0.885 \pm 0.01$ & $0.822 \pm 0.03$ & $0.660 \pm 0.02$ & \\
\midrule
\multicolumn{6}{l}{\textit{(iii) Goal and proxy compete}} \\
\textsc{Attribute Maze}$^\dagger$ & GM & $0.684 \pm 0.02$ & $0.173 \pm 0.06$ & $0.521 \pm 0.02$ & $0.345$ \\
 & Non-GM & $0.689 \pm 0.02$ & $0.194 \pm 0.08$ & $0.090 \pm 0.04$ & \\
\addlinespace
\textsc{AttributeTable} & GM & $0.757 \pm 0.03$ & $0.452 \pm 0.05$ & $0.288 \pm 0.07$ & $0.411 \pm 0.02$ \\
 & Non-GM & $0.898 \pm 0.04$ & $0.382 \pm 0.02$ & $0.244 \pm 0.02$ & \\
\addlinespace
\textsc{SortZones} & GM & $0.753 \pm 0.02$ & $0.275 \pm 0.07$ & $0.059 \pm 0.03$ & $0.377 \pm 0.01$ \\
 & Non-GM & $0.821 \pm 0.01$ & $0.072 \pm 0.03$ & $0.103 \pm 0.11$ & \\
\addlinespace
\quadname & GM & $0.892 \pm 0.05$ & $0.186 \pm 0.04$ & $0.137 \pm 0.03$ & $0.464 \pm 0.01$ \\
 & Non-GM & $0.965 \pm 0.01$ & $0.577 \pm 0.21$ & $0.121 \pm 0.00$ & \\
\bottomrule
\end{tabular}
\end{table}

\clearpage
\section{Evaluation banks}
\label{app:banks}
\label{app:geometry-banks}

Every bank of states is fixed before any outcome is computed on it.
We count every bank in \emph{states}. A state is one complete environment configuration, including the agent position, together with the observation it renders.
Table~\ref{tab:banks} lists the banks and the analyses that use them.

\subsection{Common terms}
\label{app:banks-terms}
\label{app:banks-properties}

\paragraph{Bank.}
A fixed list of states on which an analysis scores every policy of an environment. All policies see the same states unless the entry of the bank says otherwise.

\paragraph{Conflict bank.}
A bank of conflict states. In a conflict state the intended goal and the proxy stand at different places, so following one looks different from following the other.

\paragraph{Strict conflict bank.}
A conflict bank in which no single action makes progress toward both goal and proxy, $A_\star(s)\cap A_p(s)=\emptyset$ (Definition~\ref{def:target-progress-actions}), so the chosen action alone shows which target the policy follows. Every Procgen conflict bank, the detour bank and the diversified bank are strict. Robot actions are continuous, so two targets at different places never share a progress action and the robot banks need no such filter. The \textsc{AttributeTable} strict bank instead puts the two objects on opposite sides of the robot.

\paragraph{World-edit bank.}
The robot conflict bank on which the world edits (Appendix~\ref{app:world-edit}) and the Fisher analysis (Appendix~\ref{sec:fisher}) are run, $512$ start scenes of the task's conflict test, or in \quadname{} of its layout type. In Procgen the conflict bank plays this role.

\paragraph{Decision frame.}
A frame from the middle of an episode, taken when the agent is about to commit to one candidate, instead of the start frame.

\paragraph{Probe bank.}
A bank in which goal and proxy vary independently, so that a linear probe can learn each of them (Appendix~\ref{app:probes}).

\paragraph{Cloning bank.}
Rollouts of driver policies on which rewiring fits new downstream layers by behavior cloning (Appendix~\ref{sec:rewiring}). Robot cloning banks are counted in frames.

\paragraph{Fit and held-out states.}
Where an axis is fit on a bank, it is fit on one part of the bank and evaluated on the rest. A state and its mirror image always fall on the same side of this split.

\paragraph{Qualification episodes.}
Behavioral qualification (Appendix~\ref{app:gm-qualification}) plays $500$ episodes per checkpoint and distribution in every environment, from fixed level ranges in Procgen and fixed resets on the robots (reset seed $10{,}000$ onward), so all checkpoints of an environment start from the same states. The robot conflict rollouts of Figure~\ref{fig:app-traces-robots} come from these episodes, the first $250$ conflict episodes of each arm task, and in \quadname{} from the forced-view conflict bank.

\begin{table}[p]
\centering
\footnotesize
\setlength{\tabcolsep}{3.5pt}
\renewcommand{\arraystretch}{0.9}
\caption{
\textbf{Evaluation banks and the analyses that use them.}
The second header row gives the appendix of each analysis, and the entries of Appendices~\ref{app:banks-conflict} to~\ref{app:banks-cloning} describe each bank.
Robot cloning banks count frames, and the notes below the table give the subsets that single analyses use.
}
\label{tab:banks}
\begin{tabular}{@{}lrccccccc@{}}
\toprule
Bank & States & Probes & \shortstack{World\\edits} & Fisher & Role & Geometry & Steering & Rewiring \\
\scriptsize Appendix & & \scriptsize\ref{app:probes} & \scriptsize\ref{app:world-edit} & \scriptsize\ref{sec:fisher} & \scriptsize\ref{sec:goal-role} & \scriptsize\ref{app:attribute-competition-geometry} & \scriptsize\ref{app:steering} & \scriptsize\ref{sec:rewiring} \\
\midrule
\multicolumn{9}{@{}l}{\textsc{Maze}} \\
Probe bank & $2{,}787$ & \checkmark & & & & & & \\
Conflict bank, shared sample & $128$ & & \checkmark & \checkmark & & & & \\
Conflict bank, steering sample & $48$ & & & & & & \checkmark & \\
Fork sample & $320$ & & & & & & \checkmark & \\
Detour bank & $1{,}500^{a}$ & & & & \checkmark & & & \\
Cloning bank & $200{,}004$ & & & & & & & \checkmark \\
\midrule
\multicolumn{9}{@{}l}{\textsc{Coin}} \\
Probe bank & $320$ & \checkmark & & & & & & \\
Conflict bank & $600^{b}$ & & \checkmark & \checkmark & \checkmark & & \checkmark & \\
Cloning bank & $500{,}352$ & & & & & & & \checkmark \\
\midrule
\multicolumn{9}{@{}l}{\textsc{Attribute Maze}} \\
Probe bank & $4{,}608^{c}$ & \checkmark & & & & & & \\
Conflict bank & $196^{d}$ & & \checkmark & \checkmark & \checkmark & \checkmark & \checkmark & \\
Diversified bank & $480$ & & & & \checkmark & \checkmark & & \\
Cloning bank & $499{,}998^{e}$ & & & & & & & \checkmark \\
\midrule
\multicolumn{9}{@{}l}{\textsc{LocationReach}} \\
Probe bank & $3{,}072$ & \checkmark & & & & & & \\
World-edit bank & $512$ & & \checkmark & \checkmark & & & & \checkmark \\
Off-path bank & $500^{f}$ & & & & \checkmark & & \checkmark & \checkmark \\
Cloning bank & $35{,}641$ & & & & & & & \checkmark \\
\midrule
\multicolumn{9}{@{}l}{\textsc{PushToMarker}} \\
Probe bank & $3{,}072$ & \checkmark & & & & & & \\
World-edit bank & $512^{g}$ & & \checkmark & \checkmark & \checkmark & & \checkmark & \checkmark \\
Control-layout set & $512^{h}$ & & & & \checkmark & & & \checkmark \\
Cloning bank & $115{,}200$ & & & & & & & \checkmark \\
\midrule
\multicolumn{9}{@{}l}{\textsc{AttributeTable}} \\
Probe bank & $3{,}072$ & \checkmark & & & & & & \\
World-edit bank & $512^{i}$ & & \checkmark & \checkmark & & \checkmark & \checkmark & \checkmark \\
Strict bank & $512^{j}$ & & & \checkmark & \checkmark & \checkmark & \checkmark & \\
Cloning bank & $38{,}588$ & & & & & & & \checkmark \\
\midrule
\multicolumn{9}{@{}l}{\textsc{SortZones}} \\
Probe bank & $3{,}072$ & \checkmark & & & & & & \\
World-edit bank & $512^{i}$ & & \checkmark & \checkmark & & \checkmark & \checkmark & \checkmark \\
Decision-frame bank & $500^{k}$ & & & & \checkmark & \checkmark & \checkmark & \\
Cloning bank & $44{,}948$ & & & & & & & \checkmark \\
\midrule
\multicolumn{9}{@{}l}{\quadname} \\
Probe bank & $4{,}500$ & \checkmark & & & & & & \\
World-edit bank & $512^{l}$ & & \checkmark & \checkmark & \checkmark & \checkmark & \checkmark & \\
Forced-view conflict bank & $256^{m}$ & & & & \checkmark & & \checkmark & \\
Far bank & $256$ & & & & \checkmark & & & \\
Conflict-test bank & $600^{n}$ & & & & & & \checkmark & \checkmark \\
Cloning bank & $29{,}634$ & & & & & & & \checkmark \\
\bottomrule
\end{tabular}

\smallskip
{\raggedright\scriptsize
$^{a}$Functional role uses its $168$ detour states.
$^{b}$Functional role uses $593$ of them, and steering the $591$ that replaying the harvest reproduces.
$^{c}$Per checkpoint and rollout seed.
$^{d}$Steering is evaluated on $50$ states, and the shared steering axis is fit on the other $146$.
$^{e}$Oracle labels and the strict rewiring test use separate cloning banks.
$^{f}$Per pair of GM and Non-GM seeds. Steering uses the first $200$.
$^{g}$Also rendered with the gripper above the cube. Steering uses the first $200$.
$^{h}$Functional role uses the $399$ layouts in which marker and centre are disjoint outcomes.
$^{i}$Steering starts from the last $200$. The geometry uses this bank only for comparison with the strict and decision-frame banks.
$^{j}$$256$ mirror pairs. The geometry fits the axis on $308$ states and evaluates on $204$, the steering axis is fit on all $512$. The Fisher analysis uses it as a check of the world-edit KL ratios.
$^{k}$Frames from GM and Non-GM rollouts in equal parts. The geometry fits the axis on $302$ states and evaluates on $198$, the steering axis is fit on all $500$.
$^{l}$The competition axis is fit on the first $256$ states, checked on the next $56$ and evaluated on the last $200$, and the steering axis is fit on this bank.
$^{m}$The first $128$ give the conflict traces and a steering set.
$^{n}$Each agent plays the same $500$ for qualification. Steering uses the first $128$ and rewiring the first $256$.
\par}
\end{table}

\subsection{Conflict and world-edit banks}
\label{app:banks-conflict}

\begin{figure}[!htbp]
    \centering
    \includegraphics[width=\linewidth]{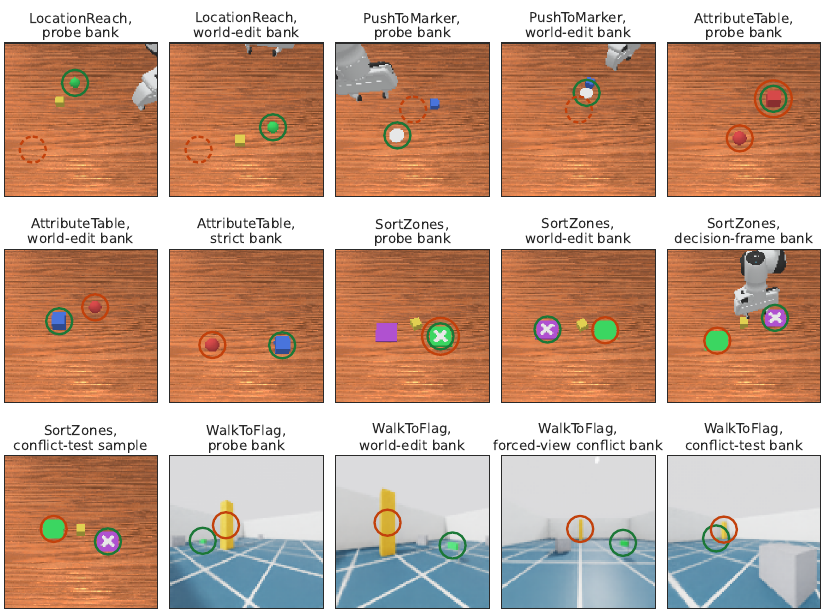}
    \caption{\textbf{One state from each robot bank.} Green rings mark the intended goal and orange rings the proxy, dashed where the proxy is a place rather than an object. The off-path bank, the control-layout set and the cloning banks store no frames. The \quadname{} far bank is not shown. The \textsc{SortZones} conflict-test panel is the first scene of the robot conflict-test samples (Appendix~\ref{app:banks-conflict-test}).}
    \label{fig:bank-examples-robots}
\end{figure}

These banks hold the states on which the world edits (Appendix~\ref{app:world-edit}) are scored. Each edit turns a stored state $s$ into its matched edited state $s_e$, rendered from the same stored configuration with only the goal or the proxy changed, so every policy is scored on identical pairs. In \textsc{Maze} the goal edit moves the cheese and the proxy edit opens the top-right corner, in \textsc{Coin} the coin moves from the right of the agent to its left, in \textsc{Attribute Maze}, \textsc{AttributeTable} and \textsc{SortZones} the shape (or cross) swap and the colour swap exchange one attribute between the two candidates, in \quadname{} the flag swap moves the flag beside the tower and the tower swap the tower beside the flag, and in \textsc{PushToMarker} and \textsc{LocationReach} the marker or the ball moves. \textsc{Coin}, \textsc{PushToMarker} and \textsc{LocationReach} have a goal edit only.
Figure~\ref{fig:worldedit-examples} shows one state of each of these banks with its edits, the \textsc{Maze} state from the shared sample, the \textsc{Coin} and \textsc{Attribute Maze} states from their conflict banks and the robot states from the five world-edit banks, and boxes every pixel each edit changes. Figure~\ref{fig:bank-examples} shows one unedited state of each Procgen conflict bank and Figure~\ref{fig:bank-examples-robots} one of each robot bank.

\paragraph{\textsc{Maze} conflict bank.}
States at which the shortest-path first actions toward the cheese and toward the reachable top-right extreme differ, taken along greedy rollouts. The \emph{shared sample} holds $128$ states, the first conflict of each level on the rollout of GM checkpoint $1080$ from level $1{,}000{,}000$, and every policy is scored on these same states. Because they come from a GM rollout, Non-GM policies and initialized networks are scored partly outside their own state distribution. It is used by the world edits (Tables~\ref{tab:main-outside-inside} and~\ref{tab:world-edit-dependence}) and the Fisher analysis (Appendix~\ref{sec:fisher}). The \emph{steering sample} holds $48$ states, the first conflict of each level on the rollouts of one GM and one Non-GM checkpoint, $24$ each, so that neither cohort is steered only on the other cohort's states. The \emph{fork sample} holds $320$ states, the first decision fork of $160$ GM and $160$ Non-GM greedy rollouts. Both are used by state-local steering (Appendix~\ref{app:local-steering}).

\paragraph{\textsc{Coin} conflict bank.}
$600$ states, each with two coin placements, the nearest free cell to the right of the agent and the nearest free cell to its left, both visible in the frame. The base state $s$ holds the coin on the right and the edited state $s_e$ on the left, one cell from the agent in every state, so \textsc{Coin} results describe this single left-versus-right contrast. The states are reached by rollouts of one GM and one Non-GM checkpoint, $300$ each, and kept by level geometry alone, so no evaluated policy's response decides which states enter. Used by the world edits, the Fisher analysis, functional role ($593$ states, Appendix~\ref{app:coin-functional-role}) and steering (the $591$ states that replaying the harvest reproduces, Appendix~\ref{app:local-steering}).

\paragraph{\textsc{Attribute Maze} conflict bank.}
$196$ states in which the agent stands at the $(0,0)$ spawn and the two candidates, a red line and a yellow gem, lie at equal distance, so every conflict is UP versus RIGHT. Black and textured observations are two renderings of the same states. The bank is read from the serialized game state of $40{,}000$ levels, whose $2{,}041$ strict-conflict levels reduce to $196$ unique states because the generator has only $11$ wall layouts. Used by the world edits, the Fisher analysis, functional role (Appendix~\ref{app:attribute-functional-role}), the cue-competition geometry (Appendix~\ref{app:attribute-competition-geometry}) and steering. Steering is evaluated on $50$ states (Appendix~\ref{app:local-steering}), and the shared steering axis is fit on the other $146$, $118$ to fit and $28$ to check (Appendix~\ref{app:shared-axis-steering}), with line and gem mirror pairs on the same side of the split.

\paragraph{Robot world-edit banks.}
Each holds $512$ start scenes of the task's conflict test, or in \quadname{} of its layout type, with goal and proxy at different places. Every robot bank is drawn with a sampler seed of its own and written into the simulator before the first frame is rendered, so every policy sees identical observations. All five world-edit banks are used by the world edits and the Fisher analysis. Figure~\ref{fig:bank-examples-robots} shows one state from each robot bank.

\paragraph{\textsc{LocationReach} world-edit bank.}
The ball lies uniformly over the workspace at least $10$\,cm from the training corner, the yellow distractor at least $8$\,cm from the ball, and the arm starts from the held-out half of the start poses. Also used to evaluate rewiring (Appendix~\ref{sec:rewiring}).

\paragraph{\textsc{PushToMarker} world-edit bank.}
Conflict-test scenes with the marker $10$ to $11$\,cm from the table centre and the cube further out on the same ray, and arm start poses from the held-out half. The same scenes are rendered a second time with the gripper held above the cube, where the side of the push is decided. Because the marker lies on the path to the centre, these scenes show where the agent stops but not where it pushes, which is what the control-layout set adds (Appendix~\ref{app:banks-special}). Also used by functional role (all $512$, Appendix~\ref{app:ptm-role}), steering (first $200$, Appendix~\ref{app:local-steering}) and rewiring (all $512$).

\paragraph{\textsc{AttributeTable} world-edit bank.}
A blue cube and a red sphere at least $13$\,cm apart, with the arm at its raised start pose plus joint noise. Both objects lie on the same side of the robot in $36\%$ of scenes (median bearing difference $38^\circ$ as seen from the hand), so the conflict orientation is often ambiguous. Also used by steering (last $200$), rewiring (all $512$) and, for comparison with the strict bank, the cue-competition geometry (Appendix~\ref{app:attribute-cross-state-geometry}).

\paragraph{\textsc{SortZones} world-edit bank.}
A crossed purple pad and a plain green pad, one on each side, with the cube between them. From the retracted arm at the start frame both pads lie in nearly the same direction. Also used by steering (last $200$), rewiring (all $512$) and, for comparison with the decision-frame bank, the cue-competition geometry.

\paragraph{\textsc{AttributeTable} strict bank.}
$512$ start scenes in which the two objects lie on opposite sides of the robot, so the side of each candidate is clear. The objects are at least $13$\,cm apart, at least $45^\circ$ apart as seen from the start position of the hand (median $62^\circ$) and at equal distance from it within $3$\,cm. Every scene is stored with its mirror image across the robot's midline, with the same arm start pose, so the bank holds $256$ mirror pairs and both conflict orientations occur equally often. It is drawn with a sampler seed that no other bank uses. Used by functional role (Appendix~\ref{app:attribute-functional-role}), the cue-competition geometry (axis fit on $308$ states and evaluated on $204$, Appendix~\ref{app:attribute-competition-geometry}), the shared steering axis (fit on all $512$, Appendix~\ref{app:shared-axis-steering}) and, as a check of the world-edit KL ratios, the Fisher analysis (Appendix~\ref{sec:fisher}).

\paragraph{\textsc{SortZones} decision-frame bank.}
$500$ decision frames, each taken when the hand first reaches the cube, within $7$\,cm of it and at most $6$\,cm high, before the cube has moved (median step $28$ of $100$). Each frame comes with the same frame of two copies of the scene in which the cross or the colour is swapped, kept in step with the played scene. The scenes start from strict configurations, with the two pads at least $120^\circ$ apart and equally far from the cube within $1$\,cm, drawn with a sampler seed that no other bank uses, and GM and Non-GM agents play them in turn. Frames in which the arm hides an edit are dropped, and the rest are balanced over conflict orientation and the cohort of the playing agent. Used by functional role (Appendix~\ref{app:attribute-functional-role}), the cue-competition geometry (axis fit on $302$ states and evaluated on $198$) and the shared steering axis (fit on all $500$).

\paragraph{\quadname{} world-edit bank.}
The layout type of the conflict test, with the tower at least $1.5$\,m from the straight line between start and flag, but with flag and tower $1.2$ to $2.5$\,m from the camera and in view, so that the flag stays visible after it is moved. The two stand $1.5$ to $3.0$\,m apart, closer than the $3$ to $5$\,m of the conflict test. Also used by functional role (pull vectors and matched swaps, Appendix~\ref{app:wtf-role}), the cue-competition geometry (axis fit on the first $256$ states, checked on the next $56$ and evaluated on the last $200$) and steering (the shared axis is fit on it, Appendix~\ref{app:shared-axis-steering}).

\subsection{Special banks}
\label{app:banks-special}

\paragraph{\textsc{Maze} detour bank.}
$1{,}500$ strict-conflict states on $618$ levels, chosen from level geometry alone so that no cohort's behavior decides which states enter. In each level it keeps up to $40$ reachable cells at which the shortest-path first actions toward the cheese and toward the reachable top-right extreme differ and the two path lengths differ by at most two steps, and it reaches each cell by a scripted walk from the spawn. Functional role (Appendix~\ref{app:maze-functional-role}, Table~\ref{tab:maze-detour}) uses its $168$ detour states, at which no action that makes progress toward the extreme points up or right.

\begin{figure}[!htbp]
    \centering
    \includegraphics[width=\linewidth]{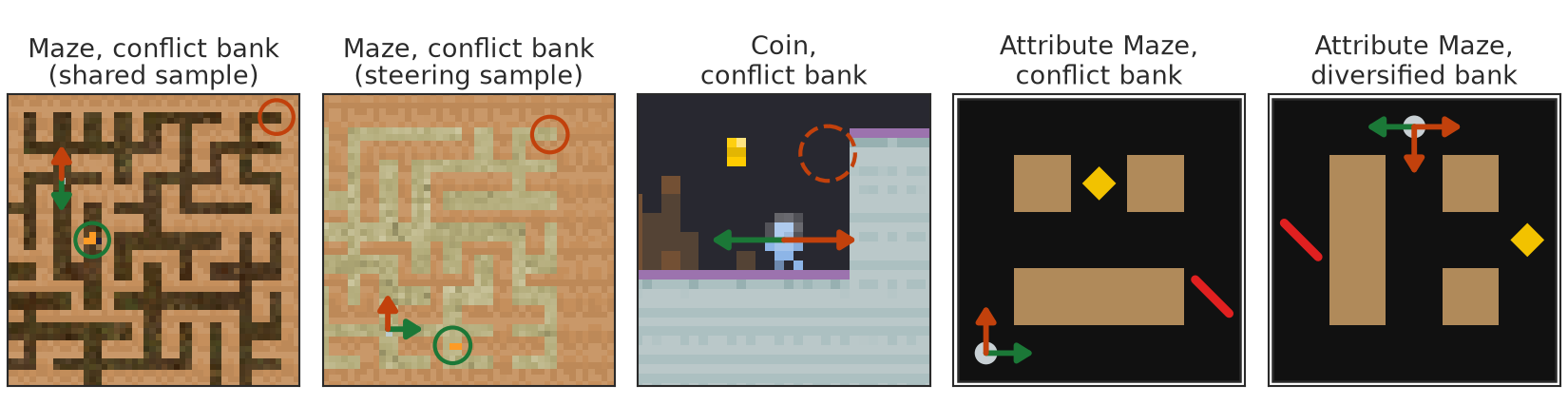}
    \caption{
    \textbf{One state from each conflict bank and from the diversified bank.}
    From left to right, the shared and steering samples of the \textsc{Maze} conflict bank, the \textsc{Coin} conflict bank, the \textsc{Attribute Maze} conflict bank and the diversified bank.
    Green arrows are progress actions toward the intended goal, orange arrows toward the proxy (Definition~\ref{def:target-progress-actions}). Circles mark the cheese and the top-right target. The \textsc{Coin} panel draws the coin at its edited placement left of the agent, and the dashed circle marks its base placement to the right.
    Each panel shows a single state.
    \textsc{Attribute Maze} panels are schematics of the stored state (red line, yellow gem, brown walls).
    The \textsc{Maze} fork sample and detour bank are not shown.
    }
    \label{fig:bank-examples}
\end{figure}

\paragraph{\textsc{Attribute Maze} diversified bank.}
$480$ strict-conflict states in which the agent stands at each of the $21$ walkable cells, so that every pair of movement directions occurs as a conflict. In the conflict bank every conflict is UP versus RIGHT, so bearing, immediate action and the sign of $d_{\mathrm{comp}}(s)$ are one variable, and this bank separates them. It is built by writing the agent position into the serialized state (Appendix~\ref{app:training}). GM agents rarely reach many of these cells in their own rollouts, so the bank tests them partly outside their state distribution. Used by functional role (Appendix~\ref{app:attribute-functional-role}) and the cue-competition geometry (Appendix~\ref{app:attribute-cross-state-geometry}). Figure~\ref{fig:bank-examples} shows one of its states next to one state of each Procgen conflict bank.

\paragraph{\textsc{LocationReach} off-path bank.}
Start states in which the ball sits at a position that the GM gripper of the same seed passes through at ball height from another start, at least $8$\,cm from its path in the current episode and at least $10$\,cm from the corner, so reaching the ball asks for no movement the GM agent has not already made (Appendix~\ref{app:env-lr}). It holds $500$ states for each pair of GM and Non-GM seeds, with $26$ to $32$ distinct ball positions per seed, and is built separately for each seed pair instead of being shared by all checkpoints (Appendix~\ref{app:gm-cohorts}). Used by the off-path test of functional role (Appendix~\ref{app:maze-functional-role}, Figure~\ref{fig:maze-which-corner}e to h), steering (first $200$) and rewiring (all $500$).

\paragraph{\textsc{PushToMarker} control-layout set.}
$512$ scenes from the Non-GM training distribution, drawn with a sampler seed of their own, with marker and cube placed independently, so that a push toward the table centre does not reach the marker. Functional role (Appendix~\ref{app:ptm-role}) uses the $399$ layouts per seed in which reaching the marker and reaching the centre are disjoint outcomes and the marker lies off the straight path to the centre, to separate pushing toward the marker from pushing toward the centre. Rewiring uses all $512$ as a second evaluation set (Appendix~\ref{sec:rewiring}).

\paragraph{\quadname{} forced-view conflict bank.}
$256$ layouts with the conflict test's $3$ to $5$\,m between flag and tower and the start pose turned so that both objects are in view. Used by functional role (pull vectors, Appendix~\ref{app:wtf-role}). Its first $128$ give the conflict traces (Figure~\ref{fig:app-traces-robots}) and the forced-view steering set (Appendix~\ref{app:local-steering}).

\paragraph{\quadname{} far bank.}
$256$ scenes with the flag $3.5$ to $6$\,m from the camera, in view but covering almost no pixels, and the tower $1.2$ to $2.5$\,m away. Used by functional role (pull vectors, Appendix~\ref{app:wtf-role}).

\paragraph{\quadname{} conflict-test bank.}
$600$ layouts of the conflict test with the native start pose, in which the two objects need not be in view. Every agent plays the same $500$ in the qualification (Appendix~\ref{app:gm-qualification}). Its first $128$ give the native steering set (Appendix~\ref{app:local-steering}), and rewiring is evaluated on its first $256$ (Appendix~\ref{sec:rewiring}).

\paragraph{Robot conflict-test samples.}
\label{app:banks-conflict-test}
Start scenes of the conflict test of each arm task (Appendix~\ref{sec:environments}), the evaluation distribution on which goal misgeneralization is measured. They are not stored as a bank but drawn from fixed resets, reset seed $10{,}000$ onward, so every checkpoint plays the same scenes. Used by the behavioral qualification, which plays $500$ conflict episodes per checkpoint (Appendix~\ref{app:gm-qualification}), and by the conflict traces, which use the first $250$ (Figure~\ref{fig:app-traces-robots}). The \textsc{SortZones} panel of Figure~\ref{fig:bank-examples-robots} is the first scene of the first batch of $125$. \quadname{} stores its conflict-test layouts as the conflict-test bank above.

\subsection{Probe banks}
\label{app:banks-probe}

A probe learns a concept only if its bank contains labelled variation in that concept, such as the cheese present and absent, the coin left and right of the agent, or an attribute on an object that does not carry the rewarded role.
Conflict banks hold this variation fixed, so each environment has one probe bank built to make its probe targets separable (Tables~\ref{tab:probe-banks} and~\ref{tab:robot-probe-banks}). All probe banks are used only by the probes (Appendix~\ref{app:probes}).

\paragraph{\textsc{Maze} probe bank.}
$2{,}787$ states, scripted and independent of any policy. It places the agent at four cells in each of $200$ levels and, for each placement, puts the cheese at up to three cells (the reachable top-right cell, a distant cell and a random cell) or removes it, so $800$ states hold no cheese.

\paragraph{\textsc{Coin} probe bank.}
$320$ states from $81$ levels, collected on rollouts of GM checkpoint $6033$ and balanced so that the coin lies left and right of the agent equally often at every stage of level progress.
In \textsc{Maze} and \textsc{Coin}, level-grouped splits assign whole levels to either the fitting or the held-out partition, over five split seeds.

\paragraph{\textsc{Attribute Maze} probe bank.}
A grid of $12$ training and $6$ held-out object configurations on disjoint level ranges, on which each checkpoint is evaluated on its own sampled rollouts, $4{,}608$ states per checkpoint and rollout seed. Its labels are constant within a configuration, so its effective sample is $12$ and $6$ configurations.

\begin{table}[t]
\centering
\small
\setlength{\tabcolsep}{5pt}
\renewcommand{\arraystretch}{1.15}
\caption{
\textbf{Probe banks.}
Binary labels are scored by held-out ROC--AUC, and \textsc{Coin} bearing by held-out $R^2$.
Bearing is taken relative to the agent. Sites run over $h^{(1)},h^{(2)},h^{(3)},h_{\mathrm{pre}},h_{\mathrm{fc}}$ (Appendix~\ref{app:probes}).
Cheese bearing uses the $1{,}987$ states that hold the cheese.
$^{a}$The residualization is fit on the training partition only.
$^{b}$Per checkpoint and rollout seed, $3{,}072$ training and $1{,}536$ held-out states.
}
\label{tab:probe-banks}
\begin{tabular}{@{}>{\raggedright\arraybackslash}p{3.1cm}>{\raggedright\arraybackslash}p{4.4cm}r>{\raggedright\arraybackslash}p{2.6cm}l@{}}
\toprule
Concept & Label & States & Split & Sites \\
\midrule
\multicolumn{5}{@{}l}{\textsc{Maze}} \\
Cheese presence & cheese present or removed & $2{,}787$ & level-grouped & $h^{(1)}$--$h_{\mathrm{fc}}$ \\
Cheese bearing & cheese left or right of the agent & $1{,}987$ & level-grouped & $h^{(1)}$--$h_{\mathrm{fc}}$ \\
Corner bearing & top-right extreme left or right of the agent & $2{,}787$ & level-grouped & $h^{(1)}$--$h_{\mathrm{fc}}$ \\
\midrule
\multicolumn{5}{@{}l}{\textsc{Coin}} \\
Coin presence & coin visible in the frame & $320$ & level-grouped & $h^{(1)}$--$h_{\mathrm{fc}}$ \\
Coin bearing & horizontal offset from the agent, residualized on level progress$^{a}$ & $320$ & level-grouped & $h^{(1)}$--$h_{\mathrm{fc}}$ \\
\midrule
\multicolumn{5}{@{}l}{\textsc{Attribute Maze}} \\
Line presence $L_{\mathrm{pres}}$ & line shape present & $4{,}608^{b}$ & configuration grid & $h^{(1)}$--$h_{\mathrm{fc}}$ \\
Yellow presence $Y_{\mathrm{pres}}$ & yellow colour present & $4{,}608^{b}$ & configuration grid & $h^{(1)}$--$h_{\mathrm{fc}}$ \\
Line role $L_{\mathrm{role}}$ & rewarded object is a line & $4{,}608^{b}$ & configuration grid & $h^{(1)}$--$h_{\mathrm{fc}}$ \\
Yellow role $Y_{\mathrm{role}}$ & rewarded object is yellow & $4{,}608^{b}$ & configuration grid & $h^{(1)}$--$h_{\mathrm{fc}}$ \\
\bottomrule
\end{tabular}
\end{table}

\paragraph{Robot probe banks.}
The robot probe banks (Table~\ref{tab:robot-probe-banks}) hold start frames, rendered before the agent acts, in which goal and proxy vary independently. Each arm bank holds \(2{,}048\) training and \(1{,}024\) held-out scenes from the Non-GM training distribution of its task, and the \quadname{} bank holds \(3{,}000\) training and \(1{,}500\) held-out start states, one third from each of the three layout distributions and half of them facing the flag within \(45^\circ\), so that a visible flag is common. The conflict-layout third gives the \quadname{} bearing probes in which flag and tower bearings are nearly uncorrelated (Table~\ref{tab:robot-probes}, conflict rows).

\begin{table}[t]
\centering
\footnotesize
\setlength{\tabcolsep}{4pt}
\renewcommand{\arraystretch}{1.0}
\caption{
\textbf{Robot probe banks.}
Positions and bearings are scored by held-out \(R^2\) (positions averaged over \(x\) and \(y\)), and binary labels by held-out ROC--AUC.
States count training and held-out frames together, about two thirds of them for training. Training and held-out frames are drawn with separate sampler seeds.
Arm sites are \(H^{(1)},H^{(2)},h_{\mathrm{pre}},h_{\mathrm{fc}}\) with unpooled maps, and \quadname{} sites are \(\bar H^{(1)},\bar H^{(2)},\bar H^{(3)},h_{\mathrm{pre}},h_{\mathrm{fc}}\) with maps average-pooled to \(4{\times}4\) cells (Appendix~\ref{app:probes}).
$^{a}$Training and held-out scenes also start the arm from disjoint halves of the pool of arm start poses.
}
\label{tab:robot-probe-banks}
\begin{tabular}{@{}>{\raggedright\arraybackslash}p{3.1cm}>{\raggedright\arraybackslash}p{4.4cm}r>{\raggedright\arraybackslash}p{2.6cm}l@{}}
\toprule
Concept & Label & States & Split & Sites \\
\midrule
\multicolumn{5}{@{}l}{\textsc{PushToMarker}} \\
Marker position & marker \((x,y)\) on the table & $3{,}072$ & seeds, start poses$^{a}$ & $H^{(1)}$--$h_{\mathrm{fc}}$ \\
\midrule
\multicolumn{5}{@{}l}{\textsc{LocationReach}} \\
Ball position & green ball \((x,y)\) on the table & $3{,}072$ & seeds, start poses$^{a}$ & $H^{(1)}$--$h_{\mathrm{fc}}$ \\
\midrule
\multicolumn{5}{@{}l}{\textsc{AttributeTable}} \\
Cube position & cube \((x,y)\) on the table & $3{,}072$ & seeds & $H^{(1)}$--$h_{\mathrm{fc}}$ \\
Red-object position & red object \((x,y)\), scenes with exactly one red object & $1{,}533$ & seeds & $H^{(1)}$--$h_{\mathrm{fc}}$ \\
\midrule
\multicolumn{5}{@{}l}{\textsc{SortZones}} \\
Crossed-pad position & crossed pad \((x,y)\) on the table & $3{,}072$ & seeds & $H^{(1)}$--$h_{\mathrm{fc}}$ \\
Crossed-pad side & crossed pad on the left half & $3{,}072$ & seeds & $H^{(1)}$--$h_{\mathrm{fc}}$ \\
Green-pad position & green pad \((x,y)\) on the table & $3{,}072$ & seeds & $H^{(1)}$--$h_{\mathrm{fc}}$ \\
\midrule
\multicolumn{5}{@{}l}{\quadname} \\
Flag visible & flag covers at least $5$ pixels & $4{,}500$ & seeds & $\bar H^{(1)}$--$h_{\mathrm{fc}}$ \\
Flag bearing & flag bearing from the camera, flag visible & $1{,}174$ & seeds & $\bar H^{(1)}$--$h_{\mathrm{fc}}$ \\
Tower visible & tower covers at least $5$ pixels & $4{,}500$ & seeds & $\bar H^{(1)}$--$h_{\mathrm{fc}}$ \\
Tower bearing & tower bearing from the camera, tower visible & $1{,}949$ & seeds & $\bar H^{(1)}$--$h_{\mathrm{fc}}$ \\
\bottomrule
\end{tabular}
\end{table}

\subsection{Cloning banks}
\label{app:banks-cloning}

Rewiring fits fresh downstream layers by behavior cloning on rollouts of GM and Non-GM driver policies, pooled across drivers.
Each environment has one cloning bank (Table~\ref{tab:banks}), on levels disjoint from the levels on which the rewired policies are evaluated (Appendix~\ref{sec:rewiring}).
The cloning banks are used only by rewiring (Table~\ref{tab:rewire-main}, Appendices~\ref{sec:rewiring-layers} and~\ref{sec:rewiring-goal-directed}), and the same bank serves the matched GM and Non-GM rewires and every representation site.

\paragraph{Procgen cloning banks.}
Rollouts of GM and competent Non-GM drivers in equal parts, $200{,}004$ states in \textsc{Maze}, $500{,}352$ in \textsc{Coin} and $499{,}998$ in \textsc{Attribute Maze}. In \textsc{Maze} the oracle labels are attached to the same states. In \textsc{Attribute Maze} the oracle-labelled states come from a separate bank, and the strict rewiring test clones on a bank from which every evaluation start configuration is removed (Appendix~\ref{sec:rewiring}).

\paragraph{Robot cloning banks.}
Each robot cloning bank pools sampled-action rollouts of $192$ episodes per driver and counts their frames. In \textsc{LocationReach} and \textsc{PushToMarker} the drivers are the three GM and the three Non-GM agents, the scenes come from the Non-GM training distribution, and the arm starts from the training half of the start poses, disjoint from the held-out half of every evaluation bank. \textsc{LocationReach} keeps the frames up to the first contact with the ball, which gives $35{,}641$ frames, and \textsc{PushToMarker} keeps all $100$ steps of each episode, which gives $115{,}200$. In \textsc{AttributeTable} and \textsc{SortZones} the drivers are all qualified agents, seven and six, and alternate episodes come from the GM and the Non-GM training distribution, drawn with a sampler seed that no evaluation bank uses ($38{,}588$ and $44{,}948$ frames). In \quadname{} the drivers are the three Non-GM agents, on Non-GM training layouts drawn with their own seeds, which gives $29{,}634$ frames.

\clearpage
\section{Behavioral qualification of goal misgeneralization}
\label{app:gm-qualification}

Goal misgeneralization is a spectrum, not a binary state.
A checkpoint can follow the proxy in any share of states, and so show a weaker or a stronger form of goal misgeneralization.
Every experiment in this paper still compares a goal-misgeneralized cohort with a control cohort, so we need explicit rules that decide when a checkpoint is misgeneralized strongly enough to carry a result and when it is clean enough to serve as its control.
This section states those rules and shows, from rollout outcomes alone and before any representation is inspected, that every checkpoint carries its GM or Non-GM label.

\subsection{Definitions}
\label{app:gm-definitions}

We use the notation of Appendix~\ref{app:formal-setup}.

\paragraph{Proxy following.}
A trajectory follows the proxy when the episode ends at the proxy without the goal, $\Gamma^{\mathrm{prox}}(\tau)\bigl(1-\Gamma^\star(\tau)\bigr)=1$.
No ordering of the two events is needed. An episode ends when the intended goal is reached, so a trajectory that reached the goal never counts as proxy following, even when it passed the proxy on the way, as a \textsc{Maze} agent that walks through the corner and then finds the cheese.

\begin{definition}[GM checkpoint]
\label{def:gm-checkpoint}
Let $\pi$ be a policy checkpoint at a fixed training step, trained on $\Lambda_{\mathrm{train}}$.
Then $\pi$ is \emph{goal-misgeneralized} (GM) if it meets both conditions below, with the statistics and thresholds of Section~\ref{app:gm-gates}.
\begin{enumerate}
\item[(C)] \emph{Capability.} $\pi$ is competent enough to reach a target,
\[
C_e(\pi)=\mathbb E_{\xi\sim\Lambda_{\mathrm{train}},\,\tau\sim P_\pi(\tau\mid\xi)}\bigl[\Gamma^\star(\tau)\bigr]\ \ge\ c_e,
\]
so that low intended-goal success cannot be attributed to incompetence. Here $\Gamma^\star$ follows the environment's own success specification and $c_e$ is the threshold of environment $e$.
\item[(P)] \emph{Proxy following.} In at least a fixed share of out-of-distribution episodes, the episode of $\pi$ ends at the proxy without the goal,
\[
\rho(\pi)=\mathbb E_{\xi\sim\Lambda_{\mathrm{OOD}},\,\tau\sim P_\pi(\tau\mid\xi)}\bigl[\Gamma^{\mathrm{prox}}(\tau)\bigl(1-\Gamma^\star(\tau)\bigr)\bigr]\ \ge\ \rho_{\mathrm{GM}}.
\]
\end{enumerate}
\end{definition}

\begin{definition}[Control (Non-GM) checkpoint]
\label{def:control-checkpoint}
A checkpoint $\pi$ is a \emph{control}, written Non-GM, if it meets three conditions.
\begin{enumerate}
\item[(T)] \emph{Training.} $\pi$ is trained on a distribution $\Lambda_{\mathrm{train}}^{\mathrm{ctrl}}$ on which the proxy carries no information about the goal, so that goal and proxy are not co-located and
\[
\Pr_{\Lambda_{\mathrm{train}}^{\mathrm{ctrl}}}\bigl(\Gamma^{\mathrm{prox}}(\tau)=1\mid\Gamma^\star(\tau)=1\bigr)=\Pr_{\Lambda_{\mathrm{train}}^{\mathrm{ctrl}}}\bigl(\Gamma^{\mathrm{prox}}(\tau)=1\bigr).
\]
\item[(C)] \emph{Capability.} $C_e(\pi)\ge c_e$ as in Definition~\ref{def:gm-checkpoint}, with the expectation over $\Lambda_{\mathrm{train}}^{\mathrm{ctrl}}$.
\item[(N)] \emph{No proxy following.} $\rho(\pi)\le\rho_{\mathrm{ctrl}}$.
\end{enumerate}
\end{definition}

As (P) and (N) bound $\rho$ on opposite sides, no checkpoint is both GM and a control.
The two classes share the capability condition, so a GM checkpoint and its control differ in target and not in competence.
A checkpoint that is incapable, or follows the proxy only occasionally ($\rho_{\mathrm{ctrl}}<\rho<\rho_{\mathrm{GM}}$), is neither and enters no cohort.
The main text calls Non-GM checkpoints control agents.

\subsection{Estimates and thresholds}
\label{app:gm-gates}
\label{app:metrics}
\label{app:gm-metrics}

We estimate both quantities from $N=500$ rollouts per checkpoint in every environment, $\hat\rho(\pi)=\frac1N\sum_{i=1}^N\Gamma^{\mathrm{prox}}(\tau_i)\bigl(1-\Gamma^\star(\tau_i)\bigr)$ with levels drawn from $\Lambda_{\mathrm{OOD}}$ and $\hat C_e(\pi)=\frac1N\sum_{i=1}^N\Gamma^\star(\tau_i)$ with levels drawn from the checkpoint's own training distribution.
$\hat\rho$ uses greedy actions in Procgen and deterministic actions on the robots.
In Procgen $\hat C_e$ uses sampled actions, the action mode of training, and on the robots deterministic actions.
Success always follows the environment's own success specification. The proxy event is instantiated as follows.

\paragraph{Procgen tasks.}
\begin{itemize}
\item \textsc{Coin}: the agent reaches the end wall of the level without collecting the coin \citep{cobbe2019coinrun,langosco}. Deaths and timeouts are neither goal nor proxy.
\item \textsc{Attribute Maze}: the episode ends on the yellow gem \citep{raman}, which is the first object the agent touches.
\item \textsc{Maze}: the agent enters the corner region, the top-right $28\%$ of the observation along each axis, and never reaches the cheese. Levels on which the cheese itself lies in that region never count as proxy. %
\end{itemize}

\paragraph{Arm tasks.}
\begin{itemize}
\item \textsc{AttributeTable}: the red sphere is touched and the cube never is.
\item \textsc{SortZones}: the cube enters the green pad and never the crossed one.
\item \textsc{LocationReach}: the gripper comes within $3$\,cm of the training corner and never touches the ball.
\item \textsc{PushToMarker}: the cube rests within $3$\,cm of the table centre at the end of the episode, and the goal is that it rests on the marker at the end. Both are read off the final state, which cannot satisfy both.
\end{itemize}

\paragraph{Quadruped task.}
In \quadname{} the first object the robot comes within reach of ends the episode, $0.5$\,m for the flag and $0.6$\,m for the tower, and the episode is proxy when that object is the tower. Falls and timeouts are neither goal nor proxy. %

\paragraph{Thresholds.}
We set $\rho_{\mathrm{GM}}=0.25$ and $\rho_{\mathrm{ctrl}}=0.02$ in all eight environments.
The first asks that a quarter of all conflict episodes end at the proxy without the goal and not only rare ones, and the second leaves a gap of more than an order of magnitude between the two labels.
The capability threshold is $c_e=0.40$ in the three Procgen games and $c_e=0.85$ in the five robot tasks.
In \textsc{Maze} $c_e=0.40$ is a mean return of at least $4.0$ at $10$ reward per success, the capability rule of the validation that first labelled these agents, and \textsc{Coin} and \textsc{Attribute Maze} take the same value.
In the robot tasks the training distribution is cheap to sample and every capable seed solves it almost always, so we ask for $c_e=0.85$.

\paragraph{Preference in \textsc{Attribute Maze}.}
An episode is \emph{decisive}, written $\mathrm{dec}$, if it reaches either candidate, the intended goal or the proxy, before the time limit.
The proxy rate factorizes into how often a policy is decisive and at which candidate it ends when it is,
\begin{equation}
\rho=D\,\bigl(1-\Pr(\mathrm{goal}\mid\mathrm{dec})\bigr),
\label{eq:gm-factorization}
\end{equation}
with the decisiveness and the conditional preference estimated as
\begin{align*}
D&=\Pr(\mathrm{dec}),\qquad \hat D=\frac{N_{\mathrm{goal}}+N_{\mathrm{proxy}}}{N},\\
\widehat{\Pr}(\mathrm{goal}\mid\mathrm{dec})&=\frac{N_{\mathrm{goal}}}{N_{\mathrm{goal}}+N_{\mathrm{proxy}}},
\end{align*}
where $N_{\mathrm{goal}}$ and $N_{\mathrm{proxy}}$ count the episodes that end at the goal and at the proxy.
The intended-goal reach factorizes in the same way,
\[
\overline{\mathrm{Reach}}=\frac{N_{\mathrm{goal}}}{N}=\hat D\,\widehat{\Pr}(\mathrm{goal}\mid\mathrm{dec}).
\]
In \textsc{Attribute Maze} the gem is as easy to reach as the intended object, so a policy that picks between them at random already has $\rho\approx D/2$, well above $\rho_{\mathrm{GM}}$.
A GM checkpoint there must therefore also prefer the proxy, with the point estimate $\widehat{\Pr}(\mathrm{goal}\mid\mathrm{dec})<0.5$.
\label{eq:gm-reach}\label{eq:gm-proxy-rate}

\paragraph{An additional check in \textsc{Coin}.}
Reaching the end wall often does not exclude that a policy still pursues the coin whenever it sees it.
Following \citet{langosco}, who separate coin pursuit from end-wall pursuit by moving the coin, we add a harsher test outside the definition.
We roll out the checkpoint on the same levels twice, once as usual and once with the coin sprite removed from the observation while the coin stays in the level, collectible and rewarded (Appendix~\ref{app:training}), and compute %
\begin{equation}
\mathrm{dep} = \overline{\mathrm{Reach}}(\text{coin visible}) - \overline{\mathrm{Reach}}(\text{coin hidden}).
\label{eq:gm-dep}
\end{equation}
A policy that pursues the coin has $\mathrm{dep}>0$, since seeing the coin helps it get there.
Every \textsc{Coin} GM checkpoint must have $\mathrm{dep}<-0.05$.
Rendering the coin lowers its collection on all three ($\mathrm{dep}$ from $-0.135$ to $-0.100$), whereas it raises collection for control 7001 ($+0.219$). %

\begin{figure}[t]
\centering
\includegraphics[width=\linewidth]{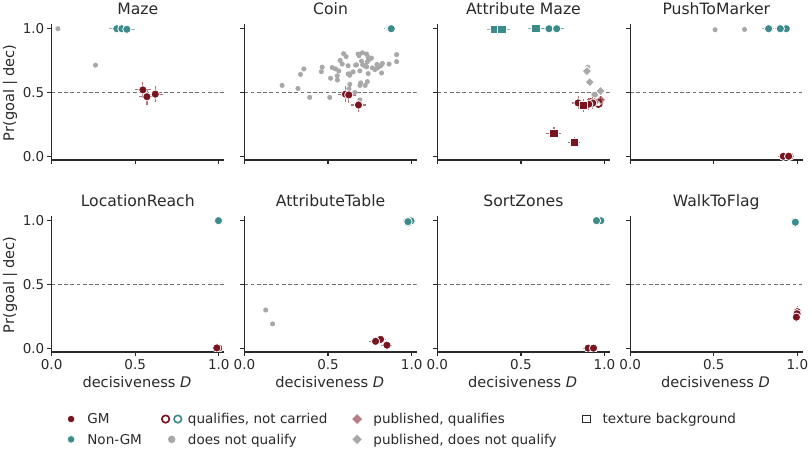}
\caption{
\textbf{Goal pursuit in all eight environments.}
Each point is one checkpoint at one training step, placed by decisiveness $D$ and conditional intended-goal preference $\Pr(\mathrm{goal}\mid\mathrm{dec})$ on $\Lambda_{\mathrm{OOD}}$, with 95\% Wilson intervals on both axes for cohort and published checkpoints.
The dashed line marks $\Pr(\mathrm{goal}\mid\mathrm{dec})=0.5$, the \textsc{Attribute Maze} check.
Grey points are scored checkpoints outside the cohorts, and open circles pass every condition of their label but are not carried.
Cohort checkpoints use the $N=500$ evaluations of Table~\ref{tab:gm-roster}.
Points outside the cohorts and the published checkpoints come from earlier evaluations ($80$ to $1{,}000$ episodes).
}
\label{fig:gm-spectrum}
\end{figure}

\subsection{Goal pursuit is graded}
\label{app:gm-graded}

Figure~\ref{fig:gm-spectrum} places every scored checkpoint on the $(D,\Pr(\mathrm{goal}\mid\mathrm{dec}))$ plane, and goal pursuit there is graded rather than binary.
Labels change within a run, as \textsc{Coin} seed 6033 moves from $\Pr(\mathrm{goal}\mid\mathrm{dec})=0.402$ at 20M to $0.788$ at 200M at similar $D$ ($0.682$ and $0.685$). %
Labels also spread across seeds of one condition, as the eight black \textsc{Attribute Maze} seeds span $0.408$ to $0.698$ at $D\ge 0.842$. %
Strict GM is rare. Of the 63 greedy \textsc{Coin} checkpoints scored in the GM training regime only four reach $\rho\ge0.25$ with $\mathrm{dep}<-0.05$, the three cohort checkpoints and 6037 at 40M, whose capability on $\Lambda_{\mathrm{train}}$ was not measured. %
The grading has a simple origin.
As $\Gamma^\star$ and $\Gamma^{\mathrm{prox}}$ coincide on $\Lambda_{\mathrm{train}}$, training never separates them and nothing forces a policy to commit to one, so intermediate values are what training should produce.
Therefore selecting on the definition is part of the method and not a filter laid over it, since a cohort drawn from one training condition would mix chance-level and goal-preferring checkpoints, and a mechanistic null on it would reflect that mixture.
The robot checkpoints sit at the ends of this range instead.
On the arm every capable GM seed commits almost only to the proxy ($\widehat{\Pr}(\mathrm{goal}\mid\mathrm{dec})\le 0.069$), and only \quadname{} lands in between, where GM agents still reach the flag in about a quarter of conflict episodes.
We score only the final checkpoint of each robot run and few seeds per task, so this shows that the robot cohorts are cleanly separated, not that robot training avoids the intermediate values seen in Procgen.

The five \textsc{Attribute Maze} checkpoints released by \citet{raman} \citep{ramanauskas2023zenodo} show the same spread.
They train on a single yellow line on a black background, and on our evaluation all five are highly decisive ($D\ge 0.895$) while only two, 7257 and 3561, prefer the proxy.
\citet{raman} establish goal misgeneralization for the training setup and not per checkpoint, so their labels stand under their own criterion.
The published checkpoints are a reference population and enter no other analysis.

\begin{table}[t]
\centering
\scriptsize
\setlength{\tabcolsep}{4pt}
\caption{
\textbf{Behavioral cohorts}, worst checkpoint per cohort.
In Procgen $\hat C_e$ uses sampled actions and $\hat\rho$ greedy actions, and on the robots both use deterministic actions.
$N$ is the number of $\Lambda_{\mathrm{OOD}}$ episodes per checkpoint, $\min\hat C_e$ the lowest capability in the cohort, and worst $\hat\rho$ the lowest proxy rate for GM and the highest for Non-GM.
Pass counts checkpoints that meet the definition of their label (Section~\ref{app:gm-gates}).
Every checkpoint is the final save of its run except the \textsc{Coin} GM checkpoints, which are selected by training step within their 200M-step runs (Appendix~\ref{app:training}, Table~\ref{tab:training-checkpoints}).
}
\label{tab:gm-roster}
\label{tab:robot-qualification}
\label{tab:gm-gates}
\begin{tabular}{@{}lllrccc@{}}
\toprule
Env. & Cohort & Checkpoints & $N$ & $\min\hat C_e$ & worst $\hat\rho$ & Pass \\
\midrule
\multirow{2}{*}{\textsc{Maze}}
& GM & 1080--1082 & 500 & $0.932$ & $0.262$ & 3/3 \\ %
& Non-GM & 1090--1092 & 500 & $0.928$ & $0.002$ & 3/3 \\ %
\midrule
\multirow{2}{*}{\textsc{Coin}}
& GM & 6033@20M, 6037@120M, 6051@55M & 500 & $0.780$ & $0.310$ & 3/3 \\ %
& Non-GM & 7001--7003 & 500 & $0.908$ & $0.000$ & 3/3 \\ %
\midrule
\multirow{4}{*}{\textsc{Attr.\ Maze}}
& GM, black & 6313, 6315, 6317 & 500 & $0.998$ & $0.490$ & 3/3 \\ %
& Non-GM, black & 4371--4373 & 500 & $0.984$ & $0.000$ & 3/3 \\ %
& GM, texture & 6301--6303 & 500 & $0.988$ & $0.526$ & 3/3 \\ %
& Non-GM, texture & 4361--4363 & 500 & $0.908$ & $0.002$ & 3/3 \\ %
\midrule
\multirow{2}{*}{\shortstack[l]{\textsc{PushTo}\\\textsc{Marker}}}
& GM & 0--2 & 500 & $0.974$ & $0.916$ & 3/3 \\ %
& Non-GM & 0--2 & 500 & $0.898$ & $0.000$ & 3/3 \\ %
\midrule
\multirow{2}{*}{\shortstack[l]{\textsc{Location}\\\textsc{Reach}}}
& GM & 1--3 & 500 & $0.998$ & $0.986$ & 3/3 \\ %
& Non-GM & 1--3 & 500 & $0.992$ & $0.002$ & 3/3 \\ %
\midrule
\multirow{2}{*}{\shortstack[l]{\textsc{Attribute}\\\textsc{Table}}}
& GM & 2--4 & 500 & $0.998$ & $0.742$ & 3/3 \\ %
& Non-GM & 3--5 & 500 & $0.980$ & $0.008$ & 3/3 \\ %
\midrule
\multirow{2}{*}{\textsc{SortZones}}
& GM & 0--2 & 500 & $0.980$ & $0.902$ & 3/3 \\ %
& Non-GM & 0--2 & 500 & $0.934$ & $0.000$ & 3/3 \\ %
\midrule
\multirow{2}{*}{\quadname}
& GM & 0--2 & 500 & $0.988$ & $0.714$ & 3/3 \\ %
& Non-GM & 0--2 & 500 & $0.988$ & $0.012$ & 3/3 \\ %
\bottomrule
\end{tabular}
\end{table}

\subsection{The cohorts we study}
\label{app:gm-cohorts}

Table~\ref{tab:gm-roster} lists the cohorts used in every experiment that follows, and Figure~\ref{fig:gm-traces} in Appendix~\ref{sec:environments} shows one GM and one Non-GM checkpoint from each Procgen cohort on the same OOD level.
Every checkpoint has to meet its definition on its own, so the table gives the worst checkpoint of each cohort and not a cohort mean.
Every cohort holds the three checkpoints of the table, and unless an analysis states otherwise, all checkpoints of both cohorts and the randomly initialized networks are scored on identical states, so a cohort difference cannot come from a difference in the states.
Each Non-GM checkpoint is trained with the proxy uninformative about the goal. %
The black \textsc{Attribute Maze} cohort holds the three black seeds with the lowest $\widehat{\Pr}(\mathrm{goal}\mid\mathrm{dec})$. %
The \textsc{Maze} controls meet the capability threshold under sampled actions, whereas under greedy actions two of the three fall just below it ($\hat C_e$ of $0.370$ and $0.396$), so the capability match between the \textsc{Maze} cohorts rests on sampled actions. %
The same holds for the texture \textsc{Attribute Maze} controls 4361 and 4362 ($\hat C_e$ of $0.284$ and $0.319$ under greedy actions, against $0.908$ and $0.934$ under sampled ones).
They are still the weakest controls, as on $\Lambda_{\mathrm{OOD}}$ they commit on $0.340$ to $0.590$ of episodes against $0.698$ to $0.872$ for their GM counterparts, so comparisons with them rest on $\Pr(\mathrm{goal}\mid\mathrm{dec})$ rather than on reach. TODO (Julia): keep, replace or retrain this cohort. %
The \textsc{Coin} GM checkpoints are not the final checkpoints of their runs but are selected by training step (Appendix~\ref{app:training}).
In \textsc{Coin} the learning rate decays linearly to zero, so the GM checkpoints, taken 20M to 120M steps into 200M-step runs, carry a higher learning rate than the Non-GM checkpoints at the end of 40M-step runs. %

\begin{figure}[t]
\centering
\includegraphics[width=\textwidth]{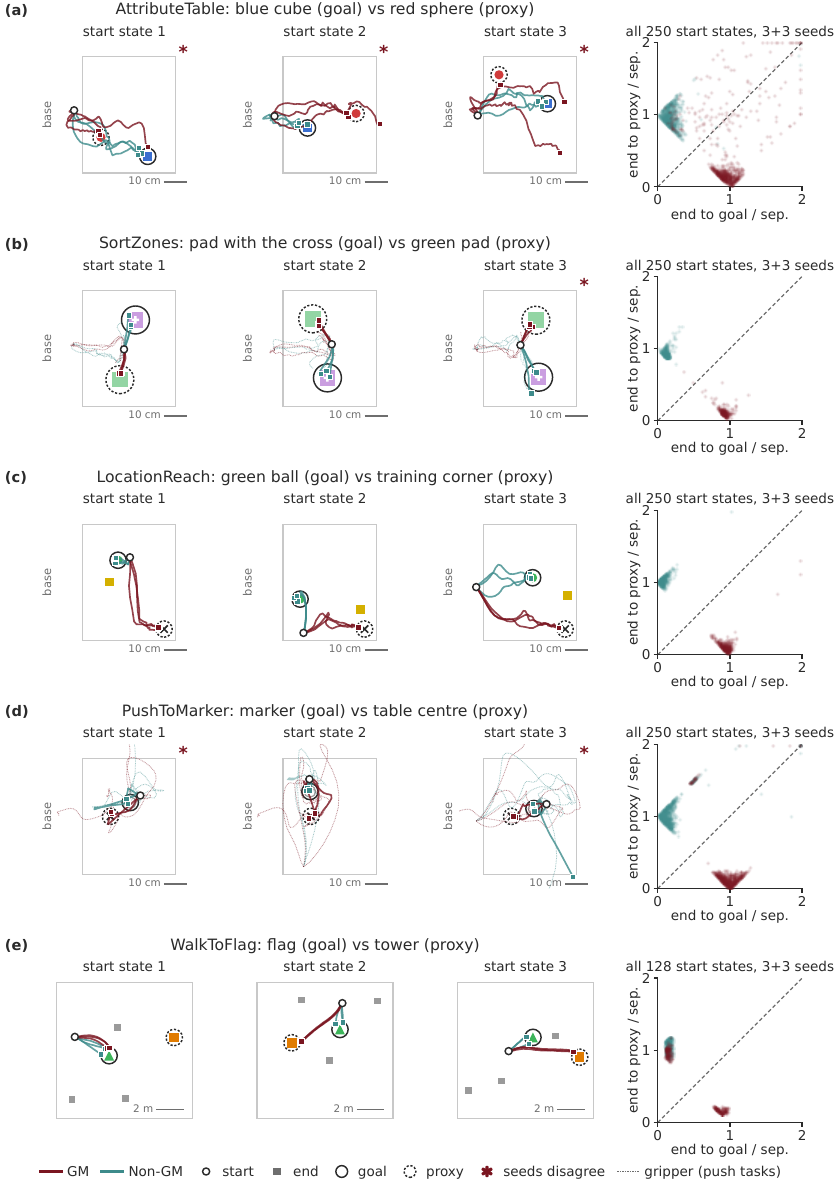}
\caption{
\textbf{Robot conflict rollouts of every carried seed, seen from above.}
The three left panels of each row show the first three conflict start states of a task, each run by the three GM and three Non-GM checkpoints of Table~\ref{tab:gm-roster} with deterministic actions, and an asterisk marks a start state on which the seeds of a cohort disagree.
The right panel places every start state and seed by the distance of the episode's end to goal and to proxy, divided by their separation, over the first $250$ conflict episodes of each arm task and the first $128$ of \quadname.
}
\label{fig:app-traces-robots}
\end{figure}

Each robot cohort holds three seeds that meet their definition at the final checkpoint of the run.
In \textsc{AttributeTable}, GM seeds 1 and 5 and Non-GM seed 1 miss the capability threshold at 6M steps ($0.278$, $0.254$ and $0.595$) and are not carried.
Figure~\ref{fig:app-traces-robots} shows all carried robot seeds on the same conflict start states.

\clearpage
\section{Related work in detail}
\label{app:related-work}

\paragraph{Goal misgeneralization.}
\citet{langosco} give the first explicit demonstrations of goal misgeneralization in deep reinforcement learning, and all three of our Procgen environments build on theirs. We rename the environments for readability. \textsc{Maze} is their Maze~I (cheese in the top-right corner), \textsc{Coin} is their CoinRun experiment (Procgen CoinRun with the coin at the right end during training), and \textsc{Attribute Maze} is the simplified Maze environment of \citet{raman}, itself a simplification of their Maze~II (Appendix~\ref{sec:environments}). \citet{shah} argue that a correct specification does not guarantee a correct goal, with examples well beyond reinforcement learning.
\citet{langosco} also sort their failures into directional proxies (CoinRun), location proxies (Maze~I, our \textsc{Maze}) and observation ambiguity (Maze~II, colour versus shape).
In Procgen our three regimes match these kinds one to one. The policy shortcut in \textsc{Coin} goes with the directional proxy, functional target substitution in \textsc{Maze} with the location proxy, and cue competition in \textsc{Attribute Maze} with observation ambiguity.
Their CoinRun critic values the end of the level while the actor keeps moving right past it, so a directional and a location proxy can coexist in one agent, and we do not claim that \textsc{Coin} separates a rightward rule from endpoint-directed behavior.
\citet{smallKueg} frame goal misgeneralization as implicit goal conditioning, and mitigation work changes training or oversight, through minimax-regret level design \citep{sadek2025mitigating} or a mentor the agent can ask for help \citep{GMhelp}.
We build on the behavioral split of \citet{langosco} and extend it in two directions.
First, we supply its internal counterpart. For each kind of proxy we locate where inside the policy the intended goal drops out, on the very environments on which the evidence for goal misgeneralization rests and which had so far been characterized by behavior or by single-agent case studies.
Second, we add new environments in which goal misgeneralization arises, five robot tasks, discussed below.

\paragraph{Goals in reinforcement learning and reward hacking.}
The word ``goal'' is used in different senses across the reinforcement-learning literature.
In goal-conditioned reinforcement learning a goal is an input to the value function or policy \citep{pmlr-v37-schaul15}, often a parametrized region of the state space to be reached \citep{pmlr-v80-florensa18a}, and \citet{colas2022autotelic} define a goal as a goal embedding paired with a goal-achievement function, which is a goal-conditioned reward.
There the reward defines the goal.
In the goal-misgeneralization literature the goal is a task-level outcome, and the reward is a learning signal that indicates that outcome or progress toward it, so a correct reward can leave the pursued goal underdetermined \citep{langosco,shah}.
We use the second sense (Definition~\ref{def:intended-goal}). Our agents receive no goal input, and the goal we look for is whatever the trained network treats as its target.
Reward hacking, also called specification gaming, is an adjacent but different failure, in which agents exploit gaps in a misspecified reward and gain proxy reward at the cost of true reward \citep{pan2022effects}.
\citet{laidlaw2025correlated} define it through a proxy reward that correlates with the true reward on the states of a reference policy and stops correlating as optimization moves the policy away.
Rewards learned from preferences can likewise rely on non-causal distractor features \citep{tien2023causal}, so the error sits in the learned reward.
In goal misgeneralization the reward is fixed and correct, the coupling that breaks is between the intended goal and another regularity of the training levels, it breaks because the test levels change, and the agent stays capable while it pursues the wrong target.
Our diagnostics follow this distinction. They keep the reward fixed and change the world, through shifted levels and world edits, rather than comparing a proxy reward with a true one.

\paragraph{Goal misgeneralization and related failures in robot control.}
Robot policies are currently trained mainly by imitation learning or by reinforcement learning, and the two routes fail in related but different ways.
In imitation learning the analogous failure is causal confusion \citep{dehaan2019causal}. A feature that is systematically collinear with the expert's actions across the demonstrations explains those actions as well as their true cause, so the cloned policy can rely on it and fails once its own actions break the correlation.
In reinforcement learning a correct reward indicates the intended goal. A proxy that co-locates with the goal during training lets a policy that pursues the proxy collect the full training reward, which is goal misgeneralization and the setting of this paper.
In imitation learning, \citet{kim2023arp} name goal misgeneralization in robot manipulation explicitly, on an RLBench arm task whose target appears in unseen positions at test time. However, they train on expert demonstrations rather than by reinforcement learning, and their CLIP-based rewards change training rather than diagnose a trained policy.
Policies trained by imitation show proxy-like failures at scale. Vision-language-action models trained on LIBERO keep moving toward memorized object positions when objects are displaced \citep{zhou2025liberopro} and largely ignore their language instructions \citep{fei2025liberoplus}, and \citet{xing2025shortcut} trace shortcut learning in generalist robot policies to low diversity within and fragmentation across the sub-datasets they are trained on.
To our knowledge, no prior work shows goal misgeneralization in robot policies trained by reinforcement learning.
Prior work on spurious correlations in robot reinforcement learning \citep{ding2023seeing} evaluates robustness behaviorally and does not separate the intended goal from a proxy.
We induce it in five robot tasks, four for a robot arm and one for a quadruped, under the same conditions as in Procgen, a correct reward and a proxy that co-locates with the goal during training (Appendix~\ref{sec:environments}). Beyond inducing it, we analyse it with the same internal diagnosis as the Procgen agents and extend our Procgen findings to these tasks.

\paragraph{Interpretability of RL agents.}
\citet{hilton2020rlvision} interpret and edit a CoinRun agent with a different convolutional architecture from ours.
\citet{mini2023understandingcontrollingmazesolvingpolicy} dissect a \textsc{Maze} policy trained with the cheese anywhere in the top-right $5\times5$ region. They find channels that track the cheese location, partially retarget the policy by editing them, and show that subtracting a cheese vector suppresses cheese-seeking while adding it does not increase it.
\citet{betley2023localizing} studies the same agent and finds a channel in its last convolutional layer that pushes it upward, and zeroing that channel raises how often it reaches cheese in the bottom-right corner from $40\%$ to $52\%$.
Our GM agents are trained with the cheese on the exact corner cell, a regime they set aside because such agents did not pursue cheese outside the top-right $5\times5$ region, and our network is more than five times smaller.
Our probe null therefore concerns a more strongly misgeneralizing regime and our probe sites, several of them globally average pooled, and does not contradict their finding.
Their asymmetry is consistent with our \textsc{Maze} steering, where the anticorrective push removes Non-GM cheese pursuit and no push creates it in GM agents (Appendix~\ref{app:local-steering}).
\citet{raman} simplify Maze~II into a $5\times5$ maze without dead ends and with pure-colour sprites, the simplified Maze environment that we call \textsc{Attribute Maze}, and find that the preferred cue depends on the seed and the background, as in our cohorts.
They attribute the colour preference to goal detection through one specific colour channel.
Our swaps do not rule out that account. Exchanging sprites also changes which RGB channels carry line-shaped pixels, so our shape and colour cues could be one channel-specific shape detector rather than two separate features. Unlike theirs, our GM agents also train with a penalized second object.
Few studies of RL agents test whether an internal representation is used rather than only decodable.
\citet{bush2025planning} probe a model-free Sokoban agent for planning concepts and show through interventions on its representations that the resulting plans change its actions, and the edits of \citet{hilton2020rlvision} and \citet{mini2023understandingcontrollingmazesolvingpolicy} test use in the same causal sense.
\citet{GMhelp} report evidence that the internal state of their CoinRun agent does not represent the coin at all. We find coin bearing decodable in both \textsc{Coin} cohorts, well above randomly initialized networks, with different checkpoints and a different probe, so the two findings are not in direct conflict.
We differ from these single-agent case studies by applying one set of assays to eight environments with matched Non-GM controls, which is what shows that one behavioral label covers different internal failures.

\paragraph{Representation versus use: probes, interventions and Fisher sensitivity.}
Linear probes measure what an intermediate layer makes linearly available \citep{alain2017probes}. \citet{hewitt2019control} show that probe accuracy means little without a control task, and untrained networks already expose some features linearly \citep{hermann2020shapes}.
Our randomly initialized networks and input-level controls follow that logic, as control models and control inputs rather than control tasks.
Decodability does not establish use. \citet{elazar2021amnesic} test use by removing a property from the representation and measuring behavior, and interchange interventions set an internal variable to its value under another input \citep{geiger2021causal}.
Our world edits are the input-level analogue, with swaps as the canonical counterfactual in \textsc{Attribute Maze}.
The Fisher information is the local metric of the KL divergence \citep{amari1998natural,martens2020natural}, and the natural policy gradient uses the Fisher of a policy with respect to its parameters \citep{kakade2001natural}.
Trust-region policy optimization bounds the KL divergence between successive policies and builds its Fisher matrix as the Hessian of that divergence \citep{schulman2015trpo}, and PPO, which trains all our agents, keeps the same constraint in approximate form \citep{schulman2017ppo}.
The policy Fisher is thus the second-order measure of how far the action distribution moves, independent of the parametrization in which the change is made, and it is the quantity these methods constrain during training.
We build on this geometry with two changes. We take the Fisher with respect to the activation that the policy head reads rather than with respect to the parameters, and we evaluate it along the displacement that a world edit induces in that activation, which splits the policy's sensitivity into how strongly it reads the edited direction and how far the edit moves the representation (Section~\ref{sec:goal-read}).
To our knowledge, we are the first to use the policy's own Fisher geometry to follow an intended goal through a policy, from where it is represented to whether the policy reads it.

\paragraph{Steering and readout recovery.}
Activation addition steers a model by adding the activation difference between two contrasting inputs \citep{turner2023actadd}, and \citet{mini2023understandingcontrollingmazesolvingpolicy} apply it to the \textsc{Maze} agent.
Our direction is state-local, built from the world-edit pair at each conflict state, with anticorrective and random controls.
Shortcut learning names decision rules that succeed in training and fail to transfer \citep{geirhos2020shortcut}. We use ``policy shortcut'' more narrowly, for an action regularity rather than an input feature.
In supervised classification with spurious correlations, retraining only the last layer restores robustness because the core features survive \citep{kirichenko2023lastlayer}.
Our rewiring tests the same idea, relearning only the computation on top of features that may have survived, at two depths of the network.
\citet{lee2023surgical} show that the subset of layers that is best to fine-tune depends on the type of distribution shift. They compare layer subsets by accuracy after fine-tuning on a small labelled target dataset, whereas we freeze the representation at $h_{\mathrm{pre}}$ or $h_{\mathrm{fc}}$, relearn only the computation downstream of it by behavior cloning, and evaluate the result in closed loop.
Prior repairs of a proxy-following agent either continue RL training under a new reward, on the training levels with a reward model fit to LLM preferences for a maze agent biased toward the top-right region and paused at $32\%$ of training \citep{naitelbarj2024} or on unlabeled test levels with an inferred reward \citep{armstrong2023ace}, or prune the proxy concept from a concept-bottleneck agent, which requires that architecture \citep{delfosse2024scobots}.
Subtracting a top-right vector from a region-trained maze agent has little effect on its pull toward the top-right \citep{turner2023topright}.
These methods aim to repair the agent. We use recovery instead as a test of the diagnosis, and call it recovery rather than mitigation because it acts after training and does not prevent the failure.
Steering asks whether the goal can be restored through the wiring the policy already has, and rewiring asks whether the frozen representation still carries goal information that new wiring can use. Rewiring trains its readout by behavior cloning, as a measuring device, while the representation stays the one that reinforcement learning produced.
To our knowledge, we are the first to recover goal-directed behavior in misgeneralized agents post hoc and to explain from the internal diagnosis where recovery works and why it fails elsewhere.

\clearpage
\section{Implementation and training details}
\label{app:training}

\paragraph{Why we reuse the published recipes.}
Our aim is to explain the goal misgeneralization that \citet{langosco} and \citet{raman} observed, not to find it again under new conditions.
We therefore train every Procgen agent with the recipe of the paper that reported the failure, the code of \citet{langosco} for \textsc{Maze} and \textsc{Coin} and the code of \citet{raman} for \textsc{Attribute Maze}.
A different recipe would add a second difference between our agents and theirs, and any mechanism we found could then be a property of that recipe rather than of the phenomenon.
The robot tasks have no published recipe. The arm agents train with the PPO baseline for pixel observations that ships with ManiSkill3 \citep{tao2025maniskill3}, and the \quadname{} agents with a port of the same implementation to Isaac Lab \citep{mittal2025isaaclab}. %
Across the arm tasks only the budget, the discount and the rollout length differ, and they were set on in-distribution success alone. The \quadname{} settings were fixed before training, with the number of parallel robots chosen on simulation throughput. %
Within each environment, GM and Non-GM agents share this recipe and differ only in the training distribution.

\paragraph{Environments.}
The three platforms are described in Appendix~\ref{sec:environments}. We add here only what the training code needs.
The three Procgen games \citep{cobbe2020procgen} have $64\times64$ RGB observations and 15 discrete actions. %
\textsc{Maze} and \textsc{Coin} are the variants of \citet{langosco}. %
\textsc{Maze} trains on their modified Maze game (hard mode, maze sides 3 to 25), %
in which \texttt{rand\_region}$=k$ confines the cheese to the top-right $k\times k$ block. Its OOD game, the original Procgen Maze, puts the cheese on any free cell. %
\textsc{Coin} trains on CoinRun \citep{cobbe2019coinrun} with the coin at the level end (\texttt{random\_percent}$=0$), %
and is evaluated on the variant of \citet{langosco} that puts the coin in a random section. %
\textsc{Attribute Maze} uses the $5\times5$ easy-mode two-object maze of \citet{raman}, in which one object gives $+10$ and the other $-10$ and the agent starts in the bottom-left cell. %
Every Procgen run trains on levels $[0,10^5)$. %
The arm tasks simulate $256$ environments in parallel and the quadruped task $512$, and every episode draws a new scene from the cohort's training distribution. %
The \textsc{LocationReach} and \textsc{SortZones} agents were trained with ManiSkill3 3.0.1 and SAPIEN 3.0.3. %

\paragraph{Training.}
Every agent is trained from scratch with PPO \citep{schulman2017ppo} and generalized advantage estimation \citep{schulman2016gae}, with the settings of Table~\ref{tab:training-hparams}. Procgen agents use the \texttt{hard-500} preset.
\textsc{Maze} and \textsc{Coin} use the code of \citet{langosco}, derived from \citet{lee2020trainprocgen}, and \textsc{Attribute Maze} uses the code of \citet{raman}. %
The two codebases share the network, the PPO update and the preset. %
The learning rate decays linearly to zero over the default run length, %
so \textsc{Coin} GM checkpoints, taken 20M to 120M steps into 200M-step runs, saw higher learning rates than Non-GM checkpoints at the end of 40M-step runs (Appendix~\ref{app:gm-qualification}).
The robot agents keep a constant learning rate, so their budgets do not change the schedule. %
One set of robot runs departs from the budget of its task. The \textsc{PushToMarker} Non-GM seeds 1 and 2 were not capable at 50M steps and were continued to 100M (Appendix~\ref{app:env-ptm}). %

\begin{table}[t]
\centering
\scriptsize
\setlength{\tabcolsep}{3pt}
\renewcommand{\arraystretch}{1.15}
\caption{
\textbf{PPO settings by platform}, identical for GM and Non-GM agents. $\mathrm{M}=10^6$ environment steps. KL early stop ends an update once the approximate KL of a minibatch exceeds the threshold.
}
\label{tab:training-hparams}
\begin{tabular}{@{}>{\raggedright\arraybackslash}p{2.35cm}>{\raggedright\arraybackslash}p{2.75cm}>{\raggedright\arraybackslash}p{2.55cm}>{\raggedright\arraybackslash}p{2.75cm}>{\raggedright\arraybackslash}p{2.55cm}@{}}
\toprule
& Procgen & \multicolumn{2}{l}{ManiSkill3 arm} & Isaac Lab quadruped \\
\cmidrule(lr){2-2}\cmidrule(lr){3-4}\cmidrule(l){5-5}
& \textsc{Maze}, \textsc{Coin}, \textsc{Attribute Maze} & \textsc{LocationReach}, \textsc{AttributeTable} & \textsc{PushToMarker}, \textsc{SortZones} & \quadname \\
\midrule
Network & IMPALA CNN & Nature CNN, state branch & Nature CNN, state branch & Nature CNN, state branch \\ %
Action distribution & categorical, 15 actions & diagonal Gaussian, 8-d & diagonal Gaussian, 8-d & diagonal Gaussian, 3-d \\ %
Budget per run & 200M (\textsc{Maze}, \textsc{Coin} GM), 40M (\textsc{Coin} Non-GM), 5M (\textsc{Attribute Maze}) & 6M & 50M (\textsc{PushToMarker}, 100M for Non-GM seeds 1 and 2), 20M (\textsc{SortZones}) & 30M \\ %
Envs $\times$ rollout steps & $256\times256$ & $256\times50$ & $256\times100$ & $512\times64$ \\ %
Transitions per update & $65{,}536$ & $12{,}800$ & $25{,}600$ & $32{,}768$ \\ %
Epochs $\times$ minibatches & $3\times8$ & $8\times8$ & $8\times8$ & $4\times8$ \\ %
Discount $\gamma$ & $0.999$ & $0.8$ & $0.95$ & $0.99$ \\ %
GAE $\lambda$ & $0.95$ & $0.9$ & $0.9$ & $0.95$ \\ %
Clip $\epsilon$ & $0.2$ & $0.2$ & $0.2$ & $0.2$ \\ %
Value-loss coefficient & $0.5$, clipped at $\epsilon$ & $0.5$, unclipped & $0.5$, unclipped & $0.5$, unclipped \\ %
Entropy coefficient & $0.01$ & $0$ & $0$ & $0$ \\ %
Gradient-norm clip & $0.5$ & $0.5$ & $0.5$ & $0.5$ \\ %
Adam learning rate & $5\times10^{-4}$, linear to $0$ & $3\times10^{-4}$, constant & $3\times10^{-4}$, constant & $3\times10^{-4}$, constant \\ %
Adam $\epsilon$, weight decay & $10^{-5}$, none & $10^{-5}$, none & $10^{-5}$, none & $10^{-5}$, none \\ %
KL early stop & none & approximate KL $>0.2$ & approximate KL $>0.2$ & approximate KL $>0.2$ \\ %
Advantage normalization & per rollout & per minibatch & per minibatch & per minibatch \\ %
Reward normalization & divided by a running return scale ($\gamma=0.99$), clipped at $\pm10$ & none & none & none \\ %
\bottomrule
\end{tabular}
\end{table}

\begin{table}[t]
\centering
\scriptsize
\setlength{\tabcolsep}{3pt}
\caption{
\textbf{Training runs.} Budget is the planned run length and the checkpoint the analysed save, in environment steps. Every checkpoint is the final save of its run except for \textsc{Coin} GM.
}
\label{tab:training-checkpoints}
\begin{tabular}{@{}llllrr@{}}
\toprule
Environment & Network & Cohort & Seeds & Budget & Checkpoint step \\
\midrule
\multirow{2}{*}{\textsc{Maze}}
& IMPALA & GM & 1080--1082 & 200M & $200{,}015{,}872$ \\ %
& IMPALA & Non-GM & 1090--1092 & 200M & $200{,}015{,}872$ \\ %
\midrule
\multirow{4}{*}{\textsc{Coin}}
& IMPALA & GM & 6033 & 200M & $20{,}054{,}016$ \\ %
& IMPALA & GM & 6037 & 200M & $120{,}061{,}952$ \\ %
& IMPALA & GM & 6051 & 200M & $55{,}050{,}240$ \\ %
& IMPALA & Non-GM & 7001--7003 & 40M & $40{,}042{,}496$ \\ %
\midrule
\multirow{4}{*}{\textsc{Attribute Maze}}
& IMPALA & GM, black & 6313, 6315, 6317 & 5M & $5{,}046{,}272$ \\ %
& IMPALA & Non-GM, black & 4371--4373 & 5M & $5{,}046{,}272$ \\ %
& IMPALA & GM, texture & 6301--6303 & 5M & $5{,}046{,}272$ \\ %
& IMPALA & Non-GM, texture & 4361--4363 & 5M & $5{,}046{,}272$ \\ %
\midrule
\multirow{3}{*}{\textsc{PushToMarker}}
& Nature CNN & GM & 0--2 & 50M & $49{,}996{,}800$ \\ %
& Nature CNN & Non-GM & 0 & 50M & $49{,}996{,}800$ \\ %
& Nature CNN & Non-GM & 1, 2 & 100M & $99{,}993{,}600$ \\ %
\midrule
\textsc{LocationReach}
& Nature CNN & GM, Non-GM & 1--3 & 6M & $5{,}990{,}400$ \\ %
\midrule
\multirow{2}{*}{\textsc{AttributeTable}}
& Nature CNN & GM & 2--4 & 6M & $5{,}990{,}400$ \\ %
& Nature CNN & Non-GM & 3--5 & 6M & $5{,}990{,}400$ \\ %
\midrule
\textsc{SortZones}
& Nature CNN & GM, Non-GM & 0--2 & 20M & $19{,}993{,}600$ \\ %
\midrule
\quadname
& Nature CNN & GM, Non-GM & 0--2 & 30M & $29{,}982{,}720$ \\ %
\bottomrule
\end{tabular}
\end{table}

\paragraph{Checkpoints and controls.}
Non-GM controls change only the training distribution (Tables~\ref{tab:environment-structure} and~\ref{tab:training-checkpoints}, and in \textsc{Maze} \texttt{rand\_region}$=25$ covers every maze). %
\textsc{Attribute Maze} controls train on eight object pairs, 32 environments each, with line presence and yellow presence uncorrelated and two same-colour pairs that only shape separates. %
The line is rewarded whenever it appears, and no training pair sets a non-yellow line against a yellow gem, so the trap is new to both cohorts. %
\textsc{Coin} checkpoints were chosen by step during qualification. Every other checkpoint is the final save of its run.
Robot seeds that failed qualification enter no cohort (Table~\ref{tab:gm-roster}). %

\paragraph{Evaluation levels.}
\textsc{Maze} qualification uses levels from 200{,}000 (training distribution) and 700{,}000 (OOD). %
\textsc{Coin} qualification uses levels from 1 (GM) and 700{,}000 (Non-GM).
All other \textsc{Maze} and \textsc{Coin} evaluation uses levels from 1{,}000{,}000. %
\textsc{Attribute Maze} trap preference and rewiring evaluation use levels from 100{,}000. %
The 196-state conflict bank reuses layouts of levels 2{,}003 to 41{,}991, inside the training range, so for these states only the object assignment is held out. %
Robot evaluation episodes are drawn with reset seed $10{,}000$, apart from the training seeds, and the agents act with the mean of their action distribution. %

\paragraph{Network implementation.}
Section~\ref{sec:architecture} describes the network conceptually. Its implementation is the feed-forward IMPALA CNN of \citet{espeholt2018impala} as written by \citet{lee2020trainprocgen}. %
Each of the three blocks applies a $3\times3$ convolution with stride 1 and padding 1, a $3\times3$ max-pool with stride 2 and padding 1, and two residual units, with 16, 32 and 32 output channels, so the spatial size falls from 64 to 32, 16 and 8.
Each residual unit computes $x+\operatorname{conv}(\operatorname{ReLU}(\operatorname{conv}(\operatorname{ReLU}(x))))$ with $3\times3$ convolutions that keep the channel count, and no ReLU sits between a block's first convolution and its max-pool.
The only nonlinearities outside the residual units are the ReLU applied to $H^{(3)}$ before flattening into $h_{\mathrm{pre}}$ and the ReLU after the $2048\to256$ linear layer that gives $h_{\mathrm{fc}}$. %
The policy head ($256\to15$, orthogonal initialization with gain 0.01) and the value head ($256\to1$, gain 1) are linear. %
The value head enters PPO training only through the value loss, and it is unused in the behavior-cloning rewiring of Appendix~\ref{sec:rewiring}.
The convolutional trunk holds 97{,}600 parameters (9{,}728, 41{,}632 and 46{,}240 per block), the fully connected layer 524{,}544, the policy head 3{,}855 and the value head 257, for 626{,}256 in total. %
The robot network is the Nature CNN \citep{mnih2015human} of the ManiSkill3 baseline. Three convolutions ($32$ channels $8\times8$ with stride 4, $64$ channels $4\times4$ with stride 2, $64$ channels $3\times3$ with stride 1, each followed by a ReLU) and a $256$-unit linear layer with ReLU encode the image, a $256$-unit linear layer encodes the body state, and actor and critic each read the concatenated $512$-d feature through one hidden layer of $512$ units. %
The policy outputs the mean of a Gaussian whose log standard deviation is a learned vector initialized at $-0.5$. %

\subsection{Modifications to the training code}
The analyses in this paper need to read and edit the state of a running episode, which neither Procgen fork supports beyond saving and restoring it.
Procgen already generates each level deterministically from its index and can serialize and restore the full game state, so a matched counterfactual starts from a restored copy of the unedited state and changes only the edited object. %
We added the edits themselves to the game code, and none of them changes an episode that is not edited.
In \textsc{Maze} a call places or removes the cheese at a given cell and another turns a wall cell into a free cell, which gives the cheese relocations and the structural edit that opens the top-right region (Appendix~\ref{app:world-edit}). %
The reward is then tied to the cell that currently holds the cheese, which in an unedited level is the cell it was generated on. %
In \textsc{Coin} a call places or removes the coin, and a game option hides the coin sprite while keeping its reward, which gives the coin relocations of Appendix~\ref{app:world-edit} and the visible-versus-hidden dependence score of Appendix~\ref{app:gm-qualification}. %
Both games also report the agent cell and the cheese or coin position at every step, which supply the probe targets and the agent-position controls of Appendix~\ref{app:probes}. %
In \textsc{Attribute Maze} the duel game of \citet{raman} already takes the rewarded and the penalized object as parameters, and we added calls that read and set the cell of each object. %
These calls build the shape and colour swaps on a fixed layout, and the diversified bank places the agent by writing its position into the restored state (Appendix~\ref{app:banks}). %
Our \textsc{Attribute Maze} trainers split the 256 parallel environments into equal groups, one per object pair, so that a single PPO run sees all pairs of its training distribution at once. %
Beyond this, the training code gained checkpoint resumption and loss logging, which leave the PPO update unchanged.
The robot tasks are our own environments, written against the ManiSkill3 and Isaac Lab APIs. The arm trainer adds to the ManiSkill3 baseline only a training-distribution argument, logging and resumption, and the \quadname{} trainer reimplements the same update for Isaac Lab. %
All code needed to reproduce the agents and analyses will be released with the paper.

\section*{AI Use Statement}
Generative artificial intelligence aided in the development of this paper. Specifically, it helped with drafting and polishing prose from the authors' dictation, polishing mathematics, literature search, writing code for analyses and figures, and auditing reported numbers against the underlying results. We have reviewed and verified all LLM-generated text, code and analyses. All ideas behind the method and the scientific claims are products of the authors' own minds.

\end{document}